\documentclass[12pt,a4paper]{report}

\title{\textbf{Geometric Deep Learning For Camera Pose Prediction, Registration, Depth Estimation, and 3D Reconstruction}}
\author{
    Xueyang Kang\thanks{Email: kangxueyang@126.com} \\
    KU Leuven, Belgium \\
    The University of Melbourne, Australia
}
\usepackage{nomencl}   
\makenomenclature

\usepackage{subcaption}
\usepackage{textcomp} 
\usepackage{pifont}   
\usepackage{booktabs}
\usepackage{amssymb,amsthm}
\usepackage{amsmath}
\usepackage{dsfont}

\usepackage[ruled, linesnumbered]{algorithm2e}
\usepackage{algpseudocode} 
\usepackage{gensymb}
\usepackage{float} 
\usepackage{algorithm2e}
\SetArgSty{textup}
\usepackage{adjustbox}
\usepackage{scalerel}
\usepackage{multirow}
\usepackage{tabularx}
\usepackage{makecell} 
\usepackage{array}

\usepackage{titlesec}
\usepackage{enumitem}
\usepackage{hyperref}
\usepackage{graphicx}
\usepackage[dvipsnames,table]{xcolor}

\newcommand{\cmark}{\ding{51}}  
\newcommand{\xmark}{\ding{55}}  
\newcommand{\revAouda}[1]{\textcolor{black}{#1}}   
\newcommand{\revPatrice}[1]{\textcolor{black}{#1}}    
\newcommand{\revAdalberto}[1]{\textcolor{black}{#1}} 
\newcommand{\revDavood}[1]{\textcolor{black}{#1}} 

\usepackage[dvipsnames]{xcolor}
\usepackage{pgf}
\usepackage{tikz} 
\usetikzlibrary{positioning, shapes.geometric, shapes,arrows,automata}

\usepackage{nomencl}
\makenomenclature

\pgfdeclarelayer{background}
\pgfdeclarelayer{foreground}
\pgfsetlayers{background,main,foreground}   

\usetikzlibrary{arrows.meta}
\tikzset{>={Latex[width=1.8mm,length=1.5mm]}}

\begin{document}


\begin{titlepage}
	
	\vspace*{0.42cm}
    \begin{center}
        \Large
        {\textbf{Geometric Deep Learning} \\ 
        \large \textbf{for Camera Pose Prediction, Registration, Depth Estimation, and 3D Reconstruction}}
        \vspace{1.0cm}
    
        Xueyang Kang \\ 
        \small ORCID: 0000-0001-7159-676X \\  
    
        \vspace{1.5cm}
        \large
        \textbf{Doctor of Philosophy} \\
        \textbf{September 2025} \\
        \vspace{0.5cm}
        Faculty of Engineering and Information Technology \\
    \end{center}


		
		
 

	\hfill\begin{minipage}{\dimexpr\textwidth}
    \vspace{1.2cm}
Submitted in fulfillment of the requirements for the awarded degree of Doctor of Philosophy at the University of Melbourne
	\vspace{0.5cm}
    \end{minipage}
		

\end{titlepage}

\maketitle

\pagenumbering{roman}

\chapter*{Preface \\ -Acknowledgements}                                  \label{ch:preface}

\section*{Preface}



I sincerely thank my collaborators and mentors for their invaluable support, guidance, and encouragement throughout this journey. I am especially grateful to my parents for their unwavering belief in me, both spiritually and financially. Their love and support have been a constant source of strength.  

A special thanks to Ariel, Henry, Alex, Cipher, and Iacopo, the undergraduate and postgraduate students who collaborated with me during my Ph.D. Their enthusiasm and dedication played a crucial role in advancing my research.  

I extend my deepest gratitude to Prof. Patrick Vandewalle (KU Leuven) and Prof. Kourosh Khoshelham (the University of Melbourne) for shaping and supporting my joint Ph.D. program. Their efforts in bridging research resources between Leuven and Melbourne provided me with an invaluable cross-disciplinary research experience, integrating theory with practical applications.  

I am also grateful to Dr. Gong Dong, Dr. Jin Wanxin, and Dr. Wang Bing for their invaluable guidance in academic writing. Their mentorship helped me meet the quality standards required for top-tier venues in computer vision and robotics, and I have learned tremendously from these talented researchers.  

Finally, I express my profound appreciation to Prof. Matthias Niessner (Technical University of Munich), who mentored me during my first Ph.D. year. His guidance was instrumental in shaping my early research on 3D deep learning, and his pioneering work inspired my academic idea growth. I am deeply grateful for his mentorship and for the knowledge I gained from his exceptional research contributions.  

\noindent Thank you all for your support in my joint Ph.D. journey.


\cleardoublepage


\begin{abstract}
Modern deep learning developments create new opportunities for 3D mapping technology, scene reconstruction pipelines, and virtual reality development. Despite advances in 3D deep learning technology, direct training of deep learning models on 3D data faces challenges due to the high dimensionality inherent in 3D data and the scarcity of labeled datasets. Structure-from-motion (SfM) and Simultaneous Localization and Mapping (SLAM) exhibit robust performance when applied to structured indoor environments but often struggle with ambiguous features in unstructured environments. These techniques often struggle to generate detailed geometric representations effective for downstream tasks such as rendering and semantic analysis. Current limitations require the development of 3D representation methods that combine traditional geometric techniques with deep learning capabilities to generate robust geometry-aware deep learning models.

The dissertation provides solutions to the fundamental challenges in 3D vision by developing geometric deep learning methods tailored for essential tasks such as camera pose estimation, point cloud registration, depth prediction, and 3D reconstruction. The integration of geometric priors or constraints, such as including depth information, surface normals, and equivariance into deep learning models, enhances both the accuracy and robustness of geometric representations. This study systematically investigates key components of 3D vision, including camera pose estimation, point cloud registration, depth estimation, and high-fidelity 3D reconstruction, demonstrating their effectiveness across real-world applications such as digital cultural heritage preservation and immersive VR/AR environments.

The first research project of this dissertation introduces a vision-based camera pose tracking system for robust camera orientation estimation in natural environments, particularly for UAV-based imaging and data collection. The proposed method uses geometric cues of skylines and ground planes from nature to enhance orientation stability and mitigate motion drift of predicted orientation. By integrating a lightweight ResNet backbone and an adaptive particle filter, the system achieves real-time performance on embedded hardware, outperforming IMU-based solutions in stability and robustness against drift.

The second part focuses on point cloud registration, a fundamental problem in 3D vision. To address the limitations of conventional feature-based registration methods, this work introduces a 2D surfel-based $\mathbf{SE(3)}$-equivariant deep learning framework. The model learns robust point cloud alignment by leveraging surfel representations extracted from RGB-D data or LiDAR scans. The proposed method demonstrates superior and robust registration accuracy through extensive evaluations of indoor and outdoor datasets, particularly in low inlier-to-outlier ratio input point clouds, making it highly applicable for robotics and mobile cases.

The third part of this dissertation explores depth estimation, a key component for generating dense 3D reconstructions. Specifically, it investigates depth prediction from focal stack images, which infers depth information based on focus/defocus cues. A novel Transformer-based network, FocDepthFormer, is proposed, integrating self-attention mechanisms with an LSTM-based recurrent module to handle focal stacks of arbitrary lengths. Compared to traditional CNN-based methods, the proposed approach achieves state-of-the-art performance across benchmark datasets, significantly improving depth prediction accuracy while reducing reliance on large-scale focal stack training data.

The final part presents a high-fidelity 3D reconstruction framework using implicit Signed Distance Fields (SDF). The wavelet-transformed depth feature is used to condition the implicit SDF model to preserve fine-grained geometric details in implicit geometry representations. By integrating wavelet-transformed depth features with a fusion in triplane latent space, the model achieves superior accuracy and detail preservation in reconstructed 3D surfaces. Extensive experiments on DTU, Tanks, Temples, and cultural heritage datasets validate the effectiveness of this approach, demonstrating its applicability to both small-scale objects and large architectural buildings.

Through extensive experimentation on public datasets, the four proposed methods consistently outperform state-of-the-art techniques across individual 3D vision tasks. This dissertation highlights how geometric deep learning bridges the gap between traditional geometry-based methods and data-driven deep learning approaches, driving advancements in 3D vision research and its practical applications in real-world scenarios, such as digital cultural heritage.

\end{abstract}
\cleardoublepage





\printnomenclature
                           
\listoffigures
\listoftables


\pagenumbering{arabic}


\chapter{Introduction}\label{ch:introduction}
%





3D vision is a rapidly evolving field that addresses real-world challenges such as pose estimation for localization, point cloud registration for global mapping, depth estimation from images to generate dense point clouds, and 3D reconstruction from casually captured photos. The primary goal of 3D vision is to infer structure and recover geometry from raw images and laser scan data. However, the reconstruction process is complex, typically involving multiple modular components, including pose estimation between frames, dense point cloud generation from depth maps, and establishing accurate correspondences between frames.

\section{Motivation}


Improving 3D vision techniques for real-world applications can create great profits for society, helping automate tasks that used to be expensive, error-prone, inefficient, or just too computationally heavy. For instance, predicting camera position and orientation, usually known as camera pose estimation is very important for applications like self-driving cars, augmented reality, and robot vision, where you need to know exactly where you are in dynamic, challenging environments for navigation. To tackle this, we have built a vision-based system that tracks camera orientation and stabilizes camera motion by using geometric hints from the geometric primitives of the skyline and ground plane, then further fused by IMU prediction by the adaptive particle filter constrained on the manifold. This makes pose estimation much more reliable and accurate in natural settings, like drone surveys, outdoor mapping, or autonomous driving.


Another key challenge in 3D vision is point cloud registration, which is crucial for tasks like mapping cities or digital twins of real properties, helping robots perceive or inspect industrial components  We have come up with a surfel-based registration framework that uses $\mathbf{SE}(3)$-equivariant features to handle input data noise and improve the learning efficiency and generalization for both translation and rotation. This improves alignment accuracy in real-world 3D scans, which is a key technique for creating digital twins, inspecting infrastructure, or automating 3D modeling through autonomous alignments of local scan data.

Depth estimation from a focal stack is a great solution for situations where LiDAR is too costly to use or camera motion is constrained in some workspace settings, like robotic grasping of small objects, or medical and biological imaging. We have developed a Transformer-based network called FocDepthFormer, which can predict depth from focal stacks of any length, making it much more flexible for different stack imaging setups. This opens up exciting possibilities for 3D photography on smartphones, microscopy, and quality control in industries where depth sensing is key but depth sensor hardware is limited.


3D reconstruction is also a critical technique for things like virtual reality, self-driving, preserving cultural heritage, and urban planning. To help boost the creation efficiency of digital 3D assets, we have created a wavelet-feature-conditioned implicit SDF model that uses multi-scale wavelet-transformed depth features to capture those geometry details more accurately for the reconstruction of meshes with fine-grained details. This leads to better, more detailed 3D models, which is a huge help for digitizing historical sites, building virtual worlds, or digital manufacturing.


By introducing various kinds of geometric constraints or geometry prior to deep learning models, our research aims to create 3D vision solutions that are more reliable, scalable, efficient, and generalizable. We are working to boost automation, cut down on manual efforts, and promote the use of 3D deep learning models with robust and good performance, to open up new possibilities across fields as diverse as robotics and the creative industries.



\section{Research Problems \& Objectives}

This thesis explores how to develop geometric deep learning models for camera pose estimation, point cloud registration, depth estimation, and 3D reconstruction, handling the key shortcomings of both traditional and deep learning-based approaches in complex real-world environments, which may be unstructured and natural. 
While deep learning offers a promising path by directly learning 3D geometric representations, existing approaches remain limited by high computational costs, the need for large-scale 3D data, and heavy reliance on data augmentation. These challenges stem from three main factors: the absence of structured geometric priors, the difficulty of training in high-dimensional 3D feature spaces, and the lack of robustness to real-world uncertainties. To address these gaps, this thesis proposes hybrid frameworks that integrate traditional geometric constraints with modern deep learning architectures, aiming to improve accuracy, generalization, and efficiency across several core 3D vision tasks.

Given that multi-view 3D reconstruction is a complex system composed of interconnected modules—such as camera pose estimation, depth prediction, and feature correspondence matching—this thesis decomposes the overall problem into a series of subtasks. Each subtask is examined in depth to develop targeted solutions, which are then progressively combined to optimize the reconstruction process. Rather than relying solely on an end-to-end black-box deep learning model, the proposed approach emphasizes subtask-level optimization, ensuring both interpretability and robustness. Finally, the concluding chapter presents the integration of these techniques into a complete 3D reconstruction framework, demonstrating how the modular solutions contribute to the performance of the overall system.

\noindent\textbf{Camera pose estimation for images using natural geometry cues and manifold constraints.} Traditional camera pose estimation depends on point feature matching and iterative optimization techniques, such as Random Sample Consensus (RANSAC) \cite{fischler1981random}. However, these approaches perform poorly in situations where visual ambiguity and noisy input lead to a low ratio of inliers to outliers, which makes pose-solving fail.

Deep learning-based pose regression models have been proposed as an alternative, but they still face significant challenges, particularly in accurately predicting rotation due to the complex nature of geometric transformations. One major limitation is that these models lack explicit constraints on the underlying geometric structure and reliable reference cues, making it difficult for them to predict accurate poses in the long-term run due to pose drift. How to leverage some new cues from nature and how to use the geometric constraints along with deep learning techniques to create a hybrid system for robust and accurate pose estimation.

Thus, the goal of Chapter 3 is to design a robust and lightweight camera pose prediction model capable of running on mobile devices, leveraging natural reference cues extracted from images and robust orientation constraints for improved stability and accuracy of camera pose prediction.


\noindent\textbf{Point cloud registration using 2D Surfel-based equivariance constraint.} Standard point cloud registration techniques align 3D point clouds by matching keypoint features. However, these methods struggle when the features are sparse, noisy, or ambiguous, making alignment unreliable and challenging. While recent learning-based approaches attempt to learn geometric representation and spatial transformations directly, they often overlook the impact of uncertainty factors and rotation feature representations, resulting in poor generalization to noisy and unseen data, particularly for the input point cloud scans with small overlaps.

To address these challenges, Chapter 4 presents a surfel-based registration method that incorporates $\mathbf{SE}(3)$-equivariant surfel constraints into deep learning architectures, ensuring robust and generalizable point cloud alignment, particularly for input scans with small overlaps or a high ratio of outliers.

\noindent \textbf{Depth prediction from the focal stack using focal geometry Constraint.} Depth estimation from a focal stack takes advantage of the way focus distance changes with scene depth. However, traditional depth from focal stack methods are limited by the requirement for a fixed number of input images, making them impractical for many real-world applications where image sequences vary in length, and both the training and testing are very inefficient. Furthermore, most existing models are built on convolutional architectures, which have limited receptive fields to capture global multi-scale features. 

Chapter 5 proposes a Transformer-based adaptive depth estimation model, which overcomes this limitation by leveraging self-attention mechanisms to extract long-range focus cues in the focal stack and perform latent fusion by LSTM to handle varying input lengths. By embedding focal geometry information into the training loss, this approach allows the model to efficiently process an arbitrary number of input images, significantly improving the performance of depth from the focal stack.

\noindent\textbf{3D reconstruction using implicit SDF with wavelet feature-based prior.} 3D reconstruction from multi-view images is progressing very fast. The key techniques for 3D reconstruction can be categorized into explicit and implicit models, where the explicit Gaussian Splatting model although fast yet struggles with continuous geometry presentation due to the discretized Gaussian blobs, and the Implicit Signed Distance Function-based methods have proven to be effective for 3D reconstruction, as they preserve continuous surface representations. However, current implicit approaches struggle to capture fine-grained geometric details, especially when dealing with multi-scale geometric structures and sharp edges. This limitation arises because existing implicit methods have difficulty encoding high-frequency geometric features due to the limited learning capability of implicit models via MLPs, leading to over-smoothed reconstructions.

To tackle this, Chapter 6 introduces a wavelet-transformed-depth-feature-conditioned implicit SDF model, which incorporates multi-resolution geometric priors using wavelet-based feature decomposition and triplane feature fusion. By leveraging the features encoded in different frequency bands, the proposed model enhances multi-scale feature representation, preserving sharp geometric details of the input images and improving the accuracy and details of reconstructed surfaces, to overcome the limitation of the implicit model learning capacity.

\section{Contributions} 





My thesis contribution focuses on combining deep learning with geometric constraints or priors for a range of 3D vision tasks, introducing new ways to represent geometric features and guide the learning process more efficiently for 3D challenges. The primary focus areas of this thesis are camera pose estimation, point cloud registration, depth prediction, and 3D reconstruction.

The first three tasks—camera pose estimation, point cloud registration, and depth prediction—form the essential building blocks of a robust reconstruction pipeline. Camera pose estimation provides accurate geometric alignment across views, point cloud registration ensures consistency in merging multiple scans of data, and depth prediction supplies dense geometric information to recover detailed scene structure. By addressing each of these subtasks individually with hybrid deep learning–geometric approaches, the thesis establishes strong foundations that directly feed into the final 3D reconstruction system.

The contributions of the early chapters, therefore, not only advance the state of the art within their respective domains but also collectively support the integration of a modular, optimized 3D reconstruction framework presented in the concluding chapter. This modular-to-system perspective highlights how solving core subtasks with principled designs can lead to a more accurate, generalizable, modular, and efficient solution for the broader reconstruction challenge. In the following sections, we break down the key methodological contributions of each chapter.

\noindent\textbf{Camera pose estimation for images using natural geometry cues and manifold constraints.} Chapter 3 presents a new vision-based orientation tracking and fusion algorithm for robust camera pose estimation in natural outdoor environments, particularly for UAV-based wild investigation applications. The approach leverages a lightweight ResNet-18 backbone, deployed on an embedded Jetson Nano, to perform real-time binary segmentation of ground and sky regions. By utilizing the skyline and ground plane as reference cues, a natural geometric primitive-based camera pose estimation framework is developed to enhance visual tracking under challenging natural conditions. Additionally, an adaptive particle filter operates on a multi-resolution manifold surface, enabling the flexible fusion of orientation estimates from both vision-based cues and IMU data, to make the camera pose tracking more robust. 

\noindent\textbf{Point cloud registration using 2D surfel-based equivariance constraint.} Chapter 4 introduces a novel point cloud registration framework that leverages $\mathbf{SE}(3)$-equivariant 2D Gaussian surfel features to improve point cloud registration accuracy and robustness. By representing local surface geometry as learned 2D Gaussian-based surfels, the method preserves explicitly $\mathbf{SE}(3)$ equivariance for learning, ensuring robustness to rigid transformations. The encoder is an adapted E2PN encoder \cite{zhu2023e2pn} to take in both point position and orientation, improving registration performance by learning both point position and orientation representations in noisy and partial point cloud scenarios. Additionally, structured huber loss refines pose regression, enhancing robustness to outliers. 

\noindent \textbf{Depth prediction from the focal stack using focal geometry constraint.} In this part, a novel Transformer-LSTM-based network for depth estimation from focal stacks, addresses limitations in conventional CNN-based depth from focal stack approaches. The model integrates a vision Transformer encoder to capture local spatial feature cues and an LSTM-based recurrent module to aggregate stack cues along the main stack dimension, which can be arbitrary lengths. This design overcomes the constraints of fixed-stack-size processing of conventional methods, improving flexibility and generalization to diverse focal stacks. Additionally, a multi-scale convolutional encoder extracts fine-grained focus/defocus features, enhancing depth prediction accuracy by learning more accurate feature representations. 

\noindent\textbf{3D reconstruction using implicit SDF with wavelet feature-based prior.} This chapter introduces a wavelet-transformed depth feature to condition the implicit SDF model for high-fidelity 3D reconstruction, addressing the problem of fine-grained geometric detail loss in deep implicit representations. By leveraging a pre-trained wavelet autoencoder trained on sharp depth maps, our approach extracts multi-scale wavelet-transformed features that efficiently capture high-frequency geometric details like edges. These features are aligned via triplane projection and fused with implicit triplane features, enhancing surface accuracy while preserving complete structures. A hybrid UNet-based fusion module further refines SDF predictions, leading to more precise isosurface extraction. 

Overall, this thesis pushes geometric learning in 3D vision applications forward, tackling current challenges in various 3D tasks. Combining traditional geometric constraints or prior knowledge with deep learning helps to boost 3D vision models that are efficient, accurate, generalizable, and robust for real-world applications like robotics, augmented reality, and self-driving tech.

\section{Thesis Outline}
This thesis is structured into eight chapters, including the four main projects from Chapter 3 to Chapter 6.

\begin{figure}
    \centering
    \includegraphics[width=1.0\linewidth]{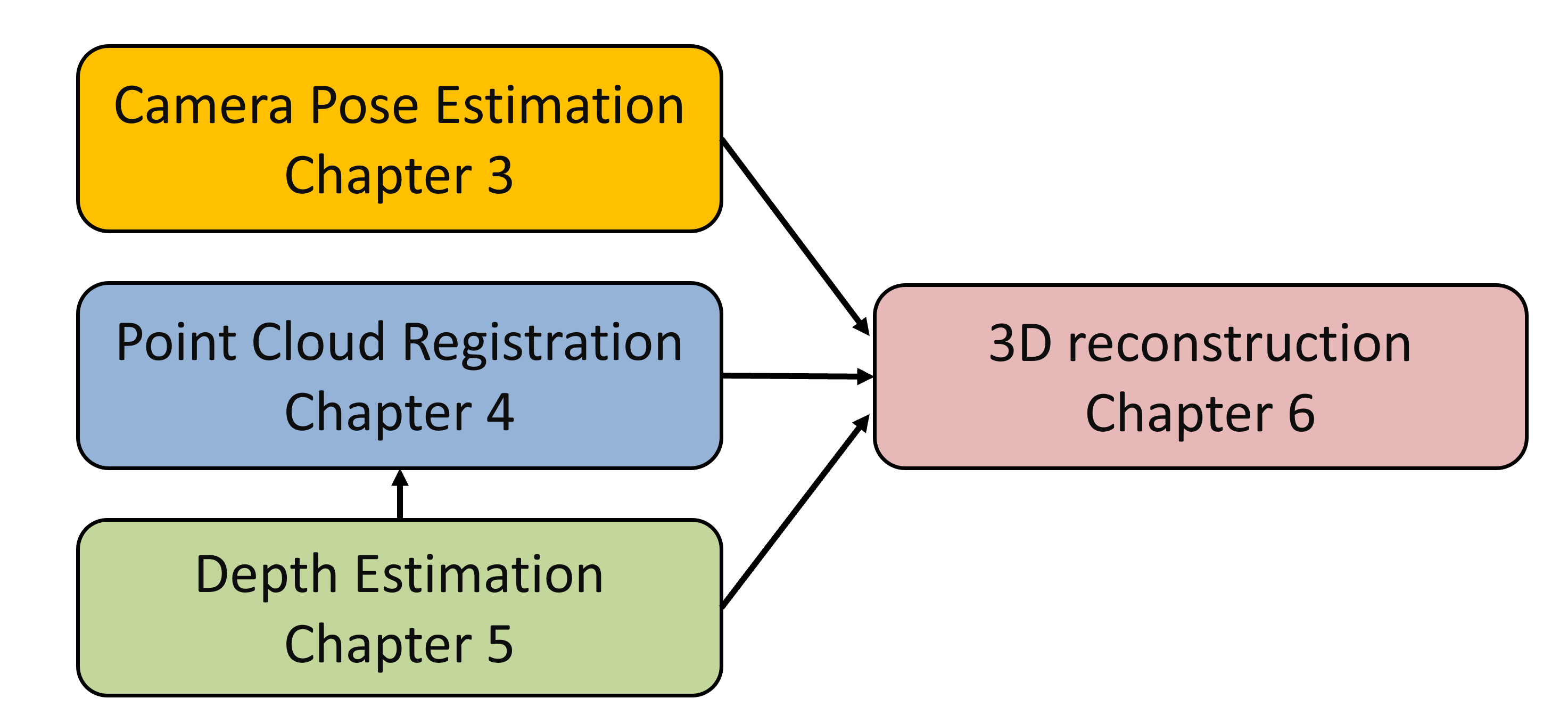}
    \caption{Chapter relationships and outline structures of dependencies, focusing on the chapters with novel contributions.}
    \label{fig:cp-structure}
\end{figure}








In the beginning, we lay out the overall research problem and define the objectives of each chapter, giving readers a big-picture view of the thesis and the limitations of existing methods. This is followed by a summary of our contributions to addressing these challenges. Chapter 2 provides a high-level overview of related work across the core problems, complementing the literature reviews in the individual chapters and helping readers contextualize the research questions within current trends.

The subsequent chapters are organized around three foundational subtasks—camera pose estimation, point cloud registration, and depth prediction—which represent the key components of a 3D reconstruction pipeline. Each subtask is studied in depth with tailored frameworks that integrate deep learning and geometric priors, yielding more accurate, robust, and efficient solutions. These contributions are not isolated but are designed to build toward the final system: the last chapter presents the complete 3D reconstruction framework, where the techniques developed in the earlier chapters are integrated and evaluated as part of a unified solution. This modular-to-system progression illustrates how targeted advances in the core subtasks collectively address the broader challenge of efficient and reliable 3D reconstruction.

From Chapters 3 to 6, each chapter builds upon the previous one, progressively increasing complexity from basic pose estimation to full 3D reconstruction. The step-by-step presentation demonstrates the strength of uniting geometric constraints with deep learning for 3D vision tasks. \revAouda{A typical 3D reconstruction pipeline consists of several modular stages: starting with camera pose estimation to align input images in a global map frame, followed by dense depth prediction and point cloud unprojection. The resulting point clouds, initially in local frames, are then registered to form a global point cloud. Given this modular structure, we decompose the reconstruction system into distinct components for focused analysis and discussion. This organization also highlights that each module can be optimized independently, rather than relying solely on end-to-end training. Modular optimization is often more practical from an engineering perspective and can effectively improve the overall quality of the final 3D reconstruction when integrated into the complete pipeline. Consequently, the overall structure of this thesis is organized according to this modular pipeline optimization strategy.} The structure outlined in Figure~\ref{fig:cp-structure} reflects this design: each chapter builds upon the previous one to form a coherent research progression, with inter-block dependencies marked by arrows to indicate how each component supports the next. Specifically, Chapter 3, which treats the subject of image pose estimation, establishes the relative spatial relationship among the frames and provides 3D reconstruction from camera poses. Chapter 4, in the case of point cloud registration, initializes 2D surfels from the 3D point cloud and predicts the relative transformation from source to target scans. Chapter 5 for depth estimation facilitates dense point cloud unprojection in the local camera frame. Finally, Chapter 6 for 3D reconstruction integrates image poses, registered point clouds to a globally consistent map, and predicted depth to generate high-fidelity 3D reconstructions. Every chapter serves as a building block upon which the later pipeline uses the earlier module to create a more complex framework for 3D vision tasks.

Chapter 7 moves from the theoretical research side of things to real-world applications, demonstrating how the methods I have developed contribute to digital cultural heritage preservation and robotic perception or other fields, and it evaluates the expected social benefits, economic values, and target business customers. This chapter explores practical deployment examples and evaluates the potential effectiveness of these methods when applied to real-world scenarios. Finally, Chapter 8 concludes the thesis by summarizing the key findings of the thesis and each project chapter, highlighting the main contributions again, and reflecting on the limitations of the approaches. It also outlines a plan for future research, suggesting potential improvements, like adapting these methods to other fields, or further enhancing model performance through novel contribution ideas.
\cleardoublepage


\chapter{Related Work}\label{ch:related-work}

Geometric Deep Learning (GDL) has emerged as a powerful technique for integrating deep learning with geometric representation or constraints, enabling more effective and efficient solutions for 3D vision tasks such as registration, depth estimation, and reconstruction. This chapter reviews prior work relevant to these topics, progressively linking traditional and modern approaches to provide a structured understanding of the 3D field.

We begin by discussing the foundations of Geometric Deep Learning, clarifying its evolution from classical deep learning applied to non-Euclidean domains to the broader incorporation of geometric constraints and representations in neural network models. Following this, we present the development of camera pose estimation, point cloud registration, depth prediction from images, and 3D reconstruction methods in the same order as the content chapters in the dissertation. This provides readers with a high-level understanding of the progression of each research problem, while each content chapter also includes a detailed review of related work and state-of-the-art methods for a more in-depth understanding of each research problem (Chapter 3 to Chapter 6). As 3D reconstruction is the most complex system, relying on the preceding three techniques, we place a lot of emphasis on its technique development timeline. For the 3D reconstruction problem, we examine traditional 3D reconstruction techniques, focusing on Structure-from-Motion (SfM) and Simultaneous Localization and Mapping (SLAM), which have been widely used in vision-based 3D applications for a long time. These methods, while robust, struggle with large-scale, unstructured, or ambiguous data, driving the transition to data-driven learning approaches.

We explore advancements in 3D Deep Learning-based reconstruction, categorizing models into explicit and implicit representations. Explicit models, such as point cloud and voxel-based networks, directly process 3D data but often face scalability challenges and artifacts from raw data. In contrast, implicit models represent 3D structures as continuous functions, enabling high-fidelity reconstruction—exemplified by neural radiance fields (NeRF) \cite{mildenhall2021nerf} and signed distance functions (SDF). More recently, Gaussian Splatting \cite{kerbl3Dgaussians} has significantly accelerated learning and rendering but at the cost of certain geometric representation continuity. This discussion follows the evolution from classical geometric methods to modern data-driven approaches, providing context for contributions of the thesis in leveraging Geometric Deep Learning for robust and efficient 3D vision applications.

\section{Geometric Deep Learning}

\begin{figure}[!thbp]   
\vspace{-0.6em}
  \includegraphics[width=\linewidth]{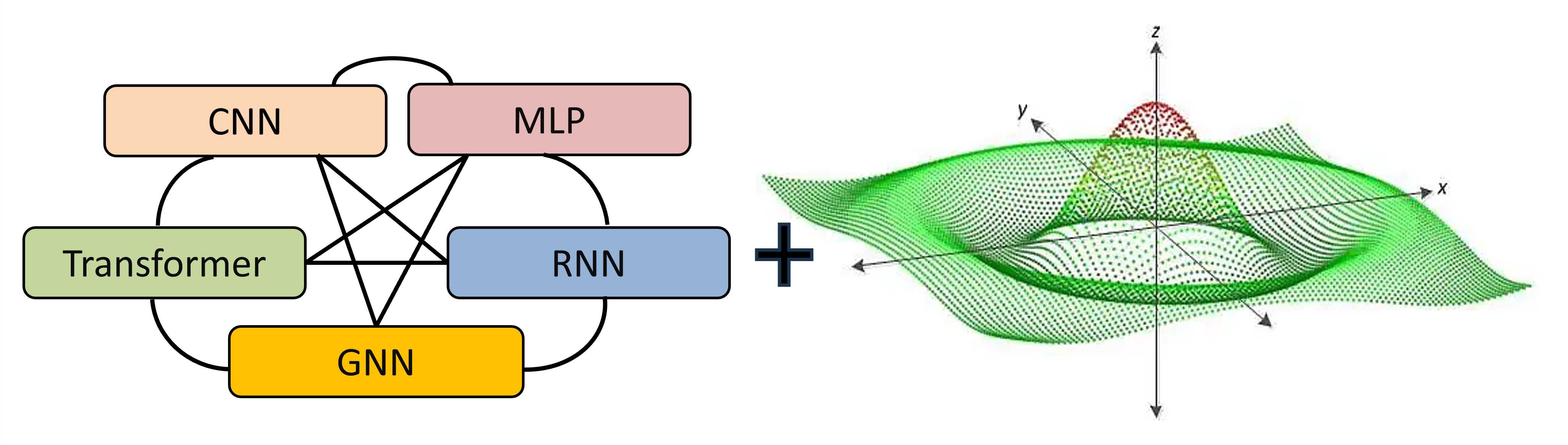}
    \caption{Geometric deep learning diagram combines various deep learning model structures and geometry constraints.}
    \label{fig:GDL-overview}
\vspace{-0.5em}
\end{figure}

Geometric Deep Learning (GDL), introduced by Bronstein \emph{et al.} \cite{bronstein2017geometric}, originally refers to deep learning in non-Euclidean spaces. As shown in Figure \ref{fig:GDL-overview}, Geometric Deep Learning emerges from combining deep learning models and non-Euclidean space distributions. Here, I adopt a broader definition, including integrating deep learning models with traditional geometric representations or constraints. This work will use Geometric Deep Learning in this generalized meaning. In essence, GDL leverages the prior of geometric primitive representations, such as points, voxels, normals, and mesh surfaces, to train neural networks to learn the general geometric representation. This approach is very promising and evolves very fast in the field of 3D vision, where high-dimensional feature representation of 3D data can be encoded into a general geometry before handling challenging problems. Unlike conventional deep learning methods that operate primarily on data commonly aligned in regular grids (e.g., images and sequences), GDL excels in handling more complex non-Euclidean structures like graphs and manifolds and facilitates the model to learn topological information underlying the input data, to handle the various down-stream tasks.


The conventional deep learning models are trained on Euclidean data structures, such as images and voxels, using convolutional neural networks (CNNs). However, many real-world problems involve data that exist on more complex structures, such as social media networks, biological or chemical molecular structures, and 3D point clouds of deformable objects. Geometric deep learning provides theoretical and practical frameworks for extending neural networks to these domains, enabling new applications and improving existing ones.

\section{Camera Pose Estimation}
\begin{figure}[!thbp]   
\vspace{-0.6em}
  \includegraphics[width=\linewidth]{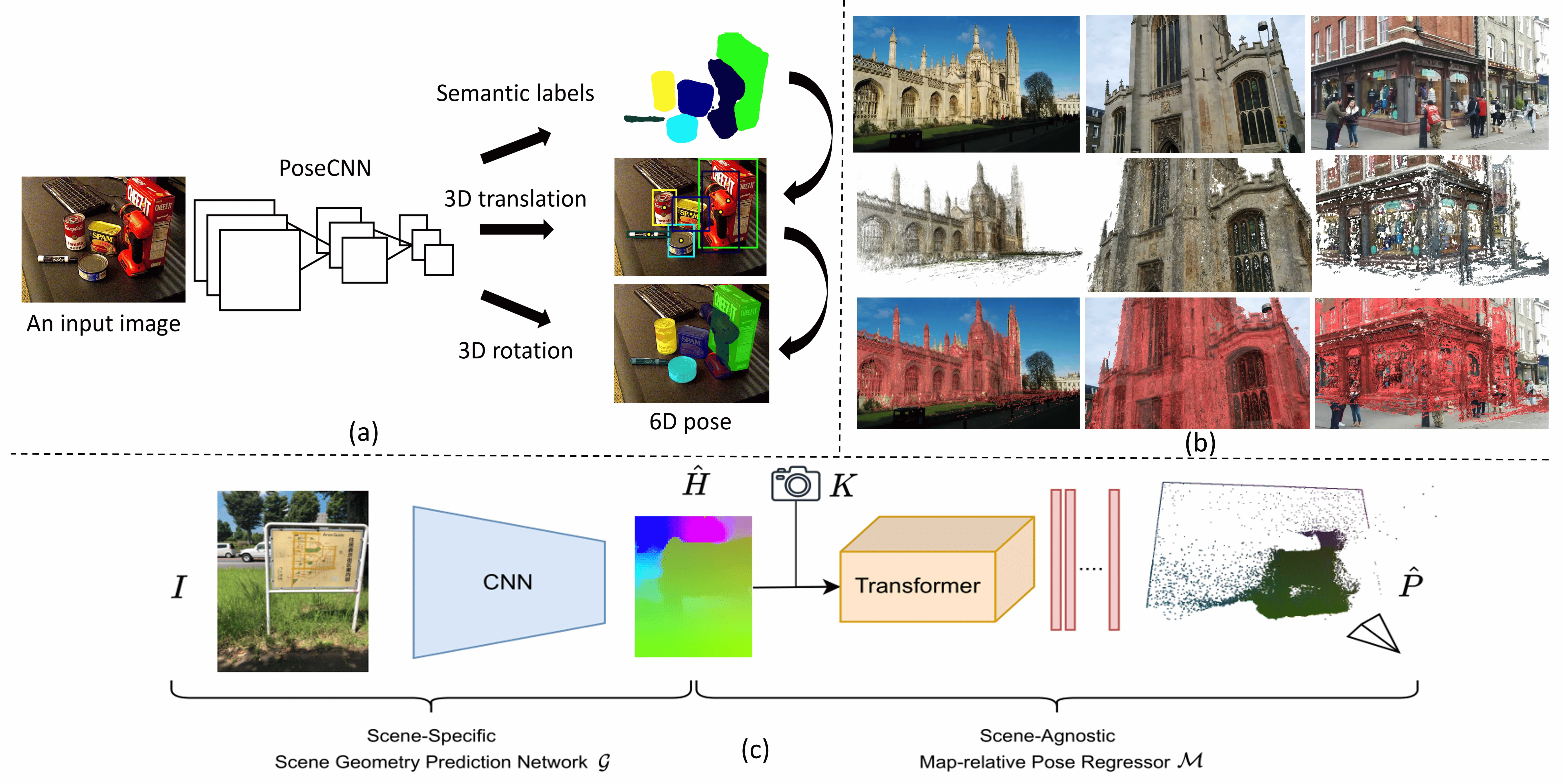}
    \caption{Deep learning methods for pose regression. (a) PoseNet: a CNN-based model for 6D camera relocalization \cite{kendall2015posenet}. (b) PoseCNN: a deep learning approach for object pose prediction \cite{xiang2017posecnn}. (c) Map-relative pose regression: a model estimating camera pose relative to a pre-built map representation by using Transformer attention of camera intrinsic and feature maps after CNN encoder \cite{chen2024map}.}
    \label{fig:pose-reg}
\vspace{-0.5em}
\end{figure}

Pose regression using deep learning models has advanced significantly, evolving from early CNN-based or ResNet-based architectures to transformer-based models. The objective is to directly predict the 6-DoF pose of an object while keeping the camera static \cite{xiang2017posecnn, mao2021tfpose, amini2022yolopose}, or to estimate the camera location while the scene remains static \cite{chen2024map, sattler2019understanding, brachmann2018learning, mahendran20173d}. As shown in Figure \ref{fig:pose-reg}, deep learning regression models are applied to both object pose estimation and visual localization. Various model structures have been explored for these tasks.

\noindent \textbf{CNN-based pose regression.} Early learning-based pose regression approaches leverage convolutional neural networks (CNNs) as encoders to extract spatial features from images and then regress object or camera poses using MLPs. Early works such as PoseNet \cite{kendall2015posenet} employed PoseCNN \cite{xiang2017posecnn} to predict the absolute pose of the camera, demonstrating robustness in challenging environments. CNNs can also be extended to 6D camera regression via 3D surface-to-image alignment regression. However, the inherent limitations of CNNs restrict feature representation accuracy due to their local receptive fields. For further analysis of the limitations of CNNs in pose regression, please refer to the work by Torsten \emph{et al.} \cite{sattler2019understanding}. Consequently, more powerful models have been explored to enhance pose estimation by expanding feature receptive fields and improving feature representation after encoding.

\noindent \textbf{Transformer-based approaches.} Recent advancements in vision transformers (ViTs) \cite{dosovitskiy2020image} have demonstrated strong performance in pose regression by capturing long-range spatial dependencies and global features within the encoder. For instance, an AutoEncoder-based transformer backbone can be used for pose regression \cite{shavit2022camera}, or a cascaded transformer can learn feature representations at multiple scales \cite{shavit2022camera}. Transformer-based encoders generalize well to pose regression across single or multiple scenes due to their superior feature representation capabilities \cite{shavit2024learning}. However, in highly challenging environments, such as natural landscapes where distinctive appearance features are difficult to identify, \emph{e.g.}, mountains, and meadows with repetitive patterns, most deep learning methods struggle to perform effectively.

\section{Point Cloud Registration}
\begin{figure}[!thbp]   
\vspace{-0.6em}
  \includegraphics[width=\linewidth]{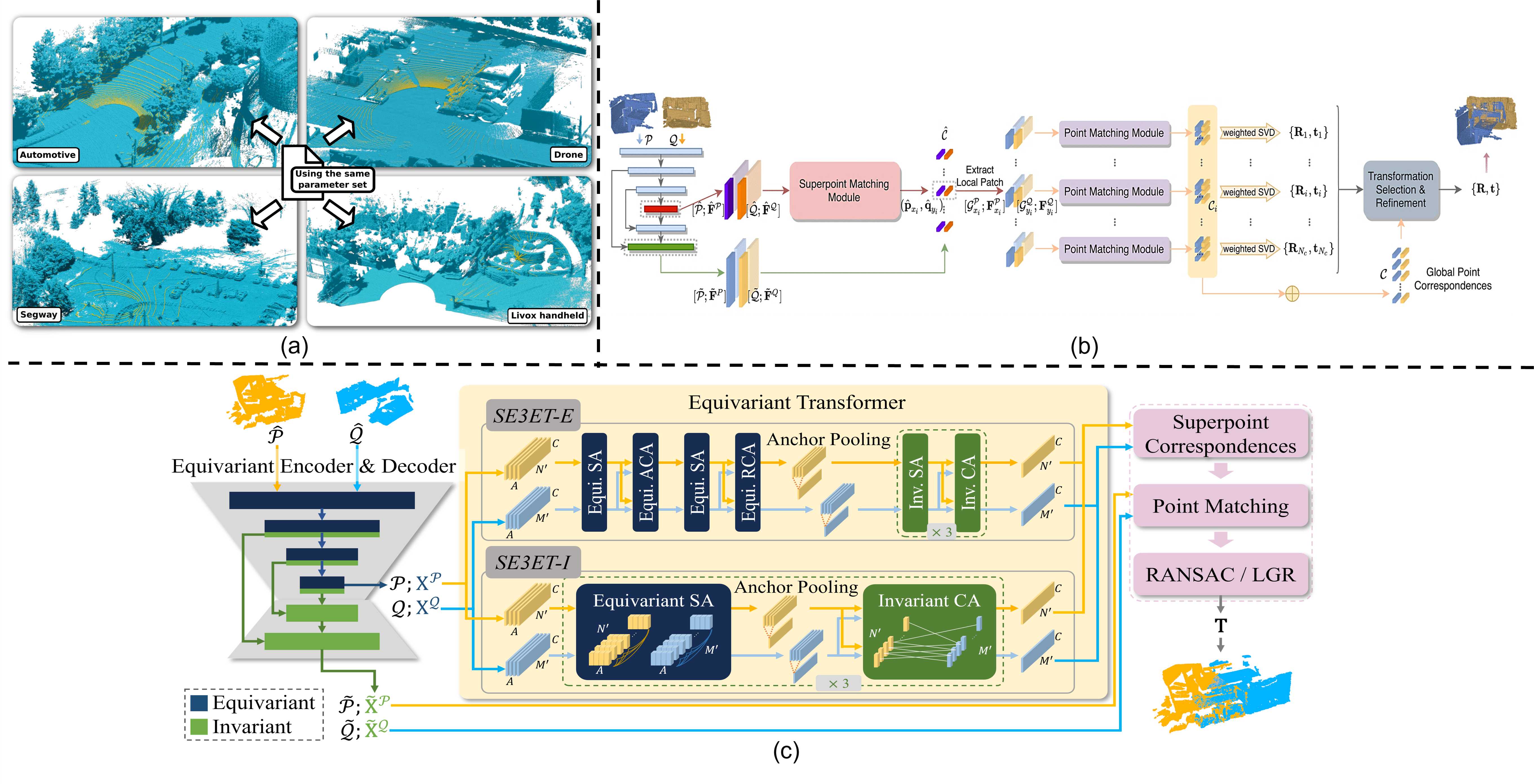}
    \caption{Deep learning methods for point cloud registration. (a) \cite{vizzo2023kiss}: an iterative approach for point cloud registration across diverse scenes, requiring only a few shared parameters. (b) Geometric Transformer with multi-level feature matching \cite{qin2022geometric}: a hierarchical method that aligns point clouds from coarse to fine to determine the final pose. (c) Equivariant Transformer SE3ET \cite{lin2024se3et}: a registration model incorporating equivariant cross-attention and invariant self-attention for robust alignment.}
    \label{fig:pt-reg}
\vspace{-0.5em}
\end{figure}

Point cloud registration is a fundamental technique in 3D vision, aiming to align two or more point cloud scans within a global mapping frame. Traditional point cloud registration methods rely on iterative optimization to identify sufficient correspondence inliers between input points and subsequently estimate the transformation pose. Deep learning-based approaches, on the other hand, are data-driven and learn to extract geometric features from input points, which can be leveraged for correspondence matching and pose regression. Recent advancements in point cloud registration have focused on leveraging equivariant feature learning to enhance learning efficiency and generalization. Representative state-of-the-art methods are illustrated in Figure \ref{fig:pt-reg}, including the traditional ICP-based approach, multi-level feature correspondence establishment via learned superpoints, and the equivariant transformer with specialized attention mechanisms designed to learn both equivariant and invariant features.

\noindent \textbf{Iterative closest point (ICP) and KISS-ICP.} Traditional registration methods, such as Iterative Closest Point (ICP) and its many variants like Generalized-ICP \cite{segal2009generalized}, remain widely used due to their simplicity and efficiency, requiring no parameter training. ICP iteratively refines the transformation between two point clouds by minimizing the point-wise distances of matched point pairs, given an initial transformation. KISS-ICP \cite{vizzo2023kiss} improves registration robustness across diverse scenes by enforcing stability constraints in optimization and using a single system configuration. Despite their efficiency, these methods struggle with noisy data, high outlier ratios, symmetric structures in point clouds, and sparsity, often requiring a good initial transformation guess, which limits their utility in complex scenarios.

In contrast, geometric feature learning-based methods for point cloud registration offer improved generalization capabilities.

\noindent \textbf{Registration through geometric feature learning.}
Deep learning methods \cite{wang2019deep, gojcic2020learning} have significantly advanced point cloud registration by learning high-dimensional geometric features directly from raw 3D point coordinates. Early approaches, such as PointNet \cite{qi2017pointnet} and its variants, have been applied to registration tasks by extracting global and local point features, improving correspondence matching by searching for matches in high-dimensional feature space to mitigate ambiguity, as demonstrated in PCRNet \cite{sarode2019pcrnet}. Subsequent works further enhance CNN-based geometric feature descriptor extraction, such as FCGF \cite{choy2019fully}, which strengthens learned feature representations for robust registration through a convolutional U-Net combined with hardest-contrastive and hardest-triplet losses.

Additionally, methods like Maximal Cliques \cite{zhang20233d} improve correspondence establishment by identifying maximal cliques in the point set. Jaesim \emph{et al.} \cite{park2017colored} introduce color information into point cloud registration to enhance robustness, while NICP \cite{serafin2015nicp} incorporates normal information in addition to point coordinates for improved alignment. Geometric Transformer \cite{qin2022geometric} further utilizes invariant self-attention and equivariant cross-attention to learn geometric feature tokens and match point tokens, enabling robust registration even in small overlap settings.

These methods demonstrate strong performance in handling noisy points and partial data scans; however, they often rely on extensive data augmentation to generalize across different scenes, making the learning process inefficient.

\noindent \textbf{Equivariant point cloud registration.}
Recent works leverage equivariant feature learning to enforce transformation consistency directly within network models, ensuring that the learned features transform correspondingly with the input points. SE(3)-equivariant models, such as E2PN \cite{zhu2023e2pn}, Equi-GSPR \cite{kang2024equi} and Equivariant Transformers SE3ET \cite{lin2024se3et}, guarantee that learned features remain invariant under rigid transformations while maintaining equivariant edge features, leading to more accurate and robust registration. By incorporating specifically designed equivariant convolution kernels or self-attention mechanisms, these models achieve superior alignment accuracy with fewer training samples and exhibit strong generalization even on unseen data.

The evolution from iterative optimization to deep learning and equivariant models highlights the growing research interest in improving accuracy, generalization, robustness, and efficiency in point cloud registration.

\section{Depth Estimation from Images}
Depth prediction from images has progressed rapidly with the emergence of image foundation models \cite{fu2024geowizard, ke2024repurposing}. Deep learning models have advanced from CNN-based networks to state-of-the-art diffusion models. These approaches estimate scene depth from a single image by leveraging relative feature scale distribution or stereo input priors constructed from a virtual camera frame, facilitating applications in 3D reconstruction and robotic mapping.

\begin{figure}[!thbp]   
\vspace{-0.6em}
  \includegraphics[width=0.98\linewidth]{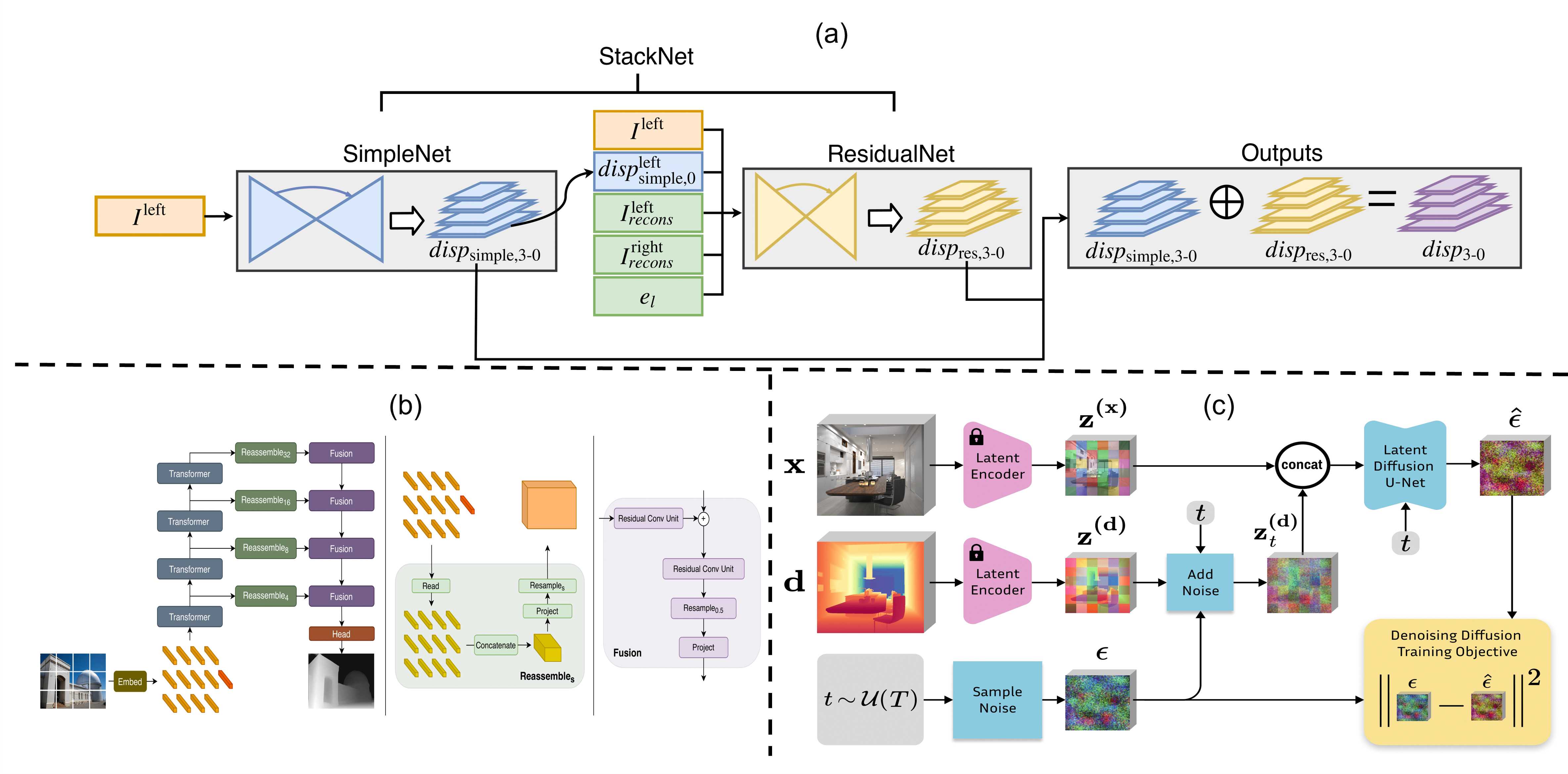}
    \caption{Deep learning architectures for depth prediction. (a) Deep Virtual Stereo Odometry: a method leveraging stereo displacements to generate a virtual stereo camera from a single image for depth estimation \cite{yang2018deep}. (b) Vision Transformer: a transformer-based model for dense depth prediction \cite{ranftl2021vision}. (c) Marigold: a diffusion-based approach for depth prediction using a generative model \cite{ke2024repurposing}.}
    \label{fig:depth-pred}
\vspace{-0.5em}
\end{figure}

\noindent \textbf{CNN-based depth prediction.}
Early deep learning methods for depth estimation leveraged convolutional neural networks (CNNs) to extract multi-scale features and infer depth maps from a single image \cite{laina2016deeper, ma2018sparse, garg2016unsupervised} or video \cite{feng2019unsupervised}. ResNet-based models, such as the pioneering work MonoDepth2 \cite{monodepth2}, demonstrated that deep networks could learn depth directly from monocular images. Later research, including monocular depth estimation via unsupervised learning \cite{casser2019depth, garg2016unsupervised}, improved prediction performance by incorporating geometric or structural constraints. Despite their success, CNN-based models struggled with predicting fine-grained details and generalizing across diverse environments.

\noindent \textbf{Virtual stereo with left-to-right displacement.}
To mitigate the scale ambiguity in monocular depth estimation, researchers introduced virtual stereo methods that synthesize a right viewpoint by learning left-to-right image transformations through network models like DVSO \cite{yang2018deep}. These methods, often based on disparity estimation, leverage epipolar constraints between left and right views to refine depth predictions. Works like StereoNet \cite{khamis2018stereonet} improve generalization by utilizing edges as an objective during training refinement. While effective, these models remain sensitive to view occlusions and textureless regions, limiting their robustness in cluttered scenes.

\noindent \textbf{Vision Transformers for depth estimation.}
Recent advancements in vision transformers (ViTs) have significantly improved depth prediction by leveraging Transformer attention mechanisms to capture global features, as seen in ViT-based dense depth prediction \cite{ranftl2021vision}. Transformer models outperform traditional CNNs in depth estimation accuracy and robustness while preserving fine-grained details in depth maps. However, their computational cost remains high, posing challenges for deployment on resource-constrained platforms like mobile hardware.

\noindent \textbf{Diffusion models for depth prediction.}
Recent breakthroughs, such as GeoWizard \cite{fu2024geowizard}, LOTUS \cite{he2024lotus}, and Repurposing Diffusion \cite{ke2024repurposing}, leverage diffusion-based foundation models, which iteratively refine depth predictions through a generative denoising process during training. Inspired by the denoising diffusion process, these models predict depth by progressively removing pixel-wise noise. Unlike deterministic depth prediction approaches, diffusion-based depth estimation iteratively captures high-frequency details, better reflecting the feature learning process from low to high frequencies. This results in sharper and more accurate depth maps. Recent advancements in diffusion models have enabled the generation of multi-view images that maintain consistency for 3D representation \cite{kang2025multi}.

The progression from CNNs to diffusion models highlights continuous improvements in depth prediction accuracy and robustness by leveraging the power of generative models. This has enabled the development of more generalizable depth estimation models for practical applications in robotics, AR/VR, and autonomous systems when dealing with unseen data.

\section{3D Reconstruction}
\subsection{Traditional 3D Reconstruction Methods}
Geometric deep learning often depends on large-scale 3D data obtained either through LiDAR scanning \cite{kang20183d}, which is efficient but costly and less scalable, or multi-view image reconstruction, which offers greater scalability. This dependence highlights the importance of understanding traditional 3D vision techniques for reconstruction, particularly Structure from Motion (SfM) and Simultaneous Localization and Mapping (SLAM). These methods have long served as the processing tool for pose estimation and 3D mapping in computer vision, and they offer robust frameworks for reconstructing 3D scenes from 2D images captured from multiple viewpoints. Even today, these traditional methods remain the foundational tool for generating camera view poses in NeRF \cite{mildenhall2021nerf} and sparse 3D point cloud for Gaussian Splatting \cite{kerbl3Dgaussians}.

\subsubsection{Structure from Motion (SfM)}
Structure from Motion (SfM) (\cite{ozyecsil2017survey, schonberger2016structure}) is a photogrammetric technique that estimates 3D structures from 2D image sequences captured at various viewpoints. Such a 3D reconstruction pipeline recovers the 3D structure of the scene visible in the multi-view images. Taking the common 3D reconstruction tool COLMAP (\cite{schoenberger2016sfm, schoenberger2016mvs}) as an example, the reconstruction process is composed of the following steps:

\begin{itemize}
    \item \textbf{Feature detection and extraction.} Distinctive feature points are first detected in each image, followed by the extraction of their feature descriptors to enable reliable matching across views.
    
    \item \textbf{Correspondence establishment.} Feature matching is performed by finding correspondences between detected features across multi-view images. A geometric verification step filters out incorrect matches. The verification ensures only geometrically consistent correspondences are retained for accurate relative pose calculation.
    
    \item \textbf{Camera pose estimation and 3D point reconstruction.} With the established correspondences, the camera poses are estimated using a linear SVD transformation, and 3D points are reconstructed by triangulating matched 2D feature points. This is achieved through incremental structure-from-motion, where new 3D points are continuously added while refining camera intrinsic and extrinsic parameters.
    
    \item \textbf{Global bundle adjustment.} To enhance pose and 3D map accuracy, a global bundle optimization is performed over all camera poses and 3D points. This optimization process minimizes the projective function of reprojection errors. The resulting sparse 3D point cloud can be further densified through post-processing techniques, to create a point cloud with uniform density.
\end{itemize}

\begin{figure}[!thbp]   
\vspace{-0.2cm}
  \includegraphics[width=0.98\linewidth]{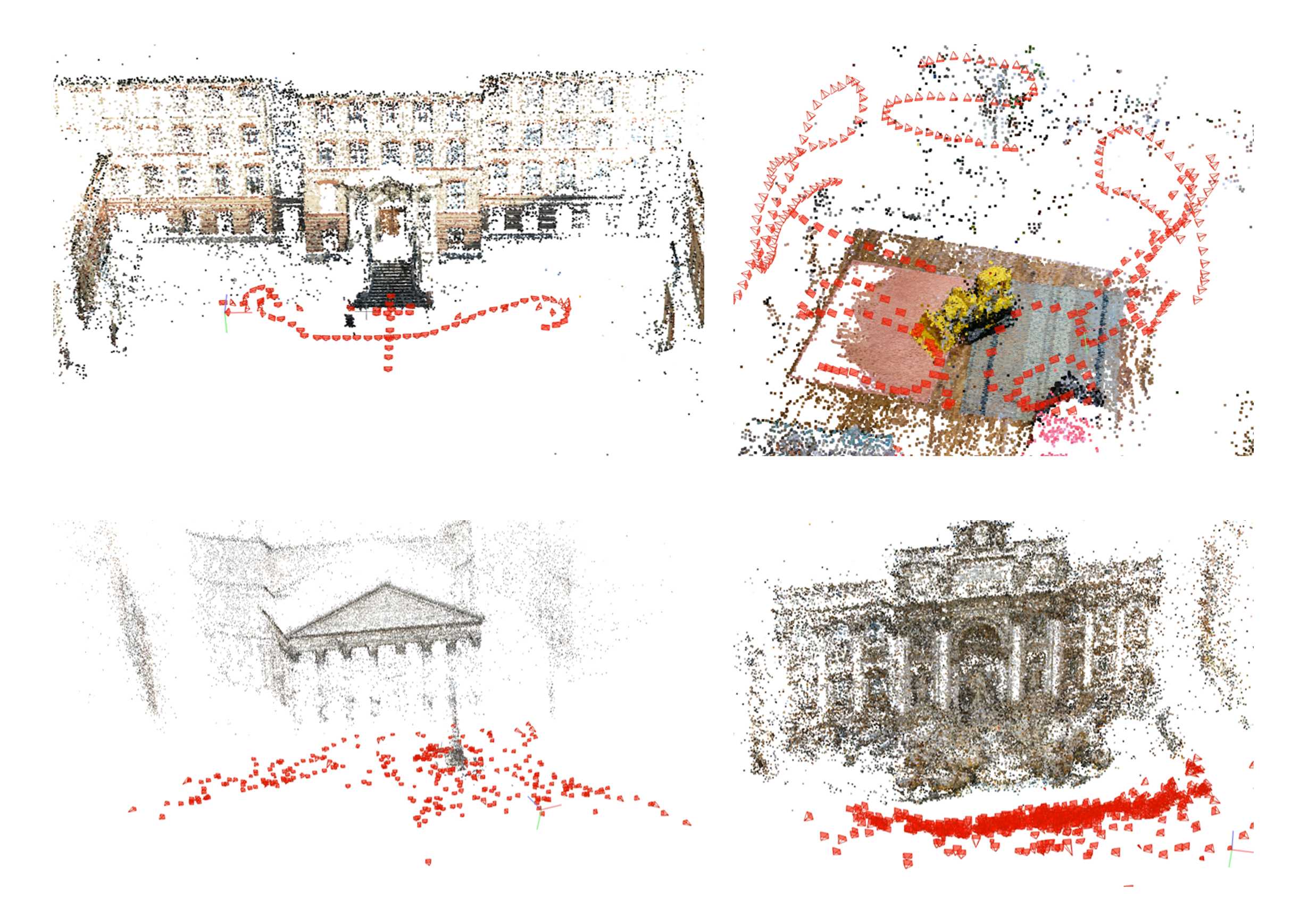}
    \caption{Illustration of structure from motion by multi-view images, image source from GLOMAP \cite{pan2024global}.}
    \label{fig:SfM}
\vspace{-0.5cm}
\end{figure}

Figure \ref{fig:SfM} depicts a fundamental Structure from the Motion approach, inferring 3D point locations through the triangulation of multi-view images. The framework can reconstruct multi-scale 3D assets, ranging from tabletop objects to large-scale buildings. 

\subsubsection{Simultaneous Localization and Mapping (SLAM)}
Simultaneous Localization and Mapping (SLAM) (\cite{thrun2008simultaneous, fuentes2015visual, kang2023integrated, xueyang2023simultaneous}) is a method employed by robotics and computer vision communities for building the map of an unknown environment incrementally, while at the same time localizing the ego-agent in the constructed map. SLAM algorithms find broad use in autonomous vehicles, drones, and augmented reality systems.

\begin{figure}[!ht]
    \vspace{-0.2cm}
    \centering
    \subfloat[ORB vision SLAM demonstration. Source image from work by Mur-Artal \emph{et al.} \cite{mur2015orb}. \label{fig:vSLAM}]{
    \includegraphics[width=0.92\textwidth]{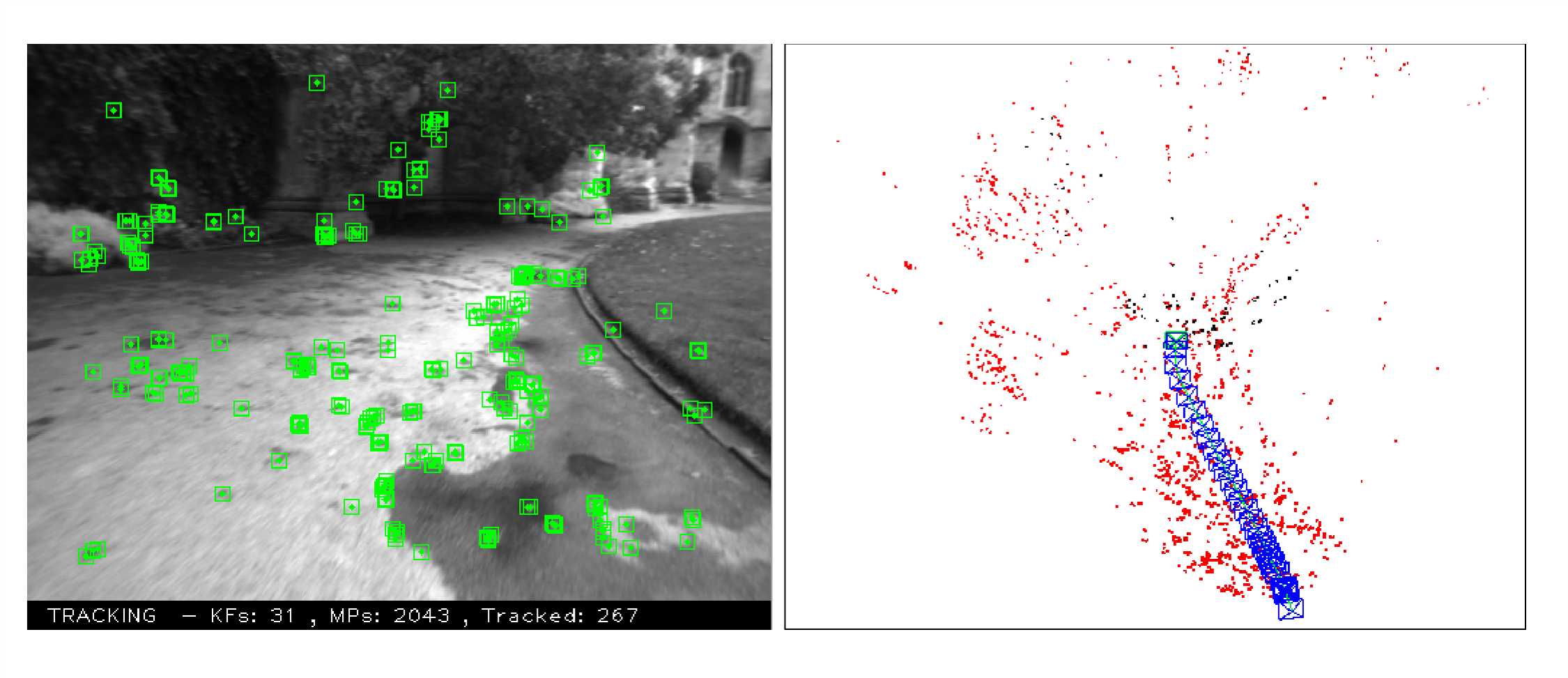}
    }
    \vspace{-0.0cm}
    \subfloat[Lidar-based Cartographer SLAM by Google. Source image from work by Hess \emph{et al.} \cite{hess2016real}. \label{fig:LidarSLAM}]{
    \includegraphics[width=0.92\textwidth]{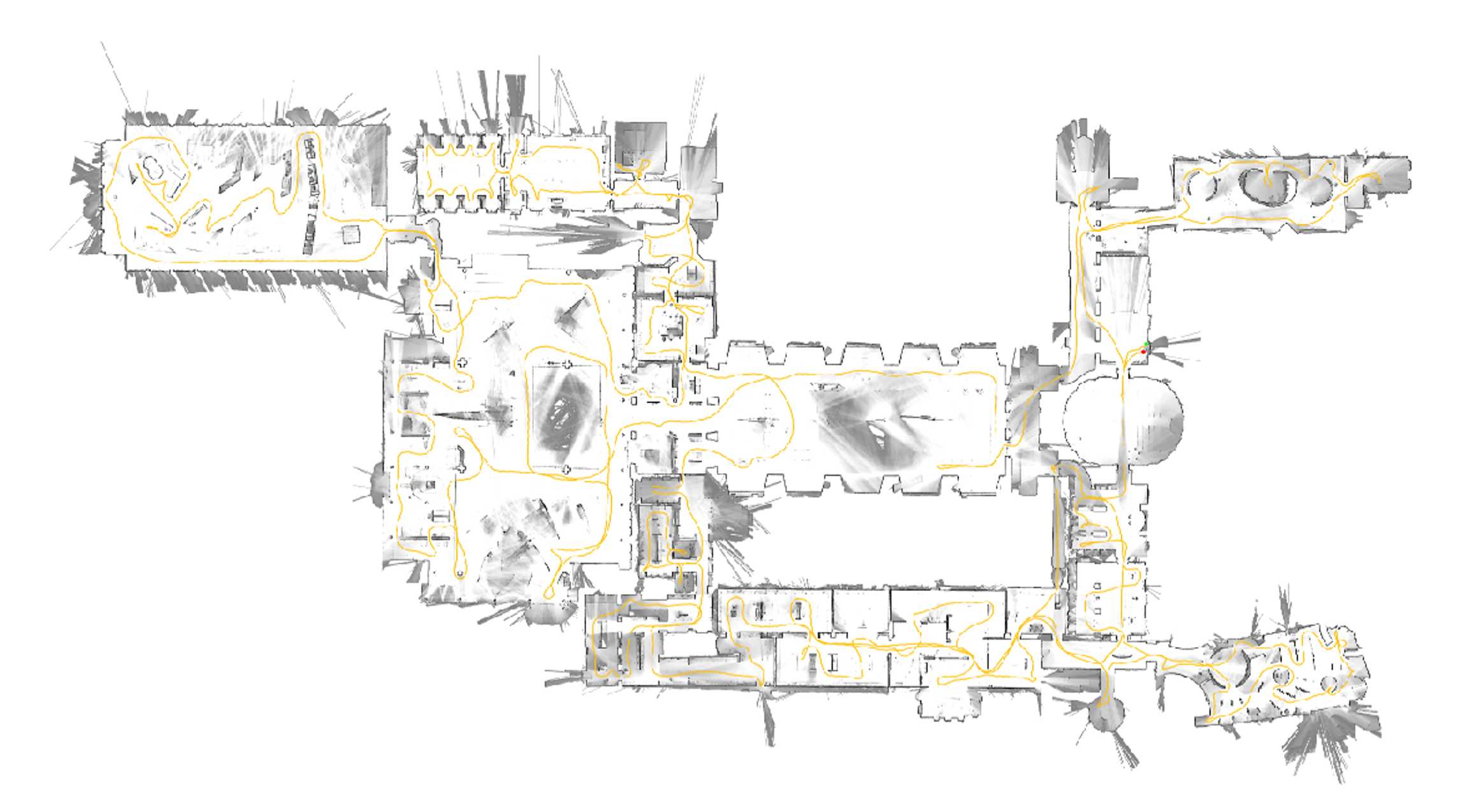}
    }
    \vspace{-0.2cm}
    \caption{Common SLAM framework: (a) visual SLAM, (b) Lidar-based SLAM.}
    \label{fig:SLAM}
    \vspace{-0.4cm}
\end{figure}

The key components of SLAM include a front-end and a back-end:
\begin{itemize}
    \item \textbf{Front-end.} Collecting raw data from various sensors, such as cameras, LiDAR, and IMU, then feature extraction and matching are performed over the raw sensor data to establish correspondences. Consequently, the position and orientation of the ego-agent are estimated based on the feature correspondences of frames.
    \item \textbf{Back-end.} A mapping module updates the map by fusing new information from the selected feature points. Once the agent has returned to a previously visited location, the triggered loop closure will correct the map misalignment accordingly by global map optimization.
\end{itemize}
In the context of SLAM classification, the classification relies mostly on the type of front-end sensor employed: (1) Vision SLAM, which utilizes camera images to aid in feature detection, motion estimation, and map construction; (2) LiDAR SLAM, which employs LiDAR sensors to scan 3D point clouds for scan matching, odometry estimation, and mapping; and (3) Sensor fusion-based SLAM, which fuses measurements from various sensors (\emph{e.g.}, cameras, LiDAR, IMU, GPS, and wheel odometry) for improving accuracy and robustness in front-end and back-end procedures. The two most representative SLAM methods are illustrated in Figure \ref{fig:SLAM}. Visual SLAM tracks image features (green patch in the left part of \ref{fig:vSLAM}) to reconstruct sparse 3D points (right part of \ref{fig:vSLAM}), while LiDAR-based SLAM generates a global grid map along the motion trajectory (Figure \ref{fig:LidarSLAM}).

\subsection{3D Deep Learning}
While traditional methods like Structure-from-Motion (SfM) (\cite{schonberger2016structure}) and Simultaneous Localization and Mapping (SLAM) have given rise to opportunities for 3D reconstruction and mapping, they are subject to severe limitations. These methods fall short in terms of computational feature matching, thus making them ineffective in processing dense 3D data. They are also prone to robustness breakdown under high outlier ratios or feature ambiguities within the input data.

In contrast, 3D deep learning models provide significant advancements by employing a large number of parameters to learn strong and precise geometric representations in parallel computation. These representations are learned through end-to-end loss backpropagation, offering general and strong geometric priors for many 3D tasks.

To provide a better understanding, let us start with a quick review of the recent 3D deep learning architectures.

\subsubsection{Explicit Model for 3D Geometry Learning}
\label{sec:explicit-model} Explicit 3D deep learning models such as PointNet \cite{qi2017pointnet} and VoxelNet \cite{maturana2015voxnet} directly process 3D data such as point clouds and voxel grids for object detection, semantic segmentation, detection \cite{kang2019robust}, and 3D reconstruction. Such models learn geometric representations automatically through neural networks under the guidance of explicit supervision signals. They have the limitation of high computational complexity, high memory, and low scalability for processing dense or large data. The effectiveness of their performance is prone to degradation due to sparse or noisy inputs and is sensitive to biases in data representation. It is necessary to overcome these difficulties for the improvement of both the efficiency and robustness of 3D deep learning models in practical applications.

\begin{figure}[!ht]
    \vspace{-0.2cm}
    \centering
    \subfloat[3D recurrent neural network (3D-R2N2) for 3D 
 shape reconstruction from sequence images. Source from work by \cite{choy20163d}. \label{fig:lstm3d}]{
    \includegraphics[width=\linewidth]{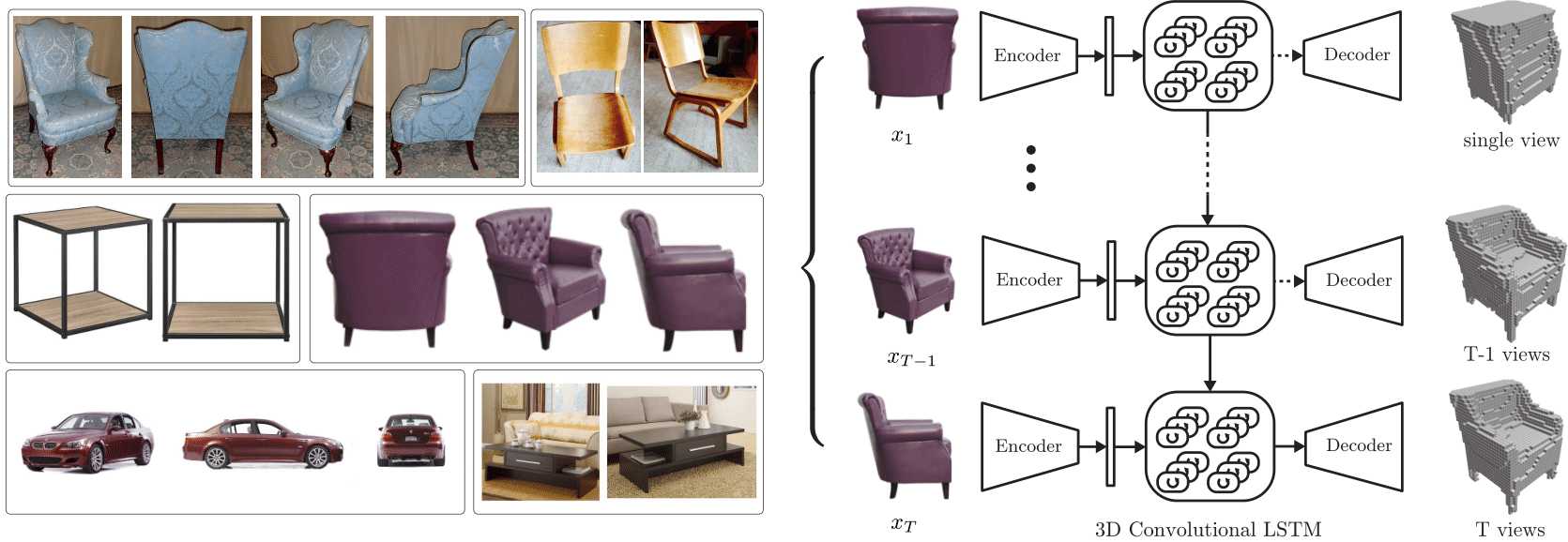}
    } \\
    \subfloat[3D Unet convolution for reconstruction. \label{3Dunet}]{
    \includegraphics[width=\linewidth]{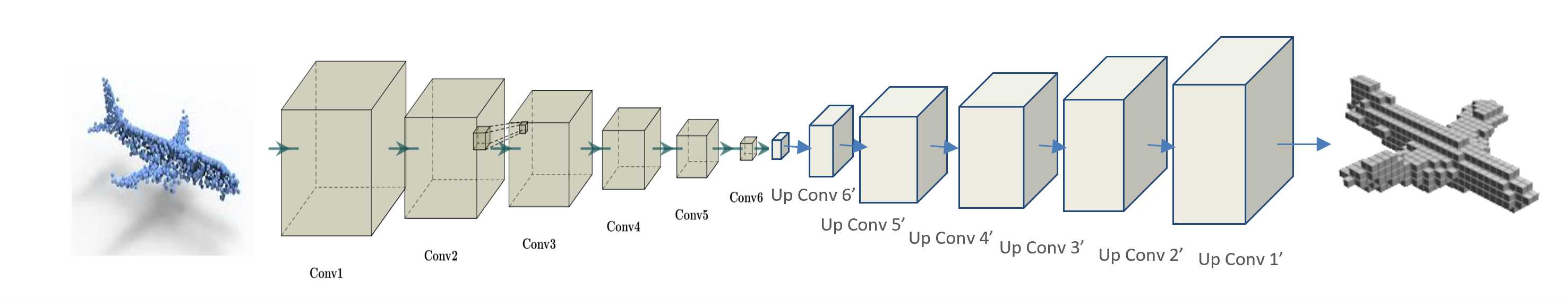}
    }
    \caption{3D reconstruction by using a 3D recurrent method (a), and explicit 3D U-Net model (b).}
    \label{fig:explicit}
\end{figure}

Explicit deep learning models for 3D reconstruction are illustrated in Figure \ref{fig:explicit}. The top row shows a classical recurrent model by \cite{choy20163d} that encodes multi-view images using LSTMs in the latent feature space. Another common approach directly takes in the point cloud to predict a shape occupancy map, as shown in Figure \ref{3Dunet}.

\subsubsection{Implicit Model for 3D Geometry Learning}
\label{sec:implicit-model}
Implicit deep learning models, such as Neural Radiance Fields (NeRF) by \cite{mildenhall2020nerf} and DeepSDF by \cite{park2019deepsdf}, represent 3D geometry and appearance through neural networks—typically Multi-Layer Perceptrons (MLPs)—instead of explicit data structures like point clouds or voxel grids. These models learn continuous volumetric representations from sparse, noisy input data by capturing the underlying geometry distribution. This enables them to generate high-fidelity 3D reconstructions and novel view synthesis by optimizing over a latent space encoding geometric information.

A key advantage of implicit models is their ability to handle complex topologies and generate continuous geometric details, and they overcome the resolution limitations of grid-based input representations. However, they also pose challenges, such as high computational demands during training with a large volume of pixel rays to query the radiance field, and the generalization ability of implicit models on diverse datasets is quite weak. With increasing computational power, these challenges are gradually being ignored through scaling law. Overall, implicit deep learning models still have their merits in 3D reconstruction and rendering while maintaining a compact model size. Next, we discuss recent developments in implicit 3D modeling.

\noindent\textbf{Implicit SDF.} 
Implicit deep learning models have changed 3D shape representation by modeling shapes as continuous functions. Unlike explicit models that rely on discrete representations such as point clouds, meshes, or voxel grids, this implicit continuous representation, learns 3D features in MLP-based weights, enabling smooth surface reconstructions. I can be scaled to complex environments with a small memory footprint. Implicit models also offer robust invariance to incomplete or noisy data, making them well-suited for robust applications in 3D shape representation and reconstruction.

A pivotal shape reconstruction work in this domain is DeepSDF by \cite{park2019deepsdf}, which introduced Signed Distance Functions (SDFs) for high-fidelity 3D shape reconstruction. This method represents shape by an implicit SdF model, allowing for smooth feature learning and robust reconstruction given partial inputs.

Building on this foundation, Convolutional Occupancy Networks (\cite{peng2020convolutional}) integrated convolutional neural networks to learn occupancy probabilities. Such an occupancy prediction-based model improves the details and accuracy of 3D geometry representation by using multi-scale triplane or volumetric features. This approach efficiently processes volumetric data, enabling high-fidelity reconstructions of complex scenes. This is achieved through grid-based latent feature interpolation in volumes or triplanes.

Next, IF-Net (\cite{chibane20ifnet}) introduced implicit function constraints in feature space, which jointly performs 3D shape reconstruction and completion. IF-Net enhanced the ability to reconstruct complex and incomplete shapes, such as partial human body scans, offering a more flexible 3D processing model.

More recently, Neural Shape Deformation Priors (\cite{tang2022neural}) incorporated learned deformable shape priors into neural networks.  Such deformable prior-guided implicit model improves the accuracy of both 3D reconstruction and shape manipulation. This method leverages a hierarchical structure, including global shape templates and local deformations encoded by different MLPs. Such a design can capture inherent shape distributions and dynamic variations more effectively.

In a nutshell, these advancements highlight the rapid evolution of implicit 3D modeling techniques. From the shape DeepSDF to the advanced Neural Shape Deformation Priors, each pushing the limits of 3D geometry reconstruction with increasing complexity and showcasing the rising influence of implicit deep learning in computer vision and graphics.

\begin{figure}[!ht]
    \vspace{-0.0cm}
    \centering
    {
    \includegraphics[trim=10 10 10 10, clip, width=\linewidth]{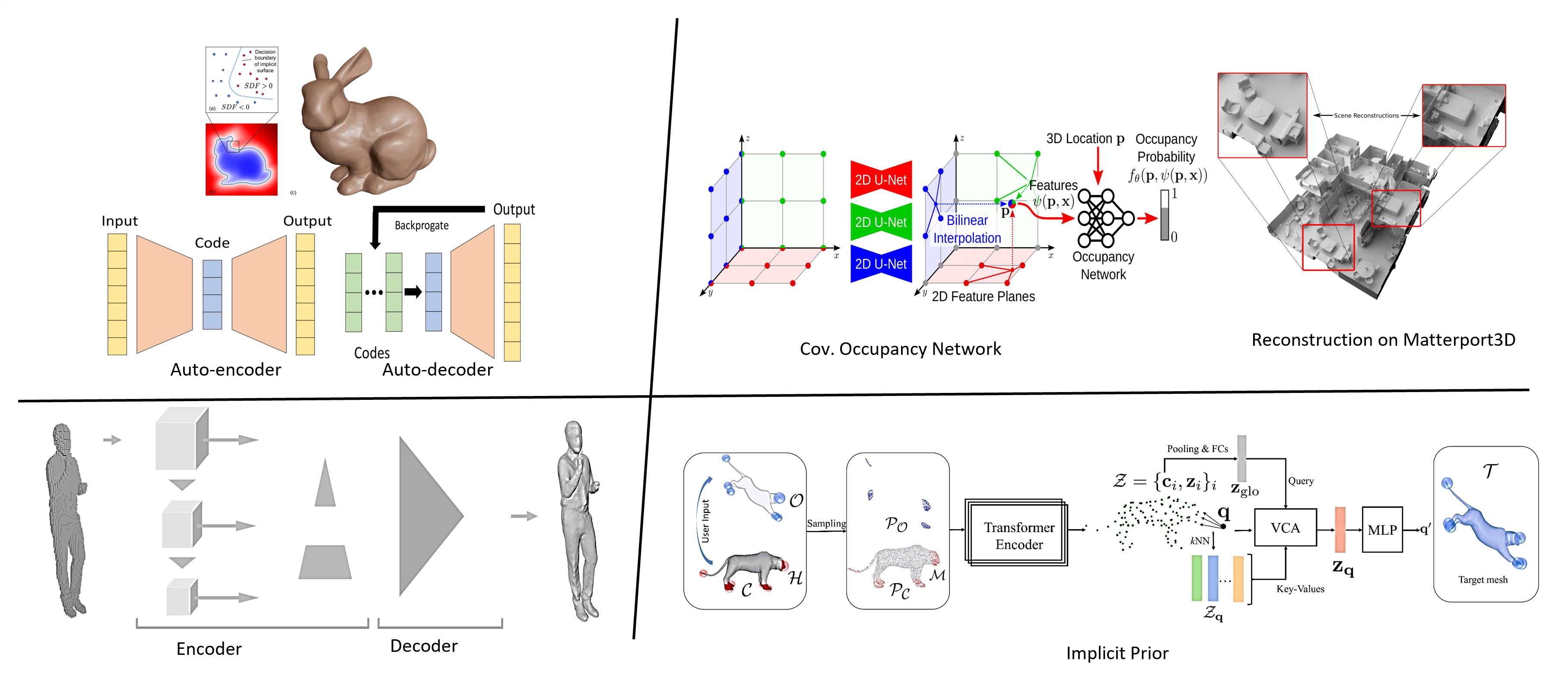}
    }
    \vspace{-0.2cm}
    \caption{Implicit SDF for a wide range of reconstruction problems, from shape to room scan, human body, and deformable animation. Source images rearranged from \cite{park2019deepsdf, peng2020convolutional, chibane20ifnet, tang2022neural}.}
    \label{fig:implicit}
    \vspace{-0.4cm}
\end{figure}

\begin{figure}[!ht]
    \vspace{-0.2cm}
    \centering
    {
    \includegraphics[width=\linewidth]{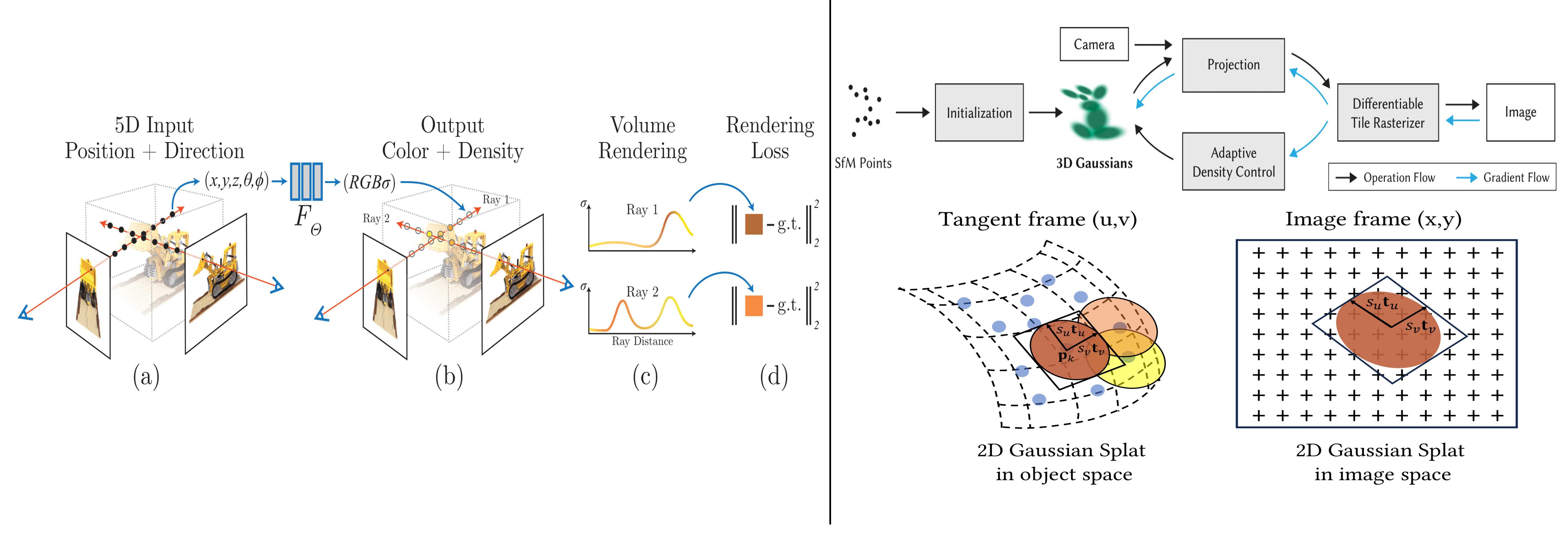}
    }
    \vspace{-0.2cm}
    \caption{Neural Radiance Field (at the left) alongside its extensions: 3D and 2D Gaussian Splatting (at the right). Source images from \cite{mildenhall2020nerf, kerbl3Dgaussians, huang20242d}.}
    \label{fig:NeRF}
    \vspace{-0.2cm}
\end{figure}

\noindent\textbf{Implicit neural radiance field.}
The transition from Neural Radiance Fields (NeRF) by \cite{mildenhall2020nerf} to 3D Gaussian Splatting by \cite{kerbl3Dgaussians} and the latest 2D Gaussian Splatting by \cite{huang20242d} marks a significant evolution in 3D scene representation and rendering, and this development is demonstrated in Figure \ref{fig:NeRF}.

NeRF (\cite{mildenhall2020nerf}) introduced a novel approach to three-dimensional scene representation by using implicit neural networks to represent volumetric geometry and color appearance. This approach enables photorealistic rendering of new viewpoints from a small set of input images by mapping three-dimensional spatial positions and viewing directions to their corresponding color and density values. Unlike implicit Signed Distance Function (SDF) models that are limited to geometry representation, NeRF combines geometric and appearance information in real-time, hence producing high-quality, continuous three-dimensional representations without the need for explicit structures like meshes or point clouds. Extensions of NeRF \cite{liu2022neural, mildenhall2022nerf} have improved the rendering of occluded regions and performance under varying lighting conditions, hence expanding its use in virtual reality, augmented reality, and 3D content creation. The compact scene representation provided by NeRF saves memory, hence improving computational efficiency and making it possible to deploy on mobile and edge devices.

Implicit NeRF suffers from complex and time-consuming training, and it often requires extensive GPU resources for large-scale scene reconstruction. To overcome this computing limitation, 3D Gaussian Splatting by \cite{kerbl3Dgaussians} is proposed to enhance real-time rendering capabilities without sacrificing rendering quality. This approach represents scenes as a collection of 3D Gaussian blobs, which are optimized efficiently and explicitly through a Gaussian feature representation. A key advantage of 3D Gaussian Splatting is its ability to achieve real-time performance through a specialized CUDA kernel implementation, so it is suitable for applications such as localization and mapping. However, Gaussian Splatting represents scenes with discrete Gaussian primitives, enabling fast rendering but lacking the continuity of implicit SDF models. Unlike SDFs, which enforce smooth and coherent surface geometry, Gaussian Splatting may produce disconnected or floating artifacts. Gaussian Splatting struggles with fine-grained topology reconstruction, making it less ideal for precise 3D reconstruction.

The latest advancement, 2D Gaussian Splatting by \cite{huang20242d}, further refines the 3D Gaussian splatting approach by improving geometric accuracy and rendering efficiency. This method projects 3D Gaussian splats into a 2D representation, and it reduces computational overhead while preserving high visual fidelity. Notably, initializing 3D Gaussians from sparse point clouds remains computationally demanding, making this refinement particularly impactful.

Overall, the transition from NeRF to 3D and even 2D Gaussian Splatting outlines a clear roadmap toward more efficient and accurate scene representations. These advancements pave the way for real-time applications in large-scale environments, enabling more complex tasks in autonomous driving and robotic perception.

\begin{figure}[!thbp]
    \vspace{-0.2cm}
    \centering
    {
    \includegraphics[width=0.96\textwidth]{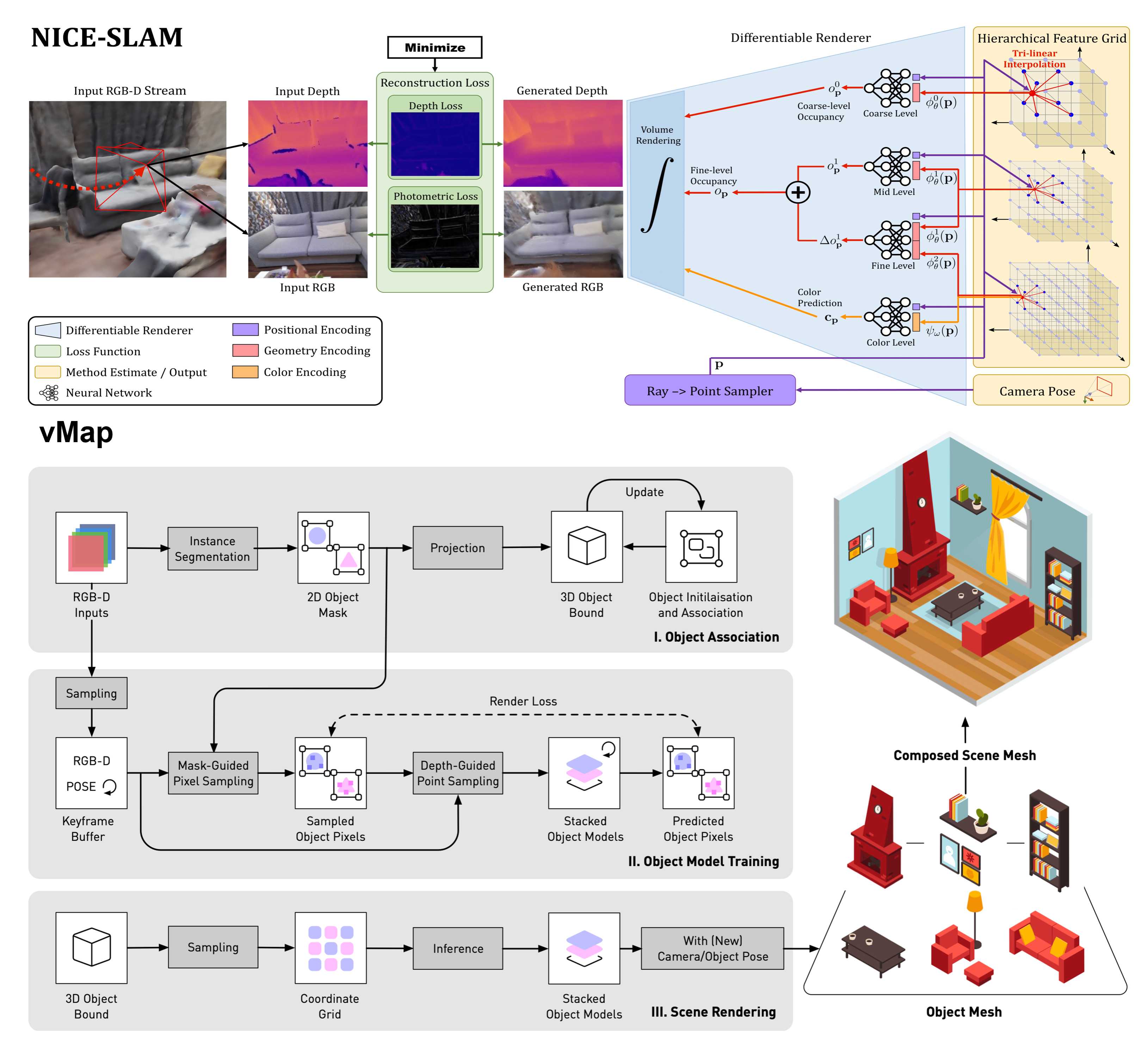}
    }
    \vspace{-0.2cm}
    \caption{The NICE-SLAM framework \cite{zhu2022nice} tracks ego-body pose and constructs a 3D mesh in real time (top row). vMap \cite{kong2023vmap} simultaneously reconstructs scene objects and renders the environment, enabling a more interactive 3D scene representation. Image from work by \cite{zhu2022nice, kong2023vmap}}
    \label{fig:implicit-SLAM}
    \vspace{-0.4cm}
\end{figure}

\noindent\textbf{Implicit Simultaneous Localization and Mapping.} 
Implicit deep learning SLAM systems achieve good performance by leveraging continuous scene representations that are obtained directly from raw color and depth images, as shown in Figure \ref{fig:implicit-SLAM}. Such systems use neural networks to map spatial coordinate information to continuous functions, thereby enabling real-time localization and mapping and producing high-quality mesh reconstructions without the need for further processing.

A major advantage of implicit models is their ability to learn autonomously robust geometric representations through end-to-end training. This allows the implicit model to better handle noisy and incomplete data, thus outperforming traditional SLAM methods under challenging conditions. Additionally, implicit models can fuse multimodal sensor data, such as images and depth maps, in a unified framework, thus the implicit model has good robustness and flexibility to use.

NICE-SLAM (\cite{zhu2022nice}) boosts SLAM (simultaneous localization and mapping) technology by leveraging neural implicit functions for real-time 3D reconstruction and pose tracking given live-streaming images. It supports scalable and robust 3D mapping by learning a continuous indoor scene representation, which easily captures complex details while being robust to noise perturbations. Additionally, its hierarchical feature interpolation improves multi-scale geometry learning, so that it effectively alleviates the over-smoothing issues of one global scene representation by leveraging multilayer perceptrons (MLP) at different feature resolution levels.

Gaussian Splatting SLAM \cite{matsuki2023gaussian} uses Gaussian splats for 3D representation, significantly improving real-time rendering and computational efficiency compared to NICE-SLAM. This approach provides an approximate geometry representation of the environment while simultaneously keeping a fast reconstruction and rendering performance.

Vectorized Object Mapping for Neural Field SLAM (\cite{kong2023vmap}) extends neural field SLAM by integrating vectorized object representations. vMap detects and registers object instances on the fly, enabling complex scene and object-level reconstruction. Combining semantic understanding with interactive scene representations enhances robot manipulation and navigation capabilities \cite{wong2025survey}. In short, the fusion of traditional geometric techniques with modern deep learning architectures has become an active field of research in recent 3D vision research. Traditional approaches offer interpretability and build upon proven mathematical models but they often struggle with scalability issues and the treatment of complex, unstructured environments. Deep learning, on the other hand, enables data-driven generalization and achieves high-quality reconstruction but often lacks geometric consistency, in addition to high training complexity. This thesis builds on these recent developments, and it aims to leverage Geometric Deep Learning, in combination with the strengths of classical geometric representations, aiming at the development of robust and efficient deep learning models for 3D vision applications.


\chapter{Camera Pose Estimation for Images Using Natural Geometry Cues and Manifold Constraint}                          
\label{ch:cp-pose-image}

\begin{center}
    \textbf{\large Abstract}
\end{center}
    
Accurate camera pose estimation from image frames is essential for 3D reconstruction. This chapter \footnote{The majority of this chapter was published as a peer-reviewed conference paper \cite{kang2023adaptive} (Kang, Xueyang, et al. ``Adaptive sampling-based particle filter for visual-inertial gimbal in the wild.'' 2023 IEEE International Conference on Robotics and Automation (ICRA). IEEE, 2023.
), with financial support from the VLIR-UOS project (Agreement No. EC2020SIN278A101).\\
\textbf{Author Contributions:} \\
\vspace{-1.2em}
\begin{itemize}
    \item \textbf{Xueyang Kang}: Idea Design, Methodology, Software, Experiment Validation, Formal Analysis, Data Curation, Writing, Review, and Editing.
    \item Ariel Herrera: Data Curation and Experiment Validation.
    \item Henry Lema: Data Curation.
    \item Esteban Valencia: Review.
    \item Patrick Vandewalle: Review, Editing, Supervision, and Funding.     
\end{itemize}
} presents a vision-based orientation tracking and fusion algorithm for drones operating in natural environments. The system can be used to mitigate motion blur and stabilize camera orientation by taking advantage of the skyline and ground plane as reference signals, ensuring high-quality image capture for downstream 3D tasks. The key contributions include:
a) A lightweight ResNet-18 backbone, trained from scratch and deployed on a Jetson Nano, for real-time segmentation of images into binary ground and sky regions.
b) A geometry-based framework that utilizes skyline and ground cues for robust visual tracking in challenging outdoor conditions.
c) An adaptive particle filter sampling technique on a multi-resolution manifold surface, enabling flexible fusion of orientation estimates from multiple frames. The proposed method was implemented and tested in real-world environments, including rooftop and drone-mounted experiments. The final experimental results demonstrate its robustness in frame-to-frame pose estimation under challenging natural conditions.

\section{Introduction}
Pose estimation from image frames is a fundamental step in 3D reconstruction. Typically, a camera follows an orbital trajectory around the target, incorporating pan rotation and elevation changes to ensure sufficient overlap between consecutive frames for reliable feature matching. This structured motion enhances pose estimation accuracy. However, abrupt tilting with minimal height change or pure rotation can severely degrade performance due to the lack of overlap between consecutive frames. Additionally, unintended jitter during motion can introduce blur, making camera stabilization essential for obtaining sharp, high-quality frames for precise pose estimation.

Motion blur degrades image quality, making feature extraction and correspondence matching significantly more challenging, often leading to incorrect matches. Reliable frame-to-frame correspondence is crucial for accurate pose estimation and consistent 3D alignment.

In UAV-based image capture, especially in mountainous regions with unpredictable rotations and motion jitter, conventional IMU-based camera pose tracking methods struggle with long-term drift and noise. To address these challenges, we propose a novel approach to estimate camera orientation by leveraging natural scene cues and a manifold-based fusion strategy. The key contributions of this chapter are as follows.

\begin{itemize}
\item A lightweight binary segmentation model trained to classify ground and sky pixels in real-time on an embedded device.
\item Geometry-based camera pose estimation method that uses the skyline and ground plane as reference cues to infer rotation angles.
\item A nonlinear particle filter with adaptive sampling resolution on the manifold surface to fuse rotation estimates from multiple sensor modalities (IMU and vision-based cues), ensuring robust orientation tracking in challenging outdoor conditions.
\end{itemize}

Our approach integrates manifold-based constraints to enforce geometric consistency in rotation estimation over a spherical domain. Traditional methods relying solely on high-precision IMUs often suffer from noise drift and instability. In contrast, our fusion framework mitigates these issues through adaptive sampling on the manifold surface, ensuring robust multi-modal integration. Using natural scene features, our method provides stable and accurate pose estimation, even in dynamic and unpredictable environments, demonstrating a significant advancement in geometry-aware deep learning for UAV-based vision tasks.

\section{Related Work}
Pose estimation from image frames is a fundamental problem in computer vision, often approached through feature matching, motion estimation, or direct pose regression. Traditional methods rely on handcrafted feature descriptors such as SIFT \cite{sift-video}, SURF \cite{surf-tracking}, or optical flow \cite{optical-flow} to establish correspondences between consecutive frames. However, these techniques are highly sensitive to motion blur and textureless regions, which degrade tracking performance \cite{matsushita2005full}.

To improve robustness, inertial sensors have been integrated into pose estimation pipelines. Some works compensate for motion jitter by fusing IMU data with vision-based tracking, as seen in humanoid robotics \cite{robot-IMU} and complex motion prediction tasks \cite{hybrid-motion}. A review of motion estimation techniques for video stabilization, many of which are applicable to pose estimation, is provided by Rawat et al. \cite{review-stabilization-tech}.

Recent advances in deep learning have significantly improved pose estimation by leveraging learned priors. Yu et al. \cite{selfie-tracking} introduced a scene representation approach that extracts structured cues such as background feature points, foreground contours, and 3D facial meshes to aid tracking. Li et al. \cite{deep-IMU-stabilization} trained a model to estimate visual odometry from raw IMU data, demonstrating the effectiveness of learning-based fusion for motion prediction.

Self-supervised methods have also been explored, where pose estimation is embedded within larger mapping and localization frameworks. Lee et al. \cite{stabilization-depth} proposed a graph-based self-supervised method, while Liu et al. \cite{hybrid-neural-fusion} introduced a dense warping field for motion compensation, synthesizing stabilized frames from sequential observations. Choi et al. \cite{self-supervised} further refined pose estimation by integrating optical flow-based motion prediction. Despite their accuracy, these approaches involve significant computational overhead, making them impractical for real-time inference on edge devices.

Pose estimation is crucial in UAV-based applications, particularly for navigation and localization in challenging environments \cite{video-surveillance, Patent1}. Several works use skyline tracking \cite{sky, carrio2018attitude, fixed-wing} to estimate camera orientation, but these methods require careful tuning and perform poorly in non-ideal conditions. Other approaches rely on geometric constraints, such as the five-point algorithm \cite{five-point} for epipolar geometry estimation or curvature alignment techniques for motion tracking \cite{boundary}. Moving object detection has been tackled using tracking filters \cite{walha2015video} and SIFT-based correspondence matching \cite{sift}. However, feature-based methods struggle in natural environments due to spurious correspondences and occlusions, limiting their reliability for pose estimation.
\section{Method}
\label{sec:method}

\begin{figure}[!th]
\centering
\includegraphics[width=0.96\linewidth]{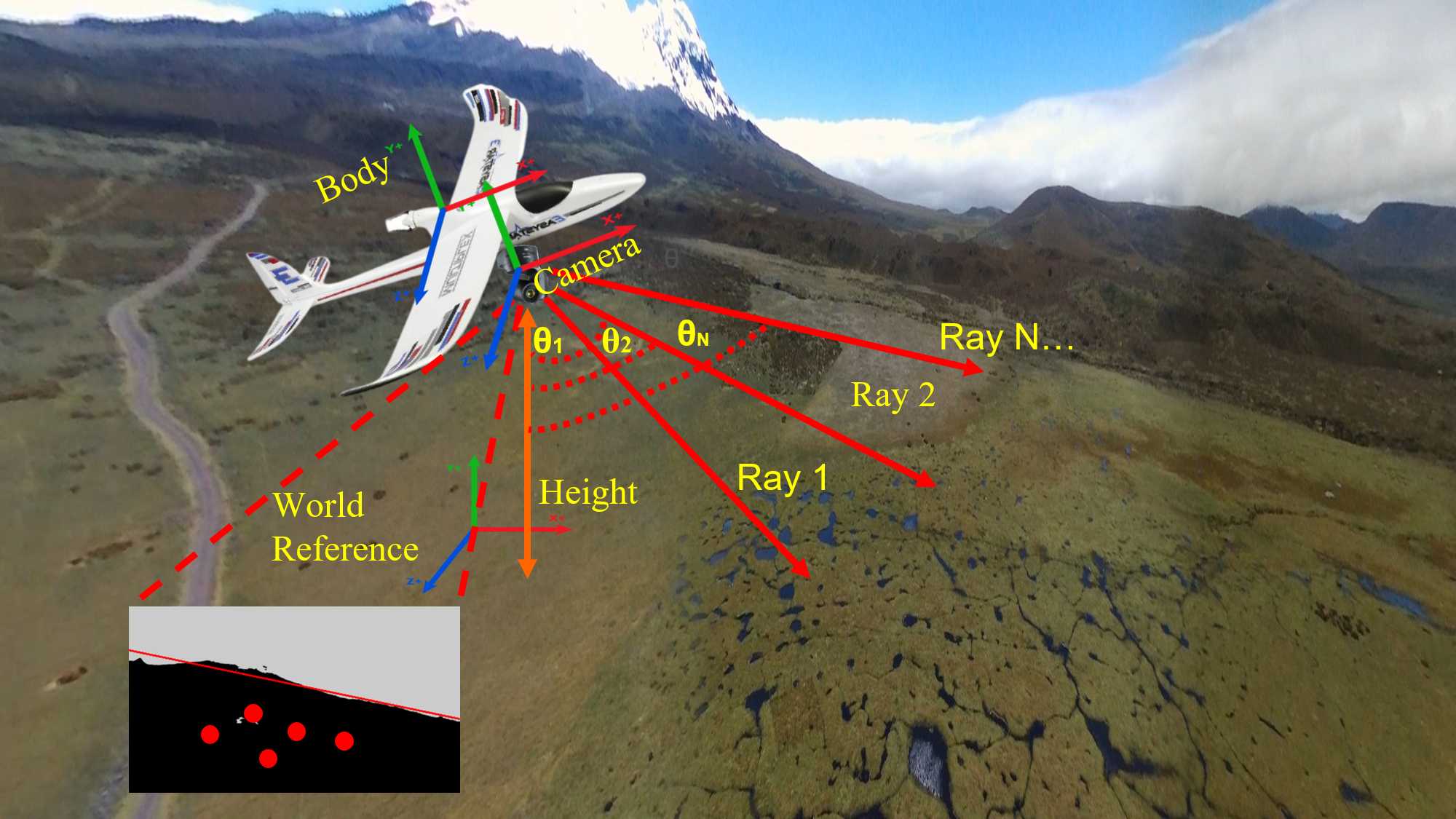}
\caption{Illustration of pose orientation estimation from image frames on the fixed-wing airplane.}
\label{fig:overview}
\vspace{-1.0em}
\end{figure}

As shown in Figure \ref{fig:overview}, the system is mounted beneath the airplane body and consists of a camera, an IMU, and a barometer. Ensuring stable and accurate camera orientation during flight is crucial, particularly in dynamic and unstructured environments where feature-based tracking often fails due to motion blur and inconsistent correspondences. To overcome these challenges, our approach leverages natural visual cues—such as the skyline and ground plane—to aid pose estimation. The lower part of Figure \ref{fig:overview} illustrates a binary mask where the skyline and ground region are distinctly segmented, providing reliable geometric references.

A key contribution of our method is the manifold-based adaptive particle filter, which fuses IMU-based orientation estimates with visual cues extracted from the segmented scene. Unlike traditional filtering approaches, which may suffer from noise drift or require precise sensor calibration, our method operates directly on the rotation manifold, ensuring smooth and consistent pose estimation. This fusion not only enhances robustness against sensor noise but also improves long-term orientation estimation stability, enabling accurate camera orientation in challenging aerial conditions.

Rays passing through the red dots, in conjunction with height measurements obtained from the barometer, enable us to accurately determine the 3D position of the ground plane using trigonometric calculations. This geometry relationship is fundamental to our approach, as it allows for precise estimation of the camera's orientation relative to the ground, thereby improving the stability of the captured imagery. The hardware diagram details the integration of the camera, IMU, and barometer, while the software diagram outlines the algorithmic pipeline that processes the visual and inertial data.


Overall, the capability of such an orientation estimation system to leverage natural environmental features, coupled with advanced geometry deep learning techniques and efficient sensor fusion, represents a significant advancement in UAV-based camera pose tracking in the natural environment. This robust design is particularly well suited for long-term operations in remote and challenging terrains, such as volcanic regions, where traditional pose estimation systems might fail due to noise drift or the lack of high-precision sensors. The integration of manifold-based constraints in our pose estimation approach ensures high precision and stability, making it an ideal solution to improve the quality of captured video and the reliability of surveillance missions.


\noindent \textbf{Hardware.} The hardware design is based on open-source hardware components. The main processing unit is a Jetson Nano, equipped with 2GB of GPU memory and Quad-core ARM A57, connected to IMU, camera, and barometer sensors. An OpenCR 2.0 driver board maps the driving command to control commands for two servo motors.

\begin{figure}[!thbp]
\centering
\includegraphics[width=0.96\textwidth]{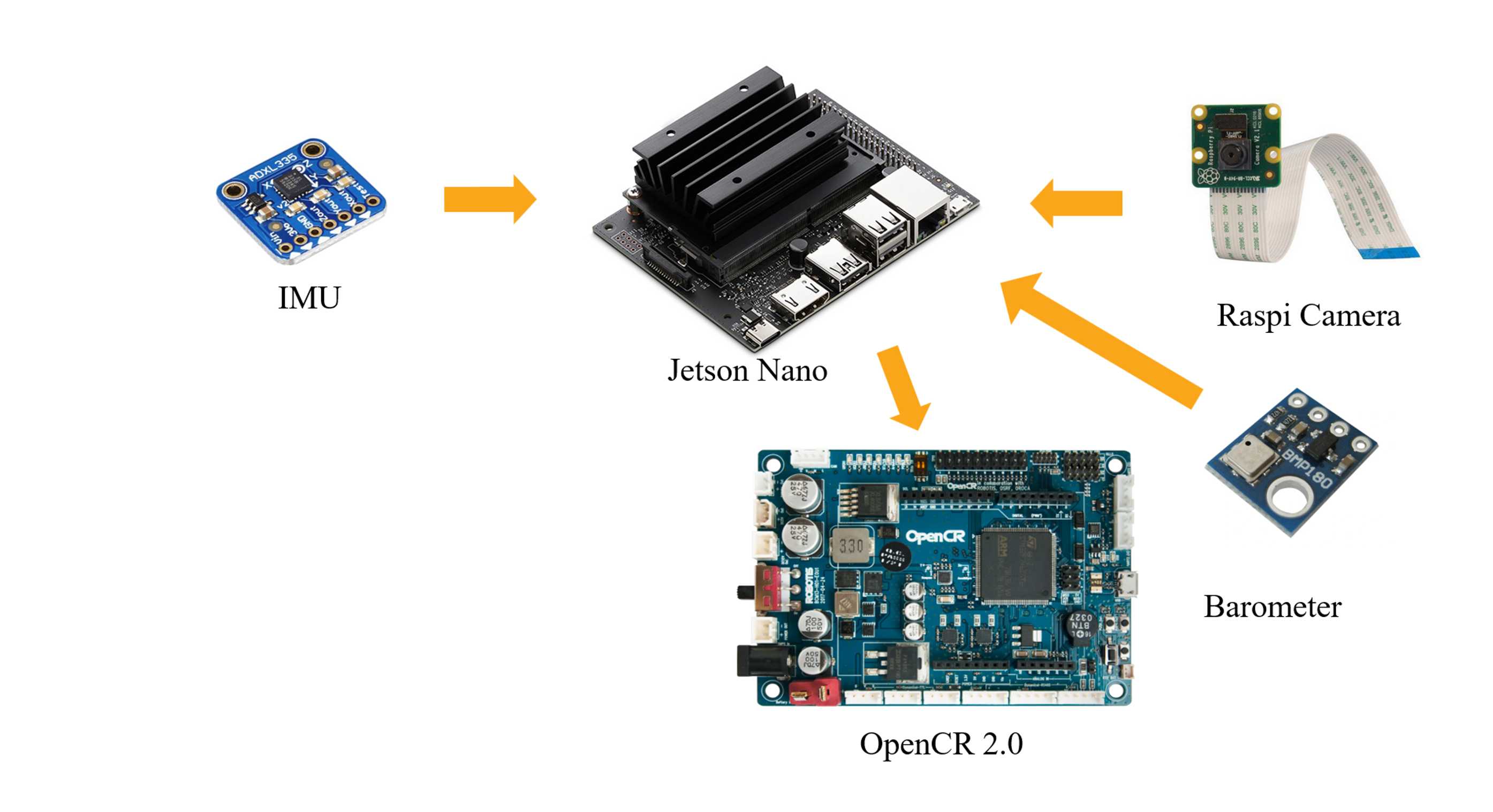}
\label{fig:hardware}
\vspace{-1.0em}
\caption{Open source hardware setup.}
\vspace{0.0em}
\end{figure}

\noindent\textbf{ROS Nodes.} The software is composed mainly of three parts: the preprocessing part including the network model and geometry primitive extraction, followed by a tracking module to align the skyline and normal of the ground plane in the current frame with those in the reference frame. The compensation angles from various pipelines are then fed into the proposed particle filter presented in the Section below to obtain fusion orientations, further as input for the controller to stabilize the camera. 

\begin{figure}[!thbp]
  \centering
  \resizebox{0.7\linewidth}{!}{
  \begin{tikzpicture}
    \node[draw,
        circle,
        text width=1.0cm,
        minimum size=0.1cm,
        fill=Rhodamine!40,
        text centered] (Video){\centering Video\\Input};
    
    \node [draw,
        fill=Goldenrod,
        minimum width=1cm,
        minimum height=1cm,
        above=0.5cm of Video,
        text centered
    ]  (SegNet) {\begin{tabular}{c} Binary Segmentation \\ \& Skyline Isolation \end{tabular}};
    
    \node[draw,
        circle,
        text width=1.0cm,
        minimum size=0.4cm,
        fill=Rhodamine!40,
        right=0.2cm of SegNet,
        text centered
    ] (Skyline){Skyline};

    \node[draw,
        text centered,
        circle,
        text width=1.0cm,
        minimum size=0.2cm,
        fill=Rhodamine!40,
        above=0.6cm of Skyline
    ] (Ground){\centering Ground\\Plane};

    \node[draw,
        circle,
        text width=0.9cm,
        minimum size=0.1cm,
        fill=Rhodamine!40,
        right=0.5cm of Video,
        text centered
    ] (IMU){\centering IMU\\Data};

    \node [draw,
        fill=Goldenrod,
        minimum width=1cm,
        minimum height=1cm,
        right=0.2cm of Skyline,
        text centered
    ]  (Perception) {\begin{tabular}{c} Geometry\\Tracking \end{tabular}};

    \node[draw,
        circle,
        text width=0.9cm,
        minimum size=0.1cm,
        fill=Rhodamine!40,
        above=0.8cm of Perception,
        text centered
    ] (Barometer){Height};

    \node [draw,
        fill=Goldenrod,
        minimum width=0.8cm,
        minimum height=0.8cm,
        below=0.75cm of Perception,
        text centered
    ]  (pf) {\begin{tabular}{c}Particle\\Filter \end{tabular}};
    
    \node[draw,
        circle,
        text width=0.9cm,
        minimum size=0.1cm,
        fill=Rhodamine!40,
        right=0.8cm of pf,
        text centered
    ] (SO2){\centering Angles\\SO(2)};
    
    
              
    \draw[->] (Video.north) -- (SegNet.south);
    \draw[->] (SegNet.east) -- (Skyline.west);
    \draw[->] (SegNet.north) -- (Ground.west);
    \draw[->] (Skyline.east) -- (Perception.west);  
    \draw[->] (Ground.east) -- (Perception.north);  
    \draw[->] (IMU.east) -- (pf.west);   
    \draw[->] (Perception.south) -- (pf.north); 
    \draw[->] (Barometer.south) -- (Perception.north);
    \draw[->] (pf.east) -- (SO2.west);   

     
     
     
  \begin{pgfonlayer}{background}
    \path (SegNet.west |- SegNet.north)+(-0.2,0.4) node (A) {};
    \path (Perception.east |- Perception.south)+(+0.1,-0.4) node (B) {};
    \path[fill=yellow!20,rounded corners, draw=black!50, dashed] (A) rectangle (B);
  \end{pgfonlayer}
  \end{tikzpicture}}
  \caption{Block diagram of the presented orientation estimation algorithm. Circular nodes in pink are signals. Rectangular boxes in yellow are ROS nodes for the algorithm, and the dashed region is the front-end perception part, including tracking of the skyline and ground plane.}
  \label{fig:ros-nodes}
\vspace{-0.7em}
\end{figure}
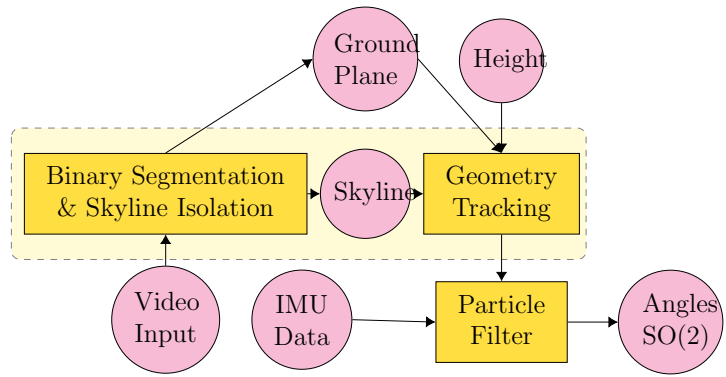  

\subsection{Perception}
\label{section:percep}
The perception part is structured into preprocessing and rotation estimation, where the rotation estimation can be further separated into two pipelines: roll and pitch prediction from skyline tracking and rotation estimation from ground plane tracking. The following subsections follow this structure.

\begin{figure}[!ht]
    \vspace{-0.2cm}
    \centering
    \includegraphics[width=0.96\textwidth]{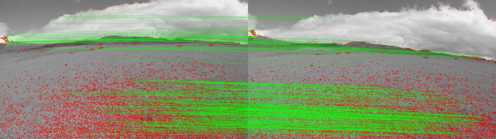}
    \caption*{(a) Correspondences based on "SIFT" \cite{lowe2004sift} feature points of neighboring image frames.}
    
    \medskip
    
    \includegraphics[width=0.48\textwidth]{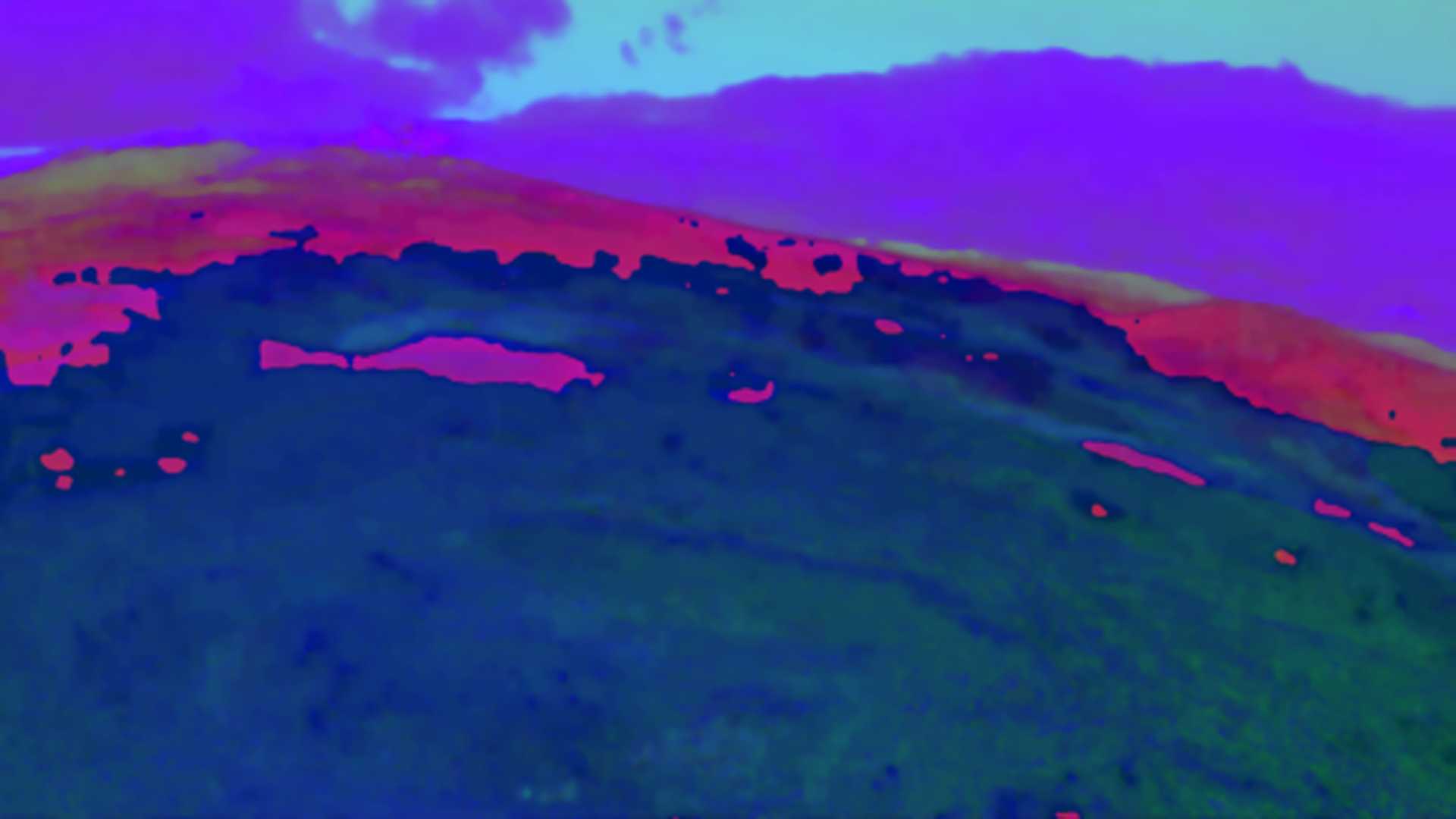}
    \hfill
    \includegraphics[width=0.48\textwidth]{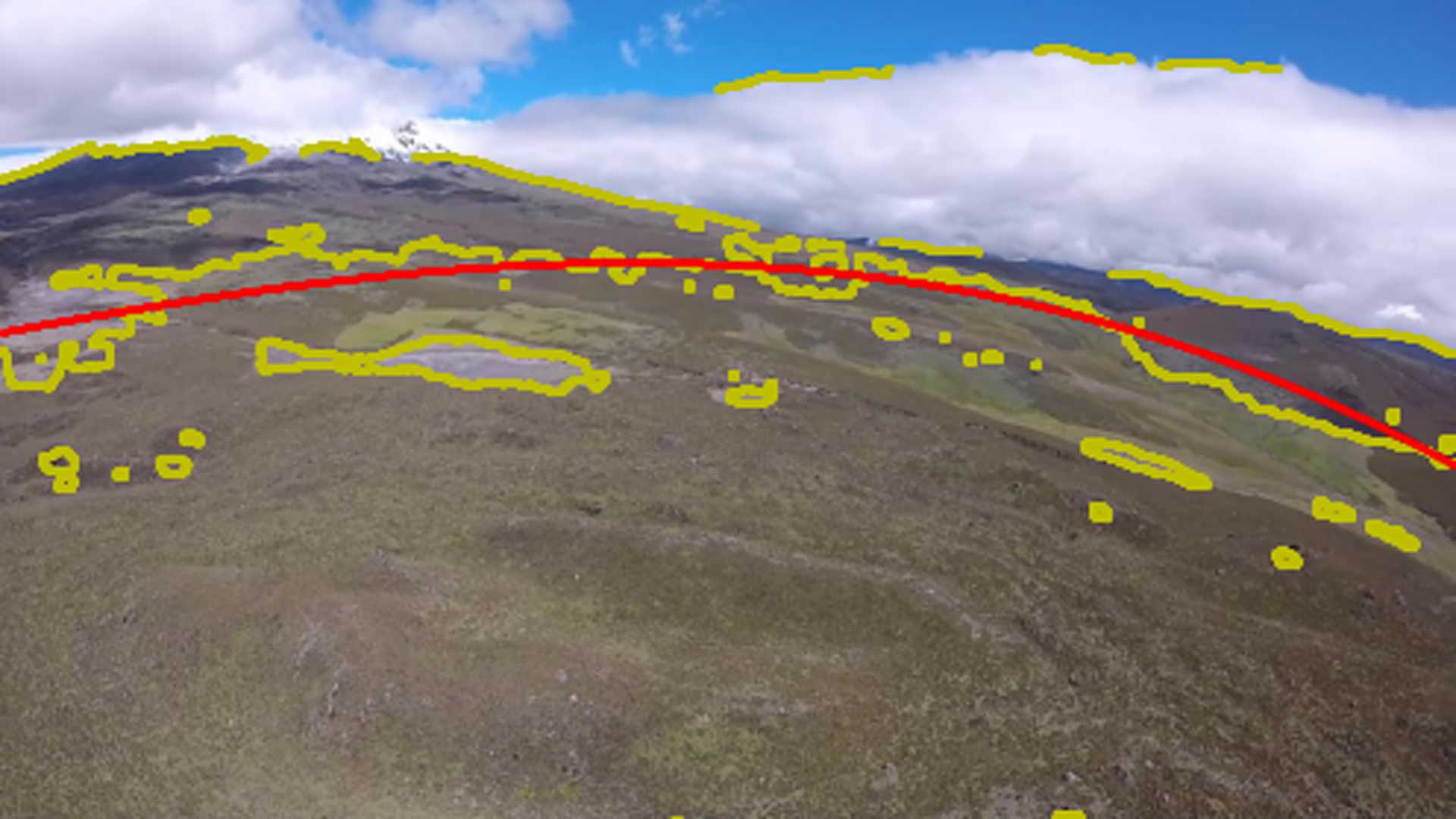}
    
    \caption*{(b) HSV image converted from a raw RGB image. \hfill (c) Boundary curve-fitting on the detected Canny edges.}
 \label{fig:failure-case}

    \vspace{-0.2cm}
    \caption{Failure case demo using traditional OpenCV pipeline.}
    \label{fig:feature-association}
\end{figure}
\begin{figure}[!ht]
    \vspace{-0.2cm}
    \centering

    \includegraphics[width=.32\linewidth]{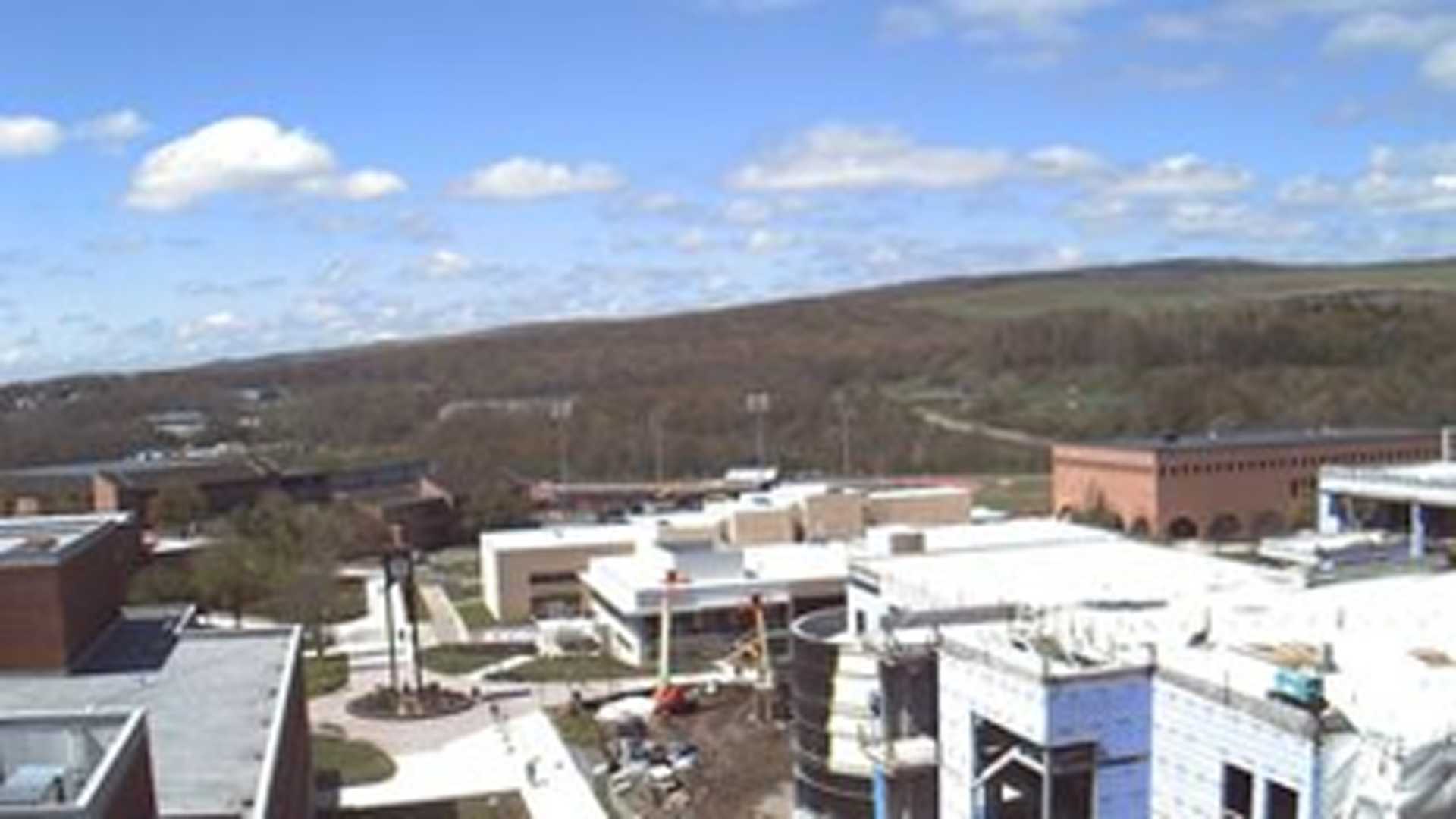}
    \hfill
    \includegraphics[width=.32\linewidth]{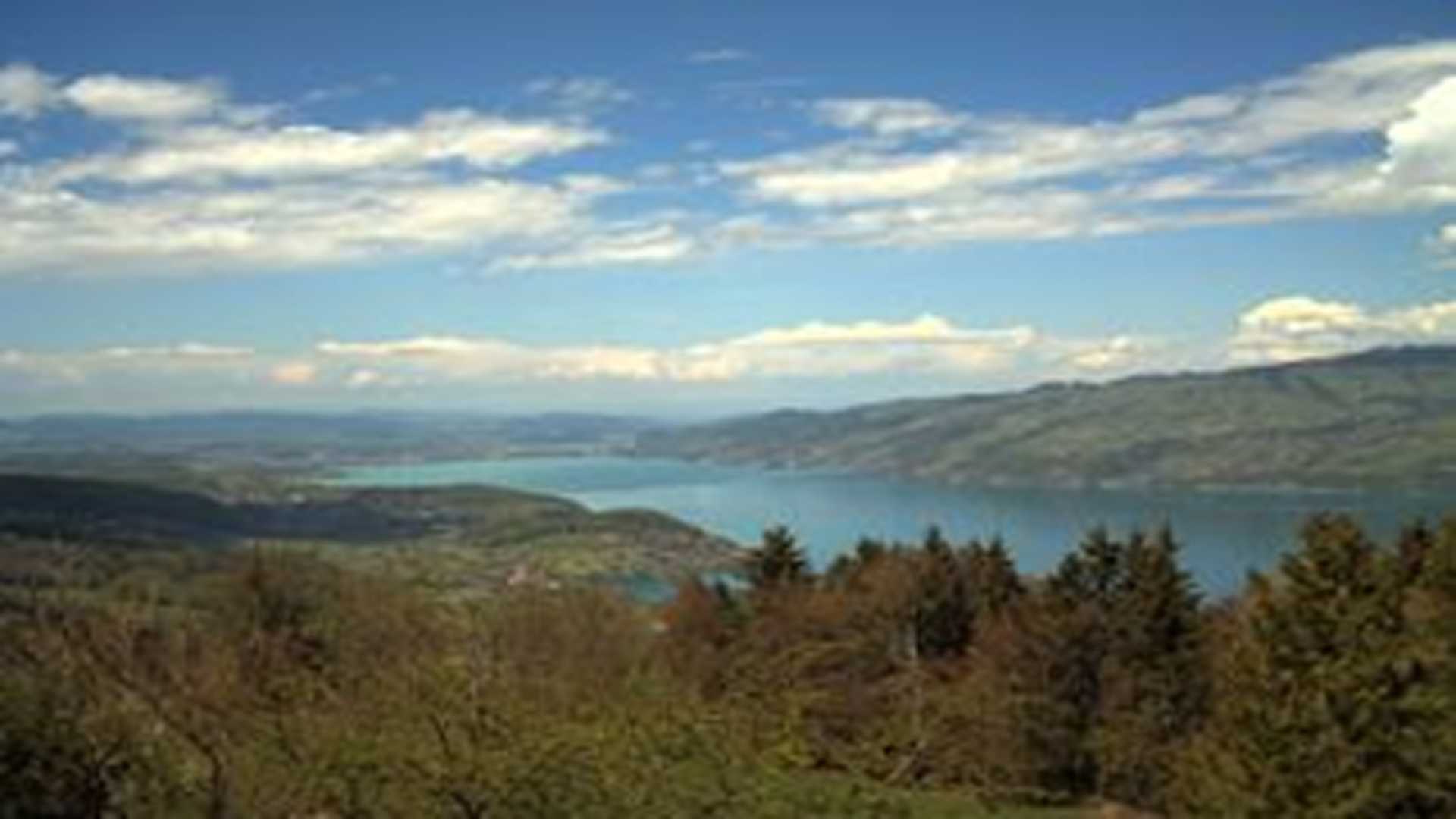}
    \hfill
    \includegraphics[width=.32\linewidth]{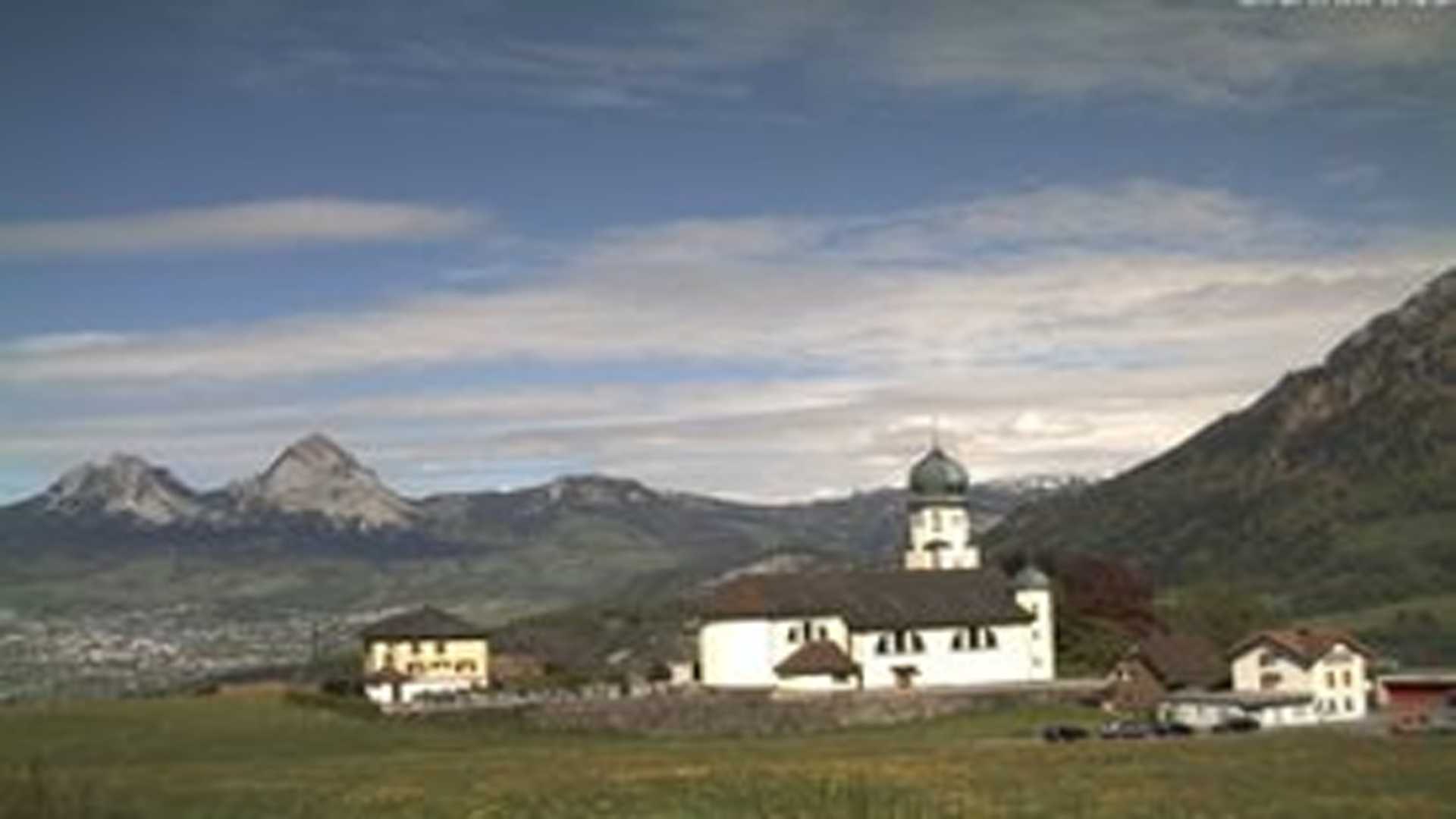}
    
    \caption*{(a) Sample training images.}

    \medskip

    \includegraphics[width=.32\linewidth]{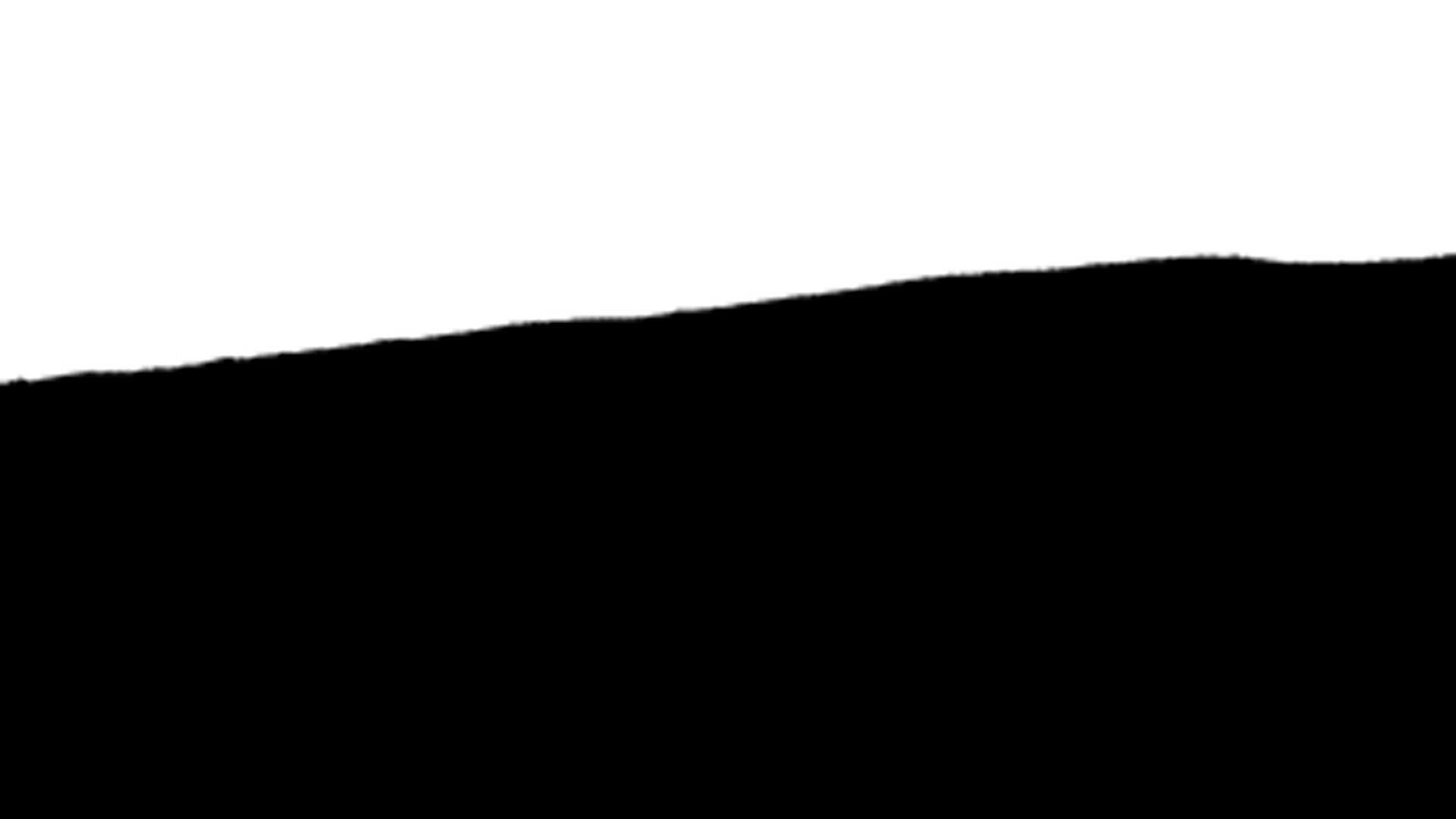}
    \hfill
    \includegraphics[width=.32\linewidth]{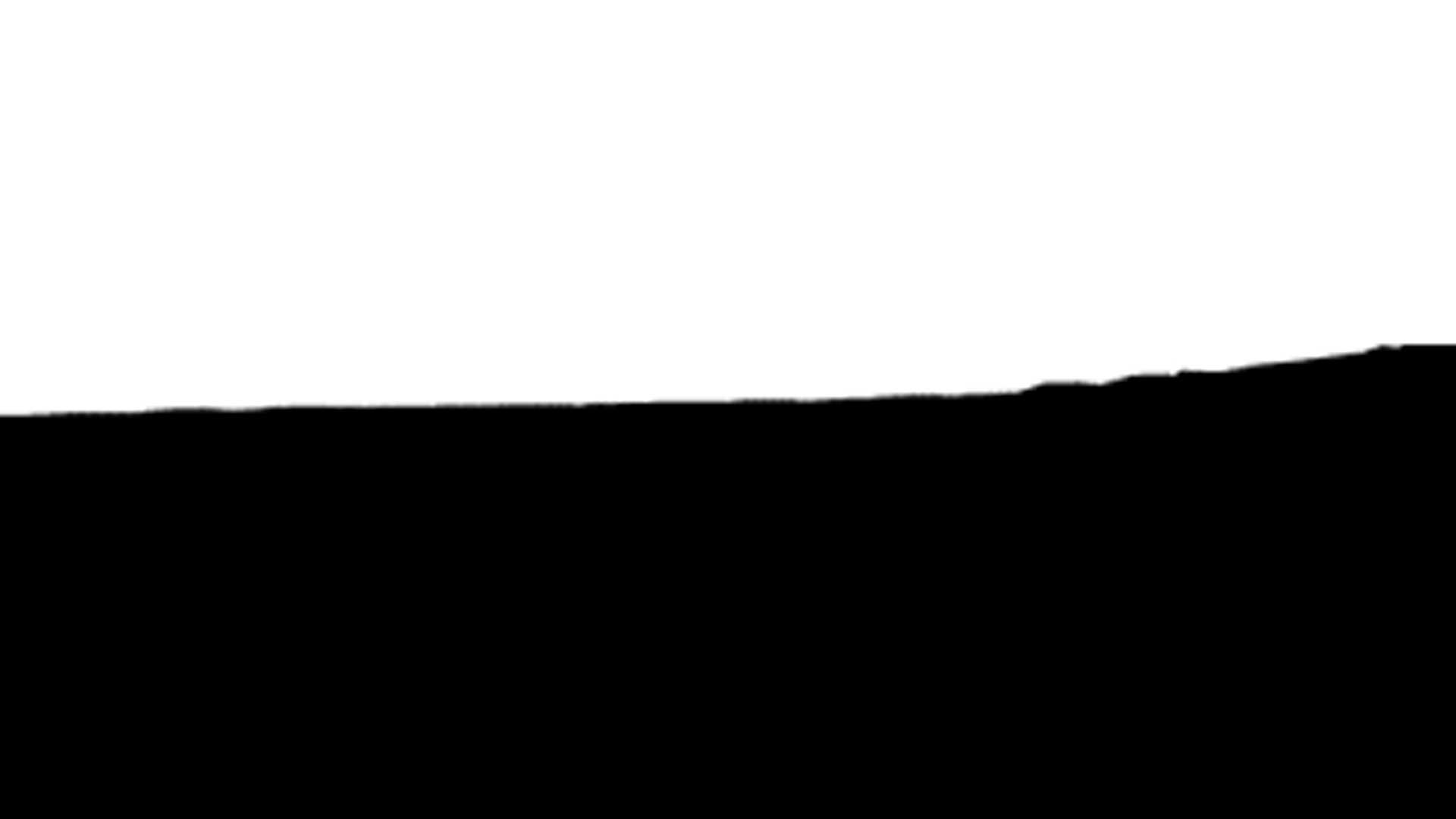}
    \hfill
    \includegraphics[width=.32\linewidth]{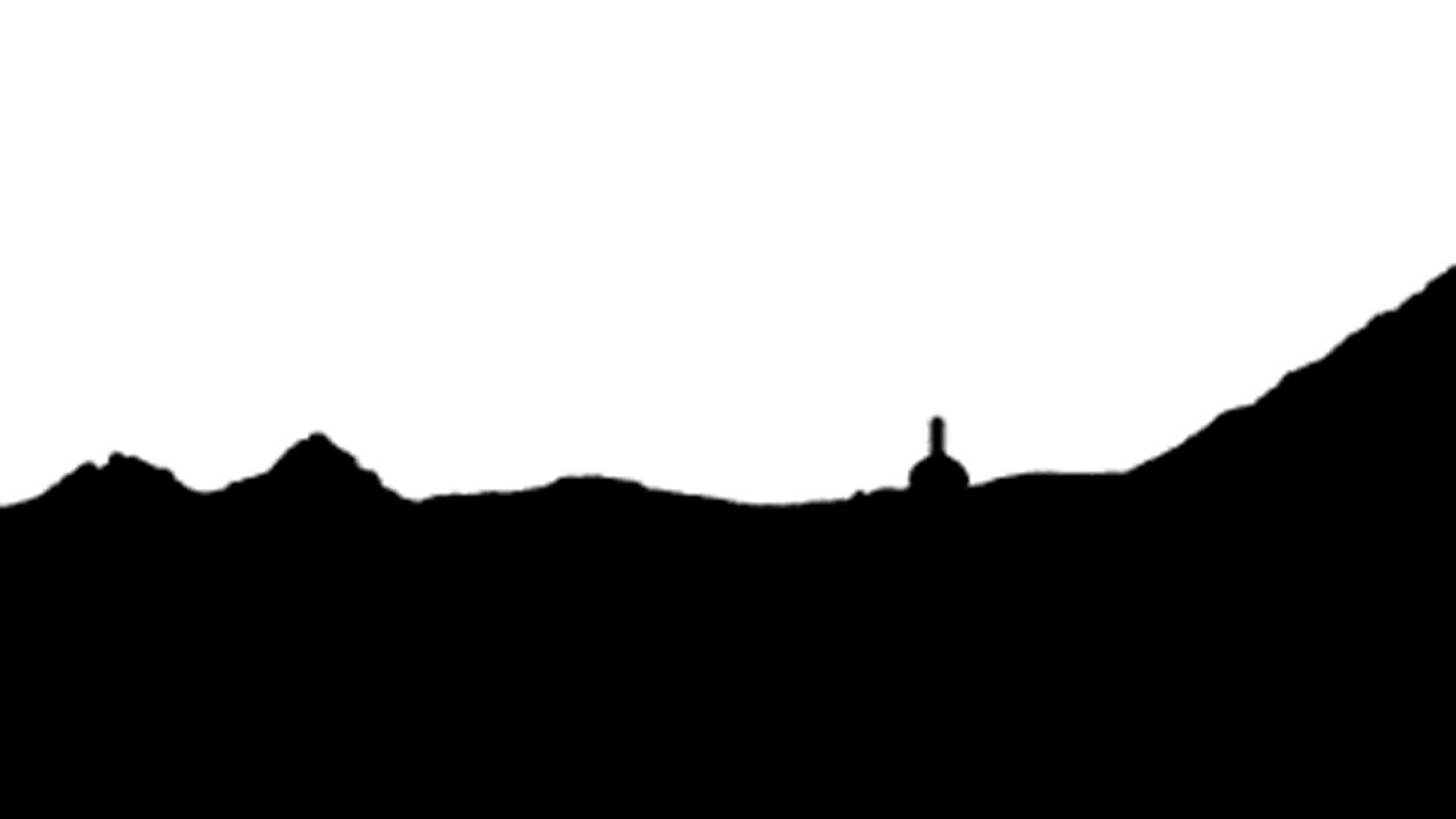}

    \caption*{(b) Corresponding ground truth masks.}
    
    \vspace{-0.2cm}
    \caption{Sample images and ground truth masks for training.}
    \label{fig:train-sample}
    \vspace{-0.5em}
\end{figure}

We first tried the general computer vision processing pipelines, but they all failed due to spurious feature candidates. As illustrated in the bottom left of Figure \ref{fig:feature-association}, the similar appearances in the grassland region and cloudy sky all pose a great challenge to correspondence search. In the top of Figure \ref{fig:feature-association}, many false correspondences are found. Canny edge detection \cite{canny} is applied to find the boundary between the ground and sky region on the HSV image; nevertheless, some brightness of the sky is cast onto the grass ground, generating the wrong boundary in the bottom right of Figure \ref{fig:feature-association}. To tackle this challenging segmentation task, we finally choose the data-driven approach by training a Resnet-18 \cite{resnet} network on the "Skyfinder" dataset \cite{dataset} first, followed by a fine-tuning on the self-collected dataset with one hundred images. Binary cross entropy loss is applied for pre-training through 100 epochs and fine-tuning with 30 epochs, respectively. The total training time is less than two hours. Some training data samples, along with ground truth masks, are presented in Figure \ref{fig:train-sample}. The model is exported into "ONNX" and optimized by "TensorRT" to convert to "FP16" precision for Jetson Nano deployment. We found the model can achieve above 90\% success rate for the segmentation on average; only under some extreme cases like overexposure does the failure happen. Additionally, our use case is for mountainous terrain. The scenario with mirror effects and reflections by water on the ground is not advisable due to the similar color distributions of the sky and the ground.

\subsection{Skyline Tracking}
\begin{figure}[!ht]
    \vspace{-0.2cm}
    \centering

    \includegraphics[width=.48\linewidth]{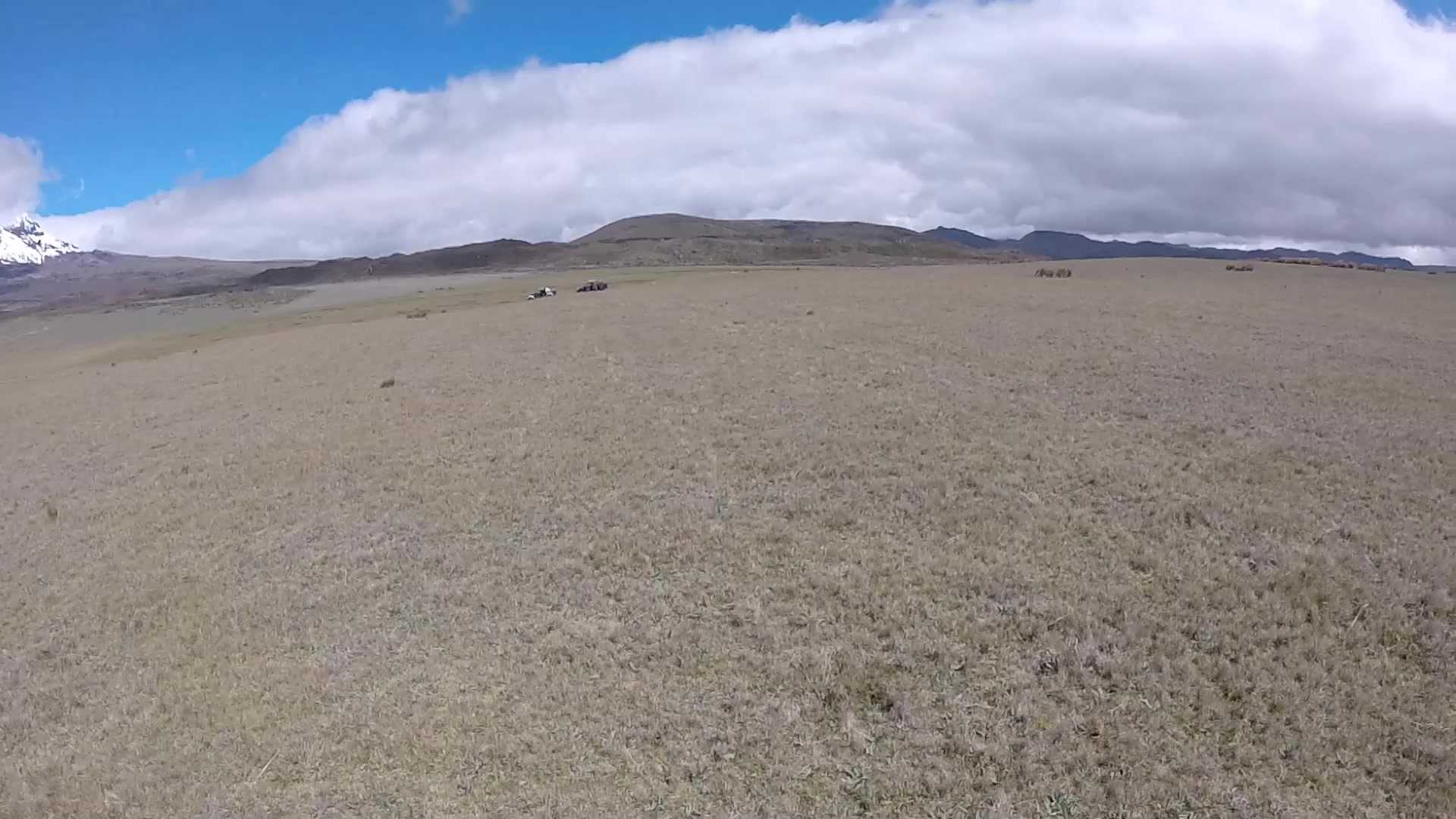}
    \hfill
    \includegraphics[width=.48\linewidth]{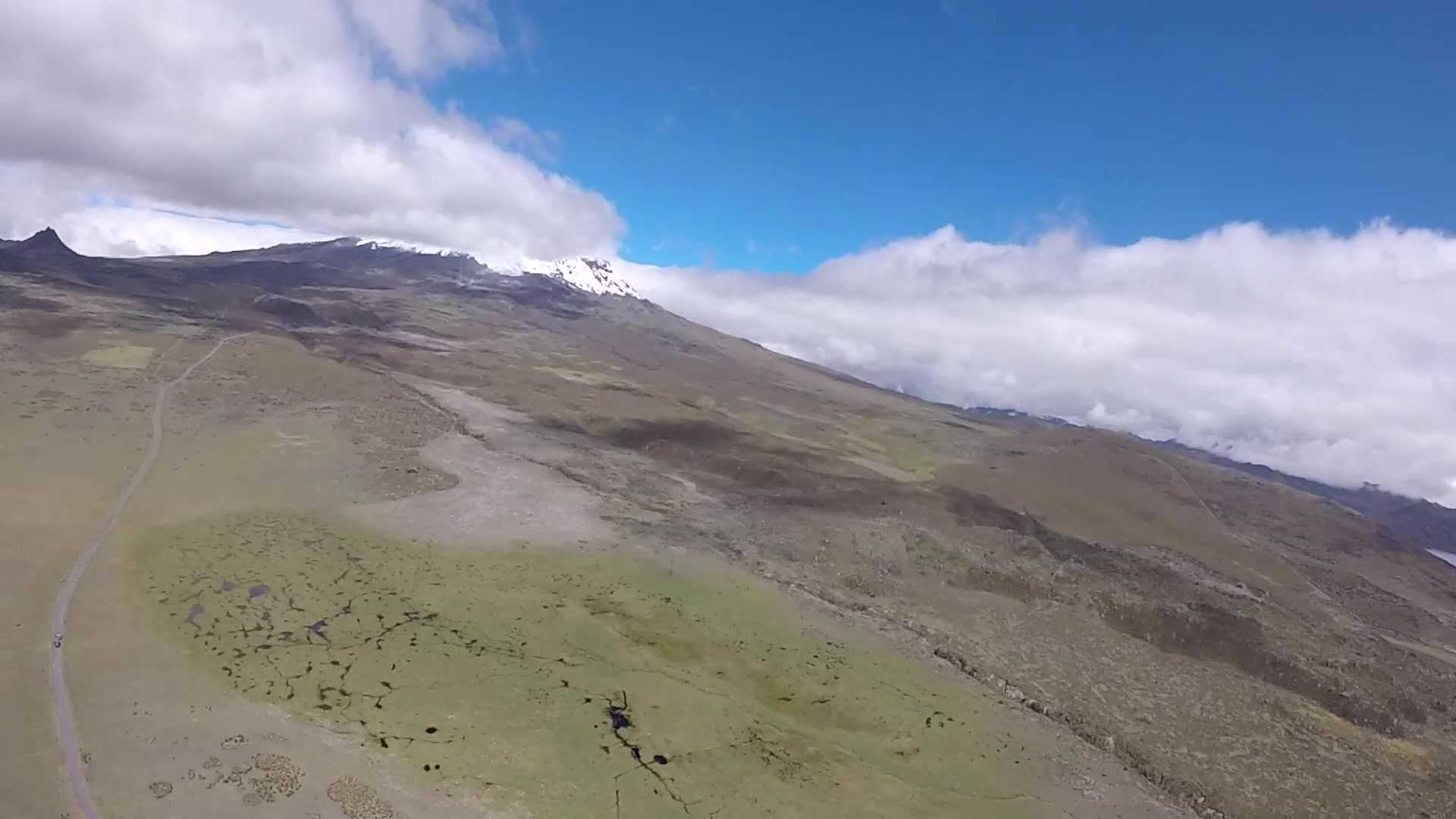}

    \caption*{(a) Reference image (left) and current image frame (right).}

    \medskip

    \includegraphics[width=.48\linewidth]{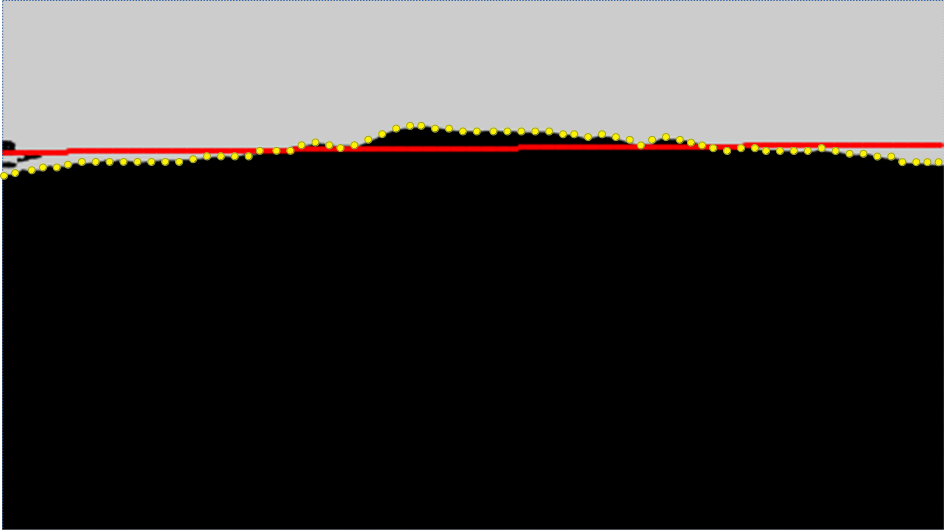}
    \hfill
    \includegraphics[width=.48\linewidth]{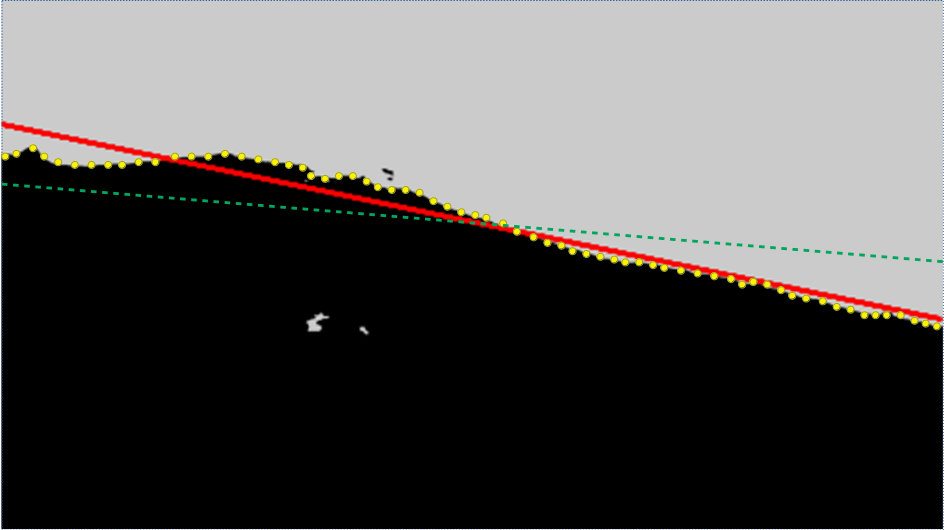}

    \caption*{(b) Reference mask with skyline (left) and current mask with skyline (right).}
    
    \vspace{-0.2cm}
    \caption{Segmented images for skyline search.}
    \label{fig:seg-samples}
    \vspace{-0.5em}
\end{figure}

Starting from the binary mask results gained from the network model, the skyline can be extracted along the boundary direction. The extracted skyline can be further considered as a cue to estimate the roll and pitch of the camera, as shown in the top of Figure \ref{fig:seg-samples}. The boundary points, represented by yellow dots at the bottom of Figure \ref{fig:seg-samples}, can be implemented firstly over the whole reference image. A straight line is fitted as illustrated in red (the bottom of Figure \ref{fig:seg-samples}), using two parameters, slope $m'$ and intercept $b'$ in Equation 1. For the subsequent images, we use a constant angular velocity model derived from skyline tracking of the previous two frames to predict the skyline position in the current frame, like the green line presented at the bottom of Figure \ref{fig:seg-samples}. The assumption is that the camera remains not upside down, due to the airplane maneuverability constraints. The boundary can always be searched along the vertical direction of the predicted skyline (green). In practice, the skyline points are further down-sampled to speed up the search. Once the current sample points of the skyline are attained, the estimated skyline (red) in the current frame can be derived via least-square at the bottom of Figure \ref{fig:seg-samples}.
\begin{flalign}
m&'x + b' = y'\\
m&x+ b = y \\
\alpha &= \arctan(m) -  \arctan(m')\\
\beta &= \arctan\left(\frac{h^{}_{1}-c^{}_{y}}{f^{}_{y}}\right)-\arctan\left(\frac{h^{}_{2}-c^{}_{y}}{f^{}_{y}}\right)
\end{flalign}
The roll $\alpha$ and pitch $\beta$ angles can be derived from Equations 3, and 4 respectively by subtracting the angles. $c_{y}$ is half of the image height, and $h^{}_{1}$ and $h^{}_{2}$ are the height of the center point of the skyline in the current image frame and reference frame respectively. $f^{}_{y}$ is the focal length $y$ of the camera.

Roll and pitch have a specific tolerance range to avoid unnecessary operations, so processing is only triggered when the movement is out of this range. Roll angle can be predicted from the slope $m$ of the skyline, followed by pitch estimation, which is on top of the image result after roll compensation. There is a total of three cases in our system, that is pure roll, pure pitch, or both happening simultaneously. Here the height shift resulting from translation is subtracted before rotation processing by barometer readings.



The orientation estimation algorithm is designed to handle roll and pitch angle changes separately. It first checks for significant roll angle changes and applies roll compensation if necessary. After that, it checks for significant pitch angle changes and applies pitch estimation, either on the roll-compensated image or directly on the raw image, depending on whether roll compensation was applied or not.
This decoupled approach can be useful in applications where roll and pitch corrections need to be applied independently, such as in image stabilization systems or camera orientation tracking. Although this compensation mechanism is designed for roll and pitch only, it can be easily adapted to the full 3D case with roll, pitch, and yaw angles.

\subsection{Ground Plane Tracking}
Ground plane tracking relies on the normal vector of the ground plane, as demonstrated in Figure \ref{fig:overview}. A set of points in the ground region of the binary mask are sampled evenly, followed by a back projection to the camera frame corresponding to Equations 5-11.
\begin{flalign}
\mathbf{K} &= 
\begin{bmatrix}
f_x & 0 & c_x \\
0 & f_y & c_y \\
0 & 0 & 1
\end{bmatrix} \\
\mathbf{\rho_{i}} &= [u, v, 1]^{T} \\
\mathbf{P_{i}} &= \mathbf{{K}^{-1}} \mathbf{\rho_{i}}, i \in (1...N) \\
\mathbf{N^{}_{G}} &= [0, 0, g^{}_{z}]^{T}\\
\cos\theta &= \frac{\mathbf{P^{}_{i}} \cdot \mathbf{N^{}_{G}}}{||\mathbf{P^{}_{i}}||\cdot||\mathbf{N^{}_{G}}||} \\
l_{i} &= \frac{h}{\cos \theta} \\
\mathbf{P'_{i}} &= l_{i}\mathbf{P_{i}}
\end{flalign}
The height $h$ is measured from the barometer, and the ray direction passing through the pixel position is $\mathbf{P_{i}}$ on the left side of Equation 7. $\mathbf{K}$ is the intrinsic matrix obtained from calibration \cite{zhang2000flexible}. $\theta$ is derived from the dot product of the gravitational vector and the ray direction. Length scale $l_i$ can be calculated by trigonometry in Equation 10. Finally, the current normal vector $\mathbf{m}$ of the ground plane is shaped by the cross product of points as below:
\begin{flalign}
    \mathbf{m} &= (\mathbf{P'^{}_{i}}-\mathbf{P'^{}_{j}})\times(\mathbf{P'^{}_{i}}-\mathbf{P'^{}_{k}})
\end{flalign}
The ground plane tracing mode is only triggered when the camera is over 300 meters above ground so that the variance of uneven grassland can be approximated by a flat plane compared to the height. 
Next, the rotation matrix to align the normal vector $\mathbf{m}$ in the current frame and the reference normal $\mathbf{n}$ at the start can be derived as follows.
\begin{flalign}
s &= \frac{\mathbf{m}}{||\mathbf{m}||} \cdot \frac{\mathbf{n}}{||\mathbf{n}||}, \\ 
\mathbf{k} &= \frac{\mathbf{m}}{||\mathbf{m}||} \times \frac{\mathbf{n}}{||\mathbf{n}||}, \\
\mathbf{k_{\times}} &= 
\begin{bmatrix}
0 & -k_{3} & k_{2} \\
k_{3} & 0 & -k_{1} \\
-k_{2} & k_{1} & 0
\end{bmatrix}  \\
\mathbf{R} &= \mathbf{I} + \mathbf{k_{\times}} + \mathbf{k_{\times}^{2}} \frac{1}{1+s}  
\end{flalign}
$\bf{k_{\times}}$ is the skew matrix, where non-zero elements are in off-diagonal positions, corresponding to the components of the cross product of $\mathbf{m}$ and $\mathbf{n}$. $s$ is a scale derived from the dot product of two vectors. The rotation matrix is calculated following Rodrigues' rotation formula in Equation 16.

In the end, the 3D Euler angles are retrieved from the rotation matrix according to Equations 17-20, in a right-hand order, "yaw$\leftarrow$pitch$\leftarrow$roll". Only roll and pitch are used for later fusion. 
\begin{flalign} \label{eq:euler}
\mathbf{R} &= 
\begin{bmatrix}
{r}^{}_{11} & {r}^{}_{12} & {r}^{}_{13} \\
{r}^{}_{21} & {r}^{}_{22} & {r}^{}_{23} \\
{r}^{}_{31} & {r}^{}_{32} & {r}^{}_{33}
\end{bmatrix}  \\
\alpha &= \arctan(r^{}_{32}, r^{}_{33})  \\
\beta &= \arctan(-r^{}_{31}, \sqrt{r_{32}^2 + r_{33}^2}) \\
\gamma &= \arctan(r^{}_{21}, r^{}_{11})
\end{flalign}

\subsection{Adaptive Particle Filter for Pose Fusion}
\label{section-pf}
The Particle Filter is easy to implement and applicable in non-linear problems, in particular for positioning \cite{pf-local}. Here, a variant of the vanilla particle filter \cite{pf-tracking}, sampling on the spherical surface adaptively is proposed in the pseudo-code below. In a real configuration, the roll and pitch are virtually constrained by a limit range due to mechanical kinematics, e.g., roll in a range from -45$^\circ$ to 45$^\circ$.


\begin{algorithm}[!htbp]
    \SetNlSty{}{}{:}  
    \SetAlgoNlRelativeSize{0}  
	\caption{Particle Filter on Spherical Surface}
		 \KwData{${S}_{I}= ({\alpha}_{I}, {\beta}_{I}), {S}_{{C}_{1,2}}= ({\alpha}_{{C}_{1,2}}, {\beta}_{{C}_{1,2}}), \Omega_{1,2,3}$.}
		\KwResult{\textbf{Output} $\bar{S}_{k} = [{\alpha}$, ${\beta}]^T$.}
	\If{new ${S}^{I}$ available}{
        Initialize the particle $s_{0}^{(j)}$ or sampling $s_{k}^{(j)}$ (on $\Omega_{1}$) from $\mathcal{N}(\tilde{\mu}_{I}|\mu_{I} + \omega\delta t,(\sigma_{I}+b))$, $j=1...N$;} 
		\If {new ${S}_{{C}_{1}}$ and ${S}_{{C}_{2}}$ both are available}
		{\For {$s_{k}^{(j)}$ in the particle set of $N$ samples}
		{$\omega_{k}^{(j)}=\omega_{k-1}^{(j)}\exp{(s_{k}^{(j)}-{S}_{{C}_{1,2}})}$; \\ 
		\If {$||s_{k}^{(j)}$-${S}_{{C}_{1,2}}||_{2} \leqslant \epsilon$}
		{Sampling $\hat{s}_{k}^{(1...m)}$ from $\mathcal{N}(\tilde{\mu}_{f}|\mu_{f}, \delta_{f})$ (on $\Omega_{3}$)\\
		initialize new weights: $\omega_{k}^{(1...m)}=\omega_{k}^{(j)}\exp{(\hat{s}_{k}^{(1...m)}-\mu_{f})}$;
		}
		}
		}
		\If {new ${S}_{{C}_{1}}$ or ${S}_{{C}_{2}}$ is available}
		{\For {$s_{k}^{(j)}$ in $N$ samples (on $\Omega_{2}$ resolution)}
		{
		 Sampling from $\mathcal{N}(\tilde{S}_{{C}_{1}}|{S}_{{C}_{1}}, \delta_{{C}_{1}})$ or $\mathcal{N}(\tilde{S}_{{C}_{2}}|{S}_{{C}_{2}}, \delta_{{C}_{2}})$, same as line 6-9; \\
		}
	    }
		$\hat{s}_{k}^{(1...m)} \cup S_{\Omega}$; \\
		Resampling $s_{k}^{(j)}$ according to $\omega_{k}^{(j)}$; \\
		$\bar{S}_{k} = \sum_{j=1}^{N+m}\omega^{(j)}_{k}\hat{s}_{k}^{(j)}$; \\
    \label{alg:pf}
\end{algorithm}
The general idea behind this adaptive particle filter in Figure \ref{alg:pf} is straightforward. The filter mainly comprises three steps: sampling from orientation measurements of the IMU ${s}_{I}^{}$ (line 2); sampling according to observation from the Computer Vision (CV) pipeline (lines 8 and 15); resampling proportionally to the updated weight of each particle (line 19). The weight in line 6 is derived from a normal distribution, as a function of the square root of the angular distance, which is between sampled cell position and sensor observation on the manifold surface. ${S}_{{C}_{1}}$ and ${S}_{{C}_{2}}$ indicate the orientation estimation from the skyline and the ground plane respectively. 

It is noteworthy that all the particle samples are generated on the discretized cells, spreading over the manifold space formed by the roll and pitch angles. Each particle is a 2D vector represented by a cell position on the spherical surface. There are three levels of cell resolution ($\Omega_{1, 2, 3}$ in Algorithm 1) from coarse to fine, where the longitudinal direction of the spherical surface represents the pitch, while the latitudinal direction is the roll. In line 2, particles are sampled from a Gaussian distribution $\mathcal{N}$, with a mean value at the IMU measurements plus a shift by a constant angular velocity propagating through a certain interval plus an offset $b$. 

When both observations from the skyline and ground plane are available (line 4), more samples will be created around those cells close to sensor measurements (line 8), and their weights are initialized by multiplication of parent weight and local weight as a function of angular distance (line 9), whereas the other particles' weights will be down-weighed by an aforementioned normal distribution as a function of angular distance. Line 7 manifests the angular distance criteria for neighboring particle cells close to observation. The sample cells meeting the criteria are used as parents to create more children particles around $\mu_{f}$ at line 8 of Algorithm 1, and $\mu_{f}$ is derived from Equation \ref{eq:fusion}, as a weighted sum of results from two Computer Vision pipelines, with each result used as a mean value of the normal distribution. A simple inverse of corresponding variance $\delta_{{C}_{1}}, \delta_{{C}_{2}}$ respectively is considered as weight for mean sum. 
\begin{flalign}\label{eq:fusion}
    \mu_{f} &= \frac{\mu_{{C}_{1}}}{\delta_{{C}_{1}}} + \frac{\mu_{{C}_{2}}}{\delta_{{C}_{2}}} \\
    \delta_{f} &= {\left(\frac{1}{\delta_{{C}_{1}}} + \frac{1}{\delta_{{C}_{2}}}\right)}^{-1}
\end{flalign}
IMU and CV observation variances are set as a constant according to practical tests. $\mu_{f}$ and $\delta_{f}$ are fused results from two CV pipelines in the same weighted sum form of Equations \ref{eq:fusion}. The same strategy repeats when a single CV observation pipeline is present (line 15), but sampling rather on a middle-level resolution.

\begin{figure}[!thbp]
\centering
\includegraphics[width=0.92\textwidth]{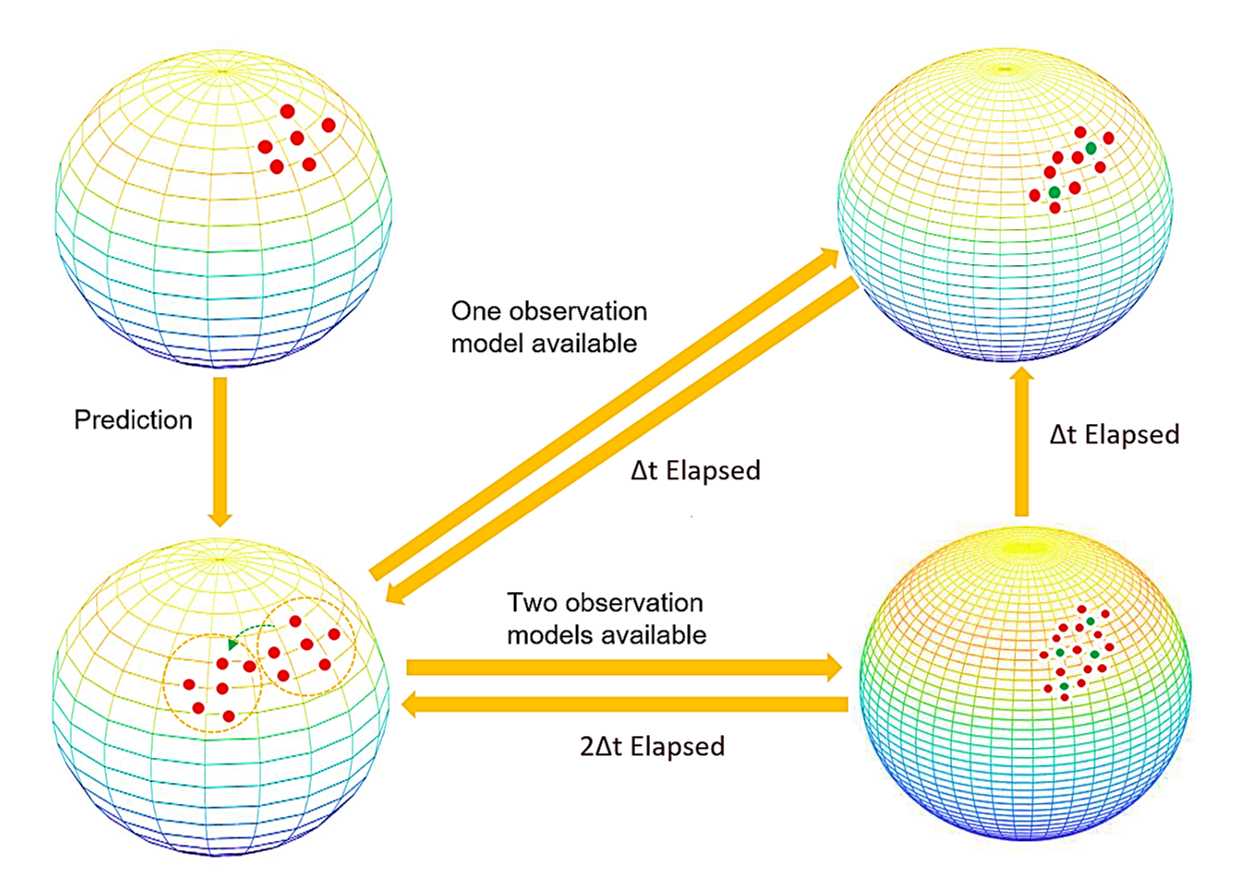}
 \caption{\centering Lifetime phases of particle filter sampling on a spherical surface.}
\label{fig:pf-state}
\end{figure}

Each particle's life cycle can be represented in four phases, as shown in Figure \ref{fig:pf-state}. Arrows between them stand for transition conditions. In our setting, the Computer Vision based orientation observation has a smaller variance compared to IMU. As aforementioned, three levels of cell resolution are employed. The top left part of Figure \ref{fig:pf-state} represents the initial state, sampled from a normal distribution centered at the measurement of timestamp $t$. The bottom left is the prediction based on the propagation of the previous particle states by extrapolation in time. When the orientation from a single pipeline, either skyline or ground plane, is available, the new samples in red are generated on finer resolution neighboring the green dots corresponding to $\mu_{f}$ in Equation \ref{eq:fusion}. Each dot is located at the center of a cell. If both skyline and ground plane pipelines are available, the highest resolution $\Omega_{3}$ is employed to generate more new samples. Each particle's timestamp of creation is kept as well. If the lapsed time exceeds a certain interval $\delta t$, the particles will be placed back at a coarse resolution, like the arrow direction from the bottom right to the top right. At a certain time point, the particles in the set have various precision. A lifetime check will be called periodically to eliminate the particles that have existed for a quite long time.

\nomenclature{$\alpha$}{Roll angle}
\nomenclature{$\beta$}{Pitch angle}
\nomenclature{$\gamma$}{Yaw angle}
\nomenclature{$m$}{Line slope}
\nomenclature{$b$}{Line intercept}
\nomenclature{$f_x, f_y$}{Focal lens distance horizontally and vertically}
\nomenclature{$c_x, c_y$}{Camera view center horizontally and vertically}
\nomenclature{$\mathbf{R}$}{Rotation matrix}
\nomenclature{$\mathbf{K}$}{Camera intrinsic matrix}
\nomenclature{$\mathbf{P}_i$}{Pixel position}
\nomenclature{$\mathbf{n}$}{Normal Vector}
\nomenclature{$\Omega_{l}$}{Manifold Surface under resolution level $l$}
\nomenclature{$s_{*}$}{Sampled particle}
\nomenclature{$\mathcal{N}(\mu_{f}, \delta_{f})$}{Gaussian distribution with mean and standard deviation}\
\nomenclature{$\mathbf{SE}(3)$}{Special Euclidean Group in three dimensions}
\nomenclature{$\mathbf{SO}(3)$}{Special Orthogonal Group in three dimensions }

\section{Experiments and Results}
\label{sec:exp}

In practice, the video is scaled down to 640$\times$480 resolution to achieve a raw frame rate of 20, while the overall frame rate scales down to 12-15 after the fusion on Jetson Nano. Intrinsics of the camera are acquired following the calibration guide of \cite{zhang2000flexible}. Extrinsic calibration between low-cost IMU (BNO055) and Raspi-camera is established using an open source tool "Kalibr" \cite{rehder2016extending}, \cite{furgale2013unified}. 
\begin{figure}[!th]
\vspace{-0.4em}
\centering
\includegraphics[width=.96\textwidth, scale =1.0]{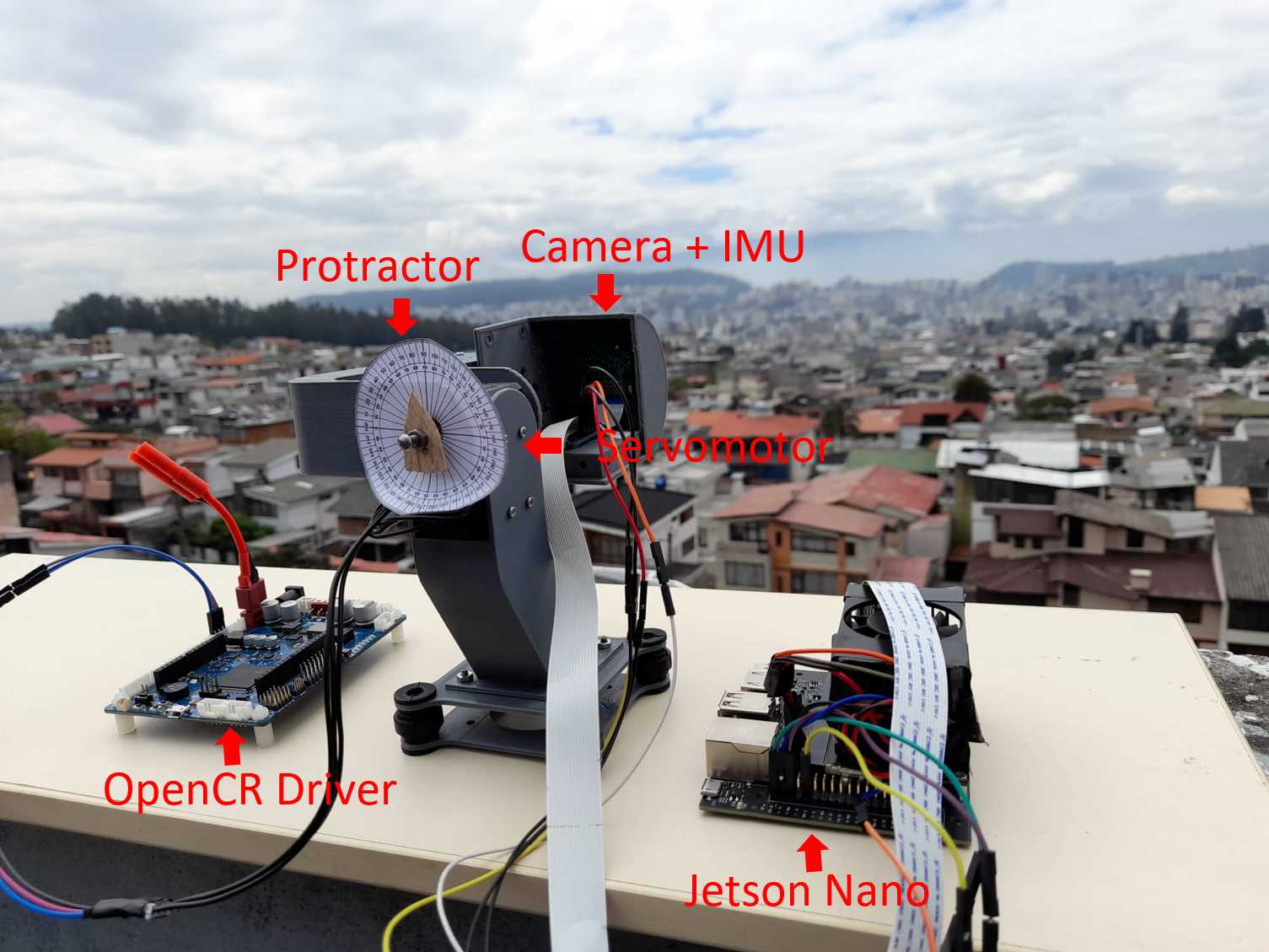}
\caption{Simulation test setup on top of the building.}
\label{fig:sim-setup}
\end{figure}
Figure \ref{fig:sim-setup} shows the simulation setup on top of the building in the landscape. There are three parts, a motor driver board, Jetson Nano, and sensors. All 3D-printed cases have enhanced connections, taking aerodynamics into account for flying efficiency. The camera on the pose estimation system is placed forward facing the landscape. Then the system cases along with the sensors are attached to a pole end (not in the view of Figure \ref{fig:sim-setup}). The other end of the pole is controlled manually to simulate random rotation. Here, the ground truth roll and pitch angles are read from the servo motors, and the protractor in the figure is only adopted for verification of the test. In the demo test, we use the fused estimation from our particle filter to steer the motors. Closed-loop PID controller is leveraged for actuation. All the following test sequences were recorded from a static position at the start. Furthermore, it is guaranteed the ground plane should be orthogonal to the gravitational vector at the start. This configuration remains unchanged for the real UAV test on the drones. 

Open-source datasets for orientation estimation research mostly overlapped with SLAM research, and the SLAM datasets were often captured indoors or within urban regions, rarely including unpopulated areas, viewed from the top. We thus recorded sequences by using the aforementioned setup in a real mountain landscape on a tall building roof nearby, which should be quite similar to the camera view on the airplane. During recording. Each pose configuration of the system is kept on par with angle readings from motors containing hall sensors as ground truth. For comparisons, the SOTA visual-inertial frameworks "ORBSLAM3" \cite{ORBSLAM3_TRO}, "R-VIO" \cite{huai2022robocentric}, and "DM-VIO" \cite{qin2018vins} were selected at first, but we found these algorithms are dedicated to 6-DoF visual odometry and are all relying on feature points extracted from the images, which are not suitable for the challenging feature-less scenes of our datasets, so we compare our fusion algorithm against the IMU filtered by quaternion-based Madgwick, CV only pipelines, like skyline or ground respectively.
\begin{table}[!thbp]
\centering
\caption{Average RMSE of roll and pitch angles (in radians). The frame-wise error bigger than radians 0.3 occurring over the half sequence length is considered a failure. The sequence test period is indicated in the brackets next to the test type.}
\label{cmp-result}
\begin{adjustbox}{width=\linewidth}
\begin{tabular}{c c c c c c c}
\hline
  &Sequence& IMU (Madgwick Filter) & Skyline only & Ground plane only & Fusion\\
 \hline
\multirow{3}{*}{Roll test (125s)} & test01 &0.0121 & 0.0182 & 0.0344 & \bf{0.0090}\\
& test02 & 0.0147 & 0.0236 & 0.0457 & \bf{0.0126}\\
& test03 & 0.0162 & 0.0325 & 0.0593 & \bf{0.0152}\\
\hline
\hline
\multirow{3}{*}{Pitch test (127s)} & test01 & 0.0147 & 0.0196 & 0.0325 & \bf{0.0118}\\
& test02 & 0.0174 & 0.0214 & 0.0291 & \bf{0.0139}\\
& test03 & 0.0208 & 0.0241 & \rule{0.6cm}{1.5pt}
 & \bf{0.0165}\\
\hline
\hline
\multirow{3}{*}{Mixed test (960s)} & test01 & 0.386 & 0.0713 & 0.0674 & \bf{0.0451}\\
& test02 & 0.415 & 0.0651 &  \rule{0.6cm}{1.5pt} & \bf{0.0584}\\
& test03 & 0.491 & 0.0742 & \rule{0.6cm}{1.5pt} & \bf{0.0617}\\
\hline
\end{tabular}
\end{adjustbox}
\end{table}

Here, the RMSE results of my model are compared to other popular SLAM baseline models, which are based on visual-inertial odometry.
\begin{table}[!thbp]
\vspace{-0.6em}
\centering
\caption{Comparison results of our method and baseline method of average RMSE of roll and pitch angles (in radians).}
\label{baseline-results}
\begin{adjustbox}{width=0.92\linewidth}
\begin{tabular}{c c c c c c c}
\hline
  &Sequence&ORB3\cite{ORBSLAM3_TRO}&R-VIO\cite{huai2022robocentric}&DM-VIO\cite{qin2018vins}&Ours\\
 \hline
\multirow{3}{*}{Roll test (125s)} & test01 &0.01011 & 0.02071 & 0.00942 & \bf{0.00862}\\
& test02 & 0.00994 & 0.03904 & \bf{0.00300} & 0.00824\\
& test03 & 0.01309 & 0.06566 & \bf{0.00324} & 0.00945\\
\hline
\hline
\multirow{3}{*}{Pitch test (127s)} & test01 & \bf{0.01207} & 0.02828 & 0.04666 & 0.01657\\
& test02 & 0.04254 & \rule{0.6cm}{1.5pt} & \rule{0.6cm}{1.5pt} & \bf{0.03446}\\
& test03 & 0.00914 & \rule{0.6cm}{1.5pt} & \rule{0.6cm}{1.5pt} & \bf{0.00875}\\
\hline
\hline
\multirow{3}{*}{Mixed test (9200s)} & test01 & \bf{0.01845} & 0.03975 & \rule{0.6cm}{1.5pt} & 0.01972\\
& test02 & 0.01828 & 0.05635 & \rule{0.6cm}{1.5pt} & \bf{0.01804}\\
& test03 & 0.01617 & 0.03982 & \rule{0.6cm}{1.5pt} & \bf{0.01615}\\
\hline
\end{tabular}
\end{adjustbox}
\vspace{-1.2em}
\end{table}
Regarding the roll-only test, the proposed method (Ours) generally performs better or on par with the other methods. It shows lower error values compared to ORB3 and R-VIO and is comparable to or slightly better than DM-VIO. As for pitch only test, the proposed method again shows competitive performance. In some cases, it outperforms other methods, especially in the test02 sequence where R-VIO does not provide results (indicated by dashed lines). In the mixed test, the proposed method maintains its competitive margins over the baseline methods, showing lower or comparable error values compared to the other methods. It performs consistently across different sequences. It should be noted that dashed lines indicate that R-VIO failed to produce results in several test sequences, particularly in the pitch and mixed tests. This indicates robustness issues with the R-VIO method in such challenging scenarios. In general, the proposed method demonstrates consistent performance across all test sequences, often showing lower error rates compared to the other methods. This suggests a more robust and reliable performance. Although DM-VIO and ORB3 also show good performance, the proposed method often achieves the lowest error values, indicating that it might be the best-performing method overall in this comparison.
This summary highlights the strengths of the proposed method in terms of performance consistency and robustness across various test conditions compared to existing methods.

We can conjecture from Table \ref{cmp-result} that our fusion approach consistently outperforms the other baseline approaches without fusion by a considerable margin on all sequences. The sequences cover three movement patterns: sequences with pure roll, pure pitch movement, and random rotation on both axes, and each case is implemented at different angular speeds, ordered from three levels, 3, 9, and 15 degrees per second. The good performance with the lowest RMSE in most tests can be attributed to our good assumptions of the environment, skyline, and ground plane in the wild. The IMU results in the table are from the 6-DoF Madgwick quaternion filter \cite{madgwick}, without the use of a magnetometer. This is because in our case, the mountain region with active volcanoes is affected by the disturbances from the earth's magnetism field change. The filtered IMU results are prone to drift, as presented in the mixed test of 16 minutes, and the errors are nearly one order of magnitude bigger than roll and pitch test sequences. Either the use of skyline or the ground plane as a tracking cue can guarantee the error remaining at a lower level compared to the Madgwick filter over IMU measurements, but in some fast rotation cases, the ground planes are partially or not present in the image, which may fail. All of the results in the table justify the merits of using an adaptive particle filter over the manifold, improving the robustness, sensor redundancy, and accuracy.

\begin{figure}[!thbp]
\centering
\includegraphics[width=1.0\textwidth, scale=0.5]{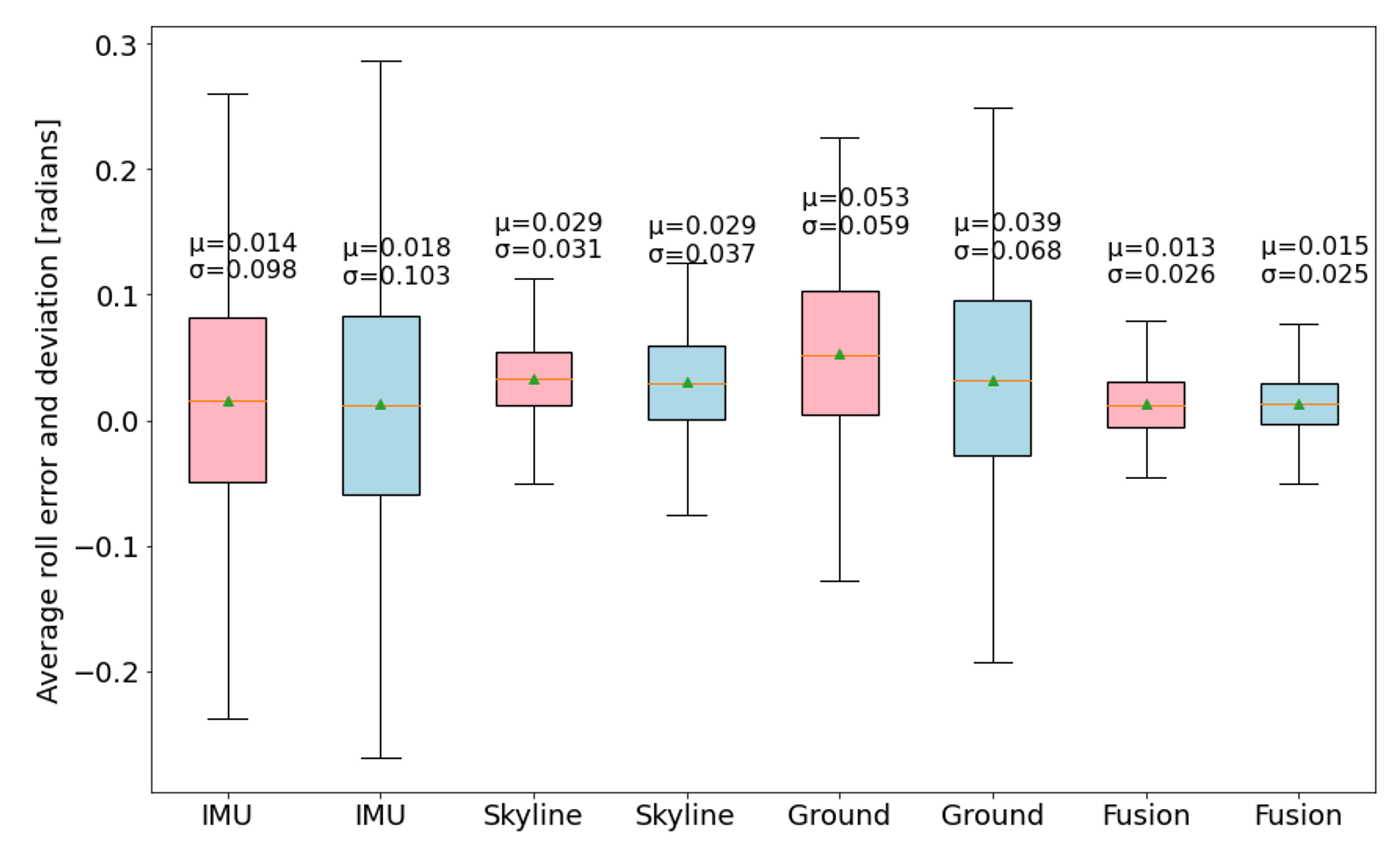}
\vspace{-1.5em}
 \caption{The Green arrow is a mean error, the orange line is a median error, and the box bounds represent the min/max errors. Roll and pitch results are in pink and blue respectively.}
\label{fig:bar-box}
\vspace{-1.3em}
\end{figure}

Figure \ref{fig:bar-box} further validates the consistency of our method. The test is repeated 10 times per sequence, and then an average error of all trials is taken over the mean error of the whole sequence. The slowest angular speed of the sequence (test01) for pure roll or pitch in Table \ref{cmp-result} is employed. The variances of our fusion results are always the smallest compared to the results of the single sensor modality.

\section{Conclusion}

In this chapter, we proposed a stand-alone gimbal system for pose estimation from image frames, leveraging natural geometric primitives such as skylines and ground plane approximations. By using two lines and their normal vectors, we derived the current frame’s rotation relative to a reference frame. A particle filter with adaptive resolution-based sampling was introduced to fuse orientation estimates from both computer vision (CV) and IMU pipelines, adapting to different phases of particle lifetime. The system was implemented on a 3D-printed gimbal platform and tested in real-time on a Jetson Nano. Comparisons with IMU-only solutions demonstrated superior accuracy and robustness against drift and disturbances.

While our approach benefits from a simplified geometric assumption, it faces challenges in extreme weather conditions, abrupt illumination changes, and scenarios where the skyline is not visible. To improve robustness in diverse environments, a hybrid system incorporating feature points alongside the skyline could be explored. Currently, the skyline is approximated as a straight line for real-time processing, but complex terrains with curved mountains challenge this assumption. Image-level matching techniques, such as iterative closest point (ICP), could help refine alignment. Additionally, integrating a fisheye camera or multiple cameras from different viewpoints would enhance the system’s robustness and adaptability across varied scenes.
\chapter{Point Cloud Registration Using 2D Surfel-Based Equivariance Constraint}                                
\label{ch:cp2}

\begin{center}
    \textbf{\large Abstract}
\end{center}

Point cloud registration is a critical task in 3D reconstruction. Traditional methods extract and match 2D or 3D feature points, but these matches are often sparse and unreliable due to a low inlier-to-outlier ratio, making registration prone to failure. Recent point cloud-based registration methods, both learning-based and non-learning-based, focus primarily on spatial alignment while often ignoring point orientations and uncertainties. Additionally, these methods require extensive transformation enhancements in training data, making these methods sensitive to noise and large rotations. To address these limitations, we propose a surfel-based point cloud registration framework \footnote{The main content of this chapter was published as a peer-reviewed conference proceeding paper \cite{kang20252d} (Kang, Xueyang et al. ``SurfelReloc: Surfel-based 3D Registration with Equivariant Features.'' CVPR 2024 workshop EquiVision proceedings. 2024).\\
\textbf{Author Contributions:} \\
\vspace{-1.2em}
\begin{itemize}
    \item \textbf{Xueyang Kang}: Idea Design, Methodology, Software, Experiment Validation, Formal Analysis, Data Curation, Writing, Review, and Editing.
    \item Hang Zhao: Data Curation and Review.
    \item Zhaoliang Luan: Review.
    \item Patrick Vandewalle: Review, and Supervision.
    \item Kourosh Khoshelham: Review, and Supervision.     
\end{itemize}
} that leverages $\mathbf{SE(3)}$-equivariant features for robust alignment. Our method initializes surfels from clustered superpixels aligned with depth maps using camera parameters or directly from LiDAR scans, learning the equivariant position and orientation representations through $\mathbf{SE(3)}$-equivariant convolutions. The model integrates an equivariant convolutional encoder, a cross-attention module for similarity estimation, a fully connected decoder, and a non-linear Huber loss for robust optimization. Experimental results on indoor and outdoor datasets demonstrate the effectiveness and robustness of our approach, outperforming state-of-the-art methods in real-world point cloud registration.

\section{Introduction}
\label{sec:intro}

Registration is a critical challenge in 3D reconstruction, shape alignment, VR/AR, and various applications, as it involves estimating relative transformations between source and target frames. Traditional 2D registration relies on finding correspondences between distinctive feature points extracted from image frames. However, repetitive or ambiguous textures, along with high outlier ratios in correspondence candidates, often make these methods unreliable. Furthermore, 2D view registration frequently results in under-constrained solutions for full $\mathbf{SE}(3)$ pose estimation due to these limitations.  

Recent progress in monocular foundation models, such as DepthAnything \cite{depth_anything_v1}, DepthAnythingV2 \cite{depth_anything_v2}, and LOTUS \cite{he2024lotus}, predict consistent depth and normal maps from color images. This enables unprojecting pixels into 3D point clouds in the camera frame, and this approach effectively lifts 2D image registration into 3D point cloud registration to avoid correspondence ambiguity of 2D views. However, these predicted depth maps inherently have uncertainty introduced by view projection geometry, making the point cloud noisy. Current point cloud registration methods, including ICP \cite{low2004linear}, DGR \cite{choy2020deep}, and PointDSC \cite{bai2021pointdsc}, neglect such uncertainties and associated point cloud normals. Additionally, while deep learning models process point clouds, they are often translation-equivariant but not rotation-equivariant, so they require extensive data augmentation during training to handle registration under large rotation, making training very inefficient.

\begin{figure}[!thbp]
\centering 
\includegraphics[width=\linewidth]{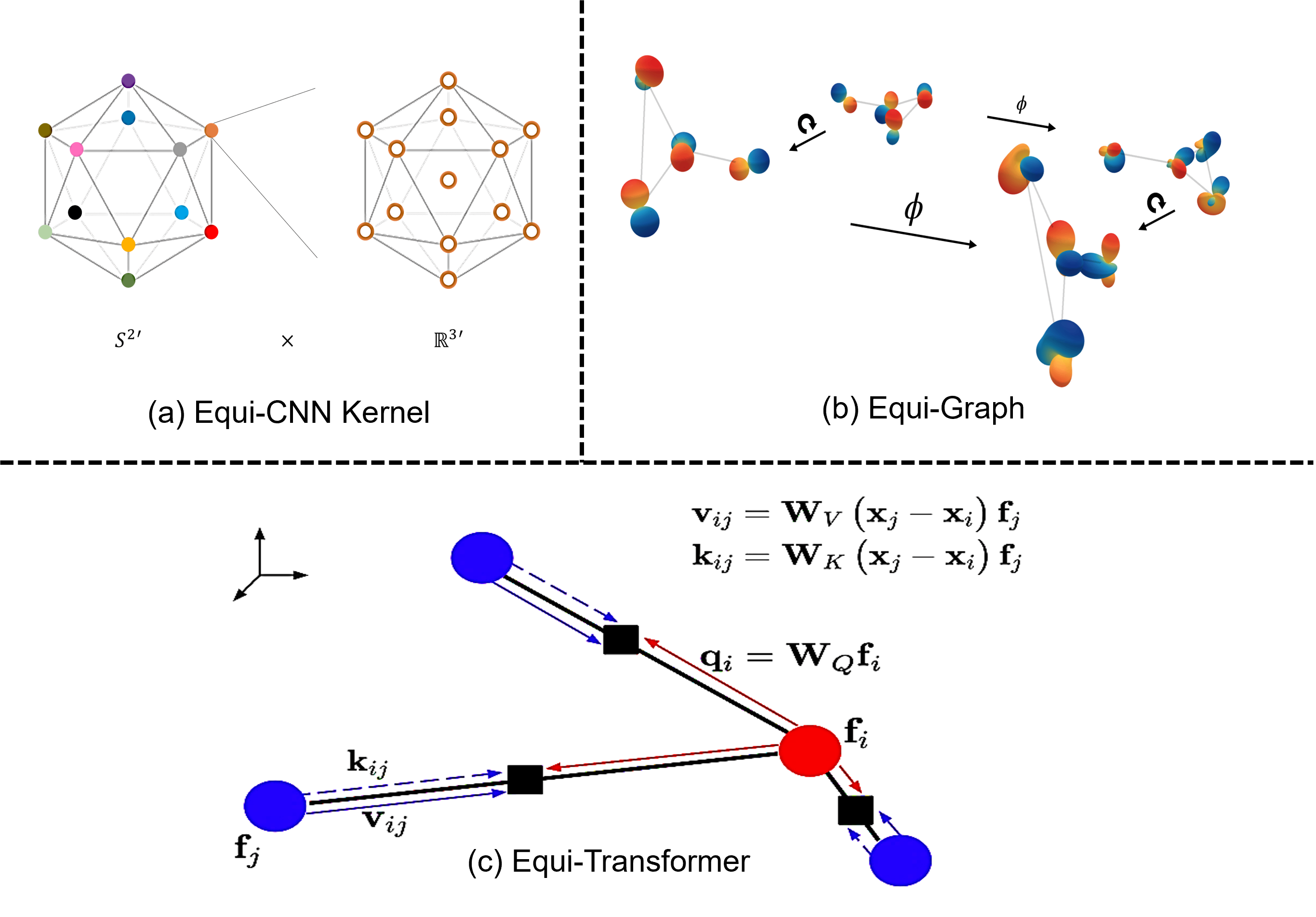}
\caption{Different Equivariant model designs. (a) is equivariant $\mathbf{SO}(3)$ kernel \cite{liu2023se} of CNN, and (b) is the equivariant graph model \cite{cesa2022program}. Lastly, (c) is the equivariant Transformer attention \cite{fuchs2020se}.}
\label{fig:equi-model-cmps}
\end{figure}

Traditional registration techniques typically use raw colored point clouds initialized from RGB-D images \cite{glocker2013real, shotton2013scene, du2020dh3d}. Rigid methods like ICP optimize point-wise match errors but are sensitive to noisy inputs and high outlier ratios, while non-rigid approaches address registration of deformable objects in dynamic environments \cite{stuckler2014multi, mitra2007dynamic, myronenko2010point} with temporal information. Learning-based methods, such as D3feat \cite{bai2020d3feat}, SpinNet \cite{ao2021spinnet}, and RoReg \cite{wang2023roreg}, aim to learn $\mathbf{SO(3)}$-equivariant features for rotation and position estimation. However, they struggle in challenging scenarios involving noisy inputs, large rotations, symmetry ambiguity existing in the input, or near-planar camera motion, often failing to resolve orientation ambiguities, in particular for transformations that are orthogonal or antiparallel. We provide a demonstration Figure \ref{fig:equi-model-cmps} to clarify the model equivariance, invariance, and its general relationship diagram, 

A model encoder function \( f \) is claimed to be equivariant if applying a transformation \( g \) to the input \( x \), followed by the function \( f \), results in the same transformation \( g' \) (an approximation for group transformation in feature space) being applied to the output. Mathematically, the equivariance can be formulated as below:

\begin{equation}
f(g(x)) = g'(f(x)).
   \label{eq:equi}
\end{equation}

In Eq. \ref{eq:equi}, the input transformation causes a corresponding transformation in the output (set of points), preserving relationships. In contrast, a function \( f \) is invariant if applying a transformation \( g \) to the input \( x \) does not change the output. Mathematically, the invariance can be stated as below: 

\begin{equation}
   f(g(x)) = f(x). 
   \label{eq:invar}
\end{equation}

Although the input undergoes the same rotation, the output remains the same, demonstrating invariance, as shown in the right side of Eq. \ref{eq:invar}.

Various equivariant model structures exist, including rotation-equivariant CNNs with steerable kernels \cite{liu2023se}, equivariant graph models with steerable features \cite{cesa2022program}, and attention-based equivariant models \cite{fuchs2021iterative} where queries, keys, and values are mapped from edge features. Each equi-model structure has its own merits, such equi-CNN is efficient, while equi-GNN and equi-Transformer have a large receptive field to learn global features.

To address these aforementioned registration challenges, we introduce a surfel-based pose regression model. Surfels, small oriented disks initialized from reprojected view point clouds or LiDAR data, serve as 2D Gaussian primitives encoding point position, normal, and uncertainty. Compared to point clouds, surfels offer superior robustness by leveraging data uncertainties, as demonstrated by recent advancements in Gaussian-based methods such as Gaussian Splatting \cite{kerbl20233d} for 3D rendering tasks. Building on previous work \cite{surfel_geometry, behley2018efficient, surfel_primitives}, we propose a novel $\mathbf{SE(3)}$-equivariant deep learning model for surfel-based registration.

Our method integrates a pipeline for initializing point clouds from 2D depth maps using camera parameters or directly from LiDAR scans \cite{izadi2011kinectfusion, huai2015real} and the depth can be predicted by a 2D monocular foundation model given a single image. The model employs a rotation-equivariant convolutional encoder based on E2PN \cite{chen2021equivariant} to extract $\mathbf{SE(3)}$-equivariant features, explicitly capturing both position and orientation. Pairwise equivariant features undergo cross-attention to compute similarity maps for establishing correspondences. These attention-enhanced features are concatenated and passed through fully connected layers to predict the relative transformation. Additionally, we introduce a differentiable $\mathbf{SE(3)}$ Huber loss function to supervise the soft correspondences, inspired by the node-wise supervision approach of PointDSC \cite{bai2021pointdsc}.

Experimental results on real-world indoor and outdoor datasets demonstrate the effectiveness and robustness of our model, particularly in large-scale scenarios. Compared to state-of-the-art methods, our approach addresses key limitations in traditional and learning-based registration methods, offering improved accuracy and scalability.

In summary, the main contribution of this work can be briefly outlined as follows:
\begin{itemize}
    \item A surfel-based initialization pipeline from depth maps or LiDAR scans for pose regression tasks.
    \item A $\mathbf{SE(3)}$-equivariant deep learning model leveraging surfel features for robust and efficient registration.
    \item A differentiable $\mathbf{SE(3)}$ Huber loss function for supervision using soft correspondences.
\end{itemize}

\section{Related Work}
\label{sec:related-work}

$\textbf{Applications of 3D Surfels.}$ SurfelMeshing \cite{schops2019surfelmeshing} has been applied to 3D mapping in indoor scenes, while other approaches have used surfels for large-scale outdoor mapping \cite{yuan2022efficient, wang2019real, behley2018efficient}. Surfels model the probability of a point by considering sensor accuracy and observation uncertainty, such as errors in pixel location due to camera projection. This data structure of surfels explicitly encodes $\mathbf{SE(3)}$ information, unlike point cloud which relies on neighboring points to derive rotation information implicitly.

\noindent$\textbf{3D Registration.}$ Registration techniques are widely used in shape alignment (\cite{choy2020deep, zhou2016fast}) and deformable target scanning \cite{bhatnagar2020loopreg}. Traditional non-learning methods like vanilla Iterative Closest Point (ICP), kiss-ICP \cite{low2004linear}, and point-to-plane ICP \cite{park2003accurate} struggled with nonlinear optimization errors. In contrast, state-of-the-art deep learning models like PointDSC \cite{bai2021pointdsc} and Deep Global Registration (DGR) \cite{choy2020deep} explore correspondences in high-dimensional feature spaces. Max-Clique \cite{zhang20233d} and GeoTransformer \cite{qin2023geotransformer} leverage the latest graph or attention learning to create the registration backbone. Additionally, research focuses on enhancing feature descriptor learning to capture neighboring geometry information, exemplified by Fully Convolutional Geometric Features \cite{choy2019fully}.

\noindent$\textbf{Equivariant Feature Representation.}$ Equivariant models have emerged as robust solutions for 3D applications such as point cloud registration and pose estimation. These models maintain $\mathbf{SE(3)}$ rotational and translational equivariance, essential for consistent alignment performance. Examples include Spherical CNNs \cite{cohen2018spherical}, group CNNs \cite{finzi2020generalizing}, and SE(3)-Transformers proposed \cite{fuchs2020se}. Unlike the state-of-the-art models that are solely translation-equivariant \cite{choy2019fully, choy2020deep}, incorporating rotation-equivariant and invariant feature characteristics can enhance learning efficiency and robustness.
D3feat \cite{bai2020d3feat}, SpinNet \cite{ao2021spinnet}, and RoReg \cite{wang2023roreg} leverage rotation-equivariant features for point cloud registration. 
Other approaches integrate equivariance learning into model structure design, such as vector neurons \cite{deng2021vector} by extending neurons from 1D scalars to 3D vectors, using steerable kernel filters \cite{cohen2018spherical, weiler2018learning, weiler20183d, sosnovik2019scale, e2cnn, jenner2022steerable}, projecting features onto spherical harmonics \cite{thomas2018tensor}, employing rotation-guided attention mechanisms \cite{hutchinson2020lietransformer, fuchs2020se, fuchs2021iterative}, and incorporating equivariant high-dimensional features into graph models for message propagation \cite{kang2024equi,du2022se, satorras2021en}.
\section{Method}

For perspective image input, we initialize surfels from the unprojected depth map and its associated normal map, where the normals are derived from the depth gradients. This process also involves determining the surfel uncertainty 1D radius based on a camera perspective projection model, accounting for both inverse depth uncertainty and camera view angle to the image center. The whole initialization process serves as a pre-processing step. 

For LiDAR point clouds, surfels can be created from the neighboring non-coplanar triplet points for normal vector estimation, and 1D uncertainty can be derived proportionally from the point density.

To ensure registration efficiency, surfels from source and target frames are downsampled before feeding into the neural network model. The model architecture, as illustrated in \ref{fig:network-structure}, consists of three components: an $\mathbf{SE(3)}$ equivariant convolution kernel-based encoder, a cross-attention, and a decoder for predicting relative transformation.

\subsection{Surfel Initialization}
The general goal of sampling is to find an optimal surfel representation of the geometry with minimal redundancy. Most sampling methods perform object discretization based on geometry parameters of the surface, such as curvature or silhouettes. This object-space discretization often results in either excessive or insufficient primitives for rendering.  

\begin{figure}[!th]
    \centering
    \includegraphics[width=0.49\textwidth]{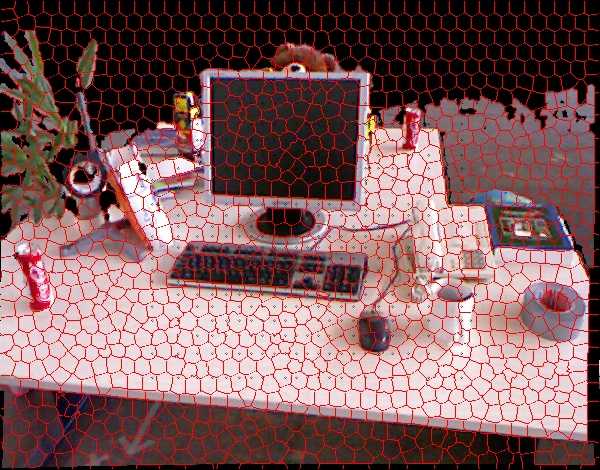}
    \hfil
    \includegraphics[width=0.49\textwidth]{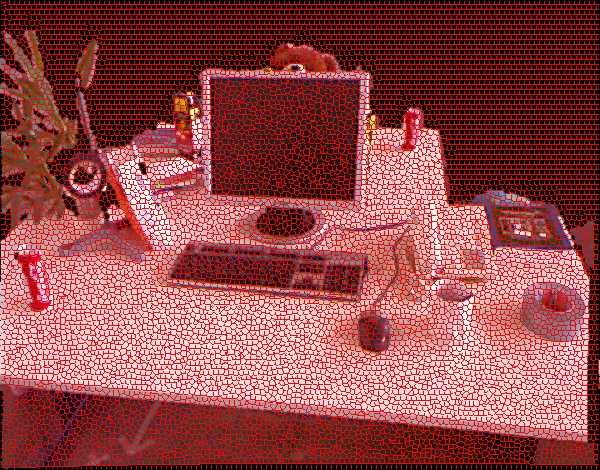}
    
    \caption{(Left) Color image decomposed into superpixels of large size. 
    (Right) Color image decomposed into superpixels of small size.}
    
    \label{fig:big-superpixel}
\end{figure}

While traditional surfel-based methods map the environment by assigning per-pixel surfels, our approach extracts surfels based on superpixels from RGB images using the SLIC algorithm \cite{slic}. This method generates 3D primitives for lower-resolution surface reconstruction compared to raw input, enabling faster processing with reduced memory consumption and noise. At the same time, it preserves critical information, such as texture and depth discontinuities. The pixels are clustered by intensity, the main parameter being the minimum number of pixels $k$ per superpixel to ensure approximately equal sizes (Figure \ref{fig:big-superpixel}). This superpixel-based approach minimizes memory overload for large-scale scenarios while reducing outliers and noise from low-quality depth maps.  

Each surfel is initialized from a single superpixel. The position $\mathbf{p^{}_i}$, normal $\mathbf{n^{}_i}$, and color $\mathbf{c^{}_i}$ are computed as the average values of the pixels within the superpixel:  

\begin{equation}\label{eq:surfel_pos}
\mathbf{p^{}_i} = \frac{1}{k} \sum_{k} \mathbf{p^{s^{}_i}_k}, \quad \mathbf{n^{}_i} = \frac{1}{k} \sum_{k} \mathbf{n^{s^{}_i}_k}, \quad \mathbf{c^{}_i} = \frac{1}{k} \sum_{k} \mathbf{c^{s^{}_i}_k},
\end{equation}

where $\mathbf{p^{s^{}_i}_k}$, $\mathbf{n^{s^{}_i}_k}$, and $\mathbf{c^{s^{}_i}_k}$ represent the position, normal, and color values of the $k^{th}$ pixel in superpixel $s^{}_i$. The surfel radius $r^{}_i$, which represents uncertainty, is calculated as the dot product between the normalized pixel direction $\mathbf{v}_i$ and the camera principal axis facing forward:  

\begin{equation}\label{eq:surfel_rad}
    \rho^{}_i = \left| \tan\left(\mathbf{v^{}_i} \cdot \mathbf{z}\right) \right|
\end{equation}

\begin{figure}[!th]
    \centering
    \includegraphics[width=0.32\textwidth]{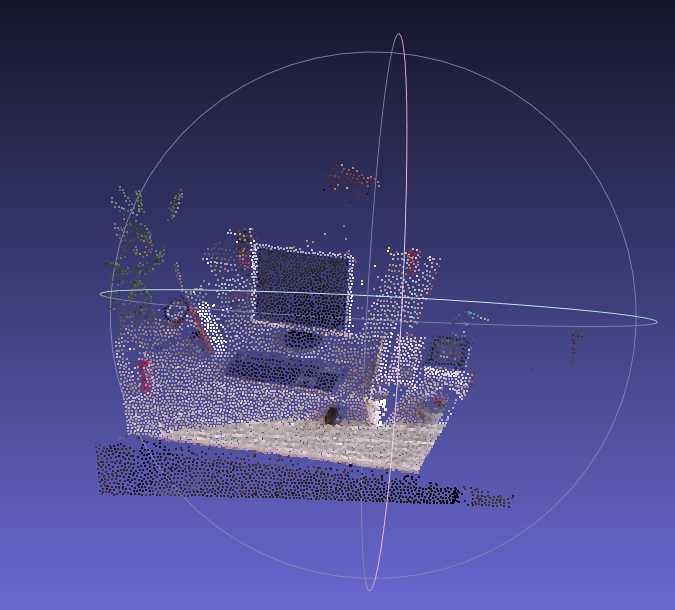}
    \includegraphics[width=0.32\textwidth]{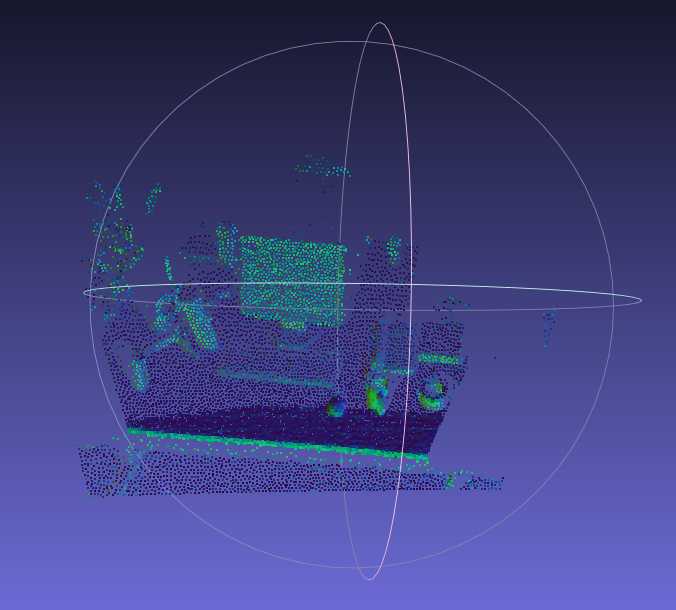}
    \includegraphics[width=0.32\textwidth]{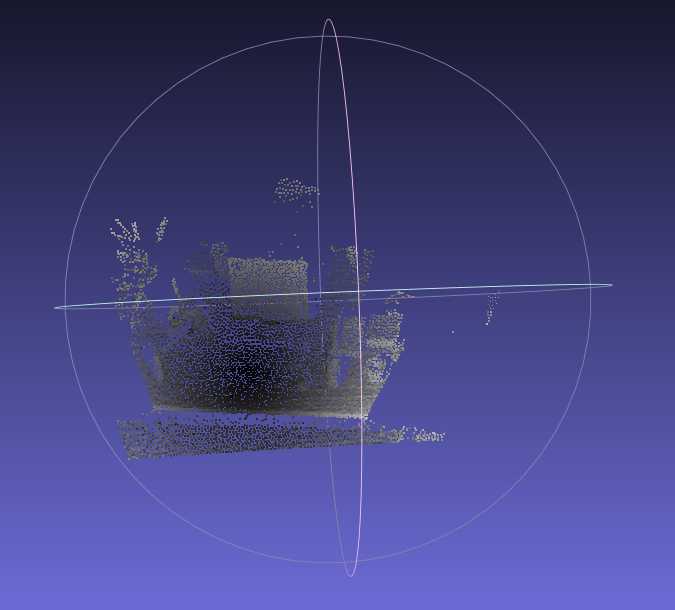}
    
    \caption{(Left) Color points cast from the depth map aligned with the color pixels. 
    (Middle) Point normals cast from the normals map aligned with the color pixels. 
    (Right) Point uncertainties are cast from the pixel-wise uncertainties aligned with the color pixels.}
    \label{fig:surfel_map_properties}
\end{figure}

This approach ensures smaller uncertainty at the center and larger uncertainty near the edges, where the viewing angle is greater. The resulting surfels have corresponding radii, as illustrated in Figure \ref{fig:surfel_map_properties}, where brighter points indicate larger radii and darker points indicate smaller radii.  

To visualize the results, we separated the properties into three different point clouds due to the limitations of available open-source libraries. These visualizations are shown in Figure \ref{fig:surfel_map_properties}.  

This way, the uncertainty is smaller in the center and larger on the sides where the angle given by the point of view is also large. The resulting surfels will have corresponding radii as shown in Figure \ref{fig:surfel_map_properties}, where brighter points correspond to large radii as opposed to dark-colored points having small radius values.

To visualize the 3D surfel results, due to the limitations in the open source libraries ready to use, we separated the different properties into three different point clouds visible in Figure \ref{fig:surfel_map_properties}.

In particular, each surfel in the source frame, indexed by $i$, consists of three main components: the 3D position $\mathbf{x}_i \in \mathbb{R}^3$, the normal vector $\mathbf{n}_i \in \mathbb{R}^3$, and a scalar radius $\epsilon_i \in \mathbb{R}$. The surfel $\mathbf{y}_j$ is in the target frame. Here, we mainly focus on clarifying the complex process of surfels created from a perspective view. The surfel center position is determined by the depth map and the camera intrinsic matrix through unprojection ($\pi$). To calculate the normal map from the depth gradient, a CUDA-based Sobel operator is applied to the depth map in parallel. Subsequently, the resulting normal map is converted into 3D normals using the camera intrinsic matrix, providing information about the orientation of the disk associated with the surfel center. The surfel radius $\epsilon_i$ is then derived by the following expression:
\begin{equation}
   \epsilon_{i} = C \frac{e^{-\hat{\rho}}}{1+e^{-\tan(\theta)}},   
\label{eq:surfel_radius}
\end{equation}
where $C$ is the normalization factor, and $\theta$ represents the view angle between the camera principal axis $\vec{o}$ and the ray $\vec{r}$ emitted from the camera center through the pixel location, as expressed in the following equation,
\begin{equation}
    \theta = \arccos\left(\frac{\vec{r} \cdot \vec{o}}{\lVert \vec{r} \rVert \lVert \vec{o} \rVert}\right).
    \label{eq:angle}
\end{equation}

The value of $\rho$ represents the inverse of the depth, which is then truncated to $\hat{\rho}$ within the inverse depth range $(\rho_{min},\rho_{max})$ of the sensor,
\begin{equation}
    \hat{\rho} = \min\left(\max\left(\rho, \rho_{min}^{}\right), \rho_{max}\right).
\end{equation}

\subsection{Network Structure}
\begin{figure}[!thbp]
\centering 
\includegraphics[width=0.96\linewidth]{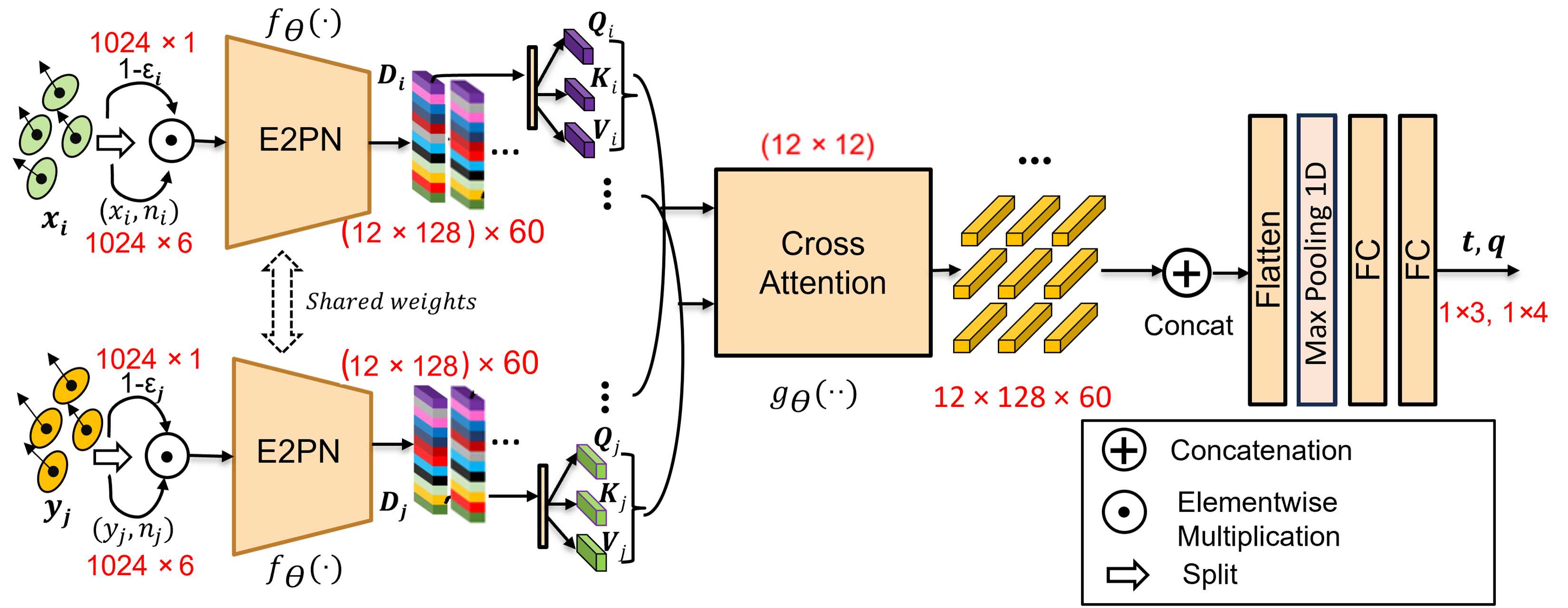}
\caption{The network structure features a shared encoder for surfels (6 dimensions plus uncertainty radius) from both source and target frames in $\mathbf{SE(3)}$ space. This encoder maps 1024 surfels with 6 dimensions (position and normal), weighted by confidence value $(1-\epsilon(\cdot))$, into 12 feature descriptors in 128-dim with 60 group rotation orders. Then each descriptor undergoes linear embedding to produce triplet token embeddings $\mathbf{Q}$, $\mathbf{K}$, and $\mathbf{V}$. Cross-attention $g_{\theta}(\cdot)$ is applied to feature descriptors from source and target frames in 12-channel dimension. The resulting tokens are in the shape of (12 $\times$ 128 $\times$ 60), where each token is in 128-dim $\times$ 60 order groups, and the attention map is 12 $\times$ 12, where each element token in the attention map is formulated via descriptor dot product. Then the features are flattened and processed through Fully-Connected (FC) layers, mapping features to relative position $\mathbf{t}{}^{}$ and relative quaternion $\mathbf{q}{}^{}$ rotation. A close-up of the E2PN module is provided below.}
\label{fig:network-structure}
\end{figure}

As exhibited in Figure \ref{fig:network-structure}, given initialized surfels of source frame $\mathbf{s}_{i}^{} \in {\mathbf{s}_{1}^{},...,\mathbf{s}_{N}^{}}$ and target frame $\mathbf{s}_{j}^{} \in {\mathbf{s}_{1}^{},...,\mathbf{s}_{N}^{}}$, all surfels are encoded by the same encoder $f_{\theta_{}}(\cdot)$. Notably, the position and normal vectors $\mathbf{n}_{(\cdot)}$, $\mathbf{p}_{(\cdot)}$ of each surfel are weighted by a factor of $(1-\| \epsilon(\cdot) \|)$ to reduce the influence of highly uncertain surfels. The encoder architecture is augmented on E2PN \cite{chen2021equivariant}, with doubled feature dimensions compared to the original point cloud input. Equivariance is maintained through a symmetric conv-kernel $\mathcal{\kappa}$ arranged in an icosahedral shape solid.

In the following convention, the symbol $'$ next to a symbol indicates discretization operation. E2PN encoder features are aligned in the spherical space $\mathbf{S}^{2'}\times\mathcal{R}^3$, where coordinates of each feature vertex in $\mathcal{R}^3$, associated with the 128-dimension feature descriptor, and $\mathbf{S}^{2'}$ signifies the discretized sphere surface. This discretized feature representation is determined by $\mathbf{SO}(3)'/\mathbf{SO}(2)'$, where $\mathbf{SO}(2)'$ is a subgroup of $\mathbf{SO}(3)$. The quotient space is defined as a group of rotations $\mathbf{R}_{i/j}$ with the same endpoint, such as the sphere's north pole after rotation. This discretization of $\mathbf{SO}(3)$ facilitates more efficient and accelerated learning. The $\mathbf{SE}(3)$ feature is constructed by extending the $\mathbf{SO}(3)$ rotation feature, incorporating translation through the concatenation of point coordinates. \\

\begin{figure}[!thbp]   
\vspace{-0.2em}
  \includegraphics[width=\linewidth]{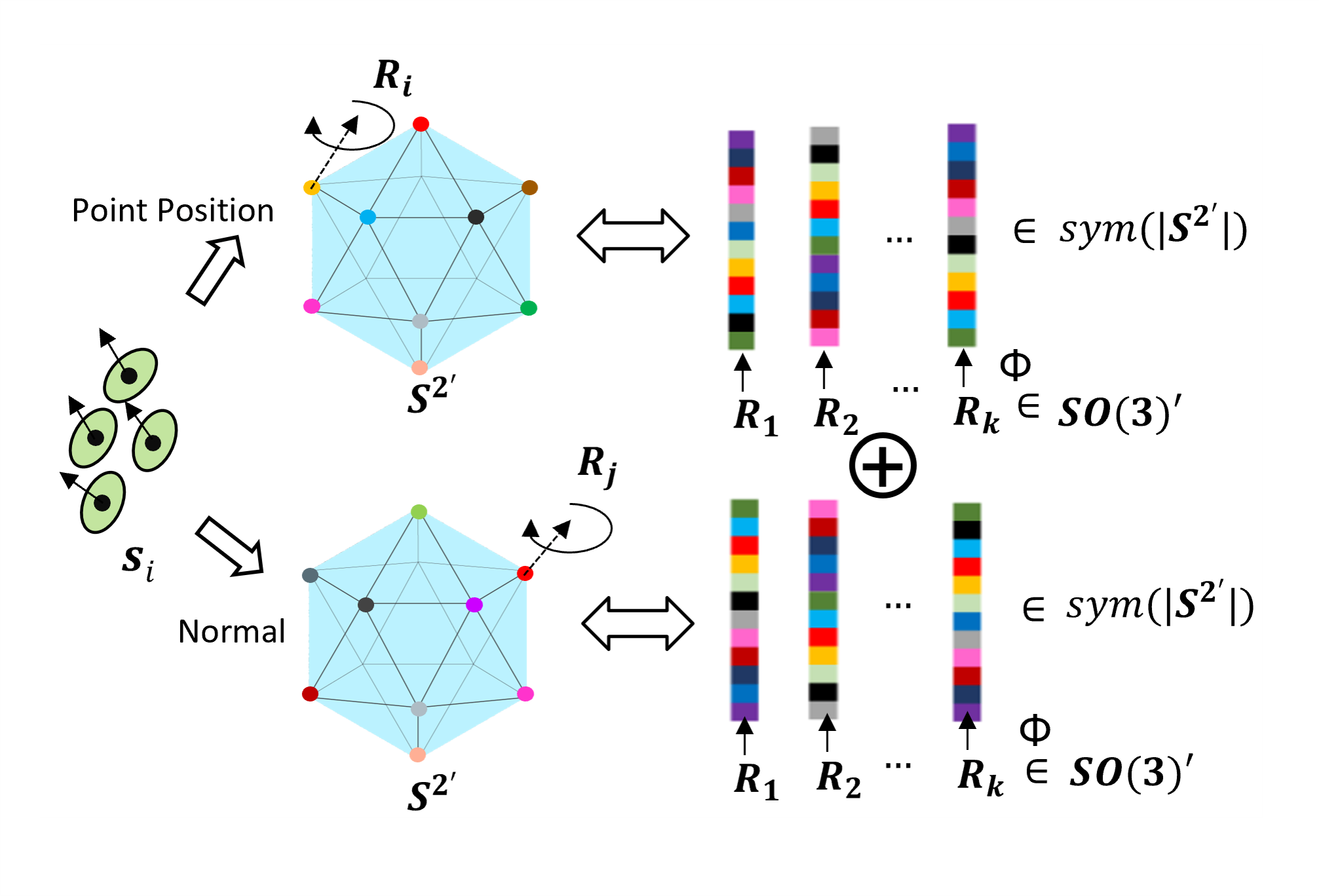}
    \caption{Recovering discreteized $\mathbf{SO(3)'}$ from the quotient feature $S^{2'}$ by permutation order.}
    \vspace{-0.0em}
    \label{fig:kernel}
\vspace{-1.0em}
\end{figure}

The surfel undergoes convolution with two distinct symmetric kernels, $\kappa_1$ and $\kappa_2$, as shown in Figure \ref{fig:kernel}, relating to point position and normal respectively. These are concatenated post-convolution. The icosahedron comprises 60 rotations, each denoted by various permutation orders $\mathbf{R}_{(\cdot)}$. The E2PN symmetric kernel selects one rotation from the 60 options to generate the output equivariant feature. This is achieved by choosing the maximum sum of each rotation feature along the 12-channel dimension. 
The Equivariant features after E2PN encoder of source and target frame can be stated as the $\mathbf{D}_i, i\in ({1,...,12})$ and $\mathbf{D}_j, j \in ({1,...,12})$, where each has 128 dimensions. Next, we use the linear layer to project each descriptor into a triplet composed of $\mathbf{Q}_{(\cdot)}, \mathbf{K}_{(\cdot)}, \mathbf{V}_{(\cdot)}$. We use the same index convention as the input in the following. $i$ states the feature of the source frame, and $j$ indicates the index of the target frame. 
The cross-attention  $g_{\theta}(\cdot)$ is then applied to calculate the attention-weighted equivariant features from the pairwise frames. Different from the normal attention-based multiplication, The feature embedding vectors (denoted as $\mathbf{Q}_{i}, \mathbf{K}_{j}$ respectively, $1\leq i, j\leq N$), and each output token feature $\mathbf{\hat{V}}_i$ is calculated as formulations below,
\begin{align}
\label{eq:atten-weight}
   \alpha_{ij} &= {\frac{\exp(\mathbf{Q}_{i}^{T}\mathbf{K}_{j})}{\sum_{j=1}^{N}\exp(\mathbf{Q}_{i}^{T}\mathbf{K}_{j})}}, \\
   \mathbf{\hat{V}}_i &= \sum_{j=1}^{N}(\alpha_{ij}\mathbf{V}_{j}),
\end{align}

The transformation applied to the input point cloud is denoted by group rotation \( g'(\cdot) \). The key idea of equivariant feature learning is that the output features of the encoder are transformed accordingly to preserve $\mathbf{SE(3)}$ equivariance $f_{\theta}(g'(x)) = g'(f_{\theta}(x))$. This ensures that the features are equivariant to the input transformations. 
The equ-features after the E2PN encoder of the source and target frames are denoted as $\mathbf{D}_i, i \in ({1,...,12})$ and $\mathbf{D}_j, j \in ({1,...,12})$, where each has 128 dimensions. Next, we use the linear layer to project each descriptor into a triplet composed of $\mathbf{Q}_{(\cdot)}, \mathbf{K}_{(\cdot)}, \mathbf{V}_{(\cdot)}$ tokens. We use the same index convention throughout the paper, where $i$ denotes the index of the source frame, and $j$ refers to the target frame. The triplet tokens, derived from feature descriptors at the 12 corners of icosahedral planoids (see Figure \ref{fig:kernel}), are aggregated from neighboring surfel coordinates. This effectively fuses the input surfels into 12 distinct regions, which are then used in the subsequent cross-attention module. 
The cross-attention $g_{\theta}(\cdot)$ is then applied to calculate the attention-weighted equivariant features (attention map in $12 \times 12$) from the pairwise frames, finally to be decoded by the fully connected layers into the transformation estimation.

\subsection{Loss Function}
Inspired by the node-wise supervision in PointDSC \cite{bai2021pointdsc}, we adapt the original binary cross-entropy loss to a non-linear Huber loss. This maps the transformed point error into $\mathcal{L}_2$ norm when it is small and into $\mathcal{L}_1$ normal when the error is large. The point position from the source frame is transformed by the predicted rotation $\mathbf{R}$ and translation $\mathbf{t}$ into $\mathbf{y}_{j^*} = \mathbf{R}\hat{\mathbf{x}}_{i^*} + \mathbf{t}$. The rotation matrix $\mathbf{R}$ is derived from the predicted quaternion $\mathbf{q}$. The Huber loss is defined as below,
\begin{equation}
\label{eq:loss}
\mathcal{L}_\text{Huber} = \begin{cases}
\frac{1}{2}(\hat{\mathbf{x}}_{i^*}-\mathbf{y}_{j^*})^2, & \text{if } |\hat{\mathbf{x}}_{i^*}-\mathbf{y}_{j^*}| \leq \delta \\
\delta(|\hat{\mathbf{x}}_{i^*}-\mathbf{y}_{j^*}| - \frac{1}{2}\delta), & \text{otherwise}
\end{cases}
\end{equation}
The threshold is set to 0.6m. The correspondence index pair $(i, j)$ is established using the nearest neighboring point search. 

\nomenclature{$\epsilon_{}$}{Uncertainty radius}
\nomenclature{$\theta$}{Inner angle of dot product}\
\nomenclature{$\rho$}{Inverse of depth value}
\nomenclature{$\vec{o}$}{Principal axis}
\nomenclature{$\vec{r}$}{Pixel ray}
\nomenclature{$\mathbf{Q}$, $\mathbf{K}$, and $\mathbf{V}$}{Query, Key and Value tokens}
\nomenclature{$\mathcal{L}$}{Loss}
\nomenclature{$\mathds{1}$}{Binary indicator}
\nomenclature{$\mathbf{D}$}{Depth map}
\nomenclature{$f(\cdot)$}{Network model function}
\nomenclature{$g'(\cdot)$}{Group transform}

\section{Experiments and Results}
\label{sec2:exp}

To evaluate the model performance, we utilized two indoor scan datasets, the outdoor dataset KITTI \cite{geiger2012we} and 3DMatch \cite{zeng20163dmatch}. These datasets include RGB-D sequence frames and extrinsic poses. Our model was trained on each dataset separately for comparison fairness against other baseline models. For ARKitScenes, we selected 10 different scenes. We further selected 100 pairs of frames out from each scene sequence by choosing the depth frames close to each other temporally, to guarantee enough overlap between the source and target scans created from the depth frame images. We employ the 3D voxel-based downsampling to generate 1024 points un-projected from the depth map of each frame for surfel initialization.

\noindent\textbf{Evaluation metrics.} We use the Rotation Error (RE) and Translation Error (TE) to evaluate the accuracy of rotation and translation separately. Furthermore, we incorporate the Registration Recall (RR) and \emph{F1 score} as registration success evaluation metrics. Our model was trained on each dataset separately for fair comparison against other baseline models. 
We employed 3D voxel-based downsampling to generate 1024 points unprojected from the depth map of each frame for the surfel initialization.
\begin{equation}
\vspace{-0.8em}
    \delta = \sqrt{\frac{1}{\mathcal{N}(\Omega)}\sum_{(\mathbf{x}_{i}, \mathbf{y}_{j})\in\Omega}\mathds{1}[\|\mathbf{R} \mathbf{x}_{i} + \mathbf{t} - \mathbf{y}_{j} \|^2 < \tau]},
    \vspace{-0.2em}
    \label{eq:rmse-rr}
\end{equation}
where $\mathcal{N}(\Omega)$ represents the total number of ground truth correspondences in the set $\Omega$. The symbol $[\cdot]$ is an indicator of whether the condition is satisfied. Based on the calculation of recall rate, the $\emph{F1 score}$ is defined as $2\cdot \frac{Precision\times Recall}{Precision + Recall}$. \\
 
\noindent\textbf{Baseline models} We compare our model with popular deep learning-based models, including PointDSC \cite{bai2021pointdsc}, Deep Global Registration \cite{choy2020deep} (DGR), Maximal Clique \cite{zhang20233d} (MAC). Additionally, we choose equivariant methods, like D3Feat \cite{bai2020d3feat}, SpinNet \cite{ao2021spinnet}, RoReg \cite{wang2023roreg} and MAC+GeoTransformer \cite{zhang20233d, qin2023geotransformer} (MAC+GeoTrans) for the two datasets, For all the point feature descriptor dependent approaches, like DGR, FCGF \cite{choy2019fully} descriptor is applied for 3DMatch \cite{zeng20173dmatch}, while FPHF \cite{rusu2009fast} descriptor is used instead for KITTI \cite{geiger2012we}. We provide quantitative evaluation results in Table \ref{tab:cmps-baselines}, and qualitative comparison results including three top performance baseline models in Figure \ref{fig:cmp-3DMatch-KITTI}. All these results exhibit the superior performance of our model over baselines. 

\subsection{Baseline Comparisons}
Our proposed model demonstrates superior performance compared to the baseline models across all test scenes, as shown in Table \ref{tab:cmps-baselines}, with consistent performance superiority over other learning models on both datasets in terms of all metrics, in particular, the proposed model has a smaller rotation error (around 8-11\% reduction) compared to the second best model of each dataset respectively.  This can be attributed to the explicit orientation signal as the input for learning, along with the good equivariant feature learning through the symmetric kernel. 

We employed 3D voxel-based downsampling to generate 1024 points unprojected from the depth map of each frame for surfel initialization.

\begin{table*}[!th]
\centering
    \caption{Evaluation results of registration approaches on 3DMatch \cite{zeng20173dmatch} (left) and KITTI \cite{geiger2012we} (right).}
    \label{tab:cmps-baselines}
\large
\vspace{-0.4em}
\begin{adjustbox}{width=\linewidth}   
     \begin{tabular}{c cccc cccc}
        \toprule
        \multirow{2}{*}{Method} & \multicolumn{4}{c}{3DMatch \cite{zeng20173dmatch}} & \multicolumn{4}{c}{KITTI \cite{geiger2012we}} \\
        \cmidrule(lr){2-5}\cmidrule(lr){6-9}
       & RE($^{\circ}$) $\downarrow$ & TE(cm) $\downarrow$ & RR(\%) $\uparrow$ & F1(\%) $\uparrow$ & RE($^{\circ}$) $\downarrow$ & TE(cm) $\downarrow$ & RR(\%) $\uparrow$ & F1(\%) $\uparrow$\\ 
        \midrule
       DGR \cite{choy2020deep} $\uparrow$ & 2.40 & 7.48 &  91.30 & 89.76 & 1.45 & 14.60 &  76.62 & 73.84 \\ 
        \hline
        D3Feat \cite{bai2020d3feat} & 2.57 & 8.16 & 89.70 & 87.40 & 2.07 & 18.92 & 70.06 & 65.31 \\ 
        \hline
       RoReg \cite{wang2023roreg} $\downarrow$ & 1.84 & 6.28 & 93.70 &  91.60 & \-\- & \-\- &  \-\- &  \-\- \\ 
        \hline
        SpinNet \cite{ao2021spinnet}  & 1.93 & 6.24 & 93.74 & 92.07 & 1.08 & 10.75& 82.83 & 80.91\\ 
        \hline
       PointDSC \cite{bai2021pointdsc} & 2.06 & 6.55 & 93.28 & 89.35 & 1.63 & 12.31 & 74.41 & 70.08\\ 
        \hline
        MAC \cite{zhang20233d} & 1.89 & 6.03 & 93.72 & 91.46 & 1.42 & 8.46 & 91.37 & 89.25 \\
        \hline
        MAC+GeoTF \cite{zhang20233d, qin2023geotransformer} & 1.74 & 6.01 & 95.02 & 91.80 & 1.37 & 8.01 & 90.59 & 88.45 \\
        \hline
        Ours & $\mathbf{1.34}$ &  $\mathbf{5.72}$ & $\mathbf{95.08}$ & $\mathbf{93.32}$ & $\mathbf{1.57}$ &  $\mathbf{6.09}$ & $\mathbf{92.05}$ & $\mathbf{90.61}$ \\ 
        \bottomrule
    \end{tabular}
\end{adjustbox}    
\vspace{-0.4em}
\end{table*}

\begin{figure}[!thb]
    \centering
    \includegraphics[width=1.0\textwidth]{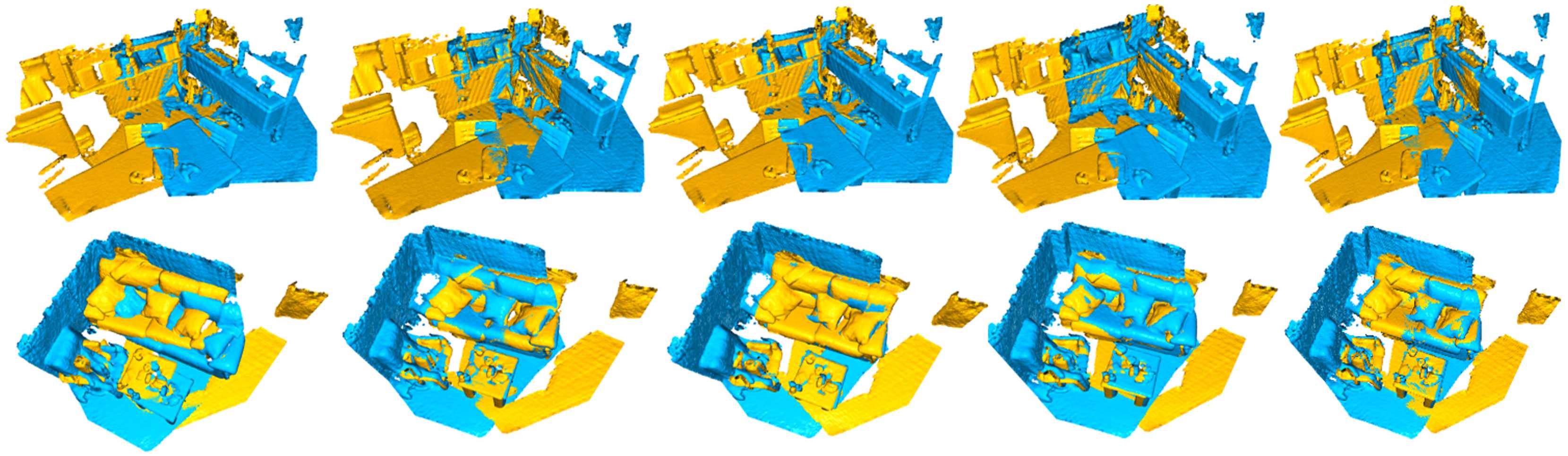}
        \hspace{-1.4em} MAX Clique \cite{zhang20233d}
        \hspace{0.0em} GeoTF \cite{qin2023geotransformer}
        \hspace{0.2em}  Equi-(GSPR\cite{kang2024equi})
         \hspace{0.6em} Ours
         \hspace{1.0em} Ground Truth\\
         \includegraphics[width=1.0\linewidth]{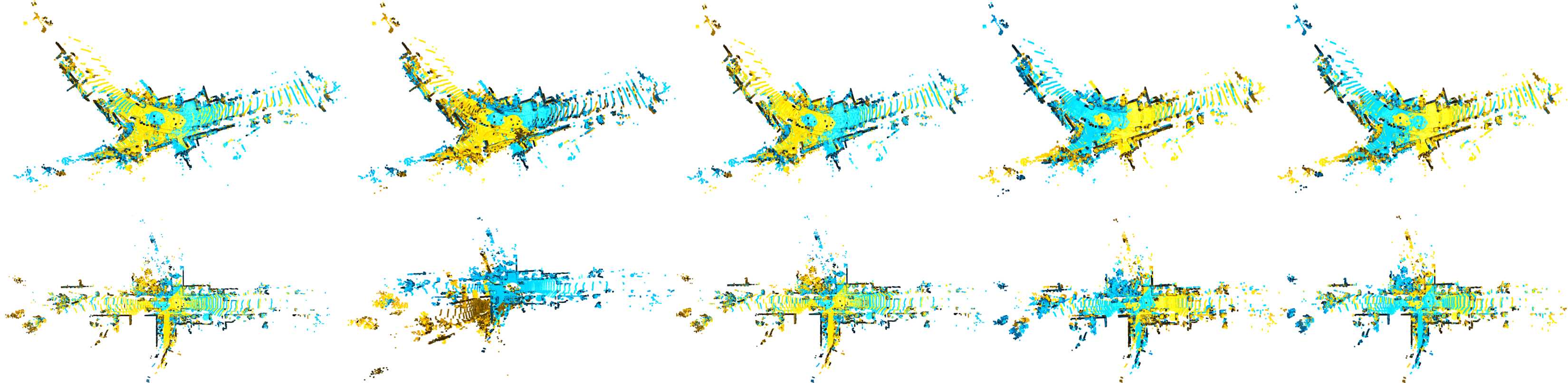}
        \hspace{1.0em} PointDSC \cite{bai2021pointdsc}
        \hspace{0.2em} MAX Clique \cite{zhang20233d}
        \hspace{0.2em}  GeoTF \cite{qin2023geotransformer}
         \hspace{1.0em} Ours
         \hspace{1.5em} Ground Truth\\
    \caption{Comparison results on KITTI \cite{geiger2012we}. For each dataset, the top three models with good performance are presented.}
    \vspace{-0.2em} 
    \label{fig:cmp-3DMatch-KITTI}
\vspace{-0.6em}
\end{figure}

In addition to the quantitative comparison results, we further provide the visual comparison results in Figure \ref{fig:cmp-3DMatch-KITTI} of the top three performance baseline models of each dataset. For better illustration purposes, we use the dense point cloud scan of source and target frames, instead of using the sparse input for training/testing to show each model's performance intuitively. While PointDSC \cite{bai2021pointdsc}, DGR \cite{choy2020deep} and RoReg \cite{wang2023roreg} exhibit good registration accuracy in some test scenes, yet they suffer from orientation ambiguity, \emph{e.g.}, in the second row, where only flat planes are dominant in the scans. PointDSC performs the worst among all the models in terms of both position and orientation accuracy. In challenging scenarios, such as the cluttered environment in the first row, DGR may even fail, resulting in an obvious misalignment, by seeing penetration into walls.

To verify the proposed model performance under varying numbers of sparsely sampled input points, we also implement the sparse tests as table below, by comparing with FCGF registration, D3Feat with or without PointDSC combination, and SpinNet. Our model has a consistent plausible performance on 3DMatch over the comparison models. In addition, more use points can boost the proposed model registration performance. In our setting. 1024 is chosen as the input point number to initialize the surfels in all the tests to get a good trade-off between accuracy and real-time performance. The surfel representation shows strong robustness even on extremely sparse points (256), with only a tiny performance drop of $2.5\%$ compared to the 4096 number of use points, while all the other baseline models demonstrate a remarkable performance difference between 4096 and 256 points.

\begin{table}[!th]
\vspace{-0.2em}
\centering
    \caption{RR results on 3DMatch with a different number of sampled points.}
    \label{tab:sample-pts}
\begin{adjustbox}{width=1.0\columnwidth}    
     \begin{tabular}{c c c c c c c}
        \toprule 
        \textbf{\#Sampled Points} & $\mathbf{4096}$ \quad & $\mathbf{2048}$  \quad & $\mathbf{1024}$ \quad & $\hphantom{}\mathbf{512}$ \quad & $\hphantom{}\mathbf{256}$ \quad & $\textbf{Average}$ \\
        \hline        
       FCGF \cite{choy2019fully} & 91.7 & 90.3 & 89.5 & 85.7 & 80.5 & 87.5\\
       \hline
       D3Feat \cite{bai2020d3feat} & 91.9 & 90.4 & 89.8 & 86.0 & 82.5 & 88.1\\    
       \hline
       PointDSC \cite{bai2021pointdsc} & 92.1 & 92.5 & 90.8 & 87.4 & 83.6 & 89.3\\
       \hline
       SpinNet \cite{ao2021spinnet} & 93.8 & 93.6 & 93.7 & 89.5 & 85.7 & 91.3\\
       \hline
       Ours & $\mathbf{95.4}$ & $\mathbf{94.8}$ & $\mathbf{94.1}$ & $\mathbf{92.4}$ & $\mathbf{87.8}$ & $\mathbf{92.9}$\\
        \bottomrule
    \end{tabular}
\end{adjustbox}
\vspace{-0.8em}
\end{table}

\subsection{Ablation Study}
1) We perform eight different types of ablation tests to verify the contribution of each design to our model performance, as shown in Table \ref{tab:ablation-cmps}. Point cloud position as input fed into various SOTA point cloud encoders, or vanilla E2PN encoders, cannot achieve performance on par with our surfel-based equi-model design. The uncertainties and Huber loss are all beneficial to improving the model performance. 2) We provide the robustness analysis of input scans perturbed by various levels of rotation and translation in Table \ref{tab:robustness-analysis}. 3) Finally, we provide model complexity comparisons for the top five baseline models and our model in Table \ref{tab:model-complexity}, as shown below, to showcase the low latency and small model size complexity of our model compared to other baselines.

\begin{table}[ht!]
\centering
    \caption{Ablation study on 3DMatch dataset.}
    \label{tab:ablation-cmps}
    \vspace{-0.6em}
\begin{adjustbox}{width=0.84\columnwidth}    
     \begin{tabular}{lccc}
        \toprule
        \multirow{1}{*}{\qquad Method} 
        & RE($^{\circ}$) $\downarrow$ & TE(cm) $\downarrow$ & RR(\%) $\uparrow$ \\ 
        \midrule
        1. Surfel w/o uncertainty weight & $1.64$ & $6.75$ & $91.73$ \\
        \hline
        2. Point cloud + vanilla E2PN & $2.24$ & $7.16$ & $89.85$ \\
        \hline
        3. Point Cloud + 3D CNN  & $12.08$ & $14.75$ & $51.47$\\%
        \hline
        4. Point Cloud + PointNet++ & $8.73$ & $10.82$ & $68.91$ \\
        \hline
        
        5. W/o attention module & $4.62$ & $18.24$ & $56.31$ \\
        \hline
        \hline
        6. $\mathcal{L}_1$ loss only & $1.96$ & $6.75$ & $87.08$\\ 
        \hline
        7. $\mathcal{L}_2$ loss only & $2.49$ & $7.93$ & $84.80$\\ 
        \hline
        \hline
        8. Full model + Huber Loss & $\mathbf{1.34}$ & $\mathbf{5.72}$ & $\mathbf{95.08}$ \\
        \bottomrule
    \end{tabular}
\end{adjustbox}    
\vspace{-0.2em}
\end{table}
Our model estimates the relative transformation between the source and target frame scans. We report the average metric errors across various translation and rotation intervals, applied to the same input scan for generating the source and target input pair.

\begin{table}[!thbp]
\vspace{-0.0em}
\centering
\caption{\small{Robustness analysis of average errors on 3DMatch.}}
\label{tab:robustness-analysis}
    \vspace{-0.6em}
\begin{small}
\begin{adjustbox}{width=0.90\columnwidth}
\begin{tabular}{c c c c c c}
\toprule
 & $5^{\circ}, 10cm$ & $25^{\circ}, 50cm$ & $50^{\circ}, 100cm$ & $75^{\circ}, 150cm$ & $100^{\circ}, 200cm$\\%
\hline
RE($^{\circ}$) $\downarrow$ & 0.18 & 0.71 & 2.37 & 3.12 &  4.60\\ 
\hline
TE(cm) $\downarrow$ & 0.51 & 1.75 & 2.86 & 6.73 & 12.59 \\
\hline
RR(\%) $\uparrow$ & 94.02 & 93.26 & 90.15 & 88.50 & 87.41 \\
\hline
\end{tabular}
\end{adjustbox}
\end{small}
\end{table}

The input consists of 100 scan pairs from the 3DMatch subset, with the source scan perturbed by the translations and rotations shown in the first row. As shown in the table, the registration error remains within 4-5\% of the input translation, and the rotation error within 3-4\%, demonstrating that the error does not scale with increased perturbations. 
\begin{table}[!thbp]
\vspace{-0.2em}
\centering
\caption{\small{Ablation study of model complexity on 3DMatch.}}
\label{tab:model-complexity}
    \vspace{-0.0em}
\begin{small}
\begin{adjustbox}{width=0.88\columnwidth}
\begin{tabular}{c c c c c c c}
\hline
 & DGR & PointDSC & Spinnet & RoReg & MAC+Trans & Ours\\%
\hline
Latency(s) $\downarrow$ & 1.26 & $\mathbf{0.08}$ & 2.84 & 22226 & 0.31 & 0.09 \\ 
\hline
Params(Mb) $\downarrow$ & 1.94 & 1.07 & 3.16 & 83.25 & 28.64 & $\mathbf{0.98}$ \\ 
\hline
\end{tabular}
\end{adjustbox}
\end{small}
\end{table}

We also provide the training loss curves within 300 epochs under the different threshold $\delta$ in Eq. \ref{eq:loss}. The threshold $\delta$ 0.5 can achieve a good trade-off between training convergence speed and accuracy (curve in blue). The Huber loss curve of the threshold bigger than 0.5 (curve in yellow and magenta), even suffers from the overfitting problem after 150 epochs. The training converges relatively slowly under the threshold $\delta$ 0.02 (in red) and 0.2 (in green) compared to 0.5, although the final loss reaches the accuracy of the same level as threshold 0.5. The small threshold $\delta$ makes the $\mathcal{L}_2$ loss range very limited, and the $\mathcal{L}_1$ loss is dominant in the learning process then.
 
\begin{figure}[!ht]   
\vspace{-0.6em}
  \includegraphics[width=0.98\linewidth]{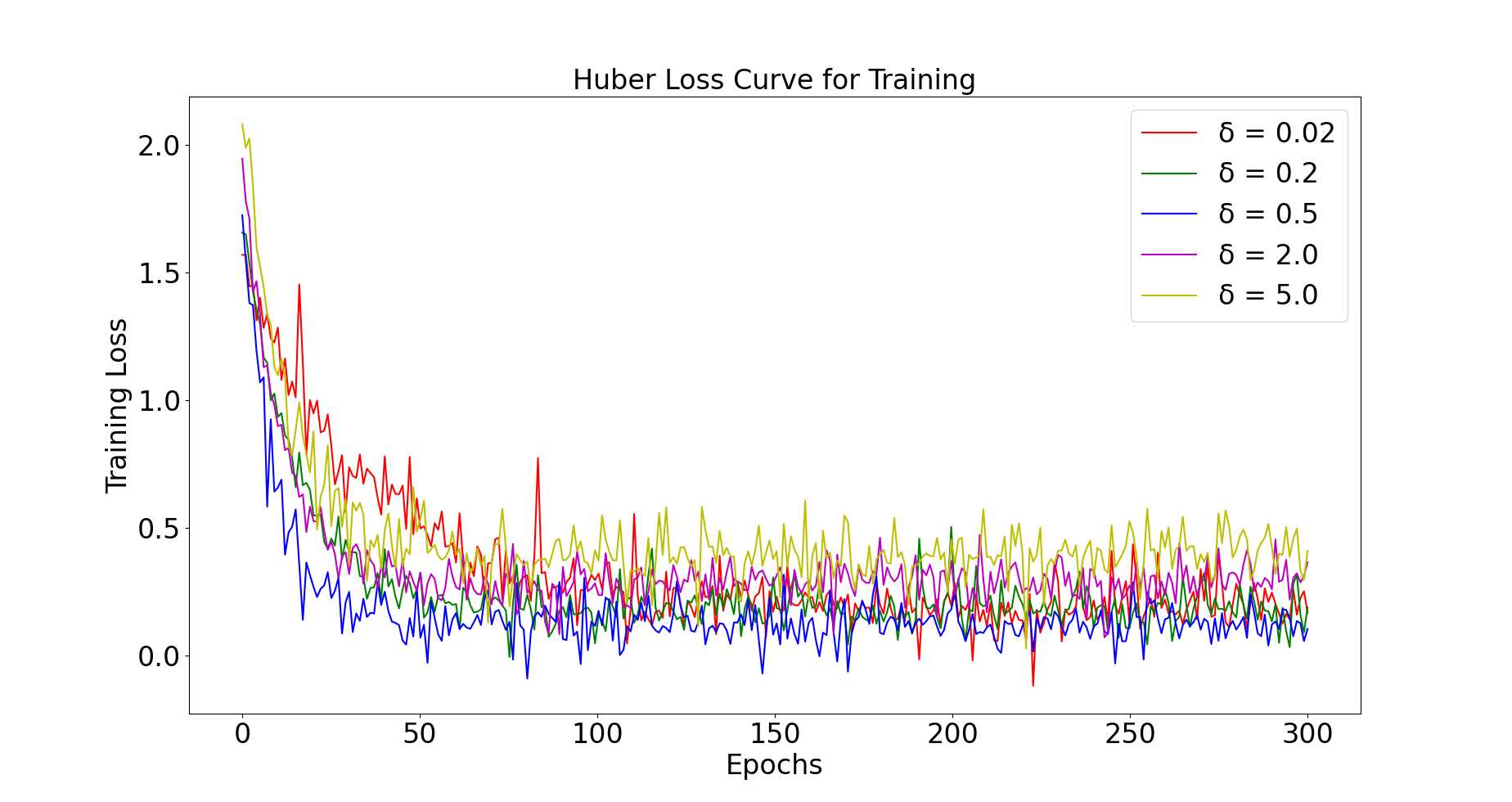}
    \caption{Huber loss learning curve under different thresholds.}
    \vspace{-0.0em}
    \label{fig:huber-loss}
\vspace{-0.5em}
\end{figure}
\section{Conclusion}

We propose a surfel-based $\mathbf{SE(3)}$ equivariant network model, incorporating surfel initialization from raw RGB-D depth maps or LiDAR point clouds. Our framework consists of a shared E2PN encoder, a cross-attention module, and an MLP-based decoder. Extensive experiments on two datasets demonstrate the model's robustness and state-of-the-art accuracy. Furthermore, the model's modular surfel representation enables generalization across diverse 3D scenes. Future work could explore leveraging surfel primitives for 3D mapping, reconstruction, and mesh conversion to enhance downstream tasks, while also improving model robustness across varying levels of point cloud sparsity and even for noisier data, including dynamic objects and uncertain information. Lastly, a more efficient and robust correspondence structure can be investigated. \revAouda{Lastly, more efficient equivariant deep learning methods can be explored. Equivariance can be achieved through structural design, such as extending neurons from 1D scalars to 3D vectors, or by integrating modules that enforce rotation-equivariant properties. However, these approaches often increase the complexity of the model. Therefore, the trade-off between implementation complexity and efficient rotation generalization should be further investigated. In addition, we acknowledge that certain objects exhibit strong view-dependent appearance characteristics, which challenge conventional point cloud registration methods that typically assume view-independence. As the main focus of this chapter and paper is on geometric modeling rather than appearance, we have chosen to leave this issue as a noted limitation and highlight it as a promising direction for future work.}
\chapter{Depth Prediction from Focal Stack Using Focal Geometry Constraint}               
\label{ch:cp3}

\begin{center}
    \textbf{\large Abstract}
\end{center}

Depth estimation from images is crucial for generating dense point clouds in the camera frame, enabling advanced 3D reconstruction tasks. While depth sensors and LiDAR provide accurate measurements, they remain expensive and limited in range. Alternatively, triangulated point clouds from sparse 2D feature points require extensive image coverage, increasing data collection challenges. Monocular depth priors in deep learning models offer a cost-effective and scalable alternative for dense point cloud generation. Depth from a focal stack is a specific challenge within depth estimation, where we use the focus differences across a sweep of images to figure out the depth of a scene for our proposed FocDepthFormer\footnote{The majority of this chapter was published as a peer-reviewed conference paper \cite{kang2024focdepthformer} (Kang, Xueyang, et al. ``FocDepthFormer: Transformer with Latent LSTM for Depth Estimation from the Focal Stack.'' Australasian Joint Conference on Artificial Intelligence. Singapore: Springer Nature Singapore, 2024). \\
\textbf{Author Contributions:} \\
\vspace{-1.2em}
\begin{itemize}
    \item \textbf{Xueyang Kang}: Idea Design, Methodology, Software, Experiment Validation, Formal Analysis, Data Curation, Writing, Review, and Editing.
    \item Fengze Han: Data Curation and Review.
    \item Abdur R. Fayjie: Data Curation and Review.
    \item Patrick Vandewalle: Review, and Supervision.
    \item Kourosh Khoshelham: Review, and Supervision. 
    \item Dong Gong: Review and Supervision.
\end{itemize}
}. Unlike broader depth estimation techniques that might use stereo vision, structure-from-motion, or deep learning, focal stack methods tap into defocus clues to recover depth. This makes them especially handy in situations where other depth hints are hard to detect. Most existing methods for depth estimation from a focal stack of images employ convolutional neural networks (CNNs) using 2D or 3D convolutions over a fixed set of images. However, their effectiveness is constrained by the local properties of CNN kernels, which restricts them from processing only focal stacks of a fixed number of images during both training and inference. This limitation hampers their ability to generalize to stacks of arbitrary lengths. To overcome these limitations, we present a novel Transformer-based network, FocDepthFormer, which integrates a Transformer with an LSTM module and a CNN decoder. The Transformer's self-attention mechanism allows for the learning of more informative spatial features by implicitly performing non-local cross-referencing. The LSTM module is designed to integrate representations across image stacks of varying lengths. Additionally, we employ multi-scale convolutional kernels in an early-stage encoder to capture low-level features at different degrees of focus/defocus. By incorporating the LSTM, FocDepthFormer can be pre-trained on large-scale monocular RGB depth estimation datasets, improving visual pattern learning and reducing reliance on difficult-to-obtain focal stack data. Extensive experiments on diverse focal stack benchmark datasets demonstrate that our model outperforms state-of-the-art approaches across multiple evaluation metrics.

\section{Introduction}
In scenarios where random camera motion is constrained, depth can be inferred by capturing a focal stack, multiple images taken at varying aperture sizes and focal distances, where each image encodes depth-dependent defocus information. However, existing depth-from-focus methods typically rely on convolutional neural networks (CNNs) with fixed 2D or 3D kernels, which struggle to generalize across focal stacks of varying lengths and fail to capture long-range dependencies effectively.

To overcome these limitations, we propose FocDepthFormer, a novel transformer-based network designed to estimate the depth of focal stacks. Our model integrates a Transformer module to capture global spatial relationships, an LSTM to aggregate information across focal stacks of arbitrary lengths, and a CNN decoder to refine depth predictions. A multi-scale convolutional encoder extracting fine-grained focus/defocus features is also used before the Transformer encoder. By leveraging pre-training before monocular RGB-depth datasets, FocDepthFormer can achieve good generalization ability without using a large volume of focal stack datasets for training.

Extensive experiments on diverse datasets demonstrate that FocDepthFormer outperforms state-of-the-art methods across multiple focal stack benchmarks. Furthermore, its strong generalization ability enables accurate depth prediction on unseen focal stack datasets, and this shows the great potential of using such a technique for direct inferences on open-world dataset.

Traditional 3D reconstruction pipelines, such as COLMAP \cite{schoenberger2016mvs}, rely on triangulating matched 2D feature points extracted from images using descriptors like SIFT \cite{lindeberg2012scale}. These methods require dense image coverage to ensure sufficient correspondences for reconstructing 3D points. For large-scale scenes, such as outdoor buildings, generating dense point clouds often requires collecting thousands of images. As a result, the 3D points reconstructed from sparse feature points can vary significantly in density across spatial regions, depending on the distribution of input images.

\begin{figure}[!thbp]
\centering
\includegraphics[width=1.0\linewidth]{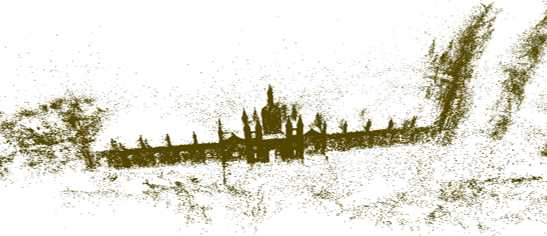}
\caption{Point cloud reconstructed from the extracted feature points of multiview image (Kings College Scene from Cambridge dataset \cite{kendall2015posenet}) by COLMAP \cite{schoenberger2016mvs, schoenberger2016sfm} tool.}
\label{fig:points-multiview}  
\end{figure} 

With the advent of deep neural networks (DNNs) and large-scale image datasets, depth can now be predicted directly from images, enabling the generation of dense point clouds from color pixels. 
When the point clouds in the local frame are aligned consistently into a global and dense map, these dense point clouds can reduce the reliance on extensive image collections. Monocular depth estimation methods \cite{yin2019enforcing,guo2018learning} have demonstrated success on benchmarks \cite{geiger2012we}, but their reliance on prior knowledge of scene textures, perspectives, and contexts often limits generalization to unseen data. Multi-view consistent cues \cite{smolyanskiy2018importance,bae2022multi} have been employed to improve generalization, but challenges persist in handling diverse scenes.

Depth estimation from RGB images remains a core paradigm for its convenience and flexibility. Researchers have explored various cues to infer depth, including image context \cite{yin2019enforcing,guo2018learning}, geometry cues \cite{smolyanskiy2018importance,zhang2019ga,godard2017unsupervised,xie2016deep3d}, and focus/defocus cues \cite{yang2022deep,gordon2019depth,ruan2021aifnet}. Among these, focus/defocus cues, which leverage sharpness variations across focal distances, offer a promising yet underexplored approach to depth estimation \cite{shane20allinfocus,carvalho2018deep,tang2017depth,maximov2020focus}.
Despite being a fundamental depth estimation problem, focal stack distinctively varies from monocular depth estimation \cite{eigen2014depth, lin2015depth, monodepth2, godard2019digging, feng2019unsupervised, hu2019revisiting, meng2021cornet, Ranftl2020, miangoleh2021boosting, ramamonjisoa2021single, hornauer2022gradient}, stereo depth or disparity estimation
\cite{godard2017unsupervised, xie2016deep3d}, and multi-frame depth estimation \cite{schonberger2016structure, gordon2019depth}. These models are trained on RGB-D image pairs or disparity maps. However, the focal stack task can not directly utilize these models as focal stack images contain out-of-focus regions and no disparity cues from motion are available.
Depth estimation from focus and defocus involves predicting the depth map from a captured \emph{focal stack} of the scene, which comprises images captured by a camera focused on different focal planes \cite{xiong1993depth}, also referred to as the depth of field control problem \cite{pentland1987new}, where focal stack images are obtained using a light field camera \cite{liu2017light}. Conventional methods \cite{suwajanakorn2015depth, surh2017noise, moeller2015variational} address this task using handcrafted features based on sharpness. However, these methods often fail in textureless scenes. To enhance feature extraction for this task, Convolutional Neural Networks (CNNs) have been employed to learn depth map prediction from focal stack \cite{hazirbas2018deep, wang2021bridging, yang2022deep, tang2017depth, carvalho2018deep, gur2019single, anwar2021deblur, he2022multi, fujimura2023deep}. Specifically, DDFFNet \cite{hazirbas2018deep}, AiFDepthNet \cite{wang2021bridging}, and DFVNet \cite{yang2022deep} leverage in-focus cues for depth estimation. DefocusNet \cite{maximov2020focus} aims to learn permutation invariant defocus cues, also known as Circle-of-Confusion (CoC). These methods utilize 2D or 3D convolutions to represent visual and focal features across spatial domains and stack channels. They either fuse stacked network 2D depth outputs \cite{maximov2020focus} or predict a single 2D depth map from the 3D feature volume \cite{yang2022deep}. Despite the potential to enlarge receptive fields with increased network depth, CNN-based models are confined to capturing features in local areas. Additionally, existing methods are limited to focal stacks with a constant number of images during both training and testing, making it challenging to generalize to stacks with an arbitrary number of images.

In this chapter, we propose a novel LSTM + Transformer-based network for depth estimation from focal stacks, which is referred to as FocDepthFormer. The core component of FocDepthFormer is a module of \emph{Transformer with latent LSTM}, which consists of a Transformer encoder \cite{dosovitskiy2020image}, an LSTM-based recurrent module \cite{hochreiter1997long} applied to the latent tokens, and a CNN decoder. The Transformer and LSTM are used to separately model the spatial and stack information. Different from the CNNs \cite{yang2022deep} restricted to local representation, the Transformer encoder captures the visual features with a larger receptive field. The self-attention mechanism in Vision Transformer (ViT) \cite{dosovitskiy2020image} facilitates the \emph{cross reference} among non-local patterns, allowing the Transformer encoder to capture more informative features to represent the sharpness and blur characteristics. Considering that the focal stacks may have arbitrary and unknown numbers of images, we utilize the LSTM in latent feature space to fuse focusing information across the entire stack for depth prediction. This differs from existing focal stack depth estimation methods \cite{yang2022deep,wang2021bridging,maximov2020focus}, as well as monocular depth estimation methods based on CNNs or Transformers \cite{agarwal2022depthformer, ranftl2021vision}, which typically handle inputs with a constant number of images. Specifically, we compactly fuse activated token features via the recurrent LSTM module after the Transformer encoder. This design allows the proposed model to handle focal stacks of arbitrary lengths with a predefined order during both training and testing, offering greater flexibility in practice.

Before inputting data into the Transformer, we employ an early-stage convolutional encoder with multi-scale kernels \cite{xiao2021early} to directly capture low-level focus/defocus features at different scales. In light of the limited availability of focal stack data, our model exhibits the capability to enhance its scene feature representation through pre-training on monocular depth estimation datasets. This is facilitated by the use of the recurrent LSTM module in our model, allowing it to take varying numbers of input images.
The main contributions of this work can be summarized as:
\begin{itemize} 
    \vspace{-0.5em}
    \item We introduce a novel Transformer-based network model designed for depth estimation from focal stack images. The model utilizes a vision Transformer encoder with self-attention to capture non-local spatial visual features, effectively representing sharpness and blur patterns. To handle an arbitrary number of input images, we incorporate an LSTM-based recurrent module. This structural flexibility enables us to pre-train the model with a monocular depth estimation dataset, mitigating the demands for focal stack data, which is both limited and expensive to collect.
    \item To fuse the stack features, the LSTM is applied, and before the fusion, we employ a grouping operation to manage recurrent complexity over tokens without scaling complexity, as the token count increases due to the larger stack size. This is achieved by applying the LSTM solely to a subset of activated embedding tokens while preserving information on other non-activated tokens through averaging.
    \item Specifically, we also propose the use of multi-scale kernels in an early-stage convolutional encoder to enhance the capture of low-level focus/defocus cues at various scales.  
    \vspace{-0.8em}
\end{itemize}

\section{Related Work}
\label{CH3-sec:rel}
\noindent\textbf{Depth from Focus/Defocus.} Depth estimation from focal stacks relies on discerning relative sharpness within the stack of images for predicting depth. Traditional machine learning methods \cite{xiong1993depth, suwajanakorn2015depth, surh2017noise} treat this problem as an image filtering and stitching process. Johannsen \textit{et al.} \cite{johannsen2017taxonomy} provide a comprehensive overview of methods addressing the challenges posed by light field cameras, laying a foundation for research in this direction. More recently, CNN-based approaches have emerged in the context of focal stacks. DDFFNet \cite{hazirbas2018deep} introduces the first end-to-end learning model trained on the DDFF 12-Scene dataset. DFVNet \cite{yang2022deep} utilizes the first-order derivative of volume features within the stack. AiFNet \cite{wang2021bridging} aims to bridge the gap between supervised and unsupervised methods, accommodating ground truth depth or its absence. Barratt \textit{et al.} \cite{shane20allinfocus} formulate the problem as an inverse optimization task, utilizing gradient descent search to simultaneously recover an all-in-focus image and depth map. DefocusNet \cite{maximov2020focus} exploits the Circle-of-Confusion, a defocus cue determined by focal plane depth, for generating intermediate defocus maps in the final depth estimation. Anwar \textit{et al.} \cite{anwar2021deblur} leverage defocus cues to recover all-in-focus images by eliminating blur in a single image. Recently, the DEReD model \cite{si2023fully} learns to estimate both depth and all-in-focus (AIF) images from focal stack images in a self-supervised way, by taking the optical model into the loop to reconstruct the defocus effects. Gur and Wolf \cite{gur2019single} present depth estimation from a single image by leveraging defocus cues to infer disparity from varying viewpoints.

\noindent\textbf{Attention-Based Models.} The success of attention-based models \cite{vaswani2017attention} in sequential tasks has led to the rise of Vision Transformer for computer vision tasks. The Vision Transformer represents input images as a series of patches ($16 \times 16$). While this model performs well in image recognition compared to CNN-based models, a recent study \cite{xiao2021early} demonstrates that injecting a small convolutional inductive bias in early kernels significantly enhances the performance and stability of the Transformer encoder. In the context of depth estimation, Ranftl \textit{et al.} \cite{ranftl2021vision} utilize a Transformer-based model as the backbone to generate tokens from images, and these tokens are assembled into an image-like representation at multiple scales. DepthFormer \cite{agarwal2022depthformer} merges tokens at different layer levels to improve depth estimation performance. The latest advancement in this domain, Swin Transformer \cite{liu2021swin}, achieves a larger receptive field by shifting the attention window, revealing the promising potential of the Transformer model.

\noindent\textbf{Recurrent Networks.} Recurrent networks, specifically LSTM \cite{hochreiter1997long}, have found success in modeling temporal distributions for video tasks such as tracking \cite{nwoye2019weakly} and segmentation \cite{xu2018youtube}. The use of LSTM introduces minimal computational overhead, as demonstrated in SliceNet \cite{pintore2021slicenet}, where multi-scale features are fused for depth estimation from panoramic images. Recent works \cite{hutchins2022block, huang2020trans} combine LSTM with Transformer for language understanding via long-range temporal attention.
\subsection{Preliminaries}
\label{sec:focal_principle}
\begin{figure}[!thbp]
\centering
\includegraphics[width=1.0\linewidth]{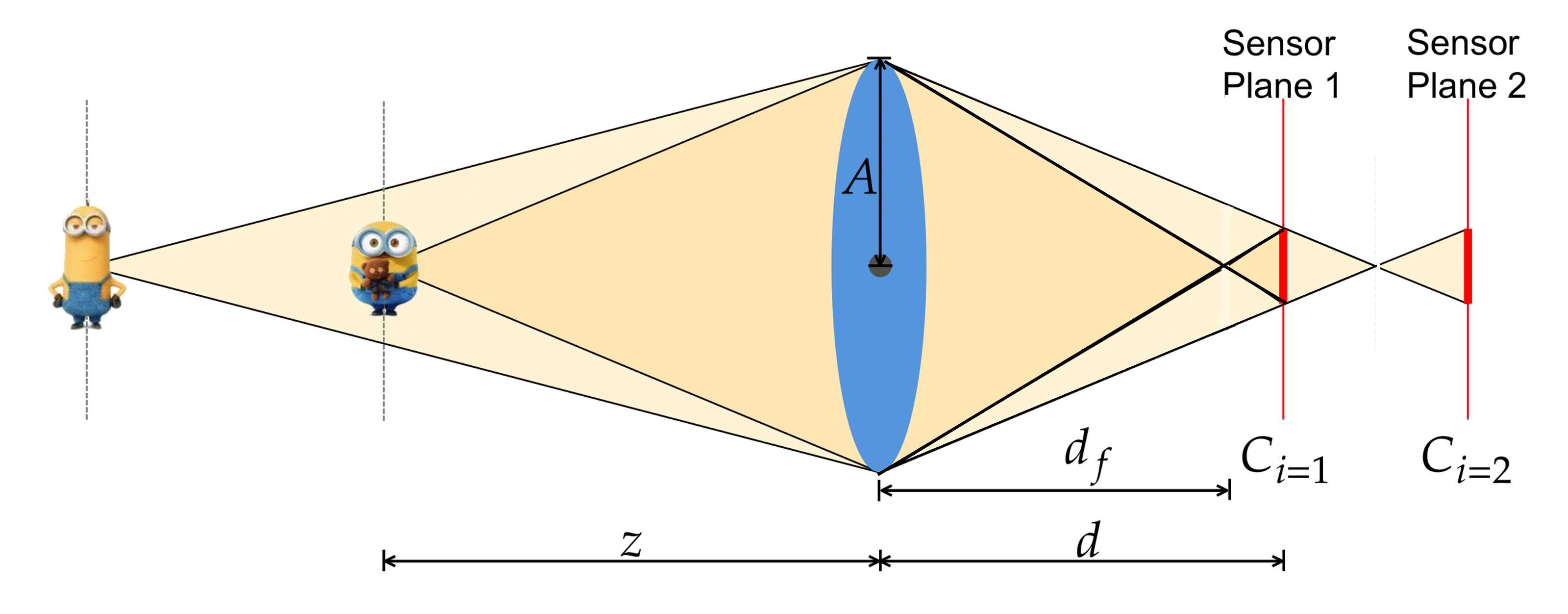}
\caption{The rays emitted from an object placed at an axial distance $z$ to the lens, converging at a distance ${d}_{f}$ behind the lens. The sensor is situated at a distance $d$ from the focal lens. The pixels are sharply imaged when the sensor is placed right at focus distance, $C$ is the CoC, which grows as the sensor position deviates from the focus plane.}
\label{fig:lens}  
\end{figure}

In the paper, we present the features of our network model, a primary factor to explain our favorable results is the good identification ability of pixel sharpness, as mentioned in our paper. Usually, the sharpness of pixels can be evaluated according to the Circle-of-Confusion, $\mathbf{C}$ (in Figure \ref{fig:lens}), defined as follows,
\begin{equation}
C_{}^{} =
\frac{f_{}^{2}}{N_{}^{}(z_{}^{}-{d}_{f})}\Big |1 -\frac{{d}_{f}}{z}\Big |,
\label{eq:Coc}
\end{equation}
where, $N$ denotes $f$-number, as a ratio of focal length to the valid aperture diameter. $C$ is the Circle of Confusion diameter (CoC). ${d}_{f}$ is the focus distance of the lens. $z$ represents the distance from the lens to the target object. In general, the range of $z$ is $[0, \infty]$. However, in reality, the range is always constrained by lower and upper bounds.

In Eq. \ref{eq:C-radius}, $C$ is zero ($C_{}^{*}$) when the image pixel is in focus, and it is a signed value, where $C>0$ indicates the camera focused in front of the sensor plane, while $C<0$ is the reverse case, with a camera focused behind the sensor plane. It can be inferred from the denominator, that CoC $C$ has two divergence points, at 0 and $d_{f}$ respectively. Our model learns the mapping relationship from CoC to depth autonomously.

\begin{figure}[!thbp]
\vspace{-2.5em}
\centering
\includegraphics[width=1.0\linewidth]{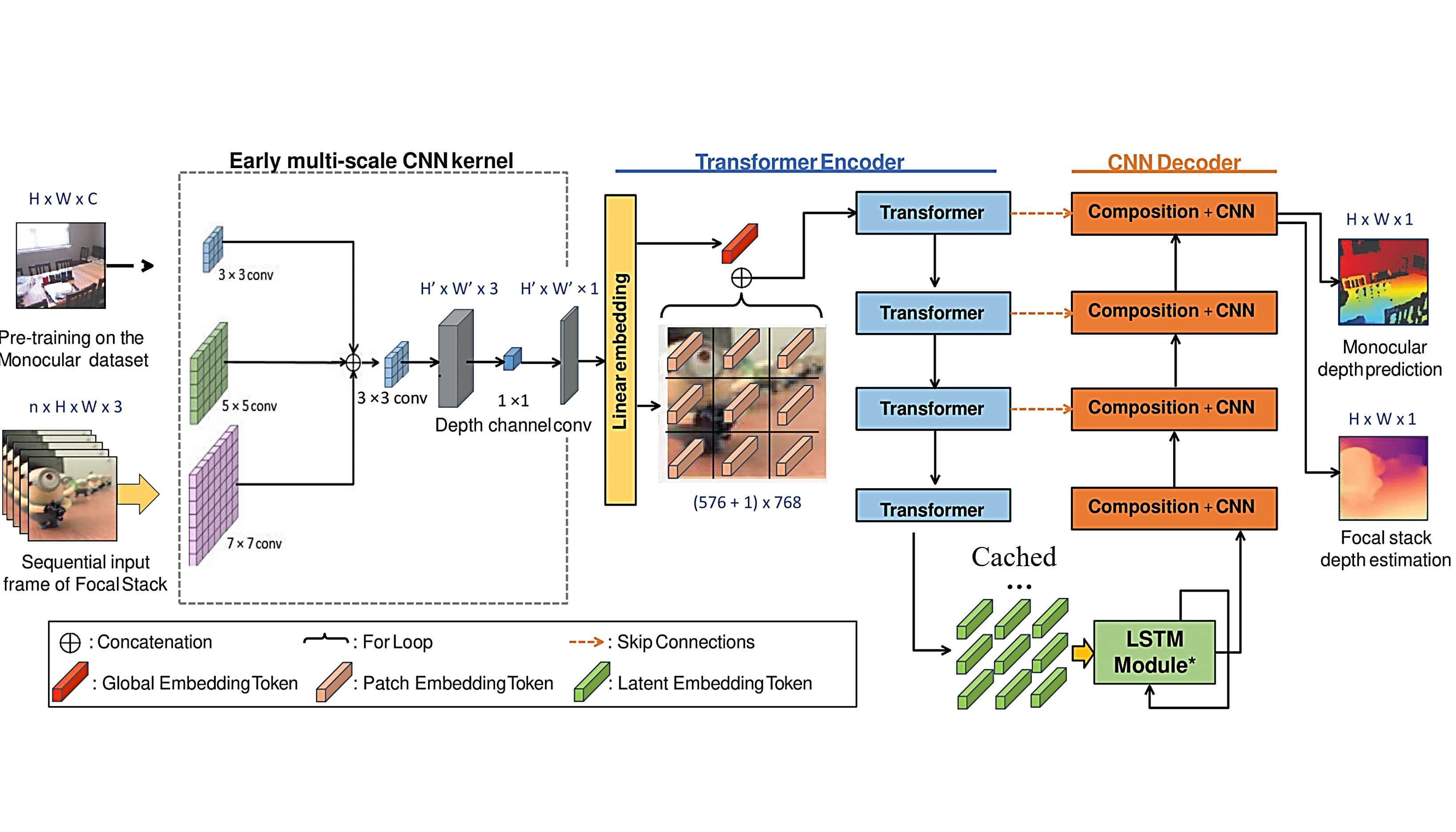}
\vspace{-2.4em}
\caption{The overview of our proposed network, FocDepthFormer, is presented with its core components: the Transformer encoder, the recurrent LSTM module, and the CNN decoder. Preceding the Transformer encoder, early-stage multi-scale convolutional kernels are depicted within the dashed line. The resulting multi-scale feature maps are concatenated and subjected to spatial and depth-wise convolution. Subsequently, the fused feature map of an image stack is divided into patches, which are then individually projected by a linear embedding layer into tokens. A red token represents a global embedding token mapped from the entire image and is summed with each patch embedding token.} 
\label{fig:model_arch}  
\end{figure}

Given a focal stack, $\mathbf{S}$, comprising $N$ images ordered from near to far by focus distance, denoted as $\mathbf{S} = (\mathbf{x}_{i})_{i=1}^N$, where each image $\mathbf{x}_{i} \in \mathbb{R}^{H\times W \times 3}$, our objective is to generate a single depth map $\mathbf{D}\in\mathbb{R}^{H\times W \times 1}$ for a stack of images. In contrast to the vanilla Transformer \cite{dosovitskiy2020image}, we initially encode each image $\mathbf{x}$ using an \emph{early-stage multi-scale kernel-based convolution} $\mathcal{F}(\cdot)$. This convolution ensures a multi-scale feature representation $\mathbf{x}'$ for the focal stack images. Subsequently, the \emph{transformer encoder} $g(\cdot)$ processes the feature maps, transforming them into a series of ordered tokens that share information through self-attention. The self-attention weights between the in-focus features and blur features, encode spatial information from each input image. The \emph{recurrent LSTM module} sequentially processes cached latent tokens from different frames of a focal stack and fuses them along the stack dimension. This stack feature fusion process is learned in the latent space by the LSTM module.
Our attention design with LSTMs enhances the model's capability to handle an arbitrary number of input images. The final disparity map is decoded (denoted as $d(\cdot)$) from the fusion feature, utilizing the aggregated tokens from all images in the stack.

\subsection{Early-stage Encoding with Multi-scale Kernels}
\label{sec:early-CNN}
To capture low-level focus and defocus features at different scales, we employ an early-stage convolutional encoder with multi-scale kernels, which is different from methods using fixed-size kernel convolution stem before the Transformer \cite{xiao2021early}. As illustrated in Figure \ref{fig:model_arch}, the early-stage encoder utilizes three convolutional kernels to generate multi-scale feature maps $f_{m}(\mathbf{x}), \{m = 1,2,3\}$. All feature maps are concatenated and merged into the feature map $\mathbf{x'}\in\mathbb{R}^{H'\times W'\times 1}$ through spatial convolution, followed by $3\times3$ and $1\times1$ convolution on the feature map depth channel:
\begin{equation}
    \mathbf{x'} = \mathcal{F}(\mathbf{x}) = \text{{Conv}}(\text{{Concat}}(f_{m}(\mathbf{x}))),
\end{equation}
where $m$ ranges from $1$ to $3$. Feature concatenation after convolutions with multiple kernel sizes preserves fine-grained details of features across varying depth scales. The first module from the left in Figure \ref{fig:model_arch}, comprising parallel multi-scale kernel convolutions followed by depth-wise convolution, ensures the model has a large receptive field beyond the $7\times7$ kernel size. This facilitates capturing more defocus features while preserving intricate details. 
\vspace{-1.0em}
\subsection{Transformer with LSTM}
\begin{figure}[!ht]
    \vspace{-0.2cm}
    \centering
    \includegraphics[width=0.48\linewidth]{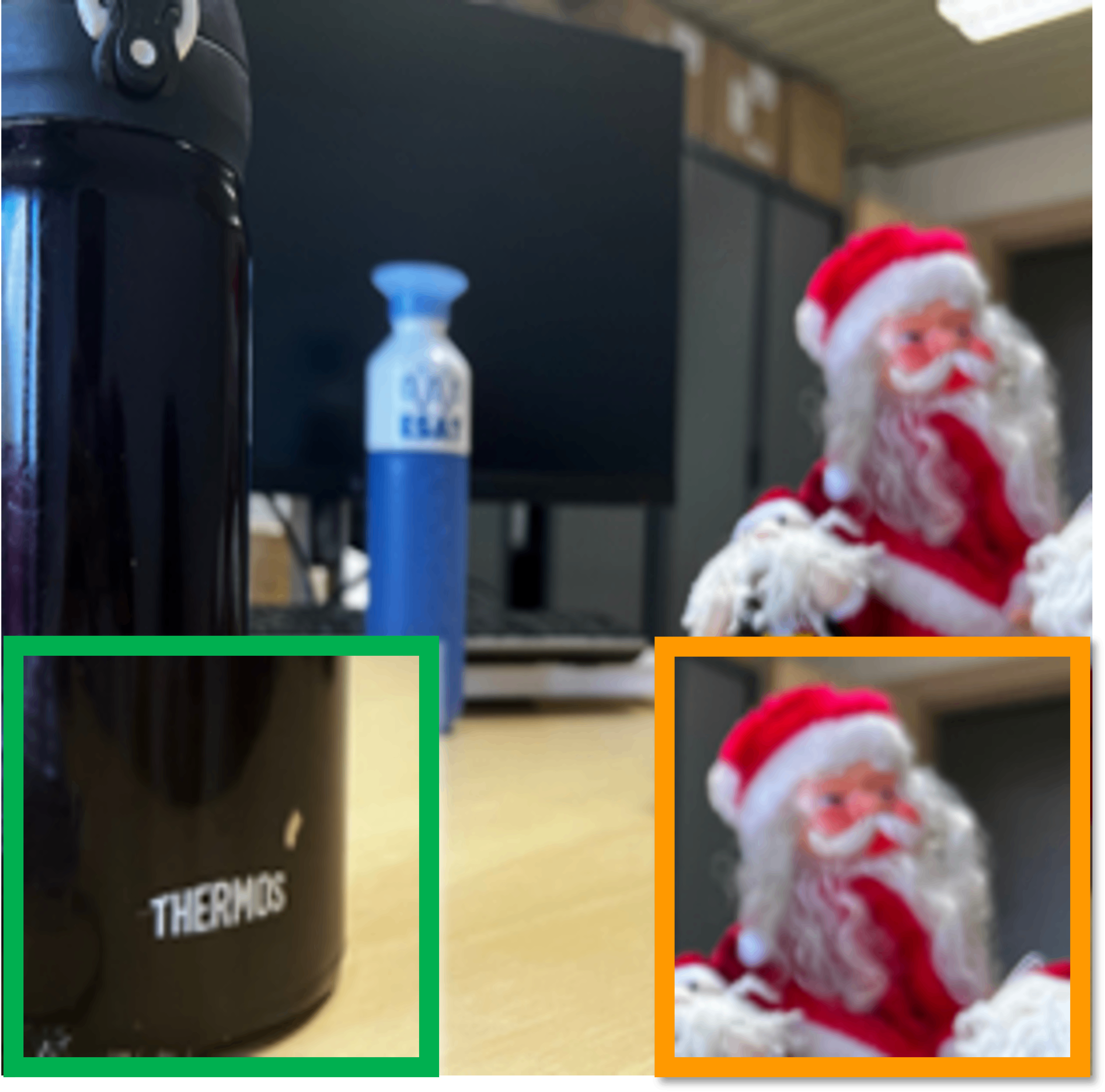}
    \hfill
    \includegraphics[width=0.48\linewidth]{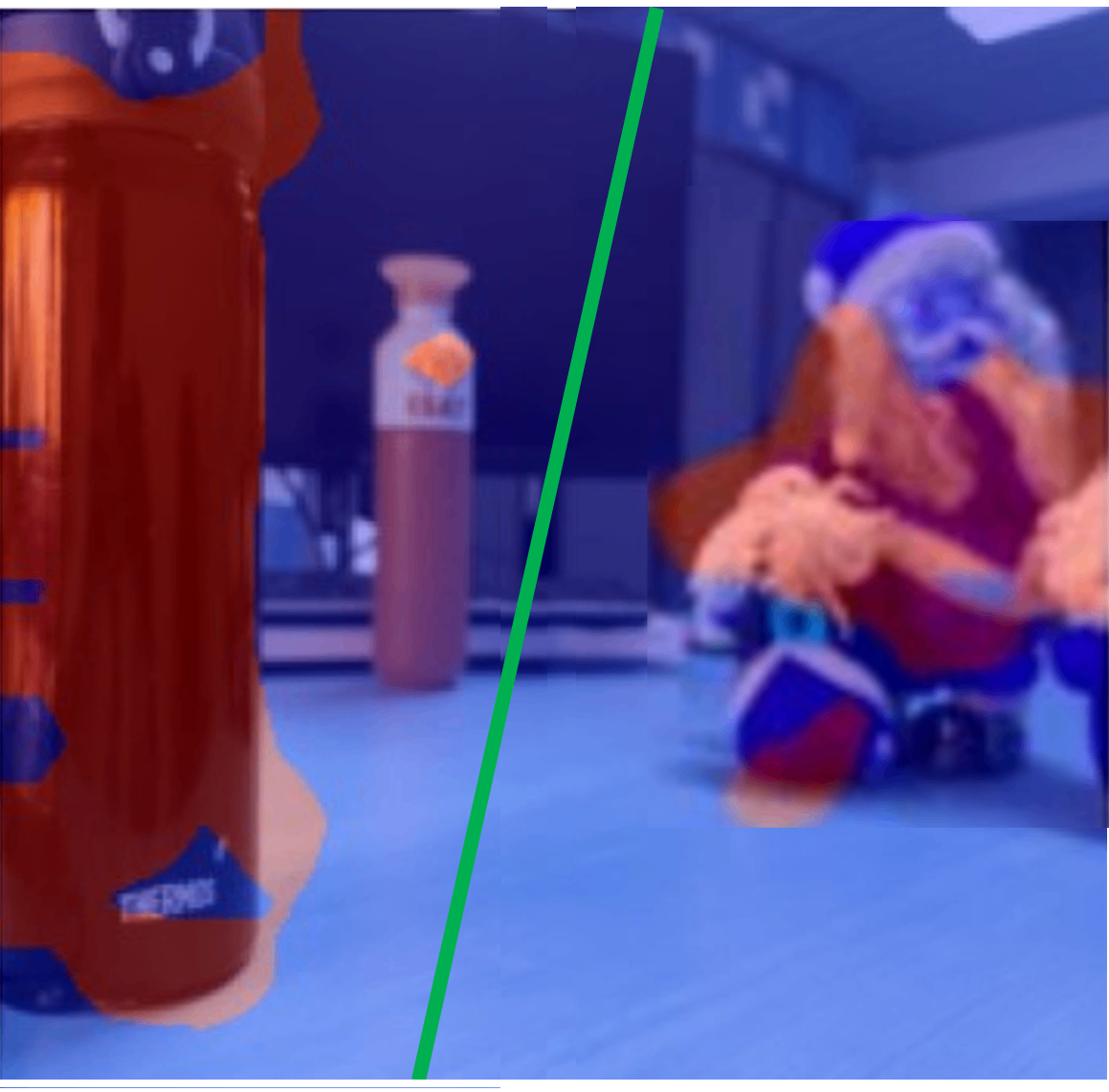}
    
    \vspace{0.2cm}
    
    \includegraphics[width=0.48\linewidth]{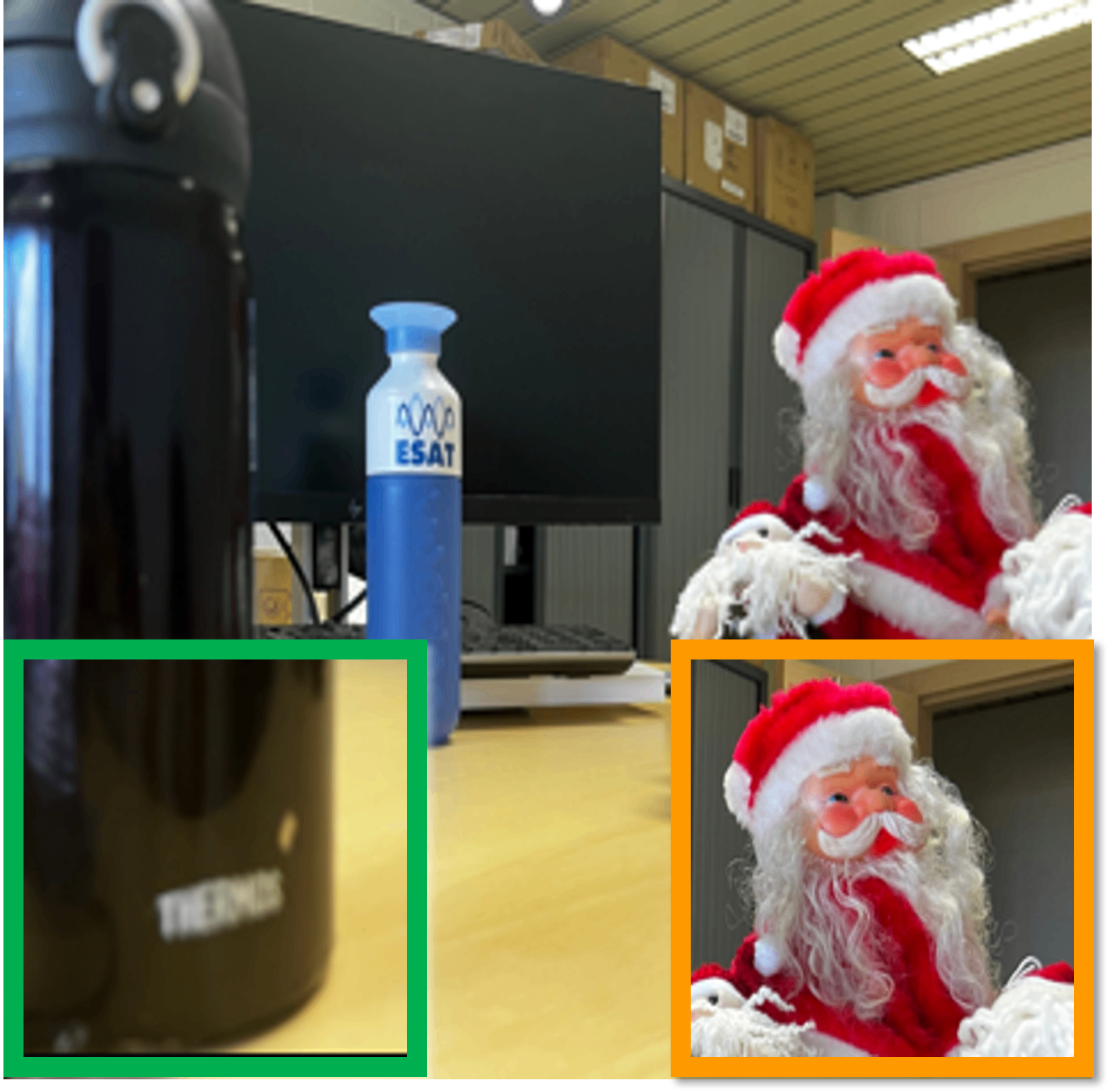}
    \hfill
    \includegraphics[width=0.48\linewidth]{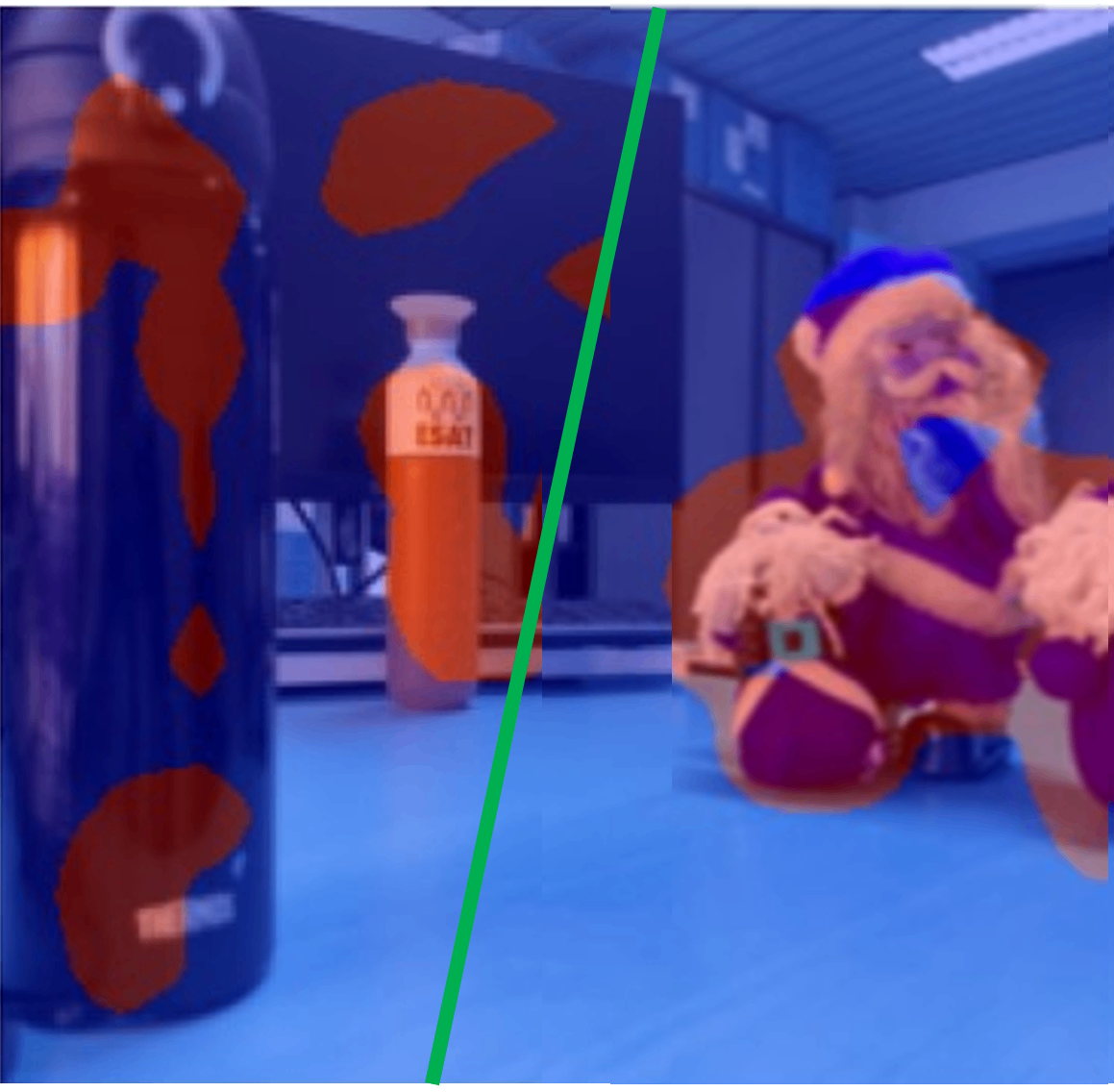}
    
    \vspace{-0.2cm}
    \caption{Comparison of Transformer attention on the two left column images. Cropped Image patches within green and orange boxes in (a) and (c) are used as the query input to calculate the self-attention map over the whole input image, respectively. In (b) and (d), the attention map on the left and right of the green line represents the attention output of the green and orange boxes, respectively. This demonstrates that the patches can selectively attend to both foreground and background areas, exhibiting the ability to differentiate the focus and defocus cues.}
    \label{fig:heatmap-cmps}
    \vspace{-0.4cm}
\end{figure}

\noindent\textbf{Transformer encoder.} 
The Transformer in Figure \ref{fig:model_arch}, denoted as $g(\cdot)$, processes the feature maps $\mathbf{x'}$ from the preceding early-stage multi-scale convolutions to generate a series of tokens $(\mathbf{t'}_{p})_{p=1}^{k}$:
\begin{equation}
    {\mathbf{t'}}_{1}^{}, {\mathbf{t'}}_{2}^{}, ..., {\mathbf{t'}}_{k}^{} = g(\mathbf{x'}).
\end{equation}
Specifically, the early kernel CNNs and Transformer encoder take the focal stack images sequentially, cache and concatenate the feature maps of a certain stack of images into $\mathbf{x'}$. Transformer uses first a linear embedding layer, which divides the feature maps $\mathbf{x'}$ into $k$ patches of size $16\times 16$. That is, $\mathbf{x'}_{p} \in \mathbf{x'}, p=1,2,..., k$, is projected by a linear embedding layer (MLP) into corresponding embedding tokens $(\mathbf{l}_{p})_{p=1}^k$, each token with a dimension of $768$ (576 in total), and all the tokens of a full stack $N$ are cached into $\{(\mathbf{l}_{p})_{p=1}^{k}\} \times N$ before LSTM to be fused at once. The Transformer's \emph{Position Embedding} encodes the positional information of the image patches of a single frame in an iterative order from the top-left of the image. An MLP layer generates the Global Embedding Token (Figure \ref{fig:model_arch}) by mapping the entire image into a global token and then adding each patch embedding token. Each linear embedding token is projected via a weight matrix $\mathbf{W}_{}^{(\cdot)}$ from dimension ${d}_{m}^{}$ into three vectors: the Query, ${\mathbf{l}}^{}_{Q}$, the Key, ${\mathbf{l}}^{}_{K}$, and the Value, ${\mathbf{l}}^{}_{V}$, with dimensions $d_Q^{}$, $d_K^{}$, and $d_V^{}$ respectively. The Queries, Keys, and Values pass through the Multi-Head Attention (MHA) units in parallel.
\begin{figure*}[!th]
\centering
\includegraphics[width=1.0\linewidth]{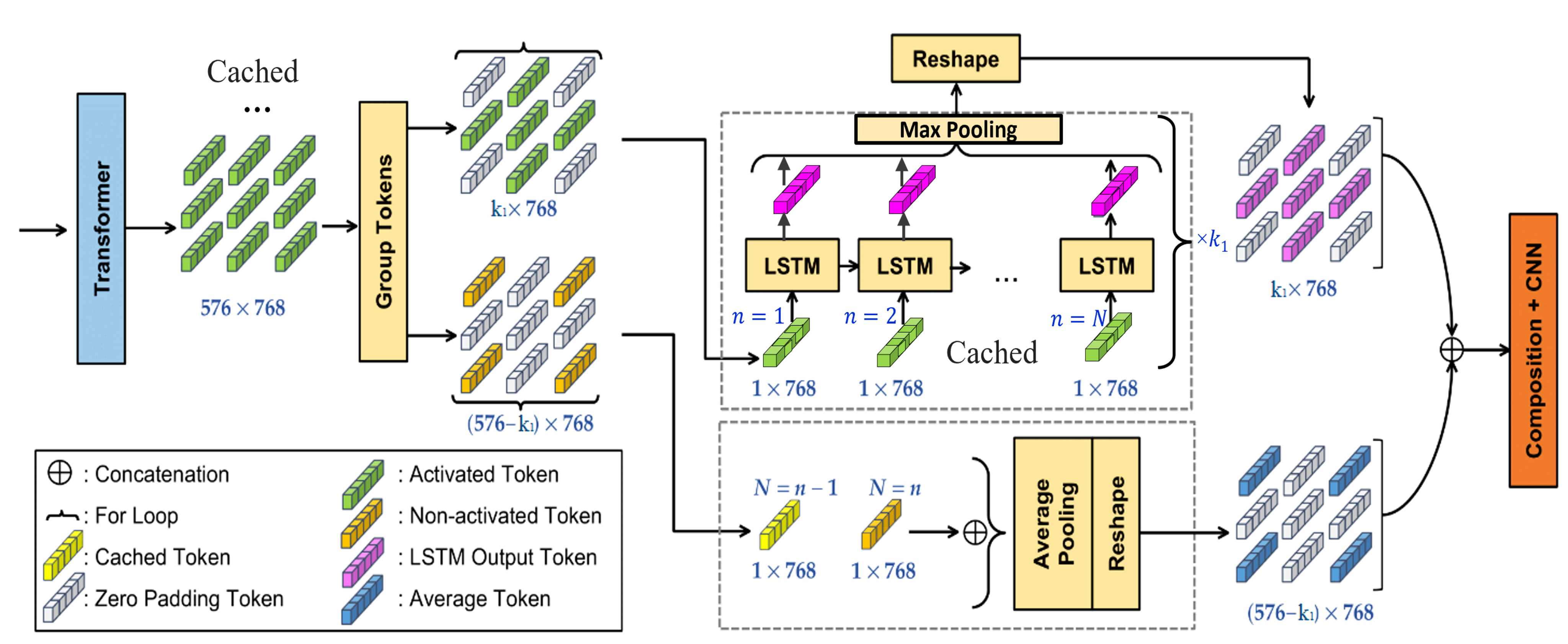}
\caption{To illustrate the LSTM module in our network, the initial step involves grouping all cached output tokens from the Transformer encoder into activated and non-activated tokens. These two groups are then individually processed, with activated tokens undergoing LSTMs followed by max pooling and non-activated tokens undergoing average pooling. Following this, the output tokens undergo reshaping and concatenation before being fed into the CNN decoder for predicting the depth map.}
\label{fig:LSTM-token}
\end{figure*}
\begin{align}
\vspace{-1.2em}
&\mathrm{MHA}({\mathbf{l}}^{}_{Q},{\mathbf{l}}^{}_{K},{\mathbf{l}}^{}_{V})= ({\mathrm{head}}_1^{} \oplus ... \oplus {\mathrm{head}}_N^{})\mathbf{W}^{O}_{}, \\
&{\mathrm{head}}_i^{}= \mathrm{softmax}\left(\frac{{\mathbf{l}}^{}_{Q}\mathbf{W}_{}^{{\mathbf{l}}^{}_{Q}}{\mathbf{l}}^{}_{K}\mathbf{W}_{}^{{\mathbf{l}}^{}_{K}}}{\sqrt{d_k^{}}}\right){\mathbf{l}}^{}_{V}\mathbf{W}_{}^{{\mathbf{l}}^{}_{V}},
\label{eq:mha}
\vspace{-0.8em}
\end{align}
where $\mathbf{W}_{}^{\mathbf{l}_Q}\in\mathbb{R}^{d_{m}\times d_{V}}$, $\mathbf{W}_{i}^{\mathbf{l}_K}\in\mathbb{R}^{d_{m}\times d_{K}}$, $\mathbf{W}_{}^{\mathbf{l}_V}\in\mathbb{R}^{d_{m}\times d_{V}}$, and $\mathbf{W}_{}^{O}\in\mathbb{R}^{d_{m}\times d_{V}}$. Following the Multi-Head-Attention modules contained in the encoder $g(\mathbf{x'})$, the resulting tokens $(\mathbf{t'}_{p})_{p=1}^{k}$ encode features that differentiate between focus and defocus cues from different stack image patches at the same spatial location within the image. This capability is illustrated in Figure \ref{fig:heatmap-cmps}. Consequently, the more in-focus and sharp features of the image patches receive increased attention in the embedding space.

\noindent\textbf{LSTM module.} To attain the flexibility of our model in handling stacks with arbitrary lengths, as opposed to the fixed length in existing methods \cite{yang2022deep, wang2021bridging, hazirbas2018deep}, we employ an LSTM to progressively fuse sharp features along the stack. The LSTM treats patch embedding tokens at the same image position as the sequential features along the stack dimension. The corresponding feature tokens $(\mathbf{t'}_p)_{p=1}^{k}$ from stack images at stack number $N$ are spatially ordered and fed into LSTM modules, which are arranged in the original spatial image order. At each position, each LSTM module incrementally fuses the latent token $\mathbf{t'}_p$ from a stack. This approach differs from existing models constrained to a 3D volume stack with a pre-defined and fixed size \cite{yang2022deep, wang2021bridging, hazirbas2018deep}. Importantly, the sequential processing in the latent space after a shared encoder for the stack images incurs limited complexity in practice.

\par
Preceding the LSTM modules, tokens associated with image patches from a single frame are categorized into activated and non-activated tokens, reflecting the informative level of the features. The $L_{2}$ norm, denoted as $\|\cdot\|$, of each embedding token is compared with a threshold of $0.4$, as shown in Figure \ref{fig:LSTM-token}. Specifically, for tokens within a single frame, only the activated tokens, identified through this threshold comparison, are forwarded to the LSTM, with the number of $k_1$ tokens. This operation reduces the computational complexity of the LSTM by applying the operation solely to a fraction of the latent tokens:
\begin{flalign}
\vspace{-1.6em}
{\mathbf{t}}^{n}_{p}, \mathbf{h}_{p}^{n} &= LSTM({\mathbf{t'}}_{p}^{n}, {\mathbf{h}}_{p}^{n-1}, c^{n}),
\label{eq:lstm-group}
\end{flalign}
this is a single LSTM layer expression, where $p=1,2,...,k_1$, and $n$ is the frame index number of a stack. We set the number of hidden layers of the LSTM module to be the same as the stack size $N$. 
The memory cell $c^{}$ undergoes continuous updates at each step, influenced by the input $\mathbf{t'}^{n}_{p}$ and the hidden state $\mathbf{h}$. Then all the LSTM layer outputs are merged by max pooling $\max\{\mathbf{t'}_{p}^{1}, ..., \mathbf{t'}_{p}^{n}\}$ into $\mathbf{t'}_{p}^{}$. For the non-activated tokens $(\mathbf{t'}_p)_{p=k-k_{1}}^{k}$, an averaging operation is performed with the corresponding cached tokens from the previous step at the same embedding position. Finally, the two groups of output tokens are arranged as the original input embedding order, yielding the final fused tokens $(\mathbf{t}_p)_{p=1}^{k}$.
\noindent\textbf{CNN decoder.} Our decoder $d(\cdot)$ follows the approach presented by Ranftl \textit{et al.} \cite{ranftl2021vision}, employing Transpose-convolutions to integrate feature maps after LSTMs. The decoder also incorporates feature maps of $i-th$ layer $g_{i}(\mathbf{x}'))$ from the encoder through skip connections, as depicted in Figure \ref{fig:model_arch}. Finally, the decoder $d(\cdot)$ predicts the depth. 
\begin{equation}
    \vspace{1.6em}
    \mathbf{\smash{\hat{D}}} = d((\mathbf{t}_p)_{p=1}^{k}, g_{i}(\mathbf{x}')),\quad i=\{1,2,3\}.
    \label{eq:depth_loss}
    \vspace{-1.5em}
\end{equation}
\subsection{Training Loss}
\label{sec:loss}
Our training loss comprises the Mean Squared Error (MSE) loss, denoted as $\mathcal{L}_{MSE}$, and a sharpness regularizer $\mathcal{L}_{log_{}}$ weighted by $\alpha$:
\begin{equation}
 \mathcal{L}_{total} = \mathcal{L}_{MSE}(\mathbf{\smash{\hat{D}_{}^{}}}, \mathbf{\smash{D_{}^{}}}) + \alpha\mathcal{L}_{\log_{}}(\delta_{\mathbf{\Delta{{\smash{\hat{D}}}_{}^{}}}}, \delta_{\mathbf{\Delta{\smash{D}}_{}^{}}}), 
 \label{eq:total_loss}
\end{equation}
where $\mathbf{\smash{D_{}^{}}}$ represents the ground truth depth, and $\mathbf{\smash{\hat{D}_{}^{}}}$ indicates the predicted depth. The $\mathbf{\Delta}$ is the laplacian operator, applied to predicted and ground truth depth images respectively. $\delta$ is the variance of the depth image. The regularizer item is formulated as $\log(\delta_\mathbf{{\Delta{\smash{\hat{D}}}_{}^{}}} /\delta_{\mathbf{\smash{\Delta{D}_{}^{}}}})$. The pixel blurriness due to out-of-focus can be described by the Circle-of-Confusion (CoC),
\begin{equation}
\sigma = \frac{C}{2\cdot r} = \frac{1}{2r}\frac{f_{}^{2}}{N_{}^{}(z_{}^{}-{d}_{f})}\Big |1 -\frac{{d}_{f}}{z}\Big |,
\label{eq:C-radius}
\end{equation}
where, $N$ denotes $f$-number, as a ratio of focal length to the valid aperture diameter, and $r$ is the CMOS pixel size. $C$ is the Circle of Confusion (CoC) diameter. ${d}_{f}$ is the focus distance of the lens. $z$ represents the distance from the lens to the target object. In general, the range of $z$ is $[0, \infty]$. However, in reality, the range is always constrained by lower and upper bounds. The goal of the model is to learn to shape a depth map from the focus/defocus features.
\subsection{Pre-training with Monocular Depth Prior}
Focal stack datasets are typically limited in size due to the high cost and challenges associated with data collection. To address the scarcity of data and fully exploit the potential of the Transformer, we can optionally pre-train the Transformer encoder on widely available monocular depth estimation datasets, such as NYUv2 \cite{silberman2012indoor}, to enhance spatial representation learning. This can be achieved by bypassing the latent LSTM module while back-propagating the gradients of the encoder $g(\cdot)$ and decoder $d(\cdot)$ weights only. The LSTM-based fusion design after Transformer allows it to process an arbitrary number of input images, making it leverage pre-training before monocular datasets. While monocular depth datasets may differ from focal stack datasets, pre-training facilitates the learning of a versatile visual representation for depth prediction, thanks to our separate spatial image and focal stack learning structure design. Notably, our model demonstrates plausible performance even in the absence of pre-training, as evidenced by the experiments presented below. For detailed parameters of each network block module, please refer to the figure below,

\nomenclature{$d_f$}{Focus distance}
\nomenclature{$CoC$}{Circle of Confusion diameter}
\nomenclature{$\mathbf{t}^{*}_{}, \mathbf{h}_{}^{*}$}{Fusion and Hidden tokens after LSTM}
\nomenclature{$\sigma$}{Confusion radius}
\nomenclature{$\mathbf{H}$}{Hessian matrix}
\nomenclature{$\Delta$}{Gradient operator}

\section{Experiments and Results}
\label{ch3-sec:exp}

The experiment is performed on four different datasets, and compared with the four baseline models for depth estimation from focal stack. The comparisons include both metric and visual results. After that, the detailed ablation study analysis is performed to evaluate each module block contribution, including data plots, visual results, and metric results.
\subsection{Experimental settings}
\label{subsec:setup}
Experiment results are composed of the 
\noindent\textbf{Datasets.} We conducted extensive evaluations of our model using four benchmark focal stack datasets: DDFF 12-Scene \cite{hazirbas2018deep}, Mobile Depth \cite{benavides2022phonedepth}, LightField4D \cite{honauer2016dataset}, and FOD500 \cite{yang2022deep}. Additionally, our model offers the flexibility of pre-training on the monocular RGB-D dataset NYUv2 \cite{silberman2012indoor}. Specifically, we conducted training on DDFF 12-Scene and FOD500 separately for the subsequent experiments, while Mobile Depth and LightField4D were employed for evaluating the model's generalizability with pre-trained on DDFF 12-Scene only. A comprehensive summary of the evaluation datasets, including their properties and captured sensor types, is presented in Table \ref{tab:dataset_summary}.

\noindent\textbf{Evaluation metrics.} The metrics used in our work for quantitative evaluations, are defined as follows,
\begin{align}
&RMSE: \sqrt{\frac{1}{|M|}\sum_{p\in \mathbf{x}}^{}\lVert {f(\mathbf{x}_{}^{})-\mathbf{D}_{}^{} \rVert}^{2}_{}}, \\
&logRMSE: \sqrt{\frac{1}{|M|}\sum_{p\in \mathbf{x}}^{}\lVert {log f(\mathbf{x}_{}^{})-\mathbf{D}_{}^{} \rVert}^{2}_{}}, \\
&absRel: \frac{1}{|M|}\sum_{p\in \mathbf{x}}^{}\frac{|f(\mathbf{x}_{}^{})-\mathbf{D}_{}|}{\mathbf{D}_{}},\\
&sqrRel  \frac{1}{|M|}\sum_{p\in M}^{}\frac{\lVert f(\mathbf{x}_{}^{})-\mathbf{D}_{}\rVert_{}^{} }{\mathbf{D}_{}},\\
&Bump:  \frac{1}{|M|}\sum_{p\in \mathbf{x}}^{}\min(0.05, \lVert \mathbf{H}_{\Delta}^{}(p) \rVert)\times 100,\\
&Accuracy (\delta): \max \Big(\frac{f(\mathbf{x}_{}^{})}{\mathbf{D}_{}},\frac{\mathbf{D}_{}}{f(\mathbf{x}_{}^{})}\Big)=\delta < threshold, \nonumber \\  & \% \enskip of \enskip \mathbf{D}_{},
\end{align}
where $\Delta = f(\mathbf{x}_{}^{})-\mathbf{D}_{}^{}$ and $\mathbf{H}$ is the Hessian matrix. The accuracy threshold of bumpiness (bump) is set at three levels ($1.25$, $1.25^2$, and $1.25^3$). 

\begin{table}[!thbp]
\centering
\vspace{-0.8em}
\caption{\small{Summary of evaluation datasets.}}
\vspace{-1.0em}
\label{tab:dataset_summary}
\begin{adjustbox}{width=0.80\linewidth}
\begin{tabular}{c c c c}
\toprule
  Dataset  &  Image source & GT type  &  Cause of defocus \\ 
  \hline
  DDFF 12-Scene \cite{hazirbas2018deep}  & Real          & Depth       & Light-field settings \\ 
  \hline
  Mobile Depth \cite{benavides2022phonedepth}  & Real          & ---      & Real\\
  \hline
  LightField4D \cite{honauer2016dataset}  & Real          & Disparity &   Light-field settings \\ 
  \hline
  FOD500 \cite{yang2022deep}    & Synthetic       & Depth       & Synthesis blendering\\ 
  \bottomrule 
\end{tabular}
\end{adjustbox}
\end{table}
\noindent \textbf{Implementation details.} For the evaluation on DDFF 12-Scene, we conducted training experiments using our model with and without pre-training on NYUv2 \cite{silberman2012indoor}, presenting results for both scenarios. We employed a patch size of $16\times16$ and an image size of $384\times384$ for the Transformer. Our network utilizes the Adam optimizer with a learning rate of $1\times10^{-4}$ and a momentum of 0.9. The regularization scalar $\alpha$ in Eq. \eqref{eq:total_loss} is set to 0.2. In terms of hardware configuration, all training and tests below were conducted on a single Nvidia RTX 2070 GPU with 8GB of VRAM.
\begin{table*}[!htbp]
\caption{Evaluation results on DDFF 12-Scene. The best results are denoted in \textbf{\textcolor{red}{Red}} while \underline{\textcolor{blue}{Blue}} indicates the second-best. $\delta = 1.25$.}
\label{tab:ddff-val}
\begin{small}
\begin{adjustbox}{width=0.96\linewidth}
\begin{tabular}{c c c c c c c c c c}
\hline
Model  & RMSE$\downarrow$ & logRMSE$\downarrow$ & absRel$\downarrow$ & sqrRel$\downarrow$ & Bump$\downarrow$& $\delta\uparrow$ & ${\delta}^{2} \uparrow$ & ${\delta}^{3} \uparrow$\\ 
\hline
DDFFNet \cite{hazirbas2018deep} & 2.91e-2 & 0.320 & 0.293 & 1.2e-2 & 0.59 & 61.95 & 85.14 & 92.98\\ 
\hline
DefocusNet \cite{maximov2020focus} & 2.55e-2 & 0.230 & 0.180 & 6.0e-3 & 0.46 & 72.56 & 94.15 & 97.92\\ 
\hline  
DFVNet \cite{yang2022deep} & 2.13e-2 & 0.210 & \underline{\textcolor{blue}{0.171}} & 6.2e-3 & 0.32 & 76.74 & 94.23 & 98.14\\ 
\hline
AiFNet \cite{wang2021bridging} & 2.32e-2 & 0.290 & 0.251 & 8.3e-3 & 0.63 & 68.33 & 87.40 & 93.96\\ 
\hline
Ours (w/o Pre-training)  & \underline{\textcolor{blue}{2.01e-2}} & \underline{\textcolor{blue}{0.206}} & 0.173 & \underline{\textcolor{blue}{5.7e-3}} & \underline{\textcolor{blue}{0.26}} & \underline{\textcolor{blue}{78.01}} & \underline{\textcolor{blue}{95.04}} & \underline{\textcolor{blue}{98.32}}\\ 
\hline
Ours (w/ Pre-training) & \textbf{\textcolor{red}{1. 96e-2}} & \textbf{\textcolor{red}{0.197}} & \textbf{\textcolor{red}{0.161}} & \textbf{\textcolor{red}{5.4e-3}} & \textbf{\textcolor{red}{0.23}} & \textbf{\textcolor{red}{79.06}} & \textbf{\textcolor{red}{96.08}} & \textbf{\textcolor{red}{98.57}}\\ 
\hline  
\end{tabular}
\end{adjustbox}
\end{small}
\vspace{-0.2em}
\end{table*}

\begin{table*}[!htbp]
\caption{Evaluation results on FOD500 test dataset. Here the first 400 FOD500 focal stacks are used for training, following the standard setting from DFVNet \cite{yang2022deep}.
The best results are denoted in \textbf{\textcolor{red}{Red}}, while \underline{\textcolor{blue}{Blue}} indicates the second-best. $\delta = 1.25$.}
\label{tab:ddff-fod}
\begin{small}
\begin{adjustbox}{width=0.96\linewidth}
\begin{tabular}{ c c c c c c c c c c}
  \hline
  Model  & RMSE$\downarrow$ & logRMSE$\downarrow$ & absRel$\downarrow$ & sqrRel$\downarrow$ & Bump$\downarrow$& $\delta\uparrow$ & ${\delta}^{2} \uparrow$ & ${\delta}^{3} \uparrow$ \\ 
  \hline
  DDFFNet \cite{hazirbas2018deep} & 0.167 & 0.271 & 0.172 & 3.56e-2 & 1.74 & 72.82 & 89.96 & 96.26\\ 
  \hline
  
  DefocusNet \cite{maximov2020focus} & 0.134 & 0.243 & 0.150 & 3.59e-2 & 1.57 & 81.14 & 93.31 & 96.62\\ 
  \hline  
  
  DFVNet \cite{yang2022deep} & \underline{\textcolor{blue}{0.129}} & \underline{\textcolor{blue}{0.210}} & \underline{\textcolor{blue}{0.131}} & \underline{\textcolor{blue}{2.39e-2}} & \underline{\textcolor{blue}{1.44}} & 81.90 & \underline{\textcolor{blue}{94.68}} & \underline{\textcolor{blue}{98.05}} \\ 
  \hline
  
  AiFNet \cite{wang2021bridging} & 0.265 & 0.451 & 0.400 & 4.32e-1 & 2.13 & \underline{\textcolor{blue}{85.12}} & 91.11 & 93.12\\ 
  \hline
  
  
  Ours (w/o Pre-training) & \textbf{\textcolor{red}{0.121}} & \textbf{\textcolor{red}{0.203}} & \textbf{\textcolor{red}{0.129}} & \textbf{\textcolor{red}{2.36e-2}} & \textbf{\textcolor{red}{1.38}} & \textbf{\textcolor{red}{85.47}} & \textbf{\textcolor{red}{94.75}} &  \textbf{\textcolor{red}{98.13}}\\ 
  \hline  
\end{tabular}
\end{adjustbox}
\end{small}
\vspace{-0.2em}
\end{table*}


\noindent \textbf{Runtime.}
We evaluated the runtime of the proposed method and baseline approaches by executing them on focal stacks from DDFF 12-Scene. Our FocDepthFormer processes a stack with 10 images sequentially in 15ms, an average of 2ms per image. In comparison, DDFFNet \cite{hazirbas2018deep} requires 200ms for each stack under the same conditions, and DFVNet \cite{yang2022deep} performs in the range of 20-30ms. 

\subsection{Baseline Comparisons}
DDFFNet \cite{hazirbas2018deep} and DefocusNet \cite{maximov2020focus} lacked pre-trained weights; therefore, we utilized their open-source codebases to train the networks from scratch. Notably, DefocusNet \cite{maximov2020focus} offers two architectures, and we chose the ``PoolAE" architecture due to its consistently good performance for comparison. Conversely, for AiFNet \cite{wang2021bridging} and DFVNet \cite{yang2022deep}, we employed the pre-trained weights provided by the authors to conduct the following evaluations.

\begin{figure}[!thbp]
\begin{tabularx}{\linewidth}{c c c c  c c c}
\centering
    \hspace{0.2cm} Input & \hspace{-0.2cm} GT & \hspace{-0.5cm} DDFF & \hspace{-1.0cm} DefocusNet & \hspace{-0.8cm}AiFNet & \hspace{-0.6cm}DFVNet & \hspace{-0.6cm} Ours  \\
\end{tabularx}
  \includegraphics[trim=0cm 0.5cm 0.5cm 0cm, width=1.0\linewidth]{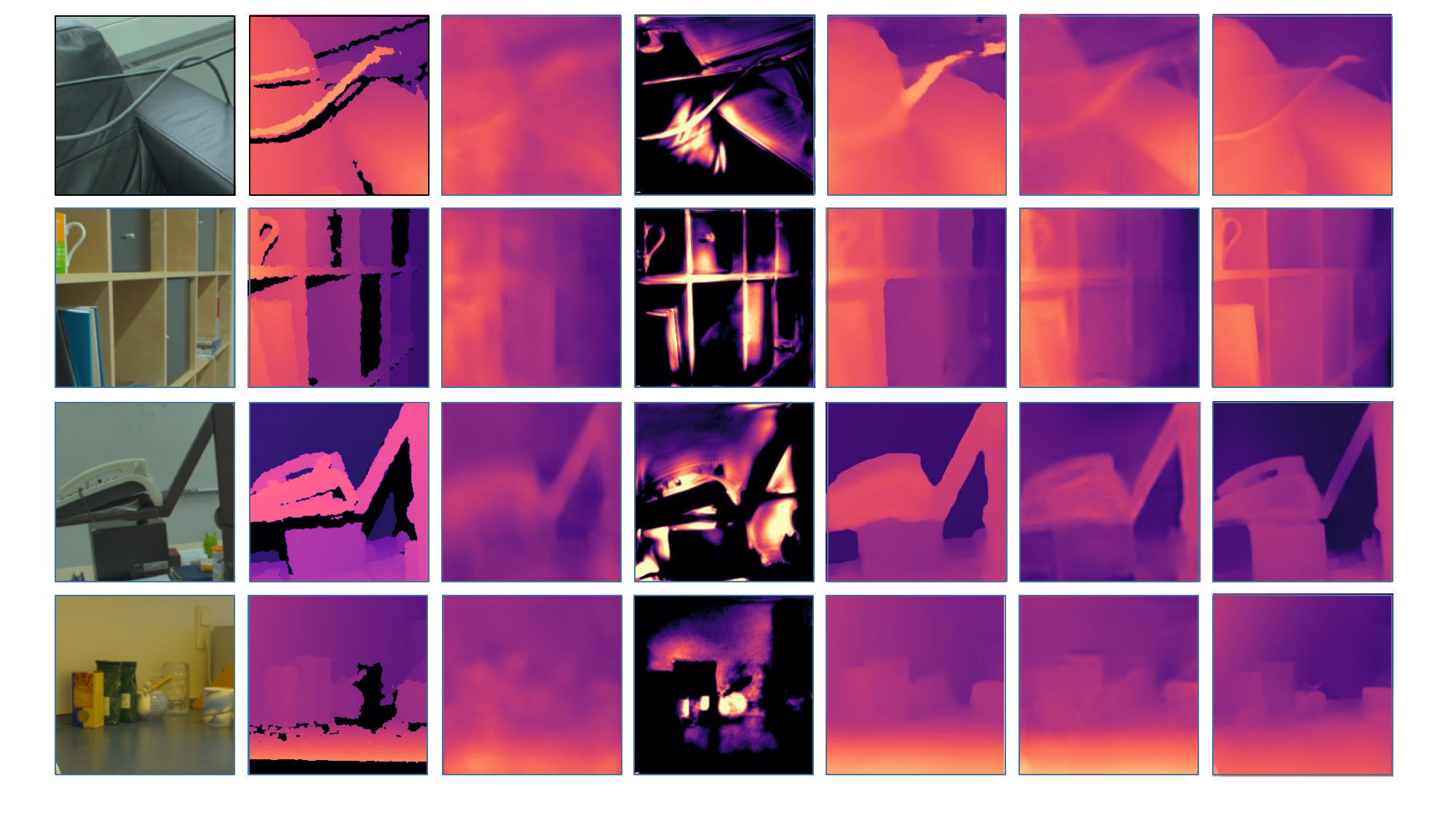}
\centering
  \caption{Qualitative evaluation of our model on DDFF 12-Scene dataset.}
  \label{fig:cmp-ddff}  
\end{figure}

\noindent \textbf{Results on DDFF 12-Scene.} Table \ref{tab:ddff-val} presents the quantitative evaluation results of our model on the DDFF 12-Scene dataset. As the ground truth for the ``test set" is not publicly available, we adhere to the standard evaluation protocol used in other comparative works, assessing the models on the ``validation set" as per the split provided by DDFFNet \cite{hazirbas2018deep}. Moreover, we demonstrate that our model, trained on DDFF-12, performs robustly on other completely unseen datasets, highlighting its generalization ability and mitigating concerns of over-fitting. The results in the table indicate that our model without pre-training outperforms prior models on all metrics, except for absRel compared to DFVNet. Specifically, our model achieves an accuracy of $78.01\%$ ($\delta = 1.25$), marking a notable improvement of $1.27\%$ over DFVNet. The Bumpiness metric also reflects this superiority with a value of $0.26$, one-third less than DFVNet. This can be attributed to the compact design of the Transformer and LSTM, efficiently learning spatial features and stack features separately. Examining the table, pre-training brings considerable advantages across all metrics, particularly for absRel and Bump, with improvements of around 7\% and 12\%, respectively, over pure training. This underscores the potential of our model. Figure \ref{fig:cmp-ddff} illustrates the qualitative performance of our model on DDFF 12-Scene. The visualizations showcase depth estimation results preserving fine-grained details such as the thin wire over the sofa (first row) and the cup handle on the shelf (second row).\\
\setlength{\tabcolsep}{15pt}
\renewcommand{\arraystretch}{1}
\noindent\textbf{Results on FOD500.} Table \ref{tab:ddff-fod} presents the quantitative evaluation of our model on the synthetic FOD500 dataset. For testing, we use the last 100 image stacks from the dataset, while the initial 400 image stacks are reserved for training. DDFFNet and DefocusNet are re-trained on FOD500 from scratch. The results highlight the consistent superiority of our model across all metrics when compared to the baseline methods. Notably, in our experiments, we observed that pre-training on NYUv2 did not provide significant benefits, likely due to the gap between synthetic and real data. Interestingly, we also noted that the proposed method can achieve satisfactory and competitive results with only a few training epochs. 
\begin{figure}[!thbp]
\begin{tabularx}{\linewidth}{c c  c c  c c c}
\centering
    \hspace{1.1cm} Input & \hspace{-0.6cm} DDFF & \hspace{-1.0cm} DefocusNet &\hspace{-1.0cm} AiFNet & \hspace{-0.8cm} DFVNet & \hspace{-0.5cm}Ours  \\
\end{tabularx}
 \centering
    \includegraphics[trim=0cm 8.8cm 0.5cm 0.0cm, width=1.0\linewidth]{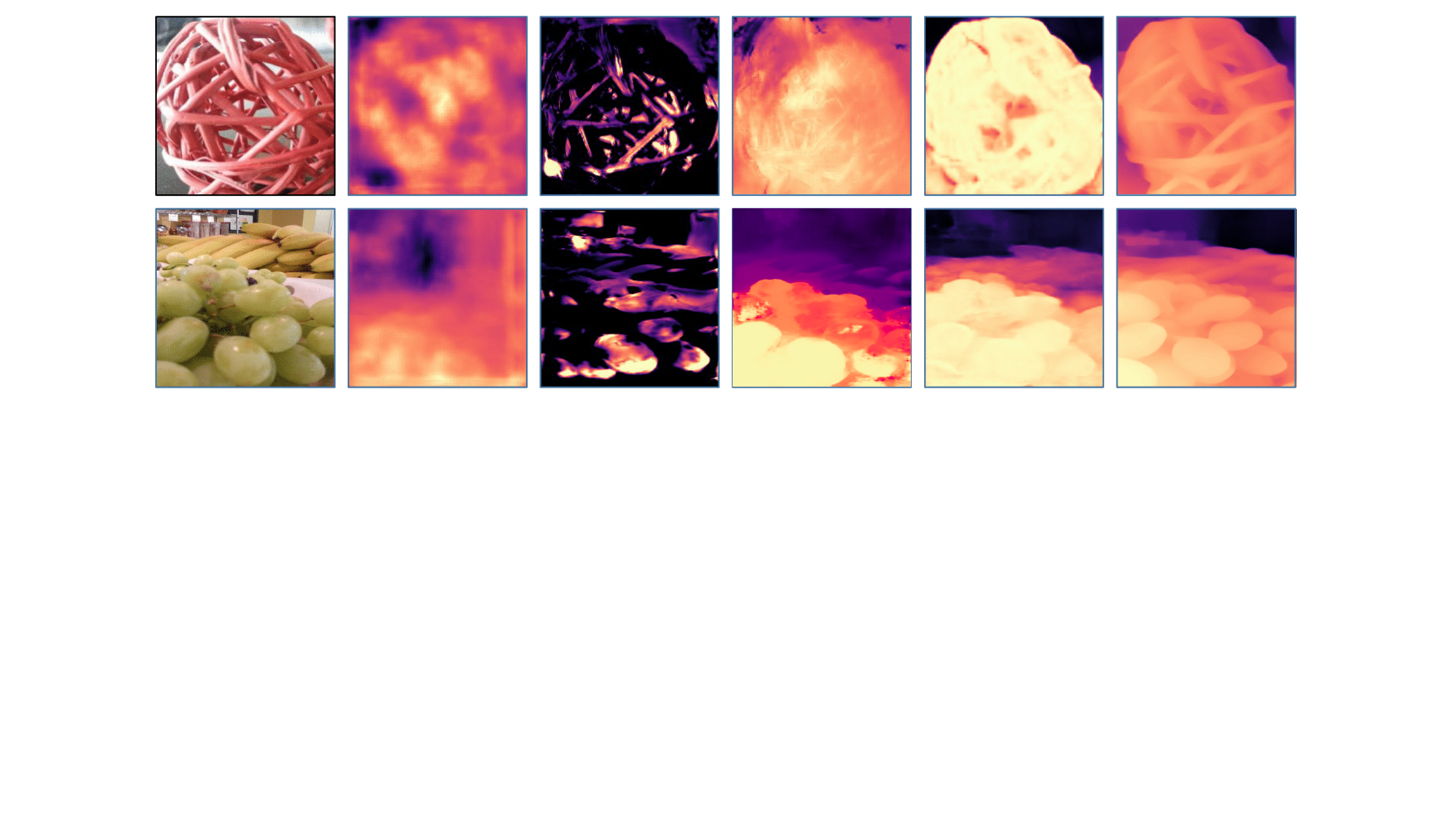}
  \caption{Qualitative evaluation of our model on Mobile Depth dataset.}
  \label{fig:cmp-mobile}  
  \vspace{-1.2em}
\end{figure}

\begin{figure*}[!thbp]
\begin{tabularx}{\linewidth}{c c c c  c c c}
    \hspace{0.2cm} Input & \hspace{-0.3cm} GT & \hspace{-0.3cm} DDFF & \hspace{-0.9cm} DefocusNet & \hspace{-0.9cm}AiFNet & \hspace{-0.7cm}DFVNet & \hspace{-0.6cm} Ours  \\
\end{tabularx}
  \centering
      \includegraphics[trim=0cm 10.8cm 0.8cm 0cm, width=1.0\linewidth]{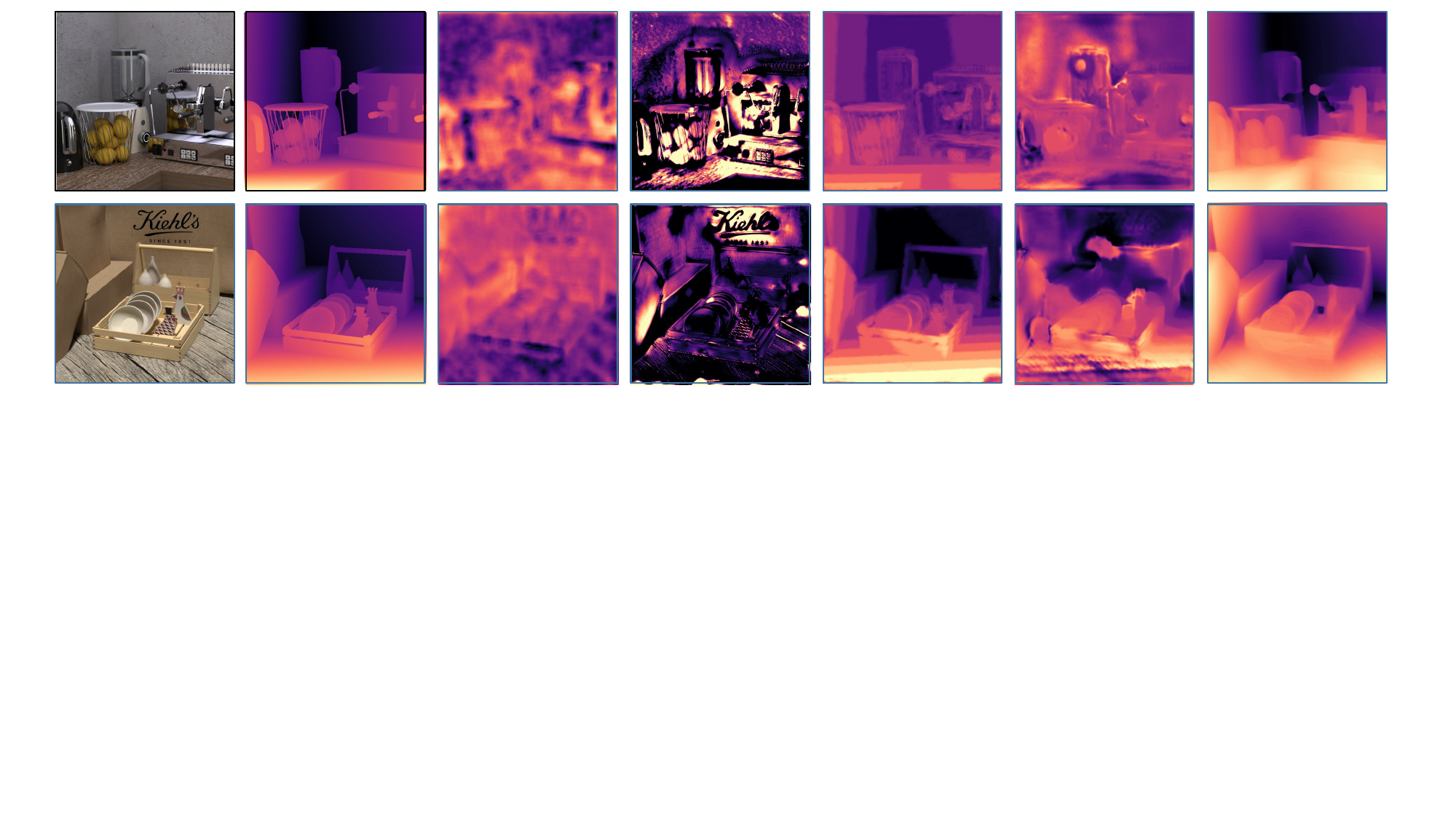}
  \caption{Qualitative evaluation of our model on LightField4D dataset.}
  \label{fig:cmps-lf4d}  
\vspace{-0.4em}
\end{figure*}
\subsection{Cross Dataset Evaluation}
To evaluate the generalizability of our model, it is initially trained on DDFF 12-Scene and subsequently evaluated on the Mobile Depth and LightField4D datasets. The Mobile Depth dataset poses a challenge as it comprises 11 aligned focal stacks captured by a mobile phone camera, each with varying numbers of focal planes and lacking ground truth. Results on the Mobile Depth dataset are illustrated in Figure \ref{fig:cmp-mobile}, showcasing the model's ability to preserve sharp information for depth prediction in complex scenes, such as the ball with many holes in the first row. Notably, the model excels in recognizing fine details in complex topological structures, such as grape granules. Additionally, our model effectively fuses depth information across a diverse range of scenes, including backgrounds like bananas in the second row. An advantage of our model is its capability to handle varying numbers of input images, a feature not supported by baseline methods that are restricted to fixed training settings. For generalization tests on the LightField4D dataset, our model, along with other baseline models, is pre-trained exclusively on DDFF 12-Scene, except for AiFNet \cite{wang2021bridging}, which is also pre-trained on LightField4D all-in-focus color images in an unsupervised manner. Visual results are presented in Figure \ref{fig:cmps-lf4d}, illustrating that our model produces more accurate depth maps for complex objects compared to AiFNet \cite{wang2021bridging}. Overall, these test results affirm that our model demonstrates comparable generalization performance to previous models without specific pre-training on specific focal stack datasets. 
We present the quantitative results for cross-dataset evaluation of our model on the LightField4D dataset in Table \ref{tab:lightfield4d}. Our model achieves a comparable performance in terms of accuracy ($58.90$\%) on this completely unseen dataset. Although the AiFNet can attain the least RMSE ($0.231$) and logRMSE($0.407$) error, our model generates smoother boundaries and fine-grained details, indicated by a lower bumpiness value ($2.53$). We use the available pre-trained AiFNet on LightField3D which uses the all-in-focus color image as supervision for evaluation. It also explains why AiFNet achieves satisfactory performance, and it further validates our model's good generability without using any supervision signal from this new focal stack dataset, while our model can achieve plausible performance only by learning a good stack distribution and spatial sharp feature representation from DDFF 12-Scene dataset. 
\begin{table}[!thbp]
\centering
\caption{Metric evaluation results on ``additional'' set of LightField4D dataset. The best results are denoted in \textbf{\textcolor{red}{Red}}, while \underline{\textcolor{blue}{Blue}} indicates the second-best.}
\begin{adjustbox}{width=0.82\linewidth}
\centering
\begin{tabular}{ c c c c c c}
  \hline
  Model  & RMSE$\downarrow$ & logRMSE$\downarrow$ & absRel$\downarrow$ & Bump$\downarrow$& $\delta(1.25)\uparrow$\\ 
  \hline
  DDFFNet& 0.431 & 0.790 & 0.761 & 2.93 & 44.39\\ 
  \hline
  DefocusNet & 0.273 & 0.471 & 0.435 & 2.84 & 48.73\\ 
  \hline  
  DFVNet & 0.352 & 0.647 & 0.594 & 2.97 & 43.54\\ 
  \hline
  AiFNet & \textbf{\textcolor{red}{0.231}} & \textbf{\textcolor{red}{0.407}} & \underline{\textcolor{blue}{0.374}} & \underline{\textcolor{blue}{2.53}} & \underline{\textcolor{blue}{55.04}}\\ 
  \hline
  Ours & \underline{\textcolor{blue}{0.237}} & \underline{\textcolor{blue}{0.416}} & \textbf{\textcolor{red}{0.364}} & \textbf{\textcolor{red}{1.54}} & \textbf{\textcolor{red}{58.90}}\\ 
  \hline  
\end{tabular}
\end{adjustbox}
\label{tab:lightfield4d}
\end{table}

\subsection{Ablation Study}
The following tests all use the DDFF 12-Scene validation set and we omit the data name for brevity.
\label{ablation}

\noindent\textbf{Transformer encoder:} Table \ref{tab:ddff-encoder-ablation} provides a comparative analysis between Transformer and CNN-based encoders, focusing on the vanilla Transformer, Swin Transformer, and the CNN-based DDFF-Net model. To ensure fairness in the comparison, we conducted experiments by excluding the early kernels from our model. The results reveal that the ViT-based model combined with LSTM has the highest accuracy in this experimental setup, showcasing an approximately 33\% improvement over the CNN-based DDFF-Net encoder in terms of RMSE.

\begin{table}[!thbp]
\centering
\caption{Metric evaluation of various encoders.}
\label{tab:ddff-encoder-ablation}
\begin{adjustbox}{width=0.74\linewidth}
\begin{tabular}{c c c c}
\hline
  & RMSE$\downarrow$ & absRel$\downarrow$ & Bump$\downarrow$ \\ 
  \hline
  ViT-base encoder &\bf{2.06e-2} & \bf{0.197} &  \bf{0.29} \\ 
  \hline
  Swin Transformer encoder &2.21e-2 & 0.205 &  0.32  \\ 
  \hline
  CNN encoder & 3.12e-2 & 0.268 & 0.46  \\   
  \hline 
\end{tabular}
\end{adjustbox}
\end{table}

We present the attention heat map of different focus or out-of-focus images patched over the whole image in Figure \ref{fig:heatmap-cmps}. The attention of the Transformer module can attend to more in-focus feature information, \emph{e.g.}, the toy on the right side of Figure (d) is with higher attention compared to Figure (b) at the same location.  It shows that the patches can attend to the fore and background regions with related focus and defocus cues of the corresponding depth field quite well. It further manifests that the proposed model based on Transformer attention can differentiate the pixel sharpness variance on the image input. Although some attention is put wrongly, as in Figure (c), where the attention on the monitor and toy is incorrect, most of the attention is distributed consistently with the input patch appearance. Additionally, the attention of the chosen channel of the Transformer encoder for visualization is not scattered around the whole image, which further discloses that the attention is mainly focused on the similar semantic, sharpness, and appearance information of the input patch.

\par
\noindent\textbf{Multi-scale early-stage kernels:} Table \ref{tab:ddff-kernel-ablation} presents a comparison of different early CNN kernel design configurations in conjunction with our Transformer encoder (ViT). The effectiveness of the proposed \emph{early-stage multi-scale kernels} encoder is evident, demonstrating robust performance. Conversely, omitting multi-scale kernels or forgoing subsequent convolutions after in-parallel convolutions leads to a degradation in model performance.
\begin{table}[!thbp]
\centering
\caption{Results of different designs of the early-stage conv. 
}
\label{tab:ddff-kernel-ablation}
\begin{adjustbox}{width=0.80\linewidth}
\begin{tabular}{c c c c}
\hline
  & RMSE$\downarrow$ & absRel$\downarrow$ & Bump$\downarrow$ \\ 
  \hline
  \emph{multi-scale kernels} w/ depth convos + ViT & \bf{2.01e-2} & \bf{0.173} & \bf{0.26} \\
  \hline
 Constant kernel size at $3\times3$ + ViT & 2.18e-2 & 0.216 & 0.31 \\ 
  \hline
  \emph{multi-scale kernels} w/o depth convos + ViT & 2.27e-2 & 0.229 & 0.29 \\ 
  \hline 
\end{tabular}
\end{adjustbox}
\vspace{-0.0em}
\end{table}

\begin{figure}[!thbp]
\centering
    \subfloat[\centering 1st image.\label{fig:rgb-1st}]{
    \includegraphics[width=0.28\linewidth]{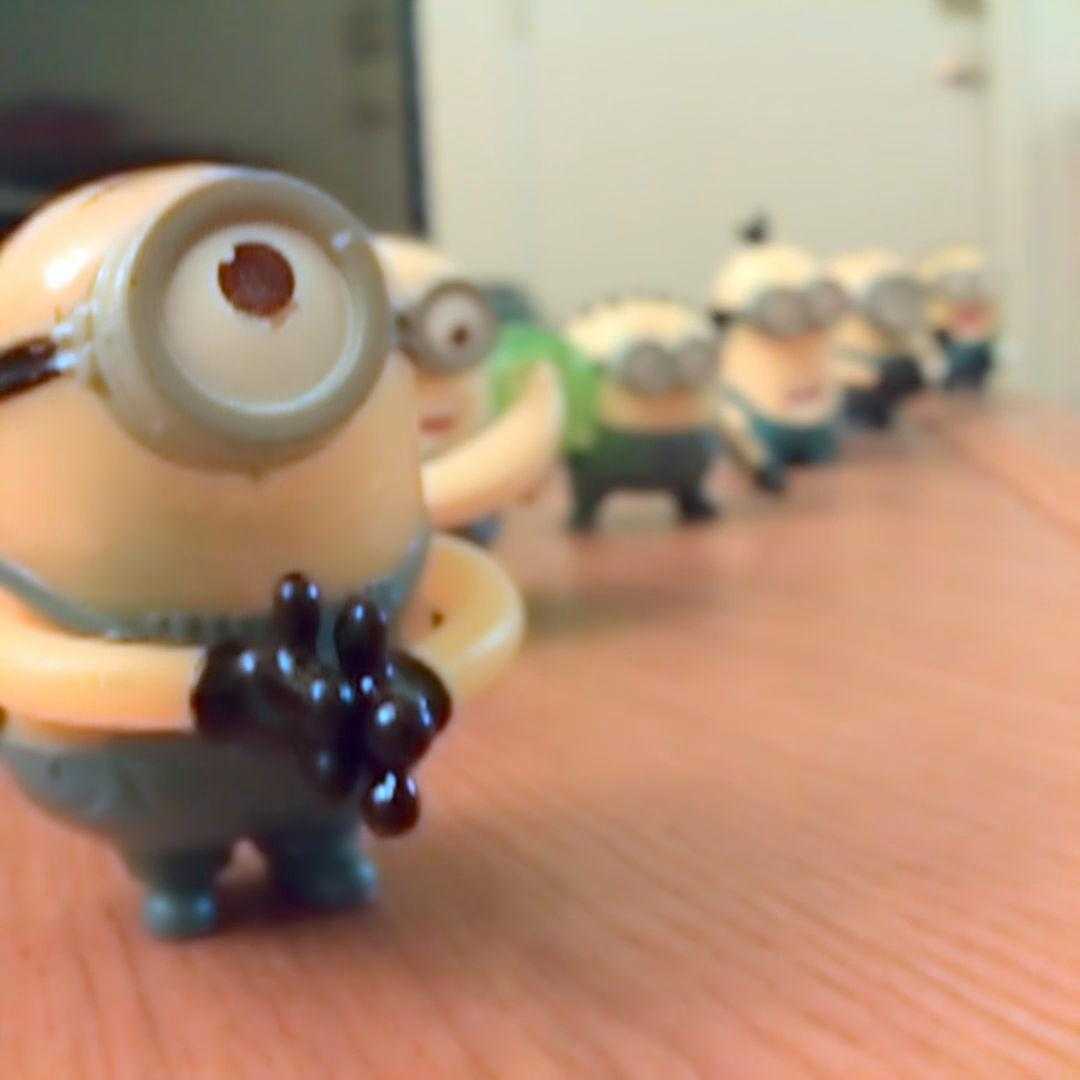}
    }
    \subfloat[\centering 5th image.\label{fig:rgb-5th}]{
    \includegraphics[width=0.28\linewidth]{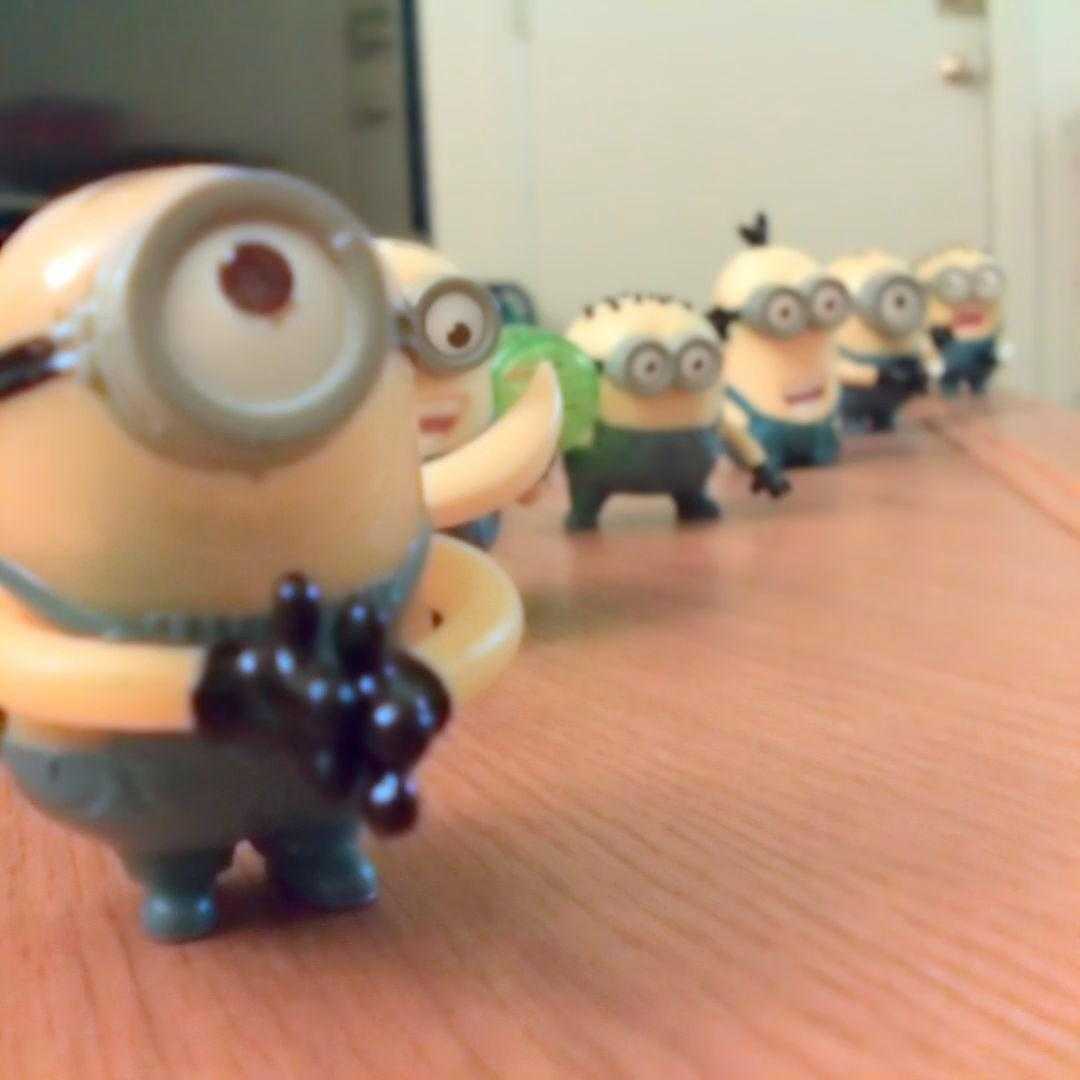}
    }
    \subfloat[\centering 10th image.\label{fig:rgb-10th}]{
    \includegraphics[width=0.28\linewidth]{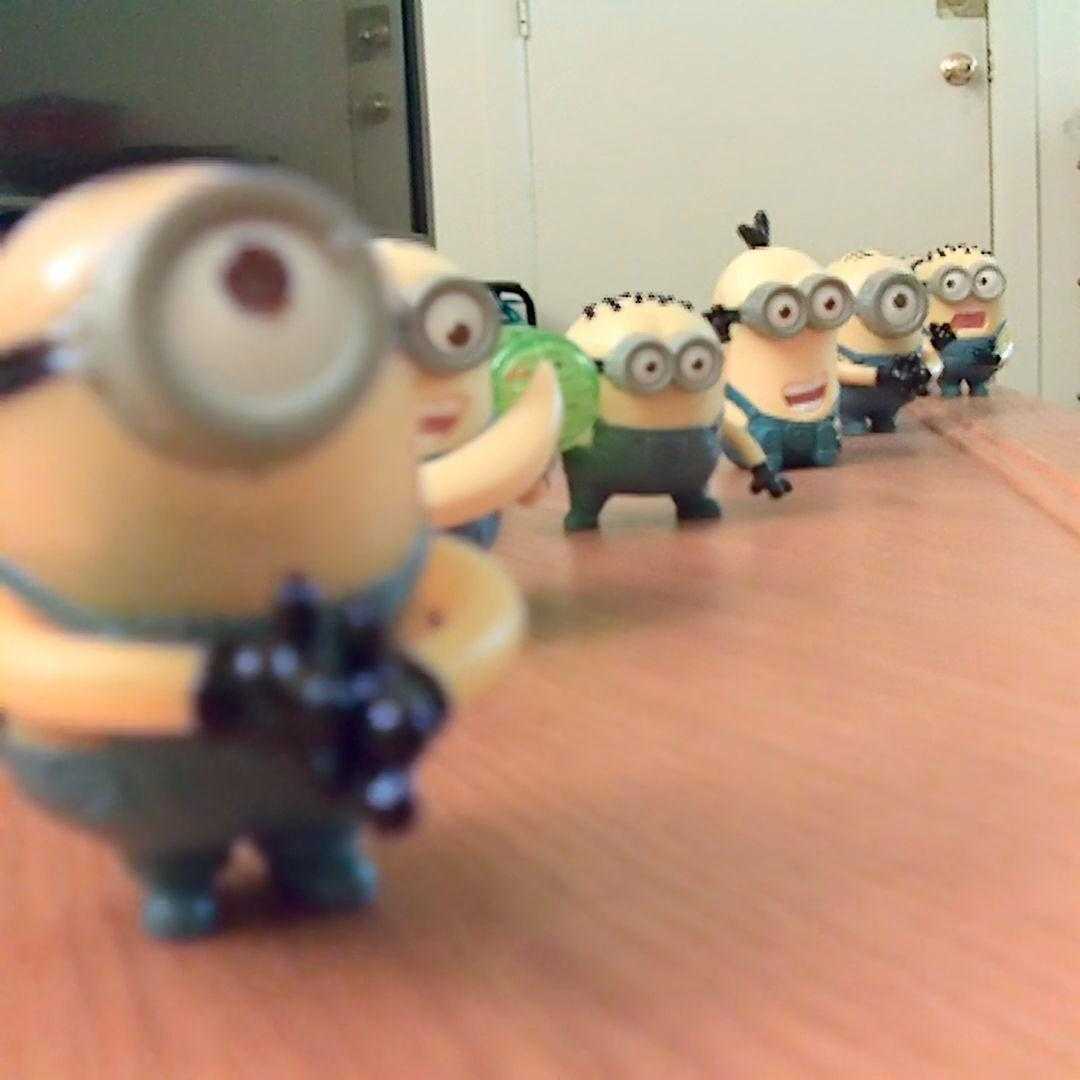}
    }
    \\
    \subfloat[\centering Depth prediction of 1st input frame.\label{fig:depth-1st}]{
    \includegraphics[width=0.28\linewidth]{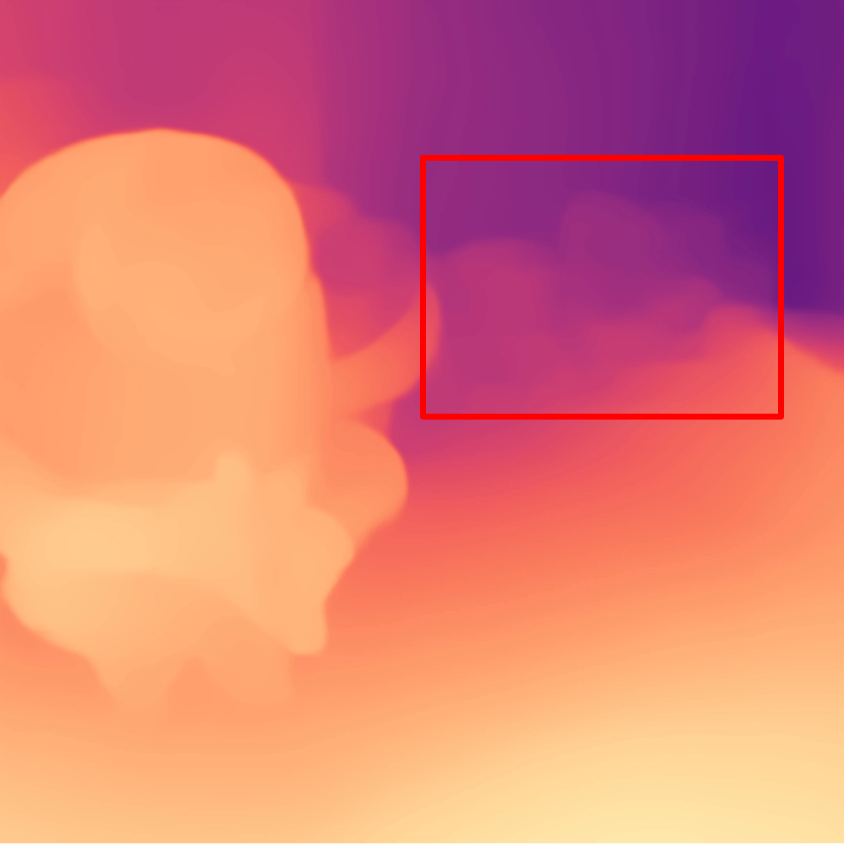}
    }
    \subfloat[\centering Depth prediction of two input frames.\label{fig:depth-2nd}]{
    \includegraphics[width=0.28\linewidth]{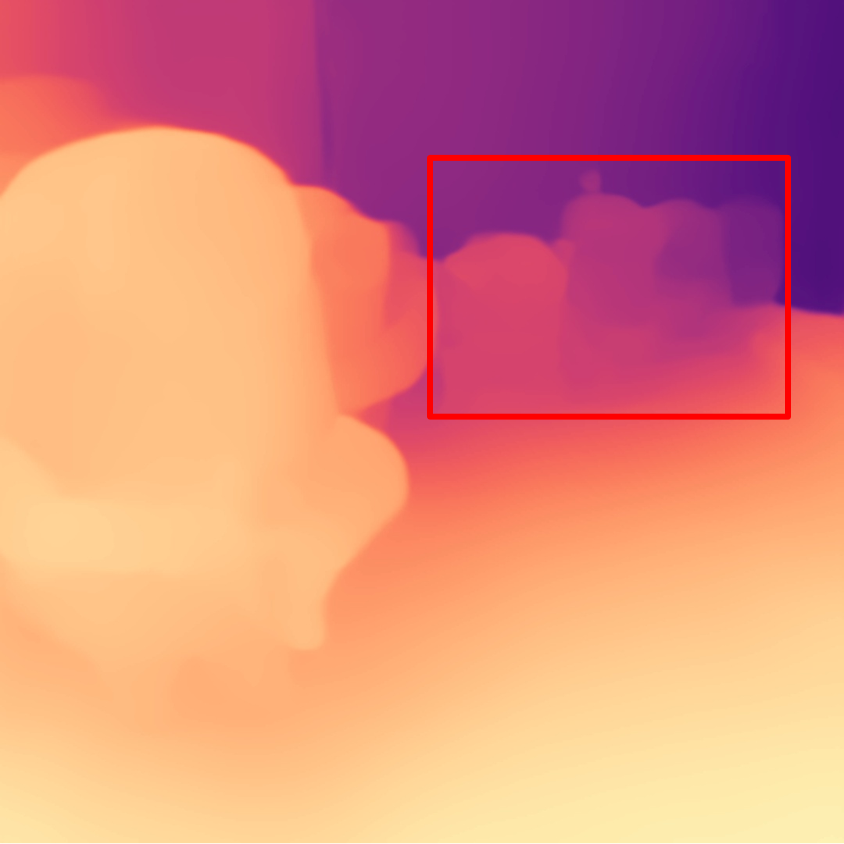}
    }
    \subfloat[\centering Fusion prediction of 3 input frames.\label{fig:depth-3rd}]{
    \includegraphics[width=0.28\linewidth]{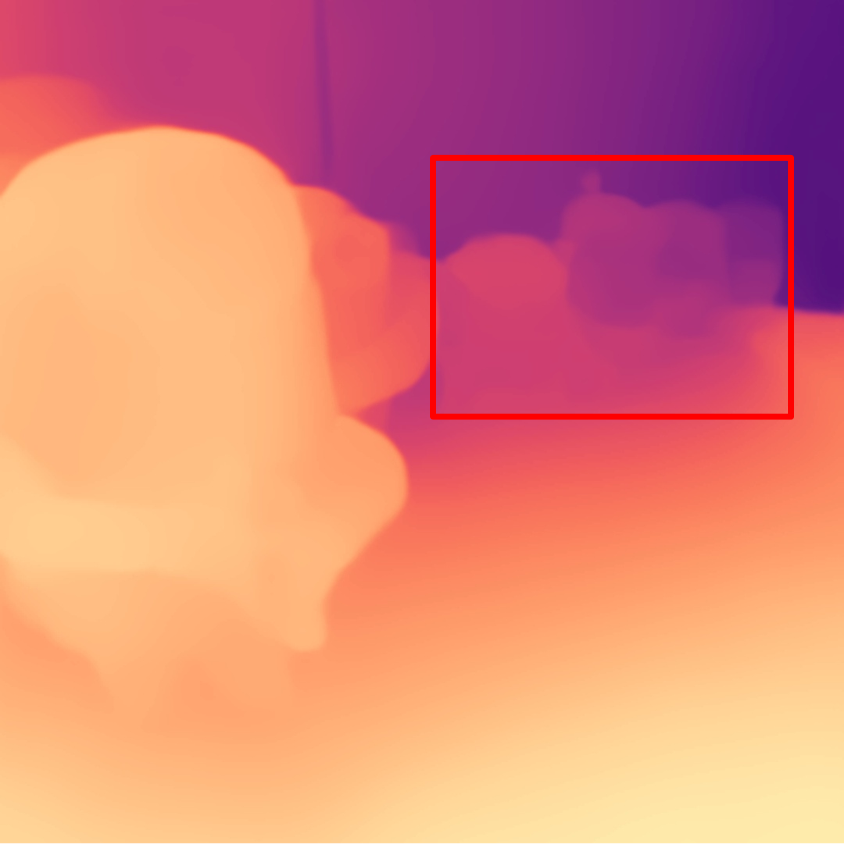}
    }    
\caption{The top row is the input, and the bottom is the output disparity map. The final disparity map is at the bottom right. The red rectangle highlights the incremental fusion results of depth information in the background.}
\label{fig:LSTM-image}
\end{figure}

\noindent\textbf{Fusion by LSTM.} The LSTM module facilitates our network to fuse each image from the focal stack incrementally, which extends the model capability to varying focal stack lengths. Figure \ref{fig:LSTM-image} depicts the fusion process of ordered input images of one stack. As the images from the stack are given sequentially, starting from the in-focus plane close to the camera, the model can fuse the sharp in-focus features from various frames to attain a final all-in-focus prediction depth map at the bottom right figure.

In table \ref{tab:ddff-LSTM-ablation}, we compare our model to its base model without the LSTM module. For the performance without LSTM, the encoder, and decoder are connected directly, and all the depth maps of each image in the stack are averaged to get the metric results of a whole stack. The results indicate the necessity and importance of LSTM in-depth estimation from the focal stack problem. It further validates that the modeling stack information separately from the image spatial features can help to improve depth prediction accuracy.

\begin{table}[!thbp]
\centering
\caption{\small{Metric evaluation on different settings for LSTM module on DDFF 12-Scene validation dataset.}}
\begin{adjustbox}{width=0.78\linewidth}
\begin{tabular}{c c c c}
\hline
  Structure design & RMSE$\downarrow$ & absRel$\downarrow$ & Bump$\downarrow$ \\ 
  \hline    
  Model w/o LSTM module &3.68e-2 & 0.324 & 0.37 \\
  \hline  
  Full model &\bf{1.92e-2} & \bf{0.161} & \bf{0.19}  \\
  \hline
\end{tabular}
\end{adjustbox}
\label{tab:ddff-LSTM-ablation}
\end{table}
To justify the model structure design, we further implement the tests by replacing The Transformer encoder with the CNN encoder, and then the features after convolution are fused by LSTM. The pure Transformer-based encoder without LSTM fusion takes the image stack as a whole, and then concatenates the feature maps to be decoded into the depth map, 
\begin{figure}[!th]
    \centering
    \subfloat[\centering Ours \label{fig:ours-lstm}]{
    \includegraphics[width=0.32\linewidth]{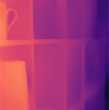}
    }
    \subfloat[\centering LSTM+CNN \label{fig:cnn=lstm}]{
    \includegraphics[width=0.32\linewidth]{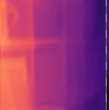}
    }
    \subfloat[\centering Transformer \label{fig:trans}]{
    \includegraphics[width=0.32\linewidth]{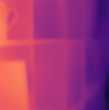}
    }
    \caption{Different model structures' comparison.}
    \vspace{-0.2em}
\label{fig:lstm-cmbs}
\end{figure}
The Transformer + LSTM design proposed in the paper can predict a more detailed feature map with fine-grained details, while the naive concatenation of feature maps after the Transformer can not achieve the equivalent performance (as reflected in the middle image) compared to our proposed structure, by combining the focus/defocus cues directly. The CNN  encode even with the help of LSTM can not learn a good image feature representation after the encoder for depth map prediction, due to the local receptive field limits, resulting in the blur effect in the final prediction. It is noteworthy that for pure Transformer without LSTM, we downsample the focal stack size to three images for these comparison experiments, as the original size of the image results in out-of-memory in pure Transformer-based encoder implementation, which takes the whole stack of images at once and processes them with multiple encoders in parallel.

\par
\begin{figure}[!th]
\centering
\includegraphics[width=\linewidth]{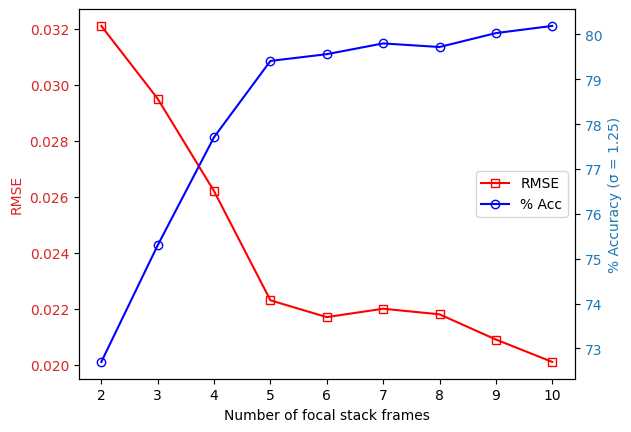}
\caption{Our model performance w.r.t. the frame size of one focal stack sample from DDFF 12-Scene test.}
\label{fig:stack-num}  
\end{figure}

\noindent\textbf{LSTM for handling arbitrary stack length.} Our proposed LSTM-based method exhibits flexibility in processing focal stacks of arbitrary lengths, a feature distinguishing it from designs that limit inputs to fixed lengths. To illustrate the advantages of our LSTM-based model, we conducted experiments using DDFF 12-scene data. The results, including the RMSE comparison between our model and DFVNet \cite{yang2022deep}, are presented in Table \ref{tab:dfv-ours-cmp}. Initially, we trained our model and DFVNet using 10-frame (10F) stacks (\textit{i.e.,} Ours-10F and DFVNet-10F). During testing, DFVNet-10F is constrained and cannot process stacks with fewer than 10 frames. In contrast, due to the fixed stack size requirements during training and testing, DFVNet must be retrained for different stack sizes (DFVNet-\#F). In comparison, our model is trained once on 10-frame stacks and can be directly tested with varying numbers of stack images. The focal stack images are ordered based on focus distances. Despite our model's initial performance being inferior to DFVNet, the learning curve of LSTM indicates rapid convergence.
\begin{table}[!thbp]
\centering
\caption{RMSE for evaluation of LSTM compared to DFVNet.}
\label{tab:dfv-ours-cmp}
\begin{adjustbox}{width=0.96\linewidth}
\begin{tabular}{ c c c c c c}
  \hline
  Model & 2 Frames & 4 Frames & 6 Frames & 8  Frames & 10 Frames\\ 
  \hline
  Ours-10F & 3.2e-2 & \textbf{2. 61e-2} & \textbf{2.18e-2} & \textbf{2.16e-2} & \textbf{2.04e-2}\\
  \hline 
  DFVNet-10F & ---- & ---- & ---- & ---- & 2.43e-2 \\
  \hline  
  DFVNet-\#F & \textbf{2.97e-2} & 2.70e-2 & 2.52e-2 & 2.47e-2 & 2.43e-2 \\
  \hline    
  \end{tabular}
\end{adjustbox}
\centering
\end{table}
We perform experiments to observe the results of our model on various focal stack sizes. Figure \ref{fig:stack-num} shows the findings of our experiments on a focal stack from the DDFF 12-Scene validation set, with multiple objects placed at varying depths. In the figure, we plot RMSE and accuracy (in percentage) ($\delta$ = 1.25) w.r.t. the number of focal stack images. We observe that the model accuracy increases with more input images from the focal stack in use, while the RMSE decreases correspondingly. Our model achieves a decent performance around six frames with a notable increase, then followed by a marginal increase after six frames.

To evaluate the impact of the token $L_2$ norm threshold, we further tested the RMSE under the various thresholds. A lower threshold smaller than 0.4 like 0.3 can improve the RMSE by a marginal increase at around 0.6\%, yet scaling the model parameter complexity a lot, because more and more tokens are activated, and they are thrown into the LSTM module for processing. The threshold lower than 0.3 can even lead to out-of-memory issues. Conversely, the large threshold can reduce the LSTM number in use but also sacrifice the accuracy raging from 0.8\% to 1.2\% as the threshold increased from 0.5 to 0.8.
\noindent\textbf{Loss Function:} Table \ref{tab:ddff-loss-ablation} presents the results obtained by employing three distinct loss functions: Mean Squared Error (MSE), Mean Absolute Error (MAE), and Regularized MSE (MSE with a gradient regularizer). The findings indicate that MSE loss consistently outperforms MAE in terms of overall performance. Notably, the inclusion of the gradient regularizer contributes to achieving the highest accuracy, as evidenced by the bumpiness metric ($0.26$).

\begin{table}[!thbp]
\centering
\caption{Evaluation for our model with different losses.}
\label{tab:ddff-loss-ablation}
\begin{adjustbox}{width=0.80\linewidth}
\begin{tabular}{c c c c}
\hline
  & RMSE$\downarrow$ & absRel$\downarrow$ & Bump$\downarrow$ \\ 
  \hline
  MSE loss &2.94e-2 & 0.280 & 0.50  \\ 
  \hline
  MAE loss &3.76e-2 & 0.372 & 0.62\\ 
  \hline
  MSE + Gradient loss  & \bf{2.01e-2} & \bf{0.173} & \bf{0.26}  \\ 
  \hline 
\end{tabular}
\end{adjustbox}
\end{table}

\noindent\textbf{Pre-training:} Table \ref{tab:pre-train} illustrates the impact of pre-training on the performance of the proposed method. The results showcase an enhancement in the model's capabilities through pre-training, leveraging the advantages of the compact design with Transformer and LSTM. Even in the absence of pre-training, our model demonstrates competitive results. Notably, attempts were made to apply pre-training to DFVNet by creating a stack from repeated monocular images of NYUv2. However, the pre-training did not yield any improvement for DFVNet and, in some instances, led to performance degradation due to the data modality gap.
\begin{table}[!thbp]
\centering
\caption{Pre-training contribution comparisons.}
\label{tab:pre-train}
\begin{adjustbox}{width=0.96\linewidth}
\begin{tabular}{c c c c c c}
\hline
    & Pre-training & RMSE$\downarrow$ & logRMSE$\downarrow$ & absRel$\downarrow$ & Bump$\downarrow$\\%
\hline
    \multirow{2}{*}{Ours} & \xmark & 2.01e-2 & 0.206 & 0.173 & 0.26   \\ 
         & \cmark & \textbf{1.96e-2} &\textbf{0.197} & \textbf{0.161} & \textbf{0.19} \\ 
\hline
    \multirow{2}{*}{DFVNet} & \xmark & 2.13e-2 & 0.210 & 0.171 & 0.32\\
    & \cmark & 2.57e-2 & 0.233 & 0.184 & 0.49\\
\hline
\end{tabular}
\end{adjustbox}
\end{table}
\section{Conclusion}
\label{ch3-conclusion}

We introduced FocDepthFormer, a novel model for depth estimation from focal stacks. The key innovation lies in its hybrid architecture, which combines a Transformer encoder for capturing global spatial relationships with an LSTM module in latent space to effectively aggregate information across focal stacks of varying lengths. This design enhances flexibility and generalization compared to traditional CNN-based approaches.

While FocDepthFormer achieves strong performance, its use of Transformers leads to a higher memory footprint than simpler CNN-based models. Future work will explore more efficient Transformer architectures to mitigate this limitation. Additionally, we aim to extend our framework to image synthesis with defocus effects, further leveraging focal stacks for advanced 3D vision applications. \revDavood{Lastly, depth prediction from focal stacks faces challenges due to the limited depth range and the need for a sufficient number of focal stack images to estimate depth with adequate resolution. Successful training relies on distinct focus and defocus signals between images, which becomes difficult in textureless scenes, such as outdoor environments or plain walls, where such signals are absent. In these cases, inferring depth is highly ambiguous. A promising direction may involve combining focal stack techniques with state-of-the-art deep prior models trained on monocular or video inputs to recover fine-grained details more robustly.}
\chapter{3D Reconstruction Using Implicit SDF with Wavelet Feature-Based Prior}                                 
\label{ch:cp4}

\begin{center}
    \textbf{\large Abstract}
\end{center}

3D reconstruction aims to recover the underlying structure of a scene from sensor data, primarily LiDAR point clouds or multi-view images. Point clouds provide a direct geometric representation but are limited by sampling density and resolution. Multi-view images, on the other hand, leverage photometric cues for dense surface reconstruction. Recent advancements in deep learning have enabled implicit representations, such as Signed Distance Fields (SDF), which model continuous surfaces and can be converted into explicit 3D geometry using techniques like Marching Cubes. Implicit SDF models process both point cloud and image inputs. For point clouds, they estimate the signed distance of query points to the nearest surface, defining the zero-crossing isosurface. For multi-view images, SDF values are predicted along sample rays, optimizing the implicit representation for view-consistent rendering. While these models offer superior topological accuracy, they struggle to capture fine-grained geometric details due to high-frequency information loss during feature extraction, leading to suboptimal multi-scale representation. To address this limitation, we propose a wavelet-conditioned implicit SDF model that enhances geometric fidelity by integrating a pretrained wavelet autoencoder optimized with sharp depth maps\footnote{The majority of this chapter is based on the paper submitted to Transactions on Machine Learning Research (under review). \\
\textbf{Author Contributions:} \\
\vspace{-1.2em}
\begin{itemize}
    \item \textbf{Xueyang Kang}: Idea Design, Methodology, Software, Experiment Validation, Formal Analysis, Data Curation, Writing, Review, and Editing.
    \item Hang Zhao: Data Curation and Review.
    \item Kourosh Khoshelham: Review and Supervision. 
    \item Patrick Vandewalle: Review and Supervision.
\end{itemize}
}. This autoencoder extracts multi-scale wavelet-transformed features, which are fused with implicit 3D triplane representations, preserving structural details more effectively. Our approach serves as a plug-and-play module that seamlessly integrates with existing implicit SDF frameworks. Extensive evaluations on DTU, Tanks, and Temples, and a cultural heritage dataset demonstrate that our model outperforms state-of-the-art implicit and explicit methods, producing more complete and detailed 3D reconstructions across various scene scales—from small objects to large architectural structures.

\section{Introduction} 


3D reconstruction is closely tied to the underlying 3D representation format. Various representations exist, each with advantages and limitations. For instance, voxel grids discretize 3D space into uniform cubic elements, making them compatible with neural networks but suffering from cubic memory complexity $O(n^3)$, limiting resolution and introducing a "Manhattan world bias" \cite{maturana2015voxnet}. Meshes represent surfaces efficiently using vertices and faces as an approximation of continuous geometry representation, but mesh also has self-intersection issues for complex shapes \cite{park2019deepsdf}. Point clouds provide a simple, lightweight representation by recoding the 3D point coordinates \cite{fan2017point}, but such discrete point cloud representation lacks connectivity information, and the underlying topology is ignored.

A more recent alternative is implicit functions, which model shapes continuously rather than as discrete elements. These functions map 3D coordinates (and optional conditions like images) to either occupancy probabilities $[0,1]$ \cite{peng2020convolutional} or signed distance values $[-D, D]$ \cite{xu2019disn}, capturing fine details while being memory efficient. 

Before the advent of deep learning, Truncated Signed Distance Functions (TSDFs) were already widely used for 3D reconstruction. At first, a ground truth (GT) mesh is voxelized accordingly, with each voxel storing the signed distance value to the nearest mesh surface. The positive TSDF values indicate points outside the surface, while negative TSDF values indicate interior regions. The distance values are usually truncated to a fixed distance range centered on the surface, ensuring a finite update to avoid numeric instability. This volumetric representation is integrated into software product kits like KinectFusion \cite{izadi2011kinectfusion} for real-time 3D reconstruction from RGB and Depth views. However, TSDF-based methods are constrained by cubic memory complexity, limiting its scalability for large-scale scenes.

To address these limitations, implicit SDFs have been used as a powerful solution for 3D reconstruction. Instead of using explicit voxel grids with TSDF values, an implicit SDF represents the watertight surface as a neural function $\mathbf{f(x)}$, where MLP layers are trained to predict signed distance values for any query point $\mathbf{x}_i$. This approach enables high-fidelity surface reconstruction, moreover, it has a lower memory footprint and is unlimited to the input resolution.

\begin{figure}[!thbp]
\centering
\includegraphics[trim=1.5cm 0.5cm 1.5cm 0.5cm, width=0.84\textwidth]{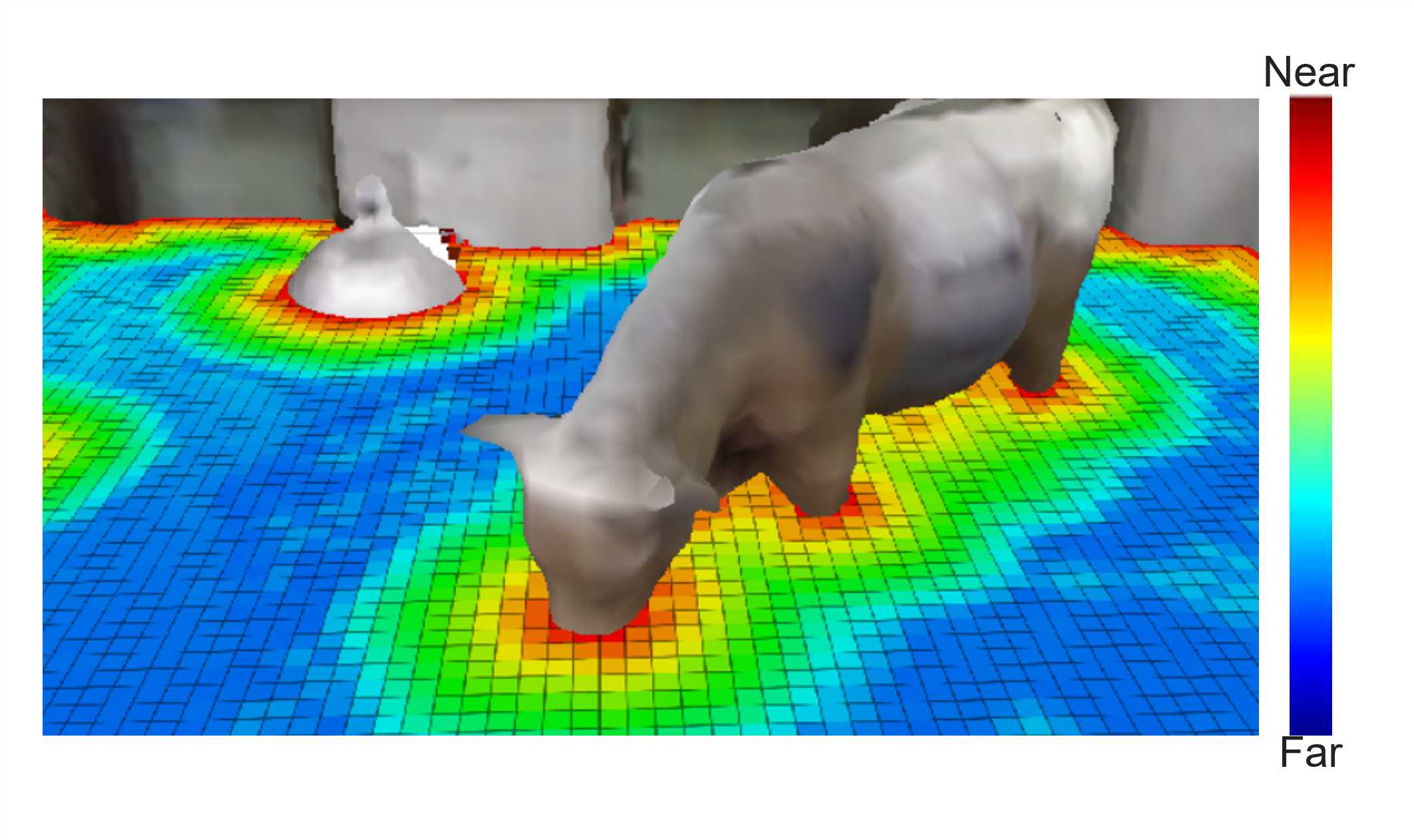}
\caption{2D cross-section of the implicit SDF volume space, where red indicates regions near the surface, and blue represents areas farther from the surface.}
\label{fig:sdf-demo}  
\end{figure}

Implicit SDFs can take in two modalities usually: point clouds and multi-view images. For point clouds, the model learns a continuous surface representation by fitting the implicit function to the sparse input point samples. For multi-view images, it leverages photometric consistency loss, along with some other geometric constraints like depth or normal losses, to learn 3D structure. The 2D cross-section snapshot in Figure \ref{fig:sdf-demo} illustrates how TSDF values are distributed in the proximity of the surface, from far (blue) to near (red) for the surface.

To efficiently generate Ground Truth (GT) SDF values for training an implicit SDF model, we primarily sample points near the surface. Points in free space, far from the surface, have large distance values, which can lead to gradient instability during training. By concentrating on points near the surface, the model learns a more stable and continuous function. Figure \ref{fig:points-multiview} illustrates this sampling strategy, with colors indicating point positions from left to right.

\begin{figure}[!thbp]
\centering
\includegraphics[width=0.80\textwidth]{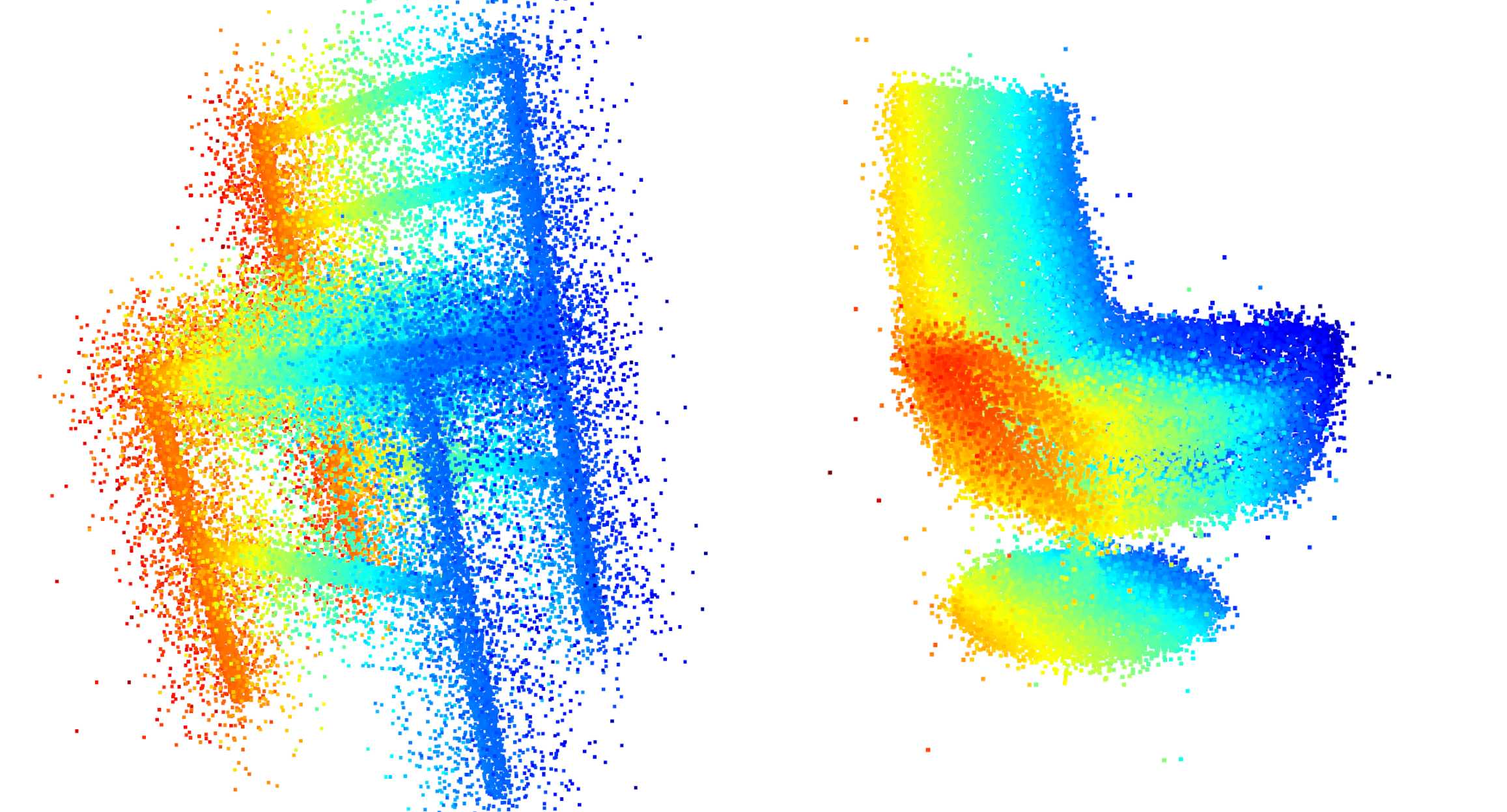}
\caption{Sampling points near the surface are generated to compute Signed Distance Values, providing training input for GT SDF calculation.}
\label{fig:points-sampling}  
\end{figure} 

\begin{figure}[!thbp]
\centering
\includegraphics[width=0.8\textwidth]{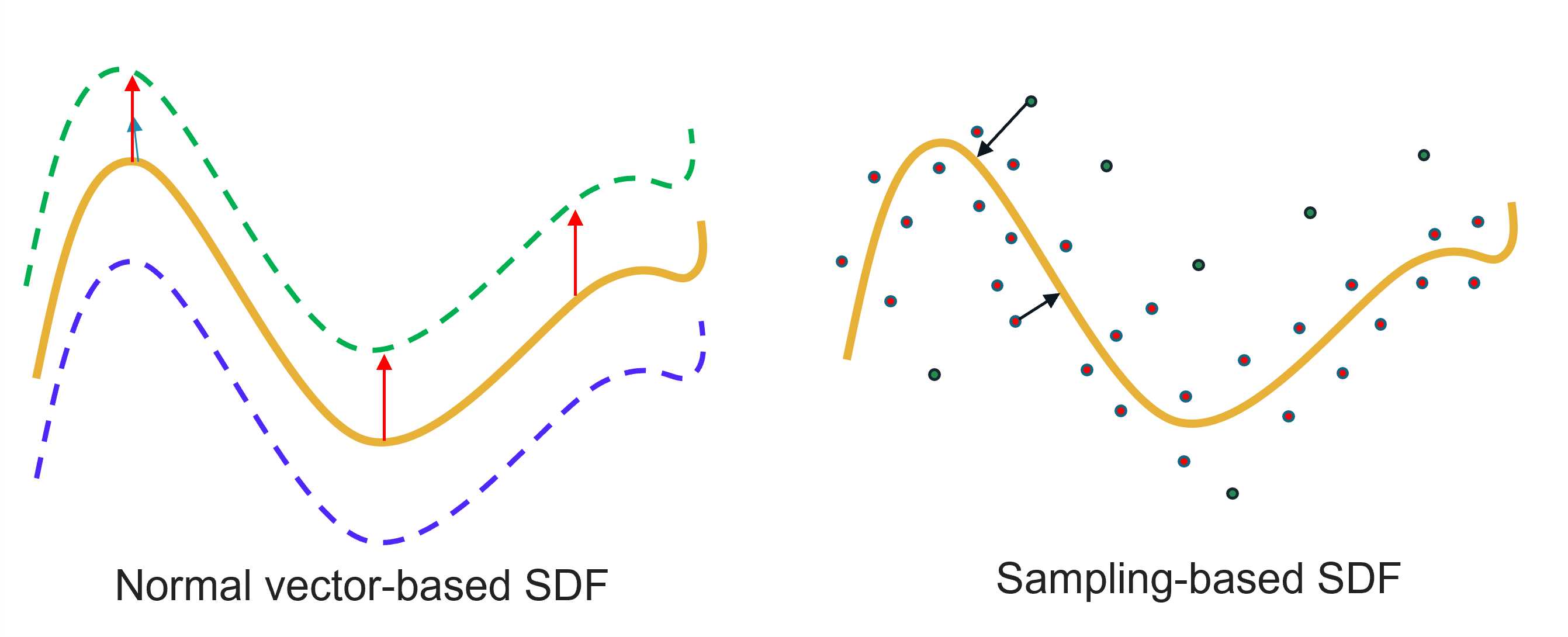}
\caption{GT SDF value generation can be achieved using two main approaches: surface normal vector-based sampling (left), where points are shifted along the normal direction and random sampling with Gaussian deviation within a narrow region around the surface.}
\label{fig:sdf-sampling}  
\end{figure}

Sampling methods in the surface region can be categorized into two approaches, as illustrated in Figure \ref{fig:sdf-sampling}. When the GT mesh is available, points are sampled by shifting along the surface normal (red arrow) and its opposite direction, scaled by a specified distance to create an envelope around the surface. The shift magnitude determines the SDF value.

Alternatively, points are randomly sampled within a predefined range around the surface, covering both interior and exterior regions. Their closest distance to the surface is then computed to obtain the corresponding SDF value.

Image-based 3D reconstruction methods, such as Structure from Motion (SfM), recover 3D structures from multi-view 2D images \cite{schoenberger2016sfm}, yet they are struggling to preserve high-fidelity details. Alternatively, reconstruction from structured light scans \cite{hu2022refractive, wang20223} or a fusion of images and LiDAR scans \cite{kang20183d, moemen20203} has seen active progress, but these point-based methods are prone to noise in scans, making it difficult to obtain a plausible reconstruction mesh. High-fidelity reconstruction with fine structural details remains a core challenge in computer vision. Recent advances in implicit representations, such as neural radiance fields (NeRF) \cite{mildenhall2021nerf}, and explicit methods such as Gaussian splatting (GS) proposed \cite{kerbl3Dgaussians}, have significantly advanced 3D applications.

Implicit models leverage photometric consistency loss to learn Signed Distance Fields (SDFs) from multi-view images \cite{hasson2020leveraging}. Unisurf \cite{oechsle2021unisurf} unifies surface and volume rendering to improve generalization, while hybrid volume-surface representations can be converted into high-quality meshes for real-time rendering, like the work \cite{yariv2023bakedsdf}. Multi-resolution hash grids further enable coarse-to-fine optimization for detailed neural surface reconstruction \cite{li2023neuralangelo}, making implicit SDF models effective for complex topologies and continuous geometry fields. \revPatrice{Image-based implicit reconstruction models recover geometry using multi-view photometric consistency under the Lambertian assumption of uniform light reflection. However, for reflective surfaces such as glass or transparent materials, the reconstruction pipeline should also explicitly model external glass planes or mirror surfaces \cite{whelan2018reconstructing, li2020through, qiu2023looking, tong2023seeing} to recover geometry from appearance features better while maintaining the consistency assumption.}

Explicit Gaussian splatting (GS) represents scenes by anisotropic 3D Gaussians \cite{kerbl3Dgaussians}, enabling efficient training and real-time rendering. However, while GS offers speed, it often sacrifices geometric quality. To address this, AGS-Mesh \cite{ren2024agsmesh} incorporates meshing priors, PGSR proposed \cite{chen2024pgsr} enforces planar constraints, DN-Splatter \cite{turkulainen2024dn} utilizes depth and normal priors for Gaussian-based reconstruction, and 2D GS \cite{Huang2DGS2024} simplifies 3D Gaussian parameters to improve surface alignment. \revPatrice{Gaussian Splatting-based methods can be further enhanced to render reflective surfaces using material shaders \cite{jiang2024gaussianshader}, or to handle complex mirror reflections \cite{ye20243d, liu2024mirrorgaussian}.}

Most prior work emphasizes global shape reconstruction and coarse geometric structures. While some methods incorporate geometric priors to enhance shape representation, they often struggle with fine-grained details due to high-frequency feature loss, as current network architectures have limited band representation capacity, often requiring complex 3D prior integration. To overcome these challenges, we propose a multi-scale wavelet-based feature approach utilizing a pre-trained depth image autoencoder trained on monocular depth priors. Wavelets efficiently capture high-frequency geometric details while preserving spatial localization, unlike Fourier transforms, which lose spatial information. This property is crucial for retaining fine surface details that deep learning models often neglect due to the lack of specialized multi-scale representation. The autoencoder is trained on wavelet-transformed depth images generated by a state-of-the-art monocular depth diffusion model \cite{he2024lotus}. The extracted wavelet features are aligned with implicit 3D triplane features via triplane projection and fused to enhance SDF predictions. Our method outperforms state-of-the-art reconstruction models across diverse scenes. The main contributions of our work can be summarized as follows:

\begin{itemize}
\item \textbf{Wavelet-Transformed Depth Feature Conditioning}: We introduce a pre-trained multi-scale wavelet autoencoder for depth image reconstruction. During implicit SDF training, wavelet features extracted from depth maps condition the network, enhancing geometric detail preservation.
\item \textbf{Triplane-Aligned Wavelet Feature Projection}: A triplane projection strategy aligns 2D wavelet features with 3D implicit representations, ensuring seamless fusion and improved geometric consistency.
\item \textbf{Hybrid Feature Fusion for SDF Prediction}: A UNet-based fusion mechanism integrates implicit 3D features with wavelet-transformed depth representations, yielding more structured and accurate SDF predictions for isosurface mesh extraction.
\end{itemize}
\section{Related Work}

\noindent \textbf{Geometry Representation.} 3D geometry representation follows two main paradigms: implicit and explicit. Implicit methods model surfaces via neural radiance fields (NeRF) \cite{mildenhall2021nerf}, surface reconstruction like Unisurf \cite{oechsle2021unisurf}, or signed distance functions like BakedSDF \cite{yariv2023bakedsdf}. Explicit methods, such as Structure from Motion (SfM) \cite{schoenberger2016sfm} and Multi-View Stereo (MVS) \cite{shen2013accurate}, reconstruct 3D geometry from multi-view images. Recent advances, like 3D Gaussian Splatting \cite{kerbl3Dgaussians}, enable real-time rendering while maintaining high fidelity. Each approach balances reconstruction accuracy, efficiency, and rendering quality.  

Further refinements address aliasing artifacts, such as Mip-NeRF created \cite{barron2021mip}, and Mip-NeRF 360 created \cite{barron2022mip}, which extends NeRF-based models to large-scale unconstrained environments akin to NeRF in the Wild \cite{meshry2019neural}. Implicit SDF models reconstruct shapes from single images akin to DISN \cite{xu2019disn} and enhance local geometry with SDF priors \cite{chabra2020deep}.  

Recent work integrates SDFs with diffusion models for high-fidelity shape generation from text or single image input \cite{shim2023diffusion, zheng2023locally, chou2023diffusion, li2023diffusion, cheng2023sdfusion}. Explicit Gaussian-based methods \cite{kerbl3Dgaussians} continue evolving: AGS-Mesh \cite{ren2024agsmesh} incorporates meshing priors, PGSR \cite{chen2024pgsr} enforces planar constraints for structured Gaussian point clouds, and DN-Splatter \cite{turkulainen2024dn} integrates depth and normal supervision for improved reconstruction.

\noindent \textbf{Spectrum Techniques.} Spectral methods have long played a crucial role in computer vision. The frequency analysis of a Fourier Transform has inspired spectral convolution kernels in CNNs like the work \cite{lavin2016fast} and enabled Fourier Convolutional Neural Networks (FCNN) \cite{pratt2017fcnn}. Similarly, the Fourier transform has also been integrated into multi-head attention mechanisms for Fourier Transformers \cite{he2023fourier, nguyen2022fourierformer, buchholz2022fourier}, enhancing image feature learning. However, Fourier-based features often face training challenges due to the broad frequency distribution and the loss of locality.  

Wavelet transforms provide a more localized spectral representation, preserving spatial details lost in the standard Fourier transform. They have been widely applied in image denoising \cite{mohideen2008image, chang2000adaptive}, super-resolution \cite{guo2017deep, huang2017wavelet}, and restoration, as well as compression \cite{shen1999wavelet, rippel2017real, ma2020end} and inpainting \cite{huang2024wavedm, yu2021wavefill, figueiredo2003algorithm}. Wavelet autoencoders \cite{fujieda2018wavelet, chen2018learning, mishra2020wavelet, sadat2024litevae, schelkens2003wavelet} efficiently represent image features while reducing parameters for lightweight models.  

Recent advances extend spectral methods to 3D tasks. Periodic activation functions can be leveraged in implicit MLP to capture repeating geometric patterns \cite{sitzmann2020implicit}, while FINER \cite{liu2024finer} adapts spectral variables for feature learning. Fourier bases have been explored for implicit representations \cite{li2024learning}, with models like Bacon \cite{lindell2022bacon} and BANF \cite{shabanov2024banf} make use of progressive learning for band-limited feature capture in 3D reconstruction.  

Wavelets have also been integrated into multi-scale triplane radiance fields \cite{khatib2024trinerflet} and SDF diffusion models \cite{hu2024neural, zhou2024udiff, hui2022neural}, enhancing shape generation with fine-grained local details. Despite these advances, spectral models still face convergence challenges, and implementing wavelet decomposition in 3D feature spaces remains computationally demanding.
\section{Method}

\subsection{Point Cloud Input for Reconstruction}

\begin{figure}[!thbp]
\centering
\includegraphics[width=\linewidth]{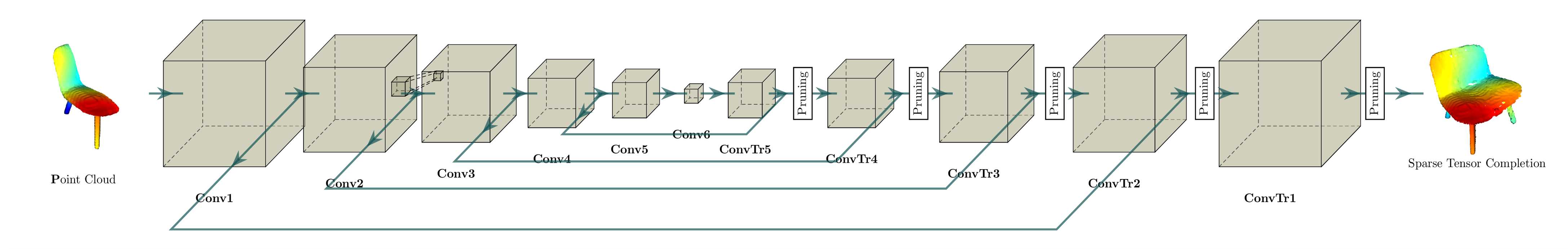}
\caption{3D CNN-based UNet for shape point cloud completion, image from Minkowski Sparse CNN demo.}
\label{fig:3D-unet}  
\end{figure}

The standard backbone for 3D reconstruction is a 3D CNN-based U-Net, comprising an encoder and a transposed decoder. Figure \ref{fig:3D-unet} illustrates a 3D U-Net architecture for shape completion and reconstruction from point cloud input. The process begins with an incomplete point cloud, voxelized into a sparse volumetric representation.

The 3D U-Net follows an encoder-decoder structure with 3D convolutional layers (Conv), transposed convolutions (ConvTr), and skip connections. The encoder progressively downsamples the input through multiple convolutional layers (Conv1 to Conv6), capturing hierarchical spatial features and compressing the data into a latent representation. The decoder then upscales the features via transposed convolutions (ConvTr1 to ConvTr5), incorporating skip connections to preserve fine-grained details. Pruning at specific layers maintains sparsity, enabling efficient processing of large-scale data.

Finally, the network reconstructs a dense, completed 3D shape. While this multi-scale feature learning improves computational efficiency, the model struggles to represent complex geometries due to the CNN kernel’s limited receptive field.

\begin{figure}[!thbp]
\centering
\includegraphics[width=0.90\linewidth]{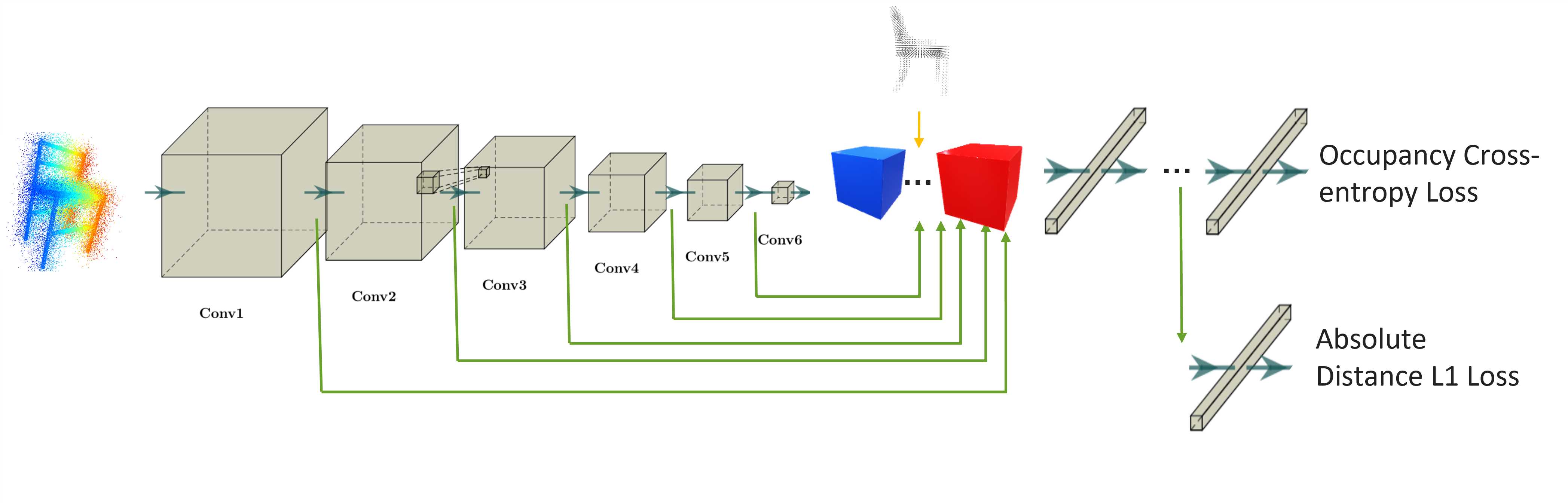}
\caption{The implicit SDF model for shape reconstruction in the training phase.}
\label{fig:shape-completion-train}  
\end{figure}

In contrast, the implicit SDF model offers greater flexibility in capturing complex topology and geometry. By leveraging implicit SDF representations, it generates continuous, watertight meshes from sparse point cloud inputs, enabling robust shape completion and reconstruction. Figure \ref{fig:shape-completion-train} illustrates the training phase of an implicit Signed Distance Function (SDF) model for mesh reconstruction from point cloud input.

The process begins with a sparse point cloud representing the target shape as raw 3D data. It is then processed through a series of 3D convolutional layers (Conv1–Conv6) to extract hierarchical features, capturing both global and local geometric details. These features are used to predict SDF values, which encode the signed distances of points to the object's surface. The blue cube in Figure \ref{fig:shape-completion-train} represents intermediate occupancy or SDF predictions, while the red cube denotes the fused 3D features from the encoder and voxelized shape points.

Finally, training is guided by two loss functions: binary cross-entropy loss on SDF values and absolute Distance L1 loss between predicted and ground truth SDF values. This dual-head learning strategy enhances reconstruction accuracy.
 
During the test/inference phase, the model processes the input sparse point cloud and predicts signed distance values and occupancy probabilities. The Marching Cubes algorithm is then applied to extract a triangular mesh surface from the predicted SDF values of the query points.

\begin{figure}[!thbp]
\centering
\includegraphics[width=0.90\linewidth]{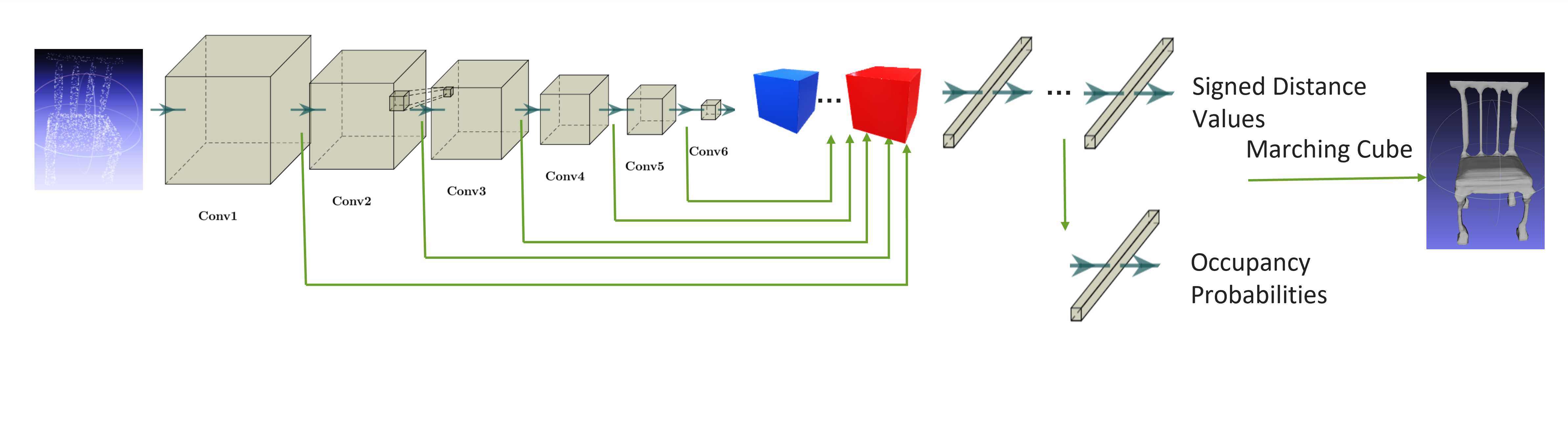}
\caption{The implicit SDF model for shape reconstruction in the test phase.}
\label{fig:shape-completion-test}  
\end{figure} 

Unlike the recent 3D Gaussian Splatting approach \cite{kerbl3Dgaussians}, which employs explicit Gaussians to learn 3D representations from multiview images, Gaussian Splatting offers fast training but relies on a discretized representation. While it can render plausible views, its geometric accuracy remains limited. To improve mesh reconstruction, methods such as 2D Gaussian-based \cite{huang20242d} and Geometry-aware Gaussian-based approaches \cite{li2025geogaussian} introduce surface-aligned geometry constraints. However, these techniques still fall short of the reconstruction performance achieved by implicit SDF, which benefits from its continuous representation. Exploring hybrid methods that integrate explicit Gaussians with implicit SDF could further enhance reconstruction quality.

\subsection{Multiview Image Input for 3D Reconstruction}

The implicit SDF model can also process image inputs by generating sample pixel rays from multiview posed images. Along these rays, sampled points are queried through the 3D implicit MLP layers for representation learning. However, MLP layers excel at capturing low-frequency features but struggle to preserve high-frequency details due to their intrinsic averaging effect across multiple layers.

To enhance reconstruction quality, feature augmentation before the decoder is crucial. One promising approach is conditioning the implicit SDF with a "codebook" representation—a pre-trained set of embedding features encoding the structural characteristics of building point clouds. 

Despite its potential, this approach faces several limitations. The pre-trained codebook is specialized for building point cloud data, which may restrict its generalizability to other 3D shapes or environments beyond its feature distribution. Additionally, the limited availability of large-scale 3D building scans makes acquiring diverse and representative training data challenging.

To overcome these challenges, future work could focus on creating more generic and adaptive codebook representations by leveraging 3D synthetic data priors or foundational 3D generation models to develop a universal geometry codebook. Incorporating additional cues, such as semantic segmentation or part-based geometry priors, could further enhance reconstruction capabilities. While this approach illustrates the value of pre-trained priors for improving 3D reconstruction from limited data, continued research is essential to make these methods more robust and broadly applicable.


Recent advancements in 2D foundational models, such as DepthAnythingV1 \cite{depth_anything_v1}, DepthAnythingV2 \cite{depth_anything_v2}, and LOTUS \cite{he2024lotus}, have achieved remarkable success in predicting sharp depth and normal maps from single image inputs. These models provide a strong geometry prior, which can be further enhanced using a frequency-based encoder to capture high-frequency features from the input images. These enhanced features are then aligned with 3D triplane features for final SDF value decoding.

\begin{figure}[!th]
\begin{center}
\includegraphics[trim=0.0cm 0.7cm 0.0cm 0.0cm, width=\linewidth]{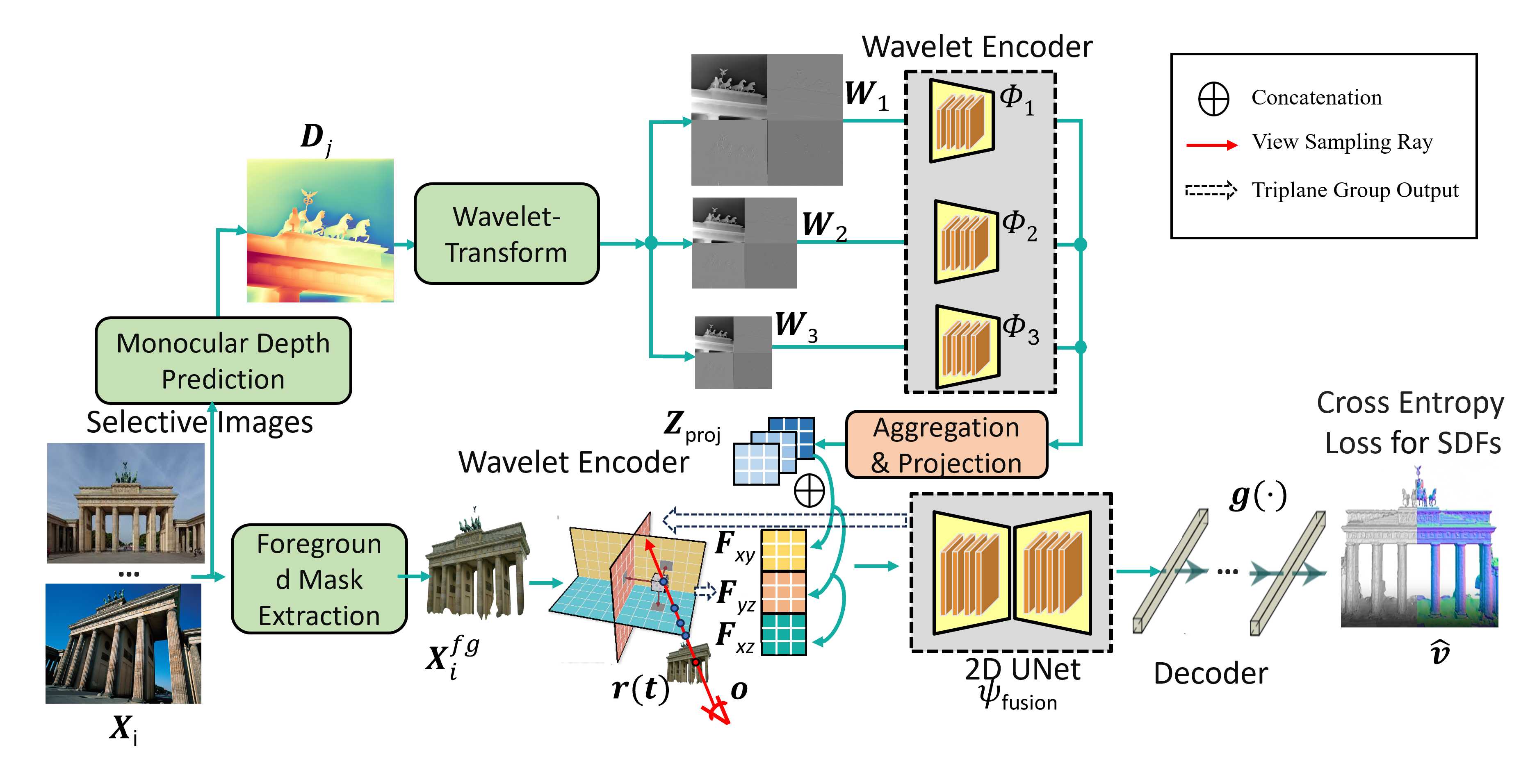}
\end{center}
\caption{Our model is based on implicit triplane feature fusion for Signed Distance Function (SDF) prediction. Given an input view image $\mathbf{X}_i$, a foreground mask $\mathbf{X}_i^{fg}$ is extracted to focus on the target region for SDF queries. For each pixel, its ray is traced from the camera view $\mathbf{C}_i$ to query the implicit triplane features $\{\mathbf{F}_{xy}, \mathbf{F}_{xz}, \mathbf{F}_{yz}\}$. Images with close-up details are processed via a monocular depth before predicting depth maps, followed by wavelet transforms in three resolutions. The transformed features $\mathbf{W}_*$ are encoded through a multi-scale wavelet feature encoder ($\Phi_1, \Phi_2, \Phi_3$) and aggregated into a fused wavelet feature map. This map is projected onto three orthogonal planes, producing $\mathbf{Z}_{proj} = \{\mathbf{Z}_{xy}, \mathbf{Z}_{xz}, \mathbf{Z}_{yz}\}$. The triplane features \cite{peng2020convolutional} and wavelet features are concatenated and further fused using a 2D U-Net $\psi$. Finally, MLPs $\mathbf{g}(\cdot)$ decode the fused features to predict SDF values. During inference, the isosurface is extracted via marching cubes to generate the mesh.}
\label{fig:overview-pipeline}
\end{figure}

As illustrated in Figure \ref{fig:overview-pipeline}, the process begins with input images, which are processed through a foreground mask to isolate the region of interest, such as a building gate. The masked images, denoted as $\mathbf{X}_i$, are then passed through a multi-scale wavelet encoder. This encoder utilizes wavelet features' spatial and multi-scale properties to capture the geometry structure and visual details of the 2D images. The wavelet-encoded features from multiple views are fused into a unified 2D feature representation, $\mathbf{C}_i$, which is used as a conditioning input to a 2D UNet architecture. The fused features are then passed through an MLP-based decoder to predict 3D signed distance values.

This approach offers several advantages. Wavelet-encoded 2D features effectively capture geometry and visual information, enabling the 2D UNet to generate more accurate 3D reconstructions. By aligning wavelet features across multiple views and using them as a conditioning input, the model leverages complementary perspectives to produce high-quality results. Compared to the codebook-based approach, this pipeline demonstrates improved generalization, as wavelet features adapt more effectively to diverse visual patterns beyond specific building structures. Moreover, the 2D UNet architecture provides robust and flexible latent feature fusion, better accommodating variations in the input images.

In summary, this wavelet encoder-based 2D-3D reconstruction framework integrates the strengths of wavelet-encoded features and deep learning-based 3D reconstruction. This results in a more versatile and generalizable system capable of handling a wide range of 3D shapes and scenes.

The proposed reconstruction method leverages implicit triplane features ${\mathbf{F}_{xy}, \mathbf{F}_{xz}, \mathbf{F}_{yz}}$, while there are learned 2D feature grids aligned with three orthogonal planes to encode both geometric and appearance information of a 3D scene. As shown in Figure~\ref{fig:overview-pipeline}, our framework utilizes these triplane representations for efficient 3D reconstruction from images with known poses. For any 3D query point along a sampled ray, features are retrieved from the three orthogonal planes and aggregated to predict the Signed Distance Function (SDF) value at that location.  

To enhance this representation, we introduce a pipeline that enriches triplane features with wavelet-encoded geometric details extracted from input images $\mathbf{X}_i$. These input images from the subset undergo monocular depth estimation and multiresolution wavelet transforms before being aggregated and fused into refined triplane characteristics $\{\mathbf{F}_{xy}^{fused}, \mathbf{F}_{xz}^{fused}, \mathbf{F}_{yz}^{fused}\}$ for high-quality SDF prediction. In essence, our method improves surface reconstruction by integrating implicit triplane features with multi-scale wavelet features. The following sections detail each stage of the method along with its mathematical formulation.

\subsection{Preliminaries}
\noindent\textbf{Implicit Neural Rendering.} Implicit NeRF encodes a 3D scene by representing its volume density and color field, leveraging multi-view posed images through volume rendering. A pixel ray $\mathbf{r}(t) = \mathbf{o} + t\mathbf{d}$ is defined, starting from the camera position $\mathbf{o} \in \mathbb{R}^3$ and traversing along the view direction $\mathbf{d} \in \mathbb{R}^3$. Radiance integration along the ray accumulates color contributions from the sampling points of each ray to generate the final pixel color. For each sampling point, volume density $\sigma$ and radiance $\mathbf{c}$ are predicted using separate MLPs. The rendered pixel color $\hat{\mathbf{C}}$ is calculated by $\mathbf{T}(t) = \exp(-\int_{t_n}^t \sigma(\mathbf{r}(u)) \, du)$, density $\sigma(t)$, and color $\mathbf{c}(t)$ over the ray, bounded by $t_n$ (near) and $t_f$ (far):

\begin{equation}
    \hat{\mathbf{C}} = \int_{t_n}^{t_f} \mathbf{T}(t) \sigma(\mathbf{r}(t)) \mathbf{c}(t) \, dt.
\end{equation}

For practical computation, the numerical quadrature-based integration \cite{alpert1999hybrid} is used to approximate continuous integral calculation.

\noindent\textbf{SDF-Based Neural Implicit Surface.} A 3D surface $\mathcal{S}$ can be implicitly represented using the zero-level-set of its signed distance function $f(\mathbf{x}): \mathbb{R}^3 \to \mathbb{R}$, with a 3D point initialized from the depth map of color image $\mathbf{X}$ as input. Here, $\mathcal{S} = \{\mathbf{x} \in \mathbb{R}^3 \, | \, f(\mathbf{x}) = 0\}$, can be seen as the zero-crossing of the signed distance function. NeuS \cite{wang2021neus} reformulates volume density rendering in NeRF into a signed distance field (SDF) representation by employing a logistic function to optimize for neural volume rendering,

\begin{equation}
    \sigma(\mathbf{x}) = \phi_s(f(\mathbf{x})),
\end{equation}

where $\phi_s(x) = s e^{-sx} / (1 + e^{-sx})^2$ is a logistic density function. It can be derived as the derivative of the sigmoid function $\Phi_s(x) = (1 + e^{-sx})^{-1}$, and is parameterized by the slope $s$. The final opaque density $\sigma(t)$ along the ray is thus given by:

\begin{equation}
    \sigma(t) = \max \left( -\frac{d\Phi_s}{dt}(f(\mathbf{r}(t))) / \Phi_s(f(\mathbf{r}(t))), 0 \right).
\end{equation}

\subsection{Preprocessing}

To get rid of clutter pixels like humans and animals existing in the random online images of landmarks collected in the wild, we further utilize a preprocessing pipeline to create cleaner images and masks for training high-fidelity reconstruction meshes.

\begin{figure}[!ht]   
    \includegraphics[width=\columnwidth]{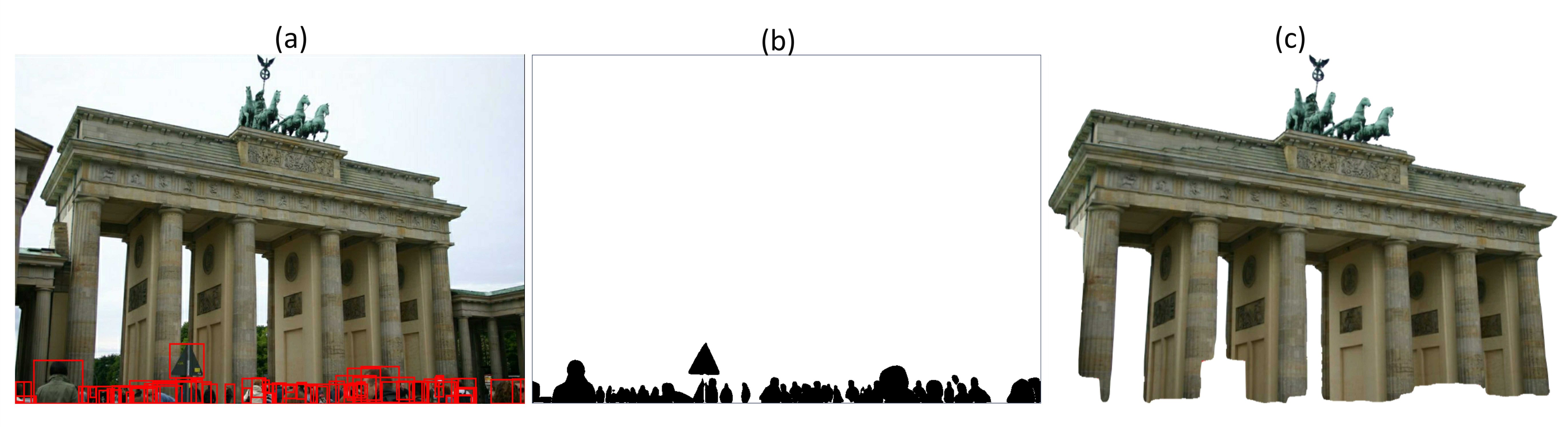}
\caption{A three-stage preprocessing pipeline for distractor removal: (a) Initial detection identifies unwanted elements like people and objects in the foreground using object detection, (b) Segment Anything Model (SAM) created \cite{kirillov2023segany} converts these detections into precise segmentation masks shown in black silhouettes, and (c) The final masked result isolates the architectural structure by removing both the detected distractors and sky background, leaving only the foreground building pixels that will be used for training the implicit model given query rays. }
\label{mask-fore}
\end{figure}


The preprocessing pipeline including distractor detection, distractor mask, background mask, effectively filters out non-architectural elements to focus the training only on the relevant structural components. The end result provides clean input data where query rays are only generated for the actual building geometry, improving the quality of the learned implicit representation.

\begin{figure}[!ht] 
    \includegraphics[width=\columnwidth]{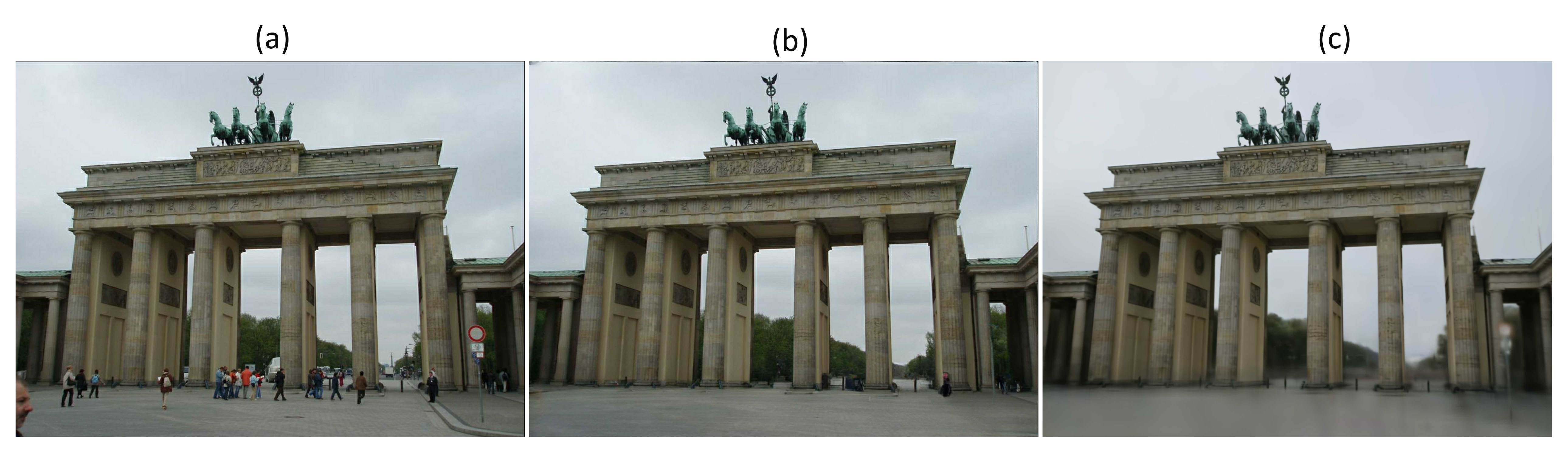}
\caption{(a) Raw image with distractor on the ground. (b) Inpainted image without distractor as training input. (c) Rendered color image predicted by conditioning on trained implicit SDF model. The whole distractor removal process on the raw image is followed by the diffusion model. In the end, our implicit model after training can render the full image without the distractor pixels.}
\label{fig:distra-removal}
\end{figure}

Furthermore, our model can also be directly used to render a clean color image by introducing a color rendering head. The images in Figure \ref{fig:distra-removal} show a comparison of removing unwanted pixels (like people) from a photo of the Brandenburg Gate in Berlin. The input to our framework is a color image (a), which contains pedestrians in front of the Berlin Gate. This image serves as the initial scene for further cleanup. In the next step, distractors are detected and removed, followed by an inpainting process to recover the missing pixels, resulting in the processed image (b). Finally, (c) presents the rendered output generated by the pre-trained implicit model, demonstrating the scene reconstruction without distractors and validating the effectiveness of our approach. Such color rendering result is implemented by introducing an additional head for color prediction conditioned on the SDF value prediction of the original implicit 3D model to justify the clean 3D representation of the implicit SDF model. This paper is still focused on the results of the 3D reconstruction instead of the results of the rendering.

Wavelet transforms are applied selectively to high-quality close-up images to optimize training efficiency for wavelet feature fusion.

The overall pipeline effectively removes the transient elements (people) while preserving and reconstructing the underlying static architecture through a combination of detection, masking, and inpainting for a cleaner 3D reconstruction.

\subsection{Model Structure}
We adopt the same implicit volumetric rendering expression as clarified in the previous section for the following model introduction. The whole model is composed of five parts, including a multi-scale wavelet feature encoder, a triplane feature query, a wavelet encoder with output feature projection onto a triplane, a triplane feature fusion, and an implicit SDF decoder. In particular, the input of the wavelet encoder is monocular depth, while all multi-view color images are used as input to the triplane feature encoder.

\noindent\textbf{Wavelet encoder for multi-scale features.} Given a selected input image $\mathbf{X}_i \in \mathbb{R}^{H \times W \times 3}$ from the original multiview images, the monocular depth before selected images predicts a depth map $\mathbf{D}_i \in \mathbb{R}^{H \times W}$.  The selection of particular close-up images is based on the quality and details that exist in the input view image, and most views are quite distant with blurry pixels, thus making it hard for the monocular depth estimation to provide accurate depth details. The depth map $\mathbf{D}_{i}$ undergoes a wavelet transform in three resolutions to produce multi-scale wavelet features $\{\mathbf{W}_1, \mathbf{W}_2, \mathbf{W}_3\}$. These are generated using a wavelet transform in three resolutions. These features are then processed by the wavelet encoder $\Phi$ with three different sizes, resulting in a final fused wavelet feature map $\mathbf{Z}_{wave} \in \mathbb{R}^{H' \times W' \times C}$ through feature aggregation:

\begin{equation}
\mathbf{Z}_{wave} = \Phi_1(\mathbf{W}_1) + \Phi_2(\mathbf{W}_2) + \Phi_3(\mathbf{W}_3),
\label{eq:wave_feat}
\end{equation}

where $\Phi_{1, 2, 3}$ are scale-specific encoding functions to extract various sized features. Usually, $C$ is four channels,  representing 2D signals through four filters, defined as $\mathbf{LL}$, $\mathbf{LH}$, $\mathbf{HL}$, and $\mathbf{HH}$. Given an input image $\mathbf{X}$, the 2D wavelet transform with specific scale decomposes the image into a low-frequency component $\mathbf{x_L}$ and three high-frequency components $\{\mathbf{x_H}, \mathbf{x_V}, \mathbf{x_D}\}$, corresponding to horizontal, vertical, and diagonal details respectively. 

We train the wavelet feature encoder using an autoencoder similar to the design of LiteVAE (\cite{sadat2024litevae}), aiming to reconstruct the original depth map from its wavelet-transformed representation. We apply a single-level wavelet decomposition independently at multiple scales of the input depth map, generating multi-scale wavelet feature maps. This allows the encoder to capture fine-to-coarse spatial details efficiently. After pretraining three separate wavelet encoders, we obtain their extracted multi-scale feature representations. To ensure alignment, We downsample the feature maps from the two higher-resolution encoders by factors of 1/2 and 1/4, respectively, to align with the smallest-scale feature map. This downsampling and aggregation follow the same process as LiteVAE \cite{sadat2024litevae}. To mitigate the loss of fine details, we retain multi-scale information by aggregating features across different resolutions. Furthermore, since each wavelet decomposition produces four sub-bands (LL, LH, HL, HH), we stack these sub-maps along the channel dimension before passing them to the subsequent processing pipeline.

\begin{figure*}[!th]
\begin{center}
\includegraphics[trim=0.0cm 0.0cm 0.0cm 0.0cm, width=\linewidth]{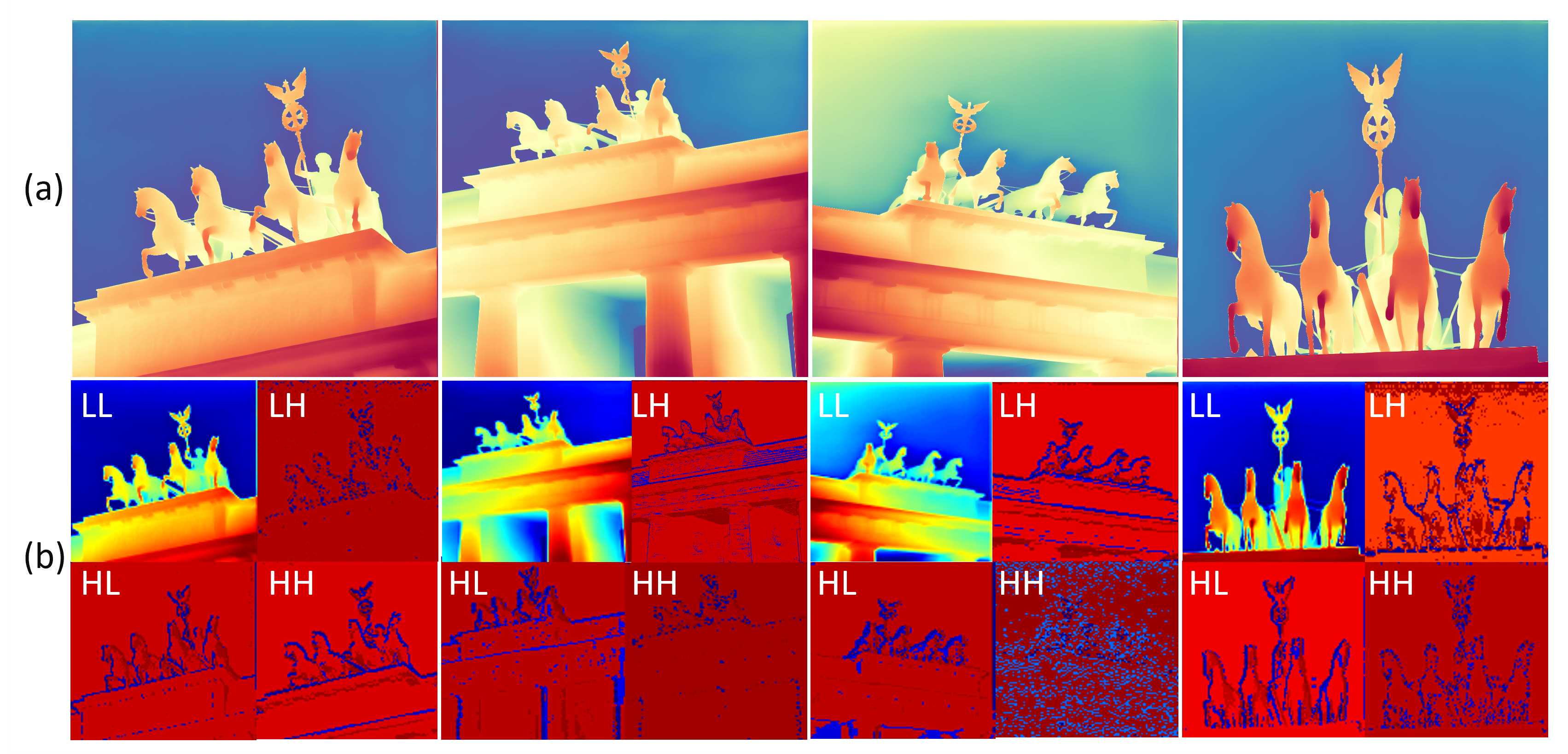}
\end{center}
\caption{Wavelet transform of the depth map in finest resolution, (a) is the original depth map, and (b) is the wavelet transform of the depth map, composed of four parts: Low-Low (LL), Low-High (LH), High-Low (HL), and High-High (HH). The wavelet transformed depth is used as input for the AutoVAE Encoder.} 
\label{fig:wavelet-depth}
\end{figure*}

We provide example results of wavelet transformed features in Figure~\ref{fig:wavelet-depth}, which demonstrates the effectiveness of wavelet transforms in preserving geometric information from depth maps. The visualization compares original depth maps $\mathbf{D}_j$ (top row) with their corresponding wavelet decompositions $\mathbf{W}_j$ (bottom row). The input depth maps, predicted by the state-of-the-art diffusion-based monocular depth estimation network of LOTUS \cite{he2024lotus}, capture detailed geometric structures and continuous depth variations. Our wavelet transform decomposes these depth maps in three resolutions into multi-scale feature representations $\mathbf{W}_j, {j=1, 2, 3}$ with three levels, where each level $j$ preserves both spatial and frequency information critical for geometric detail reconstruction. This multi-resolution representation enables the model to effectively encode both fine-grained surface details and global shape features. The wavelet transform decomposes each depth map with a specific resolution into four sub-bands (LL, LH, HL, HH), effectively capturing different frequency components. While LL retains a global structure, LH and HL emphasize horizontal and vertical details, and HH captures diagonal features. This highlights the ability to preserve and distinguish depth-specific geometry, and such spatial can also be easily aligned with the image feature map. 


\noindent\textbf{Pixel ray query for implicit triplane features.} Each pixel of the foreground masked input image $\mathbf{X}_i^{fg}$ is associated with a ray cast from the camera view $\mathbf{o} \in \mathbb{SE}(3)$. All the sampled points along query rays of each image retrieve implicit triplane features $\{\mathbf{F}_{xy}, \mathbf{F}_{xz}, \mathbf{F}_{yz}\}$ from three orthogonal planes $\{xy, xz, yz\}$ of the 3D space via ray projection, where each plane has a feature resolution of $\mathbb{R}^{H' \times W' \times C'}$, with feature channel dimension $C'$:

\begin{equation}
\{\mathbf{F}_{xy}, \mathbf{F}_{xz}, \mathbf{F}_{yz}\} = \text{Query}_{\{xy, xz, yz\}}(\mathbf{r}(t)).
\label{eq:query}
\end{equation}

\noindent\textbf{Wavelet feature projection onto triplane.} Meanwhile, the wavelet feature map $\mathbf{Z}_{wave}$ of Equation \ref{eq:wave_feat} is projected onto the three orthogonal 2D planes to match the implicit triplane feature resolution. This cosine projection generates three projected wavelet feature maps \{$\mathbf{Z}_{xy}, \mathbf{Z}_{xz}, \mathbf{Z}_{yz}$\} respectively.

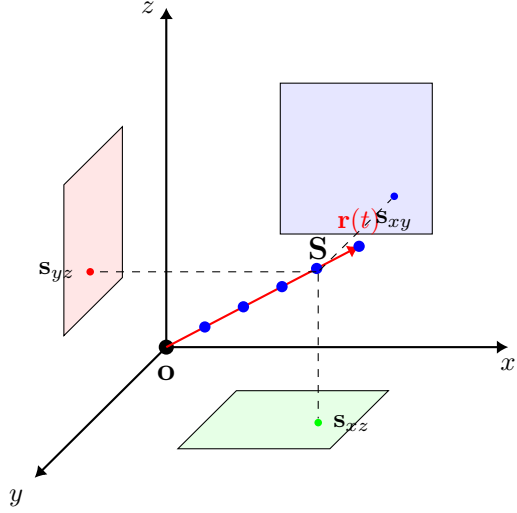
\begin{figure}[!th]
    \centering
    \begin{tikzpicture}
        \begin{scope}

            \draw[thick,->] (0,0,0) -- (4.5,0,0) node[below] {\footnotesize $x$};
            \draw[thick,->] (0,0,0) -- (0,4.5,0) node[left] {\footnotesize $z$};
            \draw[thick,->] (0,0,0) -- (0,0,4.5) node[below left] {\footnotesize $y$};

            \filldraw[fill=blue!10, draw=black] (1.5,1.5,0) -- (1.5,3.5,0) -- (3.5,3.5,0) -- (3.5,1.5,0) -- cycle; 
            \filldraw[fill=green!10, draw=black] (1.5,0,1.5) -- (1.5,0,3.5) -- (3.5,0,3.5) -- (3.5,0,1.5) -- cycle; 
            \filldraw[fill=red!10, draw=black] (0,1.5,1.5) -- (0,1.5,3.5) -- (0,3.5,3.5) -- (0,3.5,1.5) -- cycle; 

            \node[circle, fill=black, inner sep=2pt, label=below:{\textbf{o}}] (O) at (0,0,0) {};

            \draw[thick,->, red] (0,0,0) -- (4,2.8,3.8) node[pos=1.0, above] {\footnotesize $\mathbf{r}(t)$};

            \node[circle, fill=red, inner sep=0.0pt, label=above:{\textbf{S}}] (s) at (3,2,2.6) {};

            \foreach \t in {0.2, 0.4, 0.6, 0.78, 1.0} {
                \node[circle, fill=blue, inner sep=1.5pt] at ({\t*4}, {\t*2.8}, {\t*3.8}) {};
            }

            \node[circle, fill=blue, inner sep=1pt, label=below:{\footnotesize $\mathbf{s}_{xy}$}] (s_xy) at (3,2,0) {};
            \node[circle, fill=green, inner sep=1pt, label=right:{\footnotesize $\mathbf{s}_{xz}$}] (s_xz) at (3,0,2.6) {};
            \node[circle, fill=red, inner sep=1pt, label=left:{\footnotesize $\mathbf{s}_{yz}$}] (s_yz) at (0,2,2.6) {};

            \draw[dashed, black] (s) -- (s_xy);
            \draw[dashed, black] (s) -- (s_xz);
            \draw[dashed, black] (s) -- (s_yz);
        \end{scope}
    \end{tikzpicture}
    \vspace{0.5em}
    \caption{Sampling points along the pixel ray $\mathbf{r}(t)$ starting from $\mathbf{o}$ for implicit triplane feature learning via projection. For Wavelet feature projection onto triplane. The ray $\mathbf{r}(t)$ starts from the camera origin $\mathbf{o}$, then passes through a single unprojected point $\mathbf{S}$. Dashed lines represent the orthogonal projections onto the $xy$, $xz$, and $yz$ planes to obtain triplane features $s_{xy}$, $s_{xz}$, and $s_{yz}$ for a 3D point.}
    \label{fig:wavelet_projection}
\end{figure}

To incorporate wavelet features into the implicit Signed Distance Field (SDF) model, we first generate a structured 3D representation of the scene by leveraging aligned depth maps. Specifically, we reconstruct a dense unprojected point cloud in the camera frame followed by a camera-to-world transform. This transformation involves back-projecting depth pixels into 3D space using the known intrinsic and extrinsic camera parameters. The resulting 3D points are then associated with wavelet-based features aligned with 2D pixels.

Once the wavelet-enhanced feature map is obtained, it is projected onto the three feature-aligned triplane representations corresponding to the orthogonal planes defined by the normal vectors ${(1,0,0), (0,1,0), (0,0,1)}$. This projection ensures that the 3D wavelet features are properly integrated into the implicit triplane feature space. Mathematically, this process is formulated as follows:

\begin{equation} \{\mathbf{Z}_{xy}, \mathbf{Z}_{xz}, \mathbf{Z}_{yz}\} = \mathbf{P} \mathbf{Z}_{wave} \cdot \cos(\{\alpha, \beta, \gamma\}), 
\label{eq:proj-triplane}
\end{equation}

Where $\{\mathbf{Z}_{xy}, \mathbf{Z}_{xz}, \mathbf{Z}_{yz}\}$ represent the projected wavelet-enhanced features on the three orthogonal feature planes $xy, xz, yz$. The projection angles $\alpha, \beta, \gamma$ correspond to each feature plane, ensuring an optimal alignment between the wavelet features and the triplane encoding. Such an angle is the dot product between the ray direction and the axis direction. The transformation matrix $\mathbf{P}$, derived from the camera intrinsic parameters and the camera-to-world extrinsic pose, maps unprojected 3D points from the camera frame to the world coordinate system. These mapped points are then projected onto the triplane feature planes, where they serve as inputs to our method. Each pixel ray maps 3D spatial information onto a set of triplane feature representations. Given a camera pixel ray $\mathbf{r}(t)$, we analyze the sampled point $\mathbf{S}$ along the ray scaled by the predicted depth value and compute its orthogonal projections onto the three principal planes: $xy$, $xz$, and $yz$. These projections provide the corresponding triplane feature locations $\mathbf{s}_{xy}$, $\mathbf{s}_{xz}$, and $\mathbf{s}_{yz}$.




Figure~\ref{fig:wavelet_projection} illustrates the feature extraction process along a pixel ray $\mathbf{r}(t)$. The ray originates from the camera at $\mathbf{o}$, extends through the sampled point $\mathbf{S}$, and continues along its trajectory. Dashed lines indicate the orthogonal projections of $\mathbf{S}$ onto the three triplane feature planes ($xy$, $xz$, and $yz$), which are used for feature representation.

In Figure~\ref{fig:wavelet_projection}, implicit triplane features are obtained by sampling multiple points along the pixel ray uniformly, capturing continuous spatial information. Wavelet triplane features are projected in the same way as the implicit features. A key distinction is that each feature plane in the implicit approach consists of 16 channels, whereas the wavelet-based plane features are compressed into 4 channels, reducing redundancy while preserving essential spatial details.

This structured feature projection enables seamless integration of 2D wavelet-transformed depth features with 3D implicit features, leading to more accurate SDF predictions and higher-fidelity 3D reconstructions.

\noindent\textbf{Feature concatenation and fusion.} The implicit triplane features and the projected wavelet features of each feature plane are concatenated along the channel dimension and fused using a 2D U-Net. This results in three fused triplane features $\{\mathbf{F}_{xy}^{fused}, \mathbf{F}_{xz}^{fused}, \mathbf{F}_{yz}^{fused}\}$, where $\mathbf{F}_{*}^{fused} \in \mathbb{R}^{H' \times W' \times 2C}$:

\begin{align}    
\{\mathbf{F}_{xy}^{fused}, \mathbf{F}_{xz}^{fused}, \mathbf{F}_{yz}^{fused}\} = \{\psi_\text{fusion}([\mathbf{F}_{xy}; \mathbf{Z}_{xy}]), \nonumber
\\ \psi_\text{fusion}([\mathbf{F}_{xz}; \mathbf{Z}_{xz}]), \psi_\text{fusion}([\mathbf{F}_{yz}; \mathbf{Z}_{yz}]) \},
\label{eq:fused_feat}
\end{align}

where $[ ; ]$ denotes the concatenation of feature maps along the channel dimension.

\noindent\textbf{Encoder for SDF prediction.} The fused features are finally decoded by a neural network $g(\cdot)$, which predicts the SDF value $v \in \mathbb{R}$ for the given pixel ray query:

\begin{equation}
v = g(\mathbf{F}_{xy}^{fused}, \mathbf{F}_{xz}^{fused}, \mathbf{F}_{yz}^{fused}).
\end{equation}

The predicted SDF values are used to extract the isosurface via marching cubes \cite{lorensen1998marching}, producing a reconstructed 3D mesh.

\noindent\textbf{Loss function.} The total training loss $\mathcal{L}_{\text{total}}$ for the implicit model is defined as a combination of three components: the mean cross-entropy loss $\mathcal{L}_{\text{RGB}}$, the Eikonal regularizer $\mathcal{L}_{\text{Eik}}$, 
The total loss is defined as:

\begin{equation}
\mathcal{L}_{\text{total}} = \mathcal{L}_{\text{RGB}} + \lambda_{\text{Eik}} \mathcal{L}_{\text{Eik}}, 
\label{eq:3d-total_loss}
\end{equation}

The photometric loss is defined as the $L_2$ loss between rendered image $\hat{C}(\mathbf{r}_i)$ and ground truth image $C(\mathbf{r}_i)$,

\begin{equation}    
\mathcal{L}_{\mathrm{rgb}} = \frac{1}{N} \sum_{i=1}^{N} \| \hat{C}(\mathbf{r}_i) - C(\mathbf{r}_i) \|_2^2,
\label{eq:photo-loss}
\end{equation}





where the Eikon loss is applied to $M$ neighboring sampled points to regularize the smoothness of local SDF prediction.

\begin{equation}    
\mathcal{L}_{\text{Eik}} = \frac{1}{M} \sum_{i=1}^{M} \big| \|\nabla \hat{v}_i\| - 1 \big|^2 ,
\label{eq:eikon-loss}
\end{equation}





In the formulation \ref{eq:3d-total_loss}, $\mathcal{L}_{\text{Eik}}$ regularizes the gradients to enforce the local smoothness of the signed distance field. For color image rendering, we need to add another mean-squared photometric loss to leverage the supervision of RGB pixels.

\nomenclature{$\Phi_{*}$}{Wavelet feature encoder}
\nomenclature{$\mathbf{W}_*$}{Learned Wavelet Feature Output}
\nomenclature{$\sigma(t)$}{Volume density}
\nomenclature{$t_n, t_f$}{{Pixel ray near and far range}}
\nomenclature{$\phi_s(x)$}{{Logistic function}}
\nomenclature{$\mathbf{P}$}{{Projection matrix}}
\nomenclature{$\mathbf{F}_{xy}, \mathbf{F}_{yz}. \mathbf{F}_{xz}$}{{Triplane features}}
\nomenclature{$\mathbf{Z}_{xy}, \mathbf{Z}_{yz}. \mathbf{Z}_{xz}$}{{Projected triplane wavelet features}}
\nomenclature{$\psi_\text{fusion}$}{{2D Fusion Unet}}

\section{Experiments and Results}
\label{ch4-sec:exp}

\subsection{Point Cloud-based 3D Reconstruction}

At first, we present 3D point cloud completion results using a 3D U-Net backbone with partial shape point cloud inputs. The network features a CNN-based encoder and a decoder with transposed convolutional layers that mirror the encoder structure.

\begin{figure}[!thbp]
\centering
\includegraphics[width=1.0\textwidth]{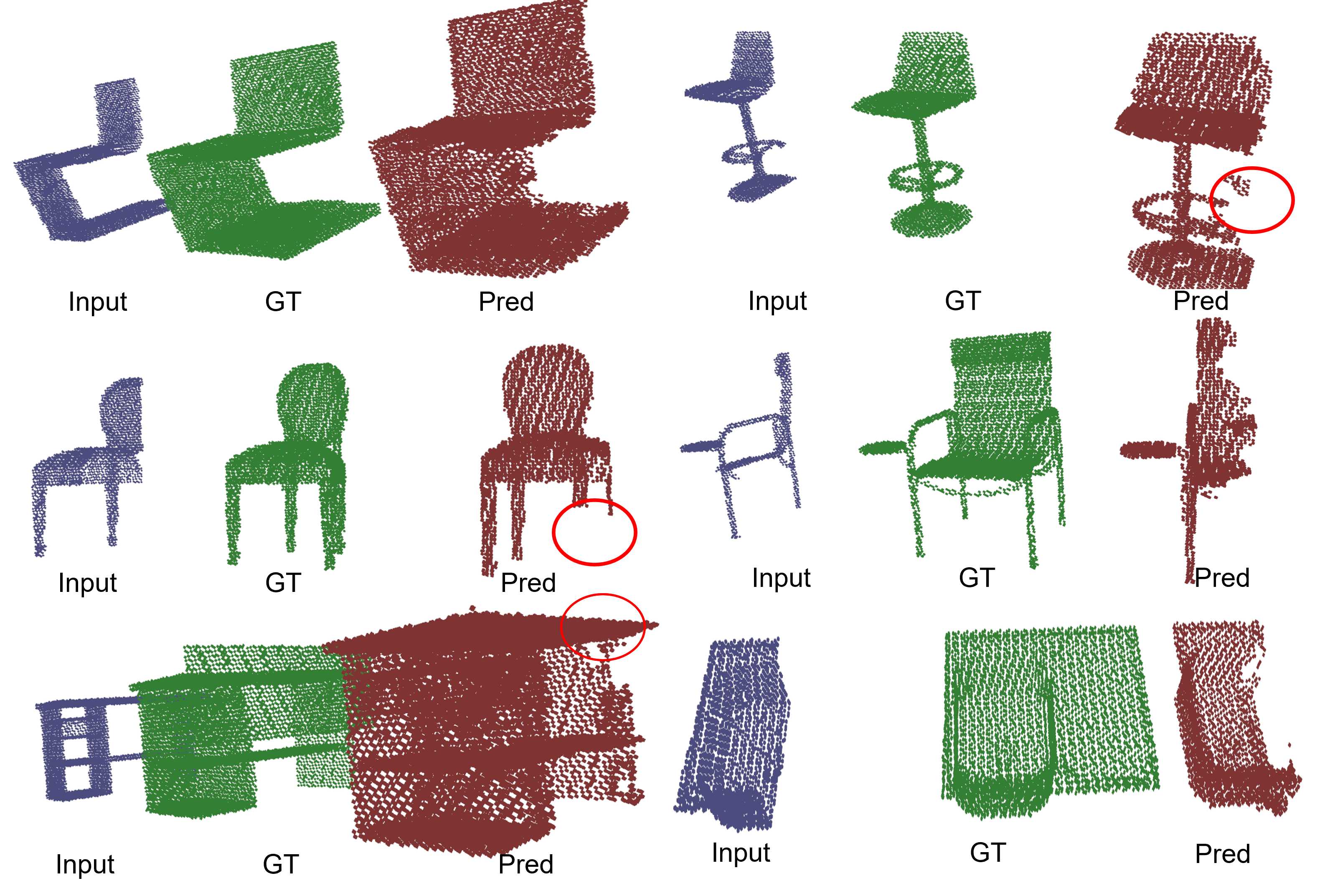}
\caption{The results of 3D UNet model for point cloud completion, where blue is the partial input, green is the GT point cloud, and red is the predicting output of 3D UNet model.}
\label{fig:shape-completion-unet-results}  
\end{figure} 

As shown in Figure \ref{fig:shape-completion-unet-results}, the model effectively completes shapes with symmetrical input, such as the chair in the top-left subfigure. However, it struggles with complex, asymmetrical structures, as indicated by the red circles. Thin elements, like chair legs, are not reconstructed accurately, highlighting the UNet model's limitations in capturing fine geometric details.

\begin{figure}[!thbp]
\centering
\includegraphics[width=0.80\linewidth]{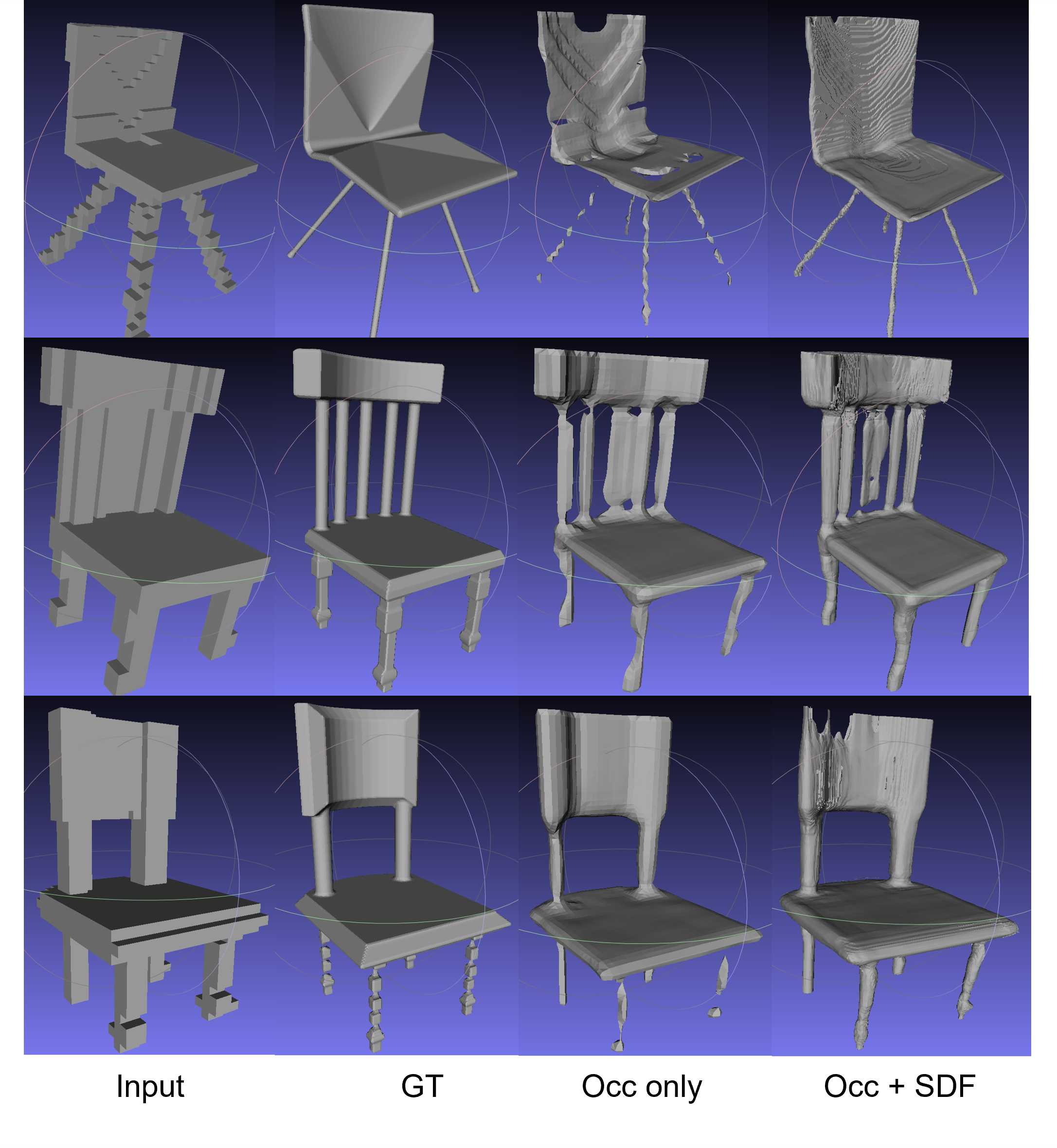}
\caption{The shape reconstruction results from 3D voxel input via implicit SDF model.}
\label{fig:shape-completion-results1}  
\end{figure}

Next, we present mesh reconstruction results using a 3D CNN encoder paired with an implicit MLP decoder, as illustrated in Figure \ref{fig:shape-completion-test}. In the last column of Figure \ref{fig:shape-completion-results1}, integrating occupancy probability and SDF loss during training enables the model to reconstruct fine details and generate smooth meshes from coarse voxels. While occupancy probability loss alone produces reasonable shapes, it fails to fully capture surface details, leading to gaps and holes, such as those in the back of the chair. This underscores the importance of combining both loss functions to enhance surface reconstruction quality.

\subsection{Image-based 3D Reconstruction}

\noindent \textbf{Datasets.}
To evaluate the general performance of our 3D reconstruction approach, we make use of a wide variety of datasets, including the DTU (\cite{jensen2014large}) dataset, which is collected from a turntable; the Tanks and Temples dataset (\cite{knapitsch2017tanks}), captured as video scans of sculptures and buildings; and the Cultural Heritage dataset (\cite{martinbrualla2020nerfw}), which features large-scale historical sites.

For DTU (\cite{jensen2014large}), we use the Chamfer distance metric calculated between the reconstructed model and the ground truth. In contrast, for Tanks and Temples (\cite{knapitsch2017tanks}), we evaluate reconstruction accuracy using the F1 score ($F_1 = 2 \cdot \frac{\text{Precision} \cdot \text{Recall}}{\text{Precision} + \text{Recall}}$), as Chamfer distance makes it difficult to differentiate the performance for some scenes. Due to the large scene scale of the Cultural Heritage dataset, obtaining ground truth (GT) meshes or pseudo-GT is challenging, so we primarily provide qualitative results. For DTU and Tanks and Temples, we uniformly sample 1,000 and 10,000 points, respectively, and compare them with the nearest 3D points from the GT mesh, which is obtained through the COMLAP tool, followed by delicate post-processing to make the GT mesh complete and smooth enough without artifacts like holes. 

\noindent \textbf{Baseline models.}
For baseline evaluation, we compare our method against several state-of-the-art implicit and explicit 3D reconstruction models. The implicit SDF baselines include VolSDF by \cite{yariv2021volume}, NeuS by \cite{wang2021neus}, Neuralangelo by \cite{li2023neuralangelo}, and BakedSDF by \cite{yariv2023bakedsdf}. The explicit reconstruction baselines include SuGaR by \cite{guedon2024sugar}, GOF by \cite{huang20242d`}, and 2DGS by \cite{Huang2DGS2024}. We provide quantitative comparisons across the DTU and Tanks and Temples datasets, while qualitative visual comparisons highlight the top three performing models. The input for all baselines is images and camera poses. 

\noindent \textbf{Implementation details.} For monocular depth estimation, we use the diffusion-based LOTUS model by \cite{he2024lotus} to predict depth maps for selected views of the heritage dataset. In contrast, for the Tanks and Temples, and DTU datasets, we process all training images. Wavelet decomposition is performed using the Fast Wavelet Transform (FWT) (\cite{mallat1989theory}) with Haar basis filters. Please note that depth is used only for training the wavelet features and not for supervising the 3D reconstruction. Furthermore, for the cultural heritage dataset, all input images undergo the same preprocessing step to remove transient pedestrians, ensuring that all baseline models are trained with the same masked image inputs. This guarantees fairness in the baseline comparisons.

 Our reconstruction pipeline is trained per scene, consistent with standard neural implicit surface reconstruction settings. Regarding the wavelet feature encoder, we emphasize that it is not trained in a scene-specific way. Rather, this module is pre-trained offline using an autoencoding objective (inspired by LiteVAE) on a collection of monocular depth predictions across a variety of generic scenes. These depth maps, along with their wavelet-transformed representations, are used to learn a compact latent prior over geometric features, analogous to learned image priors used in low-level vision tasks. This prior is then fixed and reused across all scenes during reconstruction. Our approach is thus fair for comparison, as it does not rely on specific geometry information supervision, and instead builds on learned geometric priors from the common depth prediction foundation model prior, which are estimated from monocular RGB images in a self-supervised manner. This setup maintains the generalizability of the wavelet-depth encoder and adheres to the same per-scene optimization assumptions used in all baselines by using the generalized depth wavelet feature prior.

The autoencoder for wavelet-transformed depth features consists of a ResNet encoder followed by a fully convolutional decoder, similar to LiteAutoVAE (\cite{sadat2024litevae}). We apply a Gaussian blurring loss to low-frequency sub-bands and a Charbonnier loss (\cite{barron2019general}) to high-frequency sub-bands. During implicit SDF training, the AutoVAE encoder remains frozen. The triplane feature representation is structured as $3 \times 64 \times 64 \times 16$, with an SDF decoder composed of fully connected layers. Wavelet-triplane fusion is achieved via a 2D U-Net with four downsampling and upsampling blocks, followed by a $1 \times 1$ convolution along the depth channel. The fused representation consists of three orthogonal triplane feature planes ($64 \times 64 \times 16$ each), combined with a projected wavelet feature map ($64 \times 64 \times 4$), and refined through the 2D U-Net.

The wavelet autoencoder processes four spectral channels—low-frequency, vertical high-frequency, horizontal high-frequency, and diagonal high-frequency—using ResNet blocks. Wavelet transforms are applied to Lotus-generated depth maps at three resolutions, with extracted features used to train the autoencoder. To balance fine-grained details and global features, we incorporate self-modulated convolutional layers (\cite{sadat2024litevae}). The loss function includes reconstruction, regularization, and adversarial terms (\cite{sadat2024litevae}). Features are extracted at $256 \times 256$, $128 \times 128$, and $64 \times 64$ resolutions, with higher-resolution features downsampled by $1/4$ and $1/2$ for alignment. For the cultural heritage dataset, we manually selected 100 close-up images to enhance the implicit SDF model with wavelet-transformed features.  

For the implicit SDF model, we use the Facto-SDF implementation from SDFStudio by \cite{Yu2022SDFStudio}, integrating it with the triplane feature representation as the encoder backbone.  

\textbf{Training Complexity.} Our training pipeline consists of two stages: training the wavelet encoder and training the implicit SDF conditioned on the frozen wavelet encoder. The wavelet encoder training takes approximately 8 hours on an RTX 3090. For the implicit SDF training of the DTU model, training is completed in 1-2 hours. As for Tanks and Temples, and Cultural Heritage), initial implicit SDF training on color images takes 6-8 hours due to data diversity, followed by 1-2 hours of fine-tuning with wavelet-triplane features.  

All experiments were conducted on an NVIDIA RTX 3090 GPU, ensuring efficient training and inference. This modular approach enables scalable learning across datasets of varying sizes and complexities.

\subsubsection{Baseline Comparisons}

\begin{figure}[!thbp]
\begin{center}
\includegraphics[trim=0.8cm 0.0cm 1.0cm 3.5cm, width=0.80\linewidth]{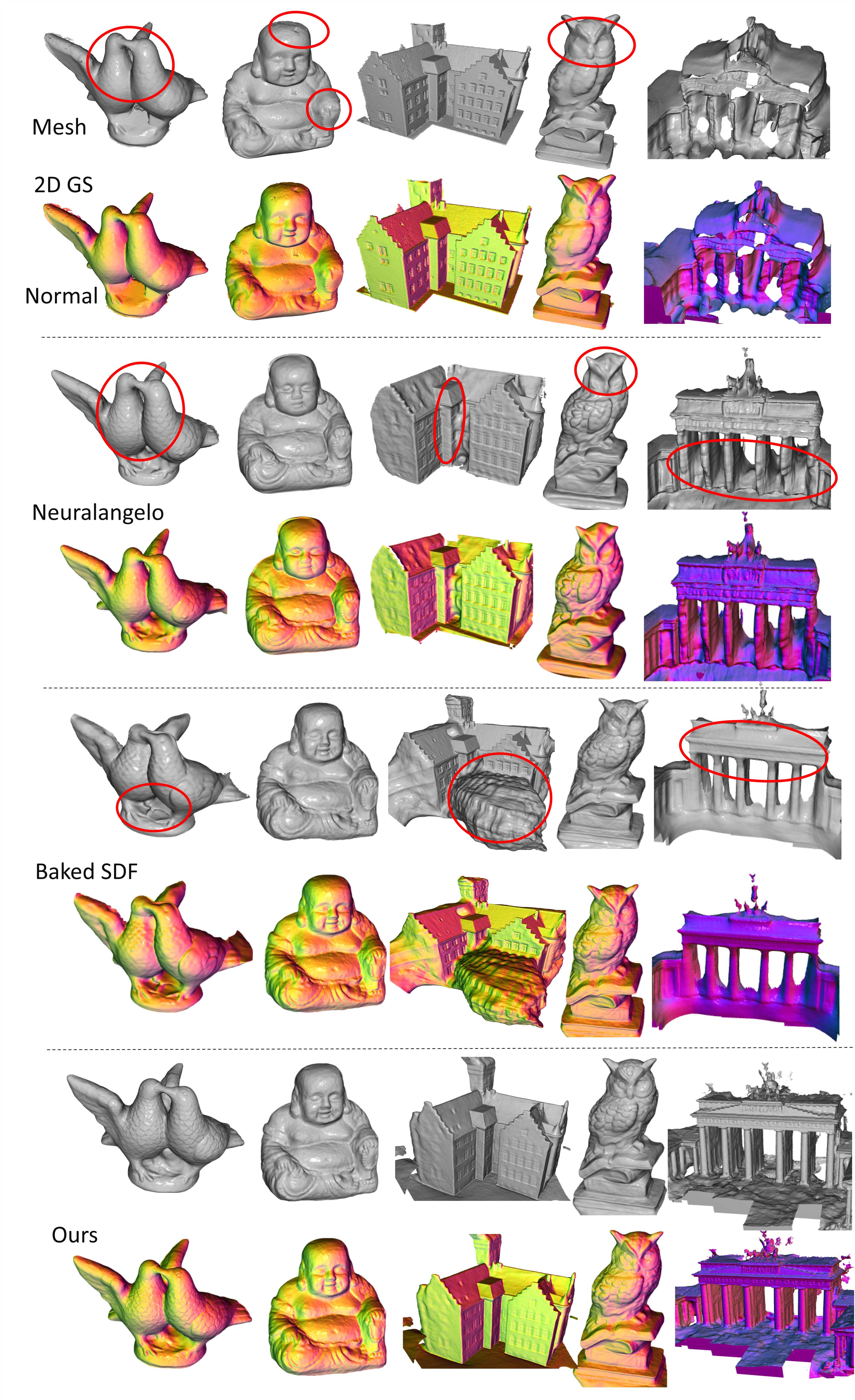}
\end{center}
\vspace{-1.6em}
\caption{Baseline comparison results on five targets from DTU (\cite{jensen2014large}), and Cultural Heritage (\cite{martinbrualla2020nerfw}) dataset. Each model result (split by a dashed line) contains mesh and normals.}
\label{fig:dtu-visual-results-part1}
\end{figure}

We first provide the qualitative comparison results of the sample targets or scenes in the three datasets, including the qualitative results of the DTU, Tank, and Temple, and the Cultural Heritage dataset in Figure \ref{fig:dtu-visual-results-part1}. These data sets are collected through cameras that point towards a target. Furthermore, the quantitative results on the DTU and Tank and Temple datasets are also provided in Table \ref{tab:cd_baselines_ours} and Table \ref{tab:tt-metrics-cd}, respectively.

As seen in Figure \ref{fig:dtu-visual-results-part1}, our model can reconstruct fine-grained details on the mesh surface, such as feature details on birds, details of clothes of happy buddha, owl, texts on the Berlin gate. We recommend that readers have a close look at the red highlighted circles. The 2D Gaussian Splatting seems to struggle to preserve details and also has obvious artifacts and holes on the mesh surface. The 2D GS also fails to reconstruct a mesh of a large-scale Berlin gate. Neuralangelo is very good at preserving some details, but still has some artifacts or obstructions, as shown on the bottom of the Berlin gate with unexpected blockings, although normals are consistent along with details of texts. BakedSDF has the worst performance, smoothing the results, with a smooth surface and loss of details, particularly obvious on the Berlin gate, which may even incur some unexpected reconstruction mesh regions in front of the house. 


Finally, we provide quantitative evaluation results in Table \ref{tab:cd_baselines_ours} and Table \ref{tab:tt-metrics-cd} using the Chamfer distance. On DTU, our model is the best, while Neuralangelo gets the second-best performance. On the Tank and Temple dataset, our model still scores best on three out of six scans, while Neuralangelo follows next.

\begin{table*}[!th]
\centering
\caption{Chamfer Distance (CD, $\downarrow$) comparison across scan IDs. \colorbox{green!50}{Green} sates the best, \colorbox{orange!50}{Orange} is second best, \colorbox{yellow!50}{Yellow} represents third best. Lower is better.}
\begin{adjustbox}{width=\textwidth}
\begin{tabular}{c c c c c c c c c c c c c c c c}
\toprule
\textbf{Method} & \textbf{24} & \textbf{37} & \textbf{40} & \textbf{55} & \textbf{63} & \textbf{65} & \textbf{69} & \textbf{83} & \textbf{97} & \textbf{105} & \textbf{106} & \textbf{110} & \textbf{114} & \textbf{118} & \textbf{122} \\
\midrule
VolSDF~\cite{yariv2021volume} 
& 1.14 & 1.26 & 0.81 & 0.49 & 1.25 & 0.70 & \cellcolor{yellow!50}0.72 & 1.29 & 1.18 & 0.70 & 0.66 & 1.08 & 0.42 & 0.61 & 0.55 \\
NeuS~\cite{wang2021neus} 
& 1.00 & 1.37 & 0.93 & \cellcolor{yellow!50}0.43 & \cellcolor{orange!50}1.10 & 0.65 & \cellcolor{orange!50}0.57 & 1.48 & 1.09 & 0.83 & 0.52 & 1.20 & \cellcolor{yellow!50}0.35 & 0.49 & 0.54 \\
Neuralangelo~\cite{li2023neuralangelo} 
& \cellcolor{green!50}0.41 & \cellcolor{orange!50}0.36 & \cellcolor{orange!50}0.35 & \cellcolor{orange!50}0.35 & 1.29 & \cellcolor{orange!50}0.54 & 0.73 & \cellcolor{yellow!50}0.52 & \cellcolor{orange!50}0.97 & \cellcolor{orange!50}0.56 & \cellcolor{orange!50}0.48 & \cellcolor{orange!50}0.73 & \cellcolor{orange!50}0.32 & \cellcolor{orange!50}0.40 & \cellcolor{orange!50}0.36 \\
BakedSDF~\cite{yariv2023bakedsdf} 
& 0.63 & 0.58 & 0.40 & 0.52 & 1.37 & \cellcolor{yellow!50}0.63 & 0.81 & 0.56 & \cellcolor{yellow!50}1.02 & 0.81 & \cellcolor{yellow!50} 0.50 & \cellcolor{yellow!50}0.82 & 0.39 & 0.45 & \cellcolor{yellow!50} 0.38 \\
SuGaR~\cite{guedon2024sugar} 
& 1.47 & 1.33 & 1.13 & 0.61 & 2.25 & 1.71 & 1.15 & 1.63 & 1.62 & 1.07 & 0.79 & 2.45 & 0.98 & 0.88 & 0.79 \\
\midrule
GOF~\cite{huang20242d`}
& 0.50 & \cellcolor{yellow!50}0.37 & \cellcolor{yellow!50}0.38 & 0.74 & \cellcolor{yellow!50}1.18 & 0.76 & 0.90 & \cellcolor{green!50}0.47 & 1.29 & \cellcolor{yellow!50}0.68 & 0.77 & 0.90 & 0.42 & \cellcolor{yellow!50}0.41 & 0.42 \\
2DGS~\cite{Huang2DGS2024} 
& \cellcolor{yellow!50}0.48 & 0.39 & 0.41 & 0.83 & 1.36 & 0.83 & 1.04 & 0.70 & 1.27 & 0.76 & 0.70 & 1.40 & 0.40 & 0.43 & 0.40 \\
Ours 
& \cellcolor{orange!50}0.45 & \cellcolor{green!50}0.34 & \cellcolor{green!50}0.32 & \cellcolor{green!50}0.34 & \cellcolor{green!50}0.97 & \cellcolor{green!50}0.52 & \cellcolor{green!50}0.54 & \cellcolor{orange!50}0.50 & \cellcolor{green!50}0.82 & \cellcolor{green!50}0.53 & \cellcolor{green!50}0.45 & \cellcolor{green!50}0.68 & \cellcolor{green!50}0.30 & \cellcolor{green!50}0.36 & \cellcolor{green!50}0.34 \\
\bottomrule
\end{tabular}
\end{adjustbox}
\label{tab:cd_baselines_ours}
\end{table*}

\begin{table*}[!thb]
\centering
\caption{Chamfer Distance (CD, $\downarrow$) results on the Tanks and Temples dataset. \colorbox{green!50}{Green} is best, \colorbox{orange!50}{Orange} is second best, and \colorbox{yellow!50}{Yellow} is third best. Lower values indicate better reconstruction accuracy.}
\begin{adjustbox}{width=\textwidth}

\begin{tabular}{l cccccc}
\toprule
\textbf{Method} & \textbf{Barn} & \textbf{Caterpillar} & \textbf{Courthouse} & \textbf{Ignatius} & \textbf{Meetingroom} & \textbf{Truck} \\
\midrule
NeuS & 0.69 & 0.68 & 0.54 & 1.62 & 0.63 & 0.90 \\
Geo-NeuS & 0.74 & 0.61 & 0.49 & 1.50 & 0.52 & 0.88 \\
Neuralangelo & 1.21 & 0.89 & 0.82 & 1.72 & 0.79 & 0.95 \\
SuGaR & \cellcolor{yellow!50}0.42 & \cellcolor{orange!50}0.53 & \cellcolor{orange!50}0.38 & \cellcolor{yellow!50}0.83 & \cellcolor{yellow!50}0.59 & \cellcolor{yellow!50}0.66 \\
2DGS & \cellcolor{orange!50}0.40 & \cellcolor{yellow!50}0.58 & \cellcolor{yellow!50}0.41 & \cellcolor{green!50}0.74 & \cellcolor{orange!50}0.53 & \cellcolor{orange!50}0.58 \\
Ours & \cellcolor{green!50}0.34 & \cellcolor{green!50}0.51 & \cellcolor{green!50}0.36 & \cellcolor{orange!50}0.76 & \cellcolor{green!50}0.49 & \cellcolor{green!50}0.52 \\
\bottomrule
\end{tabular}
\end{adjustbox}
\label{tab:tt-metrics-cd}
\vspace{-1.0em}
\end{table*}

\subsection{Ablation Study}

\begin{figure}[!th]
\begin{center}
\includegraphics[trim=0.0cm 0.0cm 0.0cm 0.0cm, width=\columnwidth]{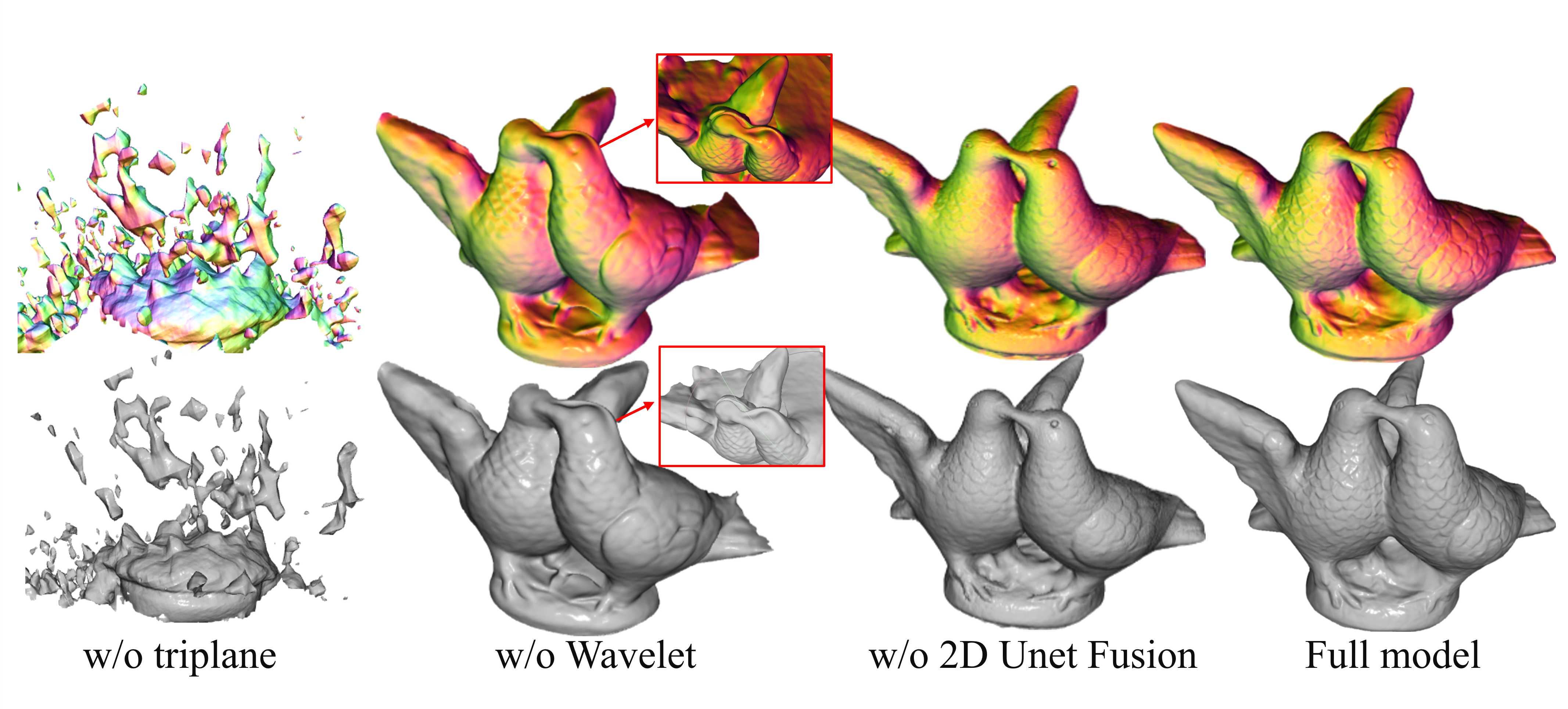}
\end{center}
\caption{Ablation study on 3D reconstruction. From left to right: (1) Removing the triplane leads to fragmented geometry. (2) Without the wavelet encoder, fine details are distorted. (3) Omitting 2D U-Net fusion results in less sharp features. (4) The full model achieves the best quality.}
\label{fig:ablation-results}
\end{figure}

We first perform ablation studies to assess the contribution of each key component in our model structure, with qualitative results in Figure~\ref{fig:ablation-results} and quantitative metrics in Table~\ref{tab:ablation-metric-cd}. 

\begin{figure}[!th]
\includegraphics[trim=0.0cm 0cm 0.0cm 1.0cm, width=\linewidth]{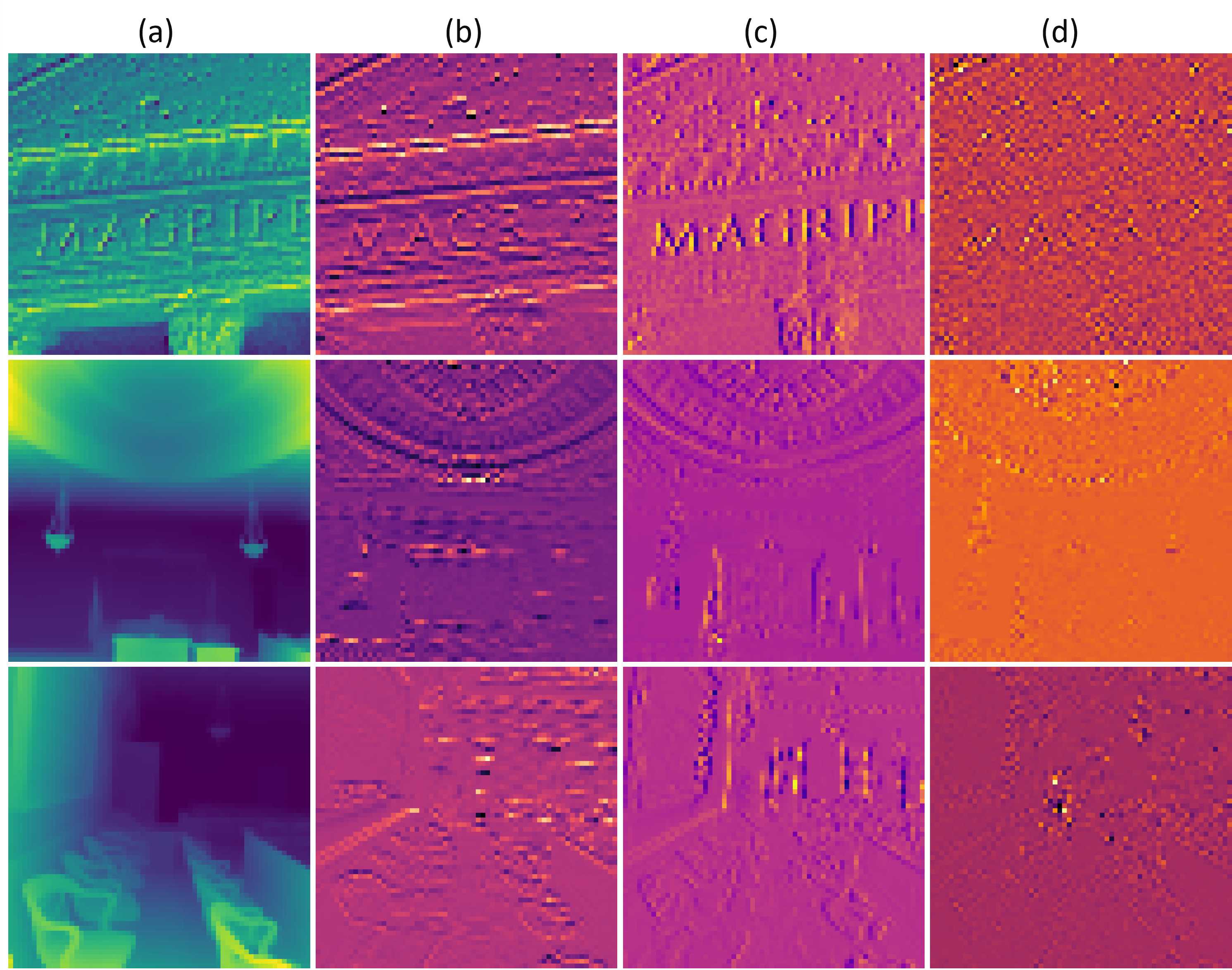}
\caption{Visualization of learned wavelet encoder feature maps at the highest resolution level. The four columns demonstrate different components of the encoded representation: (a) depth features preserving the overall geometric structure, (b) vertical gradient features capturing y-axis surface variations, (c) horizontal gradient features encoding x-axis transitions, and (d) diagonal gradient features representing cross-directional geometric patterns. Each component is processed through our wavelet encoder $\mathbf{\Phi}$ to extract orientation-specific geometric information.}
\label{fig:wavelet-feature_learned}
\end{figure}

The wavelet feature encoder can effectively capture fine-grained details from input views, such as edge features, as shown in Figure \ref{fig:wavelet-feature_learned}. The output feature maps of the learned wavelet autoencoder provide a rich representation of the input views to encode more geometric details, like the carved letters on the right side of the gate. The final reconstruction mesh details can be enhanced by the decoder conditioning on the wavelet features. Figure \ref{fig:wavelet-feature_learned} showcases the feature output of the largest wavelet encoder in an autoencoder pre-trained separately on the wavelet-transformed depth input in three resolutions, highlighting the progressive decomposition of image features across multiple frequency bands. The wavelet encoder outputs are visualized across four columns in Figure~\ref{fig:wavelet-feature_learned}. Column (a) shows the encoded depth features, preserving the overall geometric structure. Column (b) displays vertical gradient features $\mathbf{\Phi}$ that highlight edge transitions along the y-axis. Column (c) presents horizontal gradient features, capturing edge variations along the x-axis. Column (d) shows diagonal gradient features that encode diagonal directional geometric variations. This learned decomposition through our wavelet encoder enables comprehensive feature extraction at multiple orientations, crucial for accurate 3D surface reconstruction. Each component contributes specific directional information, allowing the model to capture both directional surface variations and overall geometric structure.
\begin{table*}[!thb]
\centering
\caption{Ablation study of our model on DTU~\cite{jensen2014large}, reporting Chamfer Distance (CD, $\downarrow$) to evaluate geometric accuracy. Lower values indicate better reconstruction quality.}
\begin{adjustbox}{width=\linewidth}
\begin{tabular}{l|c|c|c|c}
\toprule
\textbf{Metric (CD↓)} & \textbf{w/o Triplane Feature} & \textbf{w/o Multi-scale Wavelet Feature} & \textbf{w/ Single Scale Wavelet} & \textbf{w/o UNet Channelwise Fusion} \\
\midrule
CD $\downarrow$ & 0.87 & 0.65 & 0.60 & 0.56 \\
\midrule
\textbf{Metric (CD↓)} & \textbf{w/o Wavelet Autoencoder} & \textbf{w/o Photometric Loss} & \textbf{w/o Eikonal Loss} & \textbf{Full Model} \\
\midrule
CD $\downarrow$ & 0.60 & 0.94 & 0.55 & \textbf{0.51} \\
\bottomrule
\end{tabular}
\end{adjustbox}
\label{tab:ablation-metric-cd}
\end{table*} 

First, we evaluate the impact of removing the triplane representation, reverting to a purely MLP-based implicit SDF modeling. This results in severe surface fragmentation, geometric noise, and disconnected structures, reflected in a significantly worse Chamfer Distance (CD) of 0.87, indicating the critical role of the triplane representation in preserving global spatial coherence. Removing the multi-scale wavelet feature while retaining the triplane and decoder degrades surface quality, especially around fine structures (e.g., object edges and thin regions), increasing the CD to 0.65. Incorporating only a single-scale wavelet feature improves reconstruction to 0.60, whereas our full multi-scale wavelet pipeline further reduces CD to 0.51, confirming the importance of multi-resolution geometric encoding.

In addition, removing the 2D U-Net channel-wise fusion module and instead directly concatenating features leads to slight oversmoothing and loss of fine detail, reflected in a CD of 0.56, underscoring the benefit of learned feature fusion. On the loss design side, excluding the photometric loss produces the most degraded result (0.94 CD), demonstrating its necessity for accurate geometry recovery. Removing the wavelet autoencoder or the Eikonal regularization also leads to increased CD scores of 0.60 and 0.55, respectively, showing their contribution to feature compactness and surface smoothness.

Overall, the complete model, which combines triplane encoding, multiscale wavelet features, U-Net fusion, and all loss terms, achieves the best reconstruction accuracy with a CD of 0.51, effectively balancing global consistency with local surface precision across the DTU dataset.

\vspace{1em}

\begin{table*}[!thbp]
\centering
\caption{Extended ablation study on the DTU dataset~\cite{jensen2014large}. This table evaluates the contribution of depth, wavelet, and autoencoder components. Lower Chamfer Distance (CD) indicates better reconstruction.}
\begin{adjustbox}{width=0.86\linewidth}
\begin{tabular}{l|c}
\toprule
\textbf{Ablation Setting} & \textbf{Chamfer Distance (CD $\downarrow$)} \\
\midrule
w/o Depth Wavelet Autoencoder module & 0.60 \\
w/ Depth only (no Wavelet, no Autoencoder) & 0.72 \\
w/ Depth + Wavelet (no Autoencoder) & 0.58 \\
w/ Depth + Autoencoder (no Wavelet) & 0.65 \\
w/ Color + Wavelet + Autoencoder & 0.84 \\
w/ Full wavelet pipeline (Depth + Wavelet + Autoencoder) &  \textbf{0.51} \\
\bottomrule
\end{tabular}
\end{adjustbox}
\label{tab:ablation-depth-wavelet-cd}
\end{table*}

To further validate the design of our depth-guided wavelet feature extractor, we perform an extended ablation study isolating the roles of depth input, wavelet transform, and the autoencoder. Removing the entire module leads to a CD of 0.60, while using only depth maps without wavelet or autoencoder yields a poorer result of 0.72, indicating insufficient geometric encoding from depth alone. Incorporating depth and wavelet features without the autoencoder improves performance to 0.58, showing that multi-scale spatial cues from wavelet features significantly enhance reconstruction. However, using depth with an autoencoder but without wavelet results in a higher CD of 0.65, highlighting the essential role of wavelet transforms in the encoding structure beyond the raw depth. Notably, relying on color wavelet features alone, even with the autoencoder, produces the weakest result (0.84 CD), confirming that geometric cues from depth are critical. Our complete wavelet pipeline, which combines depth, wavelet transform, and auto-encoder, achieves the best reconstruction with a CD of 0.51, demonstrating the complementary benefits of multiscale frequency representation and learned feature compression.

We further present visualization results from the ablation study on the wavelet autoencoder module design.
\begin{figure}[!th]
\includegraphics[trim=0.0cm 0cm 0.0cm 1.0cm, width=\linewidth]{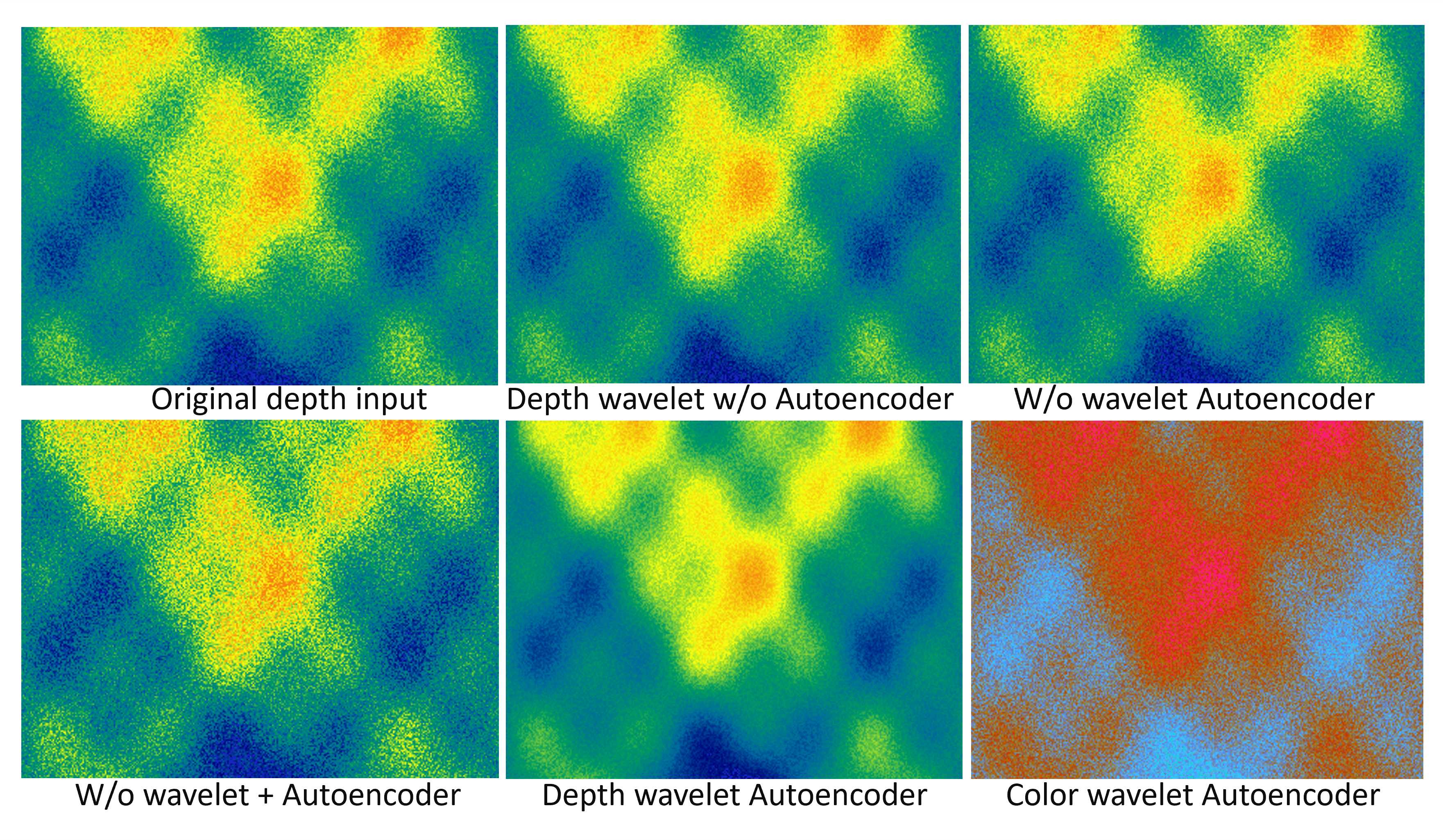}
\caption{Ablation study visualizations of different design combinations for the wavelet-transformed depth and autoencoder module.}
\label{fig:wavelet-ablation-viz}
\end{figure}

As shown in \ref{fig:wavelet-ablation-viz}, the raw depth contains noticeable local noise. Removing either the wavelet transform or the autoencoder results in feature maps that remain noisy and slightly blurred. In contrast, the second row, second column—representing the full wavelet-autoencoder pipeline—produces clean, fine-grained, and geometrically detailed feature maps. On the other hand, using color images for wavelet-based autoencoding yields a completely different feature distribution, failing to capture meaningful geometric information, as seen in the bottom-right corner of the color input example.

\section{Conclusion}

We propose an implicit SDF model that integrates wavelet-transformed depth features into a latent triplane feature space. By combining spatially decomposed wavelet representations with triplane embeddings, our approach enhances the preservation of geometric details. During inference, fused features are sampled along query rays and decoded into SDF values, enabling high-fidelity mesh reconstruction. Our model requires only monocular priors from state-of-the-art diffusion-based depth estimation models or a subset of selected heritage dataset
images. Compared to existing implicit SDF and explicit Gaussian Splatting methods, our approach achieves superior shape completeness while retaining intricate geometric details. Despite these advances, opportunities remain for further improvement. Future work could explore optimized sampling strategies to enhance computational efficiency. Additionally, integrating discrete Gaussian representations may accelerate training while maintaining high reconstruction fidelity. These extensions could expand our method's applicability to large-scale scenarios and real-time applications.

\revPatrice{A key assumption in our work is the photometric consistency of features across multi-view color images, which generally holds for Lambertian surfaces where appearance remains constant across viewpoints. However, this assumption breaks down for reflective or transparent materials due to non-Lambertian effects. To reconstruct such objects, alternative approaches are needed that explicitly model or mitigate view-dependent appearance changes. For example, incorporating uncertainty prediction to filter out reflective regions or leveraging shape priors to guide geometry inference—rather than relying solely on pixel feature matching and classical structure-from-motion—can improve robustness. While Gaussian Splatting has recently emerged as a promising prior art for handling appearance and view-dependent effects, our focus in this chapter has been on geometry rather than appearance. We therefore leave the exploration of such appearance-oriented methods and their potential integration with geometric representations as an important direction for future work.}


\chapter{Conclusion}\label{thesis:conclusion}

This dissertation presents a set of 3D geometric deep learning frameworks that integrate geometric constraints with deep learning models to address key computer vision tasks. Specifically, we explore geometric deep learning for camera pose estimation, point cloud registration, focal stack depth estimation, and implicit SDF-based 3D reconstruction. These techniques enhance mobile digital applications, providing robust, scalable, and high-quality solutions for the broader 3D vision pipeline. Beyond the academic contribution of making 3D learning representation more robust, accurate, and efficient, these models significantly affect Virtual Reality (VR), Augmented Reality (AR), and digital twin generation. Such high-fidelity 3D reconstructions and 3D asset generation enable virtual museums, digital twins, and interactive learning for a better education experience, creating immersive experiences that connect virtual with modern reality. With companies like Ubisoft, the gaming industry has already integrated reconstructed historical heritage into the entertainment industry, demonstrating great commercial value. 

This dissertation presents a clear roadmap for advancing 3D vision by integrating geometric priors and constraints (such as surfels, manifolds, and wavelet features) to condition and guide deep learning models. Many of these hybrid strategies and combinations can be adapted to solve other challenges in 3D vision. In the end, this thesis demonstrates how hybrid approaches can introduce physics-based methods to data-driven learning models, which can boost the learning model's performance.
\revAdalberto{Regarding evaluation metrics: in Chapter 3, the real-time camera pose estimation system uses algorithms involving random sampling. As a result, repeated runs on the same video sequence yield different outputs, so we performed multiple runs and reported the mean and standard deviation in the evaluation. In contrast, Chapters 4 and 5 focus on point cloud registration and depth estimation using deep learning models with fixed weights and random seeds in an offline setting, where the same input consistently produces the same output, making statistical analysis unnecessary. For 3D reconstruction, we performed multiple runs to extract meshes and selected the visually best quality mesh for metric evaluation.
}

\section{Camera pose Estimation}
For pose estimation, our manifold-based adaptive particle sampling method taps into natural geometric primitives, such as skylines and ground planes, as reliable reference points alongside IMU signals. This approach delivers outstanding accuracy and keeps drift impressively low over long-term periods, outperforming traditional camera pose estimation techniques. By providing the method with these intuitive geometric cues, it handles the correspondence uncertainty of appearance features in real-world natural settings, proving its superiority over traditional pose-tracking methods that often struggle with natural environments. Additionally, as it is quite challenging to run these models on mobile hardware with constrained computing resources and a limited power supply, we cannot run some heavy deep learning models on the hardware. In such cases, a more efficient algorithm should be preferred, or, as in our case, we directly use deep learning only for image segmentation, while the subsequent feature tracking, pose fusion, and pose estimation are non-learning-based. We implemented it in real-time on a 3D-printed gimbal platform mounted onto the polar stick to conduct the simulation in the lab with the camera pointing toward a landscape picture. Next, we put the whole system onto a UAV flying through natural outdoor environments to validate the performance in practice. The results confirm its practical use in challenging wild settings, where precision and reliability are very satisfying. This not only shows the robustness of the method but also inspires us to borrow similar ideas by leveraging natural cues or other permanent environmental features for vision tasks. Moreover, such manifold-based particle filters can better approximate the Lie group-based rotation representation and improve the pose estimation accuracy and efficiency of deep learning or partial learning models. This technique is important for aerial mapping or autonomous navigation, where steady pose estimation in natural environments is a foundational ability for such systems.

\section{Point Cloud Registration}

In point cloud registration, we introduced a surfel-based $\mathbf{SE}(3)$-equivariant network that achieves state-of-the-art accuracy on both indoor and outdoor benchmark datasets. This approach leverages a 2D surfel representation, incorporating surfel initialization from raw RGB-D depth maps or LiDAR point clouds, and consists of a shared E2PN encoder, a cross-attention module, and an MLP-based decoder. Extensive experiments across two datasets highlight its robustness and plausible accuracy. By explicitly embedding $\mathbf{SE}(3)$-equivariance into the framework, it effectively handles the geometric ambiguity in 3D data, proving its superiority over point cloud registration techniques that often fail with noisy data and input points with small overlaps. The modular surfel design also enables strong generalization across varied 3D environments, laying a solid foundation for consistent 3D global mapping. 

This method not only showcases good performance and robustness but also opens doors for broader real applications, like 3D mapping for robotics, 3D reconstruction, and even for downstream tasks like visual investigation and visual analysis in robotics or infrastructure engineering. It inspires fresh ideas too, encouraging us to explore these 2D Gaussian surfel primitives further for tackling sparse or noisy point clouds with small overlaps, including those with dynamic objects or uncertain data. Furthermore, refining correspondence and feature learning representation could boost the efficiency and performance of registration, especially under unstructured settings. Such equivariance constraints also improve the learning efficiency of geometric feature representation. This technique matters for applications like autonomous navigation or large-scale scene modeling, where reliable point cloud registration in noisy, real-world conditions is mandatory for next-level systems.

\section{Depth from Focal Stack Images }

We came up with a new model to estimate depth from focal stacks that overcomes the limitations of previous CNN-based methods. It uses a Transformer encoder to capture 2D spatial features and an LSTM module to learn depth cues across different stack images. Before the model, the multi-scale convolutional encoding further enhances detailed feature extraction, ensuring accurate depth prediction. Then we further use the lens distance constraint to create the loss as supervision for focal stack depth estimation learning.  

Through extensive evaluations, FocDepthFormer demonstrated state-of-the-art performance, outperforming prior methods across multiple benchmarks. More importantly, its ability to handle arbitrarily sized focal stacks offers practical advantages in real-world applications, where fixed stack sizes are often not flexible for training and testing. The pre-training strategy on monocular depth estimation datasets also proved effective in mitigating data scarcity, reinforcing the model’s adaptability.

Lastly, this work highlights the potential of hybrid deep learning architectures for depth estimation by merging attention and recurrence for latent feature fusion, along with focal stack constraints to differentiate focus and defocus cues better for depth feature learning. FocDepthFormer contributes to advancing 3D reconstruction techniques and lays the foundation for future research in computational photography and depth prediction for dense local viewpoint cloud generation.

\section{3D reconstruction}
Our proposed implicit SDF learning framework incorporates wavelet-transformed depth features into a triplane representation through triplane projection, achieving high-fidelity 3D reconstructions from limited image views. Such depth input for wavelet feature learning is generated by the latest monocular depth prior model, which is based on the diffusion model. By integrating spatial decomposition with neural representations, the model can help preserve intricate geometric details while maintaining global shape completeness.

In a nutshell, these contributions advance the progress of 3D reconstruction by demonstrating how such geometric constraints in high-frequency bands can be effectively integrated into deep learning frameworks. While challenges remain, including computational efficiency, extreme environmental conditions, sparse input view, and scalability to complex and large-scale scenes. This wavelet feature prior can be used as a plug-and-play module for other model backbones to boost the overall performance of the reconstruction. This research establishes a foundation for more robust, accurate, and geometrically consistent reconstruction results suitable for real-world VR/AR and robotic applications. It also inspires us to look back into the traditional spectral technique to combine the spectral features in high frequencies for reconstruction.

\section{Future Work}
While this dissertation advances 3D vision across multiple tasks, several key areas remain open for exploration. Future research can build upon these findings to enhance accuracy, efficiency, and adaptability in real-world applications.

\noindent \textbf{Robust Multi-Modal Fusion for Pose Estimation.}
Current pose estimation models rely on geometric priors and learning features, but robustness in more challenging environments with dynamics remains an issue. Future work could integrate feature tracking with lightweight neural implicit representations to improve performance under low illumination, occlusion, or dynamic scenes. We can also extend the framework to a multi-modal sensor fusion system, such as combining Lidar, event cameras to further enhance stability and precision.

\noindent \textbf{Scalable and Efficient 3D Registration.} \revAouda{The surfel-based registration method in this thesis provides accurate point cloud alignment but can struggle with large-scale scenes or cluttered environments with extremely low overlap. Incorporating Transformer-style global attention mechanisms or other probabilistic matching strategies could improve feature aggregation and robustness against sparse, noisy, or incomplete point clouds. Additionally, exploring fast, robust, memory-efficient correspondence learning could enhance registration performance for real-time SLAM and large-scale scene reconstruction. Lastly, integrating more efficient equivariant deep learning approaches, such as using 3D vector neurons \cite{deng2021vector} or modules that enforce rotation-equivariant properties, may improve the model's ability to generalize across varying orientations. However, as these designs often increase implementation complexity, a careful trade-off between rotation generalization and model design simplicity should also be considered.}

\noindent \textbf{Memory-Efficient and More Generalizable Depth Estimation.}
\revDavood{FocDepthFormer demonstrates strong depth estimation performance from focal stacks, but its high computational cost remains a bottleneck. Future work could explore lightweight Transformer architectures, knowledge distillation, or pruning techniques to enable deployment on edge devices or mobile platforms. Additionally, depth prediction from focal stacks faces inherent challenges, such as limited depth range and the need for numerous focal stack images to infer accurate disparity. The method also relies on clear focus and defocus cues, which can be difficult to extract in textureless or ambiguous scenes, like natural outdoor environments or plain walls, where such cues are absent. To address this, integrating focal stack techniques with state-of-the-art deep prior models trained on monocular or video input may help recover fine-grained depth details more robustly. Furthermore, adopting self-supervised learning paradigms could reduce dependence on large-scale, labeled focal stack datasets and improve generalization to unseen domains.}

\noindent \textbf{Real-Time and More Robust Implicit Reconstruction for Reflective Surface.}
Implicit SDF-based reconstruction offers high-quality 3D modeling, but real-time applications still require significant improvement in training and inference speed. Future work could focus on accelerating SDF optimization with diffusion models, real-time differentiable rendering, or even explore introducing the wavelet feature into the Gaussian Splatting method to leverage multi-band frequency features in the learning process. \revPatrice{Furthermore, addressing challenges in reconstructing reflective surfaces, such as glass or mirrors, by explicitly modeling these materials or using learned uncertainty masks could improve robustness in real-world scenes.} Lastly, extending the method to dynamic and deformable objects could enable real-time tracking of complex shapes and adaptive reconstruction, unleashing applications in robotics, AR/VR, and interactive digital modeling.

Beyond these individual research directions, this work also opens up broader opportunities for the fields:


Interactive AI for the virtual museum and efficient AI for a mobile robotic agent, connected to the Vision Language Model for real deployment on complex tasks, enable smarter robotics, where adaptive reconstruction enhances robotic interaction with complex dynamic environments.

Ultimately, this thesis demonstrates how geometric priors and deep learning models can be seamlessly integrated to advance 3D vision. Future research should focus on making these methods more efficient, generalizable, robust, and scalable, pushing the boundaries of robotics, AR/VR, digital twins, and other 3D vision fields beyond.
\cleardoublepage


\chapter{Valorization Plan}\label{ch:conclusion}
This chapter outlines a concrete valorization plan for the novel geometric deep learning techniques developed in this research, focusing on four key areas: pose estimation, point cloud registration, depth estimation from focal stacks, and 3D reconstruction using implicit SDF models. While these advancements have broad applications in autonomous systems, robotics, and AR/VR, this work emphasizes their significant impact on digital cultural heritage, exploring practical use cases and deployment strategies.

First, we present a pose estimation system leveraging geometric primitives such as skylines and ground planes to derive camera orientation from image frames. By integrating an adaptive particle filter, the approach enhances robustness against sensor drift and environmental disturbances, demonstrating real-time feasibility on embedded hardware.

Next, we introduce a surfel-based $\mathbf{SE(3)}$-equivariant model for point cloud registration. Using a modular surfel representation, our method enables state-of-the-art alignment performance across diverse 3D scenes, offering potential extensions for real-time mapping and reconstruction.

For depth estimation, we propose FocDepthFormer, a hybrid Transformer-LSTM model that effectively aggregates focal stack images to recover fine depth details. While achieving high accuracy, the model's computational efficiency can be further optimized for large-scale applications such as defocus-based image synthesis.

Finally, we present an implicit SDF model that fuses wavelet-transformed depth features with triplane embeddings to improve shape reconstruction. Our approach surpasses explicit Gaussian Splatting in preserving fine geometric details, with future directions including optimized sampling and hybrid implicit-explicit representations for real-time performance.

Together, these contributions provide a robust foundation for applying geometric deep learning in cultural heritage digitization, offering scalable solutions for high-fidelity 3D reconstruction, localization, and scene understanding.

\label{ch:cp5}

\section{Introduction}
This chapter presents a comprehensive valorization plan for the geometric deep learning techniques developed in the previous chapters, with a primary focus on applications in digital cultural heritage. The plan outlines this research's social, practical, and academic impacts, structured around four core aspects: pose estimation from image frames, point cloud registration, focal stack depth estimation, and 3D reconstruction using an implicit SDF model. These techniques address critical challenges in 3D computer vision while advancing Virtual Reality (VR) and Augmented Reality (AR) technologies, with significant contributions to cultural preservation and education.

By bridging cutting-edge geometric deep learning research with practical applications, this work enhances digital preservation, interactive education, virtual tourism, archaeological analysis, and digital heritage restoration. High-fidelity 3D reconstructions enable the creation of digital twins for historic artifacts and buildings, while immersive learning tools enhance user interactive experience with cultural heritage. 

Before discussing the detailed valorization plan, let us first recap the key techniques developed in this thesis:

\noindent\textbf{Pose estimation for the camera via natural cues.} Accurate pose estimation is crucial for 3D data capture and reconstruction, especially in challenging outdoor environments where traditional feature matching is unreliable when the inlier/outlier ratio is low. This research utilizes natural geometric cues, such as skylines and ground plane approximations, to estimate camera orientation. By extracting and analyzing these features, the system integrates inertial sensor data with vision-based estimation for robust orientation fusion. This method enhances stability in outdoor settings and significantly improves pose accuracy, making it particularly valuable for VR/AR applications in uncontrolled wild environments.


\noindent\textbf{Point cloud registration.} Aligning multiple point cloud frames is essential for constructing complete 3D scans. This research introduces novel algorithms leveraging data equivariance and uncertainty modeling to enhance registration accuracy. These techniques facilitate large-scale reconstructions of sculptures and buildings by merging partial scans into consistent models.

\noindent\textbf{Focal stack depth estimation.} Depth estimation from focal stacks captures fine details by analyzing images taken at varying focal distances. This approach enables high-precision 3D reconstruction of paintings, textiles, and calligraphy, offering a flexible method for digitizing cultural artifacts and heritage.

\noindent\textbf{3D reconstruction of buildings.} Chapter 6 explores 3D reconstruction techniques for architectural structures, enabling detailed virtual models for historical preservation and visualization. These reconstructions support VR/AR applications for virtual tourism, interactive education, and heritage conservation, providing an accessible platform for studying architectural and cultural history.

Each of these contributions advances VR/AR technologies, mobile robotics, and self-driving, but here we will focus on the demo case for cultural heritage applications, facilitating digital preservation and interactive education, to give readers a more concrete understanding of the usability of these techniques. By integrating geometric deep learning with the digital cultural industry, this research paves the way for a cross-disciplinary field to use Geometric Deep Learning for history, culture, arts, and the gaming industry.
\section{Use case of Pose Estimation for Camera Frame}


Capturing stable, high-quality images and videos with UAV-mounted cameras, while simultaneously estimating the camera pose in real-time, is crucial for 3D applications such as reconstruction and 3D mapping in dynamic outdoor environments where wind and other disturbances can cause significant jitter. Chapter 3 presents an advanced camera pose estimation technique that can stabilize the camera by leveraging sensor fusion and natural geometric cues like skylines as reference. By accurately determining the camera orientation in real-time, the system compensates for unwanted movements, ensuring smooth, jitter-free video footage for downstream computer vision tasks. This improves data quality, enhances image quality and pose precision for mapping and surveying, and increases operational efficiency and stability by reducing the need for post-processing, ultimately making data collection in the wild more reliable for 3D applications like reconstruction.

\begin{figure}[!thbp]
\centering
\includegraphics[width=0.92\linewidth]{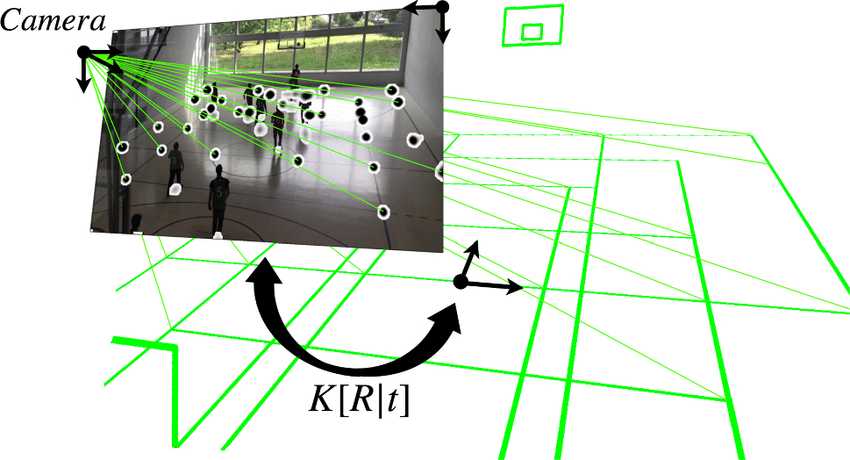}
\caption{Camera pose is calculated from the live-streaming images, demo image used from \cite{citraro2020real}.}
\label{fig:camera-pose-calc}  
\end{figure}

Figure \ref{fig:camera-pose-calc} illustrates pose estimation from camera images based on tracking features of images. These 2D features are projected into the 3D space to determine the camera pose.

\subsection{Validation}
\begin{figure}[!thbp]
\centering
\includegraphics[width=0.96\linewidth]{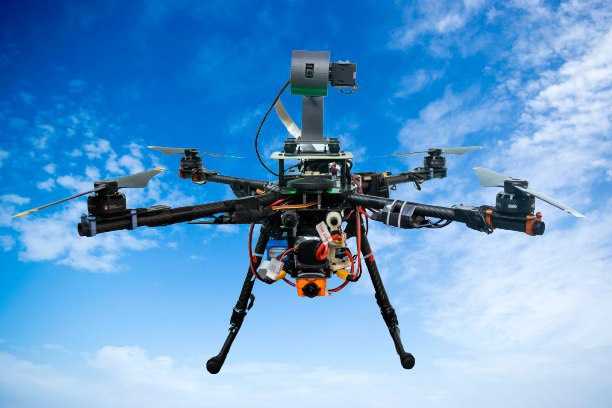}
\caption{Experimental Hardware Test Setup: Our camera pose estimation system is mounted on top of a UAV for real-world testing in a natural environment.}
\label{fig:gimbal-drone}  
\end{figure}

A real demo of using a UAV equipped with a gimbal system, including our camera pose estimation framework for a flight to collect scan images of a building,g is provided in Figure \ref{fig:gimbal-drone},

\begin{figure}[!thbp]
\centering
\begin{subfigure}{0.49\linewidth}
    \centering
    \includegraphics[width=\linewidth]{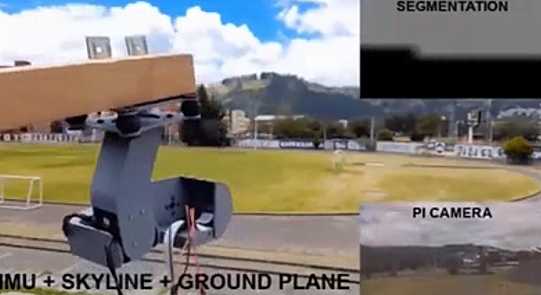}
    \caption{Camera pose estimation system tested on the ground by mounting it on a stick.}
    \label{fig:ground-test}
\end{subfigure}
\hfill
\begin{subfigure}{0.49\linewidth}
    \centering
    \includegraphics[width=\linewidth]{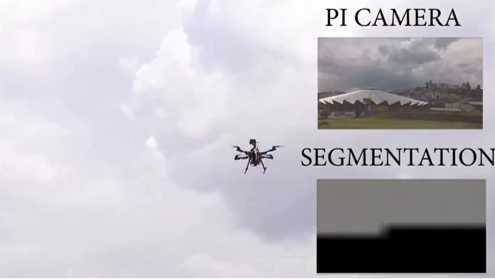}
    \caption{Camera pose estimation system tested in the air by mounting it on a UAV}
    \label{fig:air-test}
\end{subfigure}
\caption{Camera pose estimation system tested both on the ground and in the air.}
\label{fig:gimbal-setup}
\end{figure}

We conducted two tests to demonstrate practical applicability: one on the ground, where the hardware was mounted on a wooden stick facing the landscape from a rooftop for easy Ground Truth collection (Figure \ref{fig:ground-test}), and another in the air using a UAV. In the aerial test, the onboard camera transmitted live images remotely while a real-time deep learning model generated skyline boundary segmentation output (Figure \ref{fig:air-test}).

The camera pose tracking system can estimate pose robustly and quickly over long periods, without orientation drift. This demonstrates the reliable performance of our proposed tracking system, even when using a gimbal-based controller, to stabilize the camera orientation robustly.

\subsection{Target Customers}

This research work targets customers who require precise camera pose estimation for UAV-based 3D reconstruction and image capture in natural environments like mountains. For example, companies in aerial surveying and 3D mapping, such as those providing global mapping data for environmental monitoring and analysis, can leverage our efficient geometry-based orientation tracking system to improve the quality of camera pose for 3D reconstructions in challenging outdoor settings. Although the demo case in this chapter focuses on digital cultural heritage, it has applications beyond this field and can be extended to many other fields. For instance, Drone manufacturers and robotics companies, particularly those focused on autonomous navigation in outdoor terrains, can integrate our method to enhance motion stabilization and reduce orientation drift, ensuring high-quality image capture for downstream tasks. Additionally, entertainment industries like filmmaking and natural life documentation, which often need to operate drones in unpredictable natural conditions, can benefit from the system's ability to mitigate motion blur and stabilize camera orientation using natural cues like skylines and ground planes. The proposed approach, with its real-time implementation on cheap embedded devices like the Jetson Nano, offers a practical and affordable solution for the software or as a complementary tool to the sensor fusion system for pose estimation. This can also be applied to cross-disciplinary research, such as for archaeology researchers at KU Leuven.

\subsection{Economic Value}

The primary economic value of our camera pose estimation system lies in its ability to boost the efficiency and quality of UAV-based imaging, creating opportunities in outdoor environments. By licensing this technology or partnering with drone manufacturers and aerial imaging companies in regions like Asia and Europe, we can expect the fast-growing drone market projected to reach billion-dollar valuations by 2030, according to the drone market report \cite{kipponen2020defining}. This research work offers a software toolkit solution bundled with drone manufacturing firms, driven by demand in surveying, environmental monitoring, and autonomous navigation in the wild. Our method design and real-time capabilities on embedded hardware cut the need for costly computational resources, offering a cost-effective solution for small- and medium-sized companies in these sectors. Additionally, its applications in filmmaking, natural life documentation, and virtual tourism, where reliable, high-quality 3D reconstructions are key for immersive experiences to support industry economic goals through sustainable data collection practices. By tackling the natural challenges of pose estimation, such as drift and noise in outdoor settings, this research delivers scalable, robust solutions that industries can adopt to enhance operational performance, reliability, and the overall value of drone applications.

\section{Use case of Point Cloud Registration}
Archaeology involves intricate engineering to excavate artifacts. Fragile materials such as porcelain can degrade over time due to many environmental factors. Reconstructing these artifacts from numerous fragments is labor-intensive, akin to solving a complex puzzle. To address this, each fragment is scanned and advanced registration techniques, as described in Chapter 4, are used. These techniques, known for their robust performance under uncertainty, enable the precise and robust assembly of fragments into a complete 3D model, significantly enhancing the efficiency and accuracy of artifact recovery.

\subsection{Validation}

\begin{figure}[!thbp]
\centering
\includegraphics[width=0.5\linewidth]{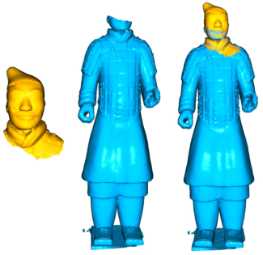}
\caption{Registration of source and target scans of Terracotta Warriors \cite{wang2024enhancing}.}
\label{fig:debris}  
\end{figure}

Figure \ref{fig:debris} shows the registration of head scans in the body scan of terracotta warriors based on the overlapping regions, based on the proposed registration technique. This task is crucial for archaeologists and historians aiming to reconstruct and understand cultural artifacts from fragmented remains for digital recovery. The intricate process involves identifying and matching numerous scanned parts to recreate a complete and accurate representation of the original artifact, thereby preserving and interpreting historical and cultural heritage.

\subsection{Target Customers}
This point cloud registration research targets customers who require a robust and accurate 3D alignment for large-scale reconstruction from Lidar scan tasks in diverse environments. Archaeologists can use this technique to assist in assembling cultural fragments by aligning scanned pieces within computer simulation software. Robotics companies developing autonomous navigation systems can leverage our surfel-based $\mathbf{SE(3)}$-equivariant framework to improve spatial alignment performance in real-world indoor and outdoor settings, ensuring reliable mapping for applications like building global 3D maps for city blocks. Industries involved in 3D mapping and surveying, such as those in architecture or urban planning, can benefit from our framework's ability to handle noise and large rotations, producing consistent point cloud registrations for detailed digital twins of buildings or objects. Additionally, augmented reality (AR) developers creating immersive experiences can use our approach to align point clouds from LiDAR scans or depth maps, enabling the integration of virtual objects into real-world scenes for interaction. By addressing the limitations of traditional and learning-based methods, our framework offers a scalable solution for these fields, ensuring high accuracy in challenging registration scenarios.

\subsection{Economic Value}
The economic value of our surfel-based point cloud registration system lies in its ability to achieve robust and efficient 3D alignment, unlocking opportunities in high-growth industries. Licensing this technology to robotics and AR companies presents a promising billion-dollar market, particularly in the entertainment and education sectors. Our model design enhances robustness by reducing sensitivity to noise and large rotations through explicit $\mathbf{SE(3)}$-equivariant features, making registration more reliable even when the inlier-to-outlier ratio is low. Additionally, this technique mitigates the need for extensive scan transformation augmentations during training, offering a cost-effective solution for small- and medium-sized enterprises.  

Beyond these applications, our system supports 3D mapping for architecture and surveying, enabling the creation of digital twins that reduce project costs and promote sustainable digital development. By addressing key challenges in point cloud registration, this research provides a high-precision tool that industries can leverage to enhance operational efficiency and drive innovation in 3D reconstruction applications.



For assembling complex cultural heritage fragments, advanced algorithms such as PuzzleFusion++ \cite{wang2024puzzlefusion} can be used. These methods utilize iterative techniques to accurately search, match, and align fragments, particularly valuable when dealing with pieces that have minimal overlapping areas and irregular shapes that make candidate matching spurious and challenging. The algorithm begins by denoising and verifying the scanned fragments to have a clean input for geometry representation learning. In each iteration, the method incrementally refines the fit of the pieces, leveraging geometric and topological features to ensure consistency and accuracy between the pieces. After refinement, the pieces are fitted more precisely, forming a coherent and accurate reconstruction of the original artifact. The ability to reconstruct artifacts digitally not only aids in preservation and study but also allows for virtual restoration and exhibition, making cultural heritage more accessible to the public, furthermore, the complete assembly digital scan can help the archaeologist to recover the full artifact by indicating the positions of each piece in the complete scan. 

\section{Use case of Depth Estimation from Focal Stack}
Depth estimation from the focal stack technique can be used to extract 3D information from cultural paintings or sculptures, as artists have created these 2D paintings according to the rules of perspective projections.

\begin{figure}[!thbp]
\centering
\includegraphics[width=0.94\linewidth]{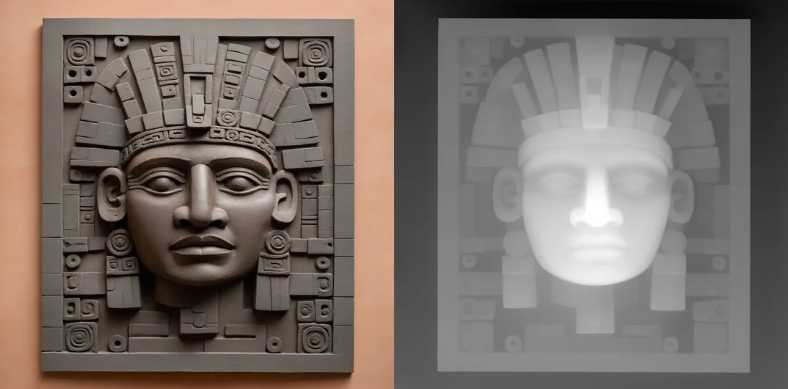}
\caption{Depth estimation for the planar sculpture of Marigold. The original image resource is from the \href{https://www.reddit.com/r/StableDiffusion/comments/18ge6cq/the_first_double_image_is_sd_on_the_left_with_a/}{public website link}.}
\label{fig:sculpture-depth}  
\end{figure} 

Figure \ref{fig:sculpture-depth} presents an example to demonstrate the application of depth estimation from monocular image techniques in the 3D modeling of art and sculpture. Depth estimation from the image is particularly well-suited for creating 3D models of intricate details and sculptures with planar constraints or primarily one-sided feature distributions. Unlike monocular depth estimation, the focal stack depth estimation technique captures a sweep of images of the target at varying focus distances and then uses algorithms to analyze the areas of sharp focus in each image to reconstruct depth information without requiring camera motion. Focal stack imaging offers several key advantages for 3D digitization when camera motion is prohibited: as a passive perception technique, it enables non-invasive capture without physical contact, which is essential for preserving fragile artwork compared to active methods like LiDAR scanning. The approach creates high-resolution detail by utilizing multiple focused images with large stack sizes, capturing fine features such as intricate decorations and facial expressions. It excels at handling complex geometries with varying depth planes by differentiating subtle depth changes along stack dimensions, as evidenced by the accurate 3D feature representation, which is critical for the precise digital reproduction of sculptures without the need for camera motion, simply by adjusting the focal lens geometry distances continuously to reconstruct fine details.

\subsection{Validation}

\begin{figure}[!thbp]
\centering
\includegraphics[width=0.94\linewidth]{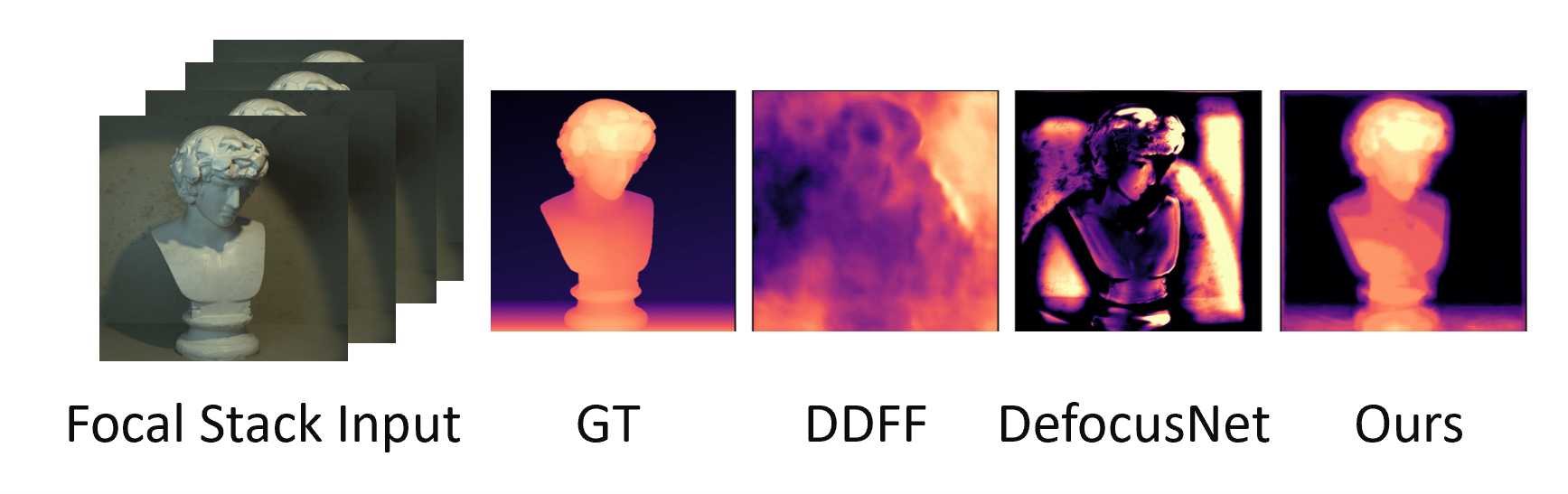}
\caption{Depth prediction from focal stack technique applied to heritage sculpture scanning.}
\label{fig:inpainting}  
\end{figure} 

The sweep of focal stack images of a sculpture in Figure \ref{fig:inpainting} demonstrates how modern 3D computer vision techniques can extract 3D information from 2D focal stack images, especially those created with strong perspective principles by artists. The 3D reconstruction of historic artwork relies on visual analysis. Perspective analysis inverts spatial relationships by examining how artists employed pinhole camera models to create depth; the depth information can be inferred by our method through the focus/defocus cues in focal stack images, as the target scan details may be at the millimeter level, such as painting strokes, as fine-grained details. Thus, traditional viewpoint change-based methods for 3D reconstruction are not applicable, as the pose error is usually several orders of magnitude larger.

This application of depth estimation to paintings not only provides interesting insights into the use of perspective and camera focal geometry but also opens up possibilities for creating 3D visualizations or augmented reality experiences based on classical artworks. It represents an intersection of art history and modern 3D computer vision technology. For instance, recent depth estimation techniques like ``Depth Anything'' \cite{yang2024depth} and ``Marigold'' \cite{ke2024repurposing} already show the great potential of the latest deep learning models applied to such challenging problems through the powerful learning representation ability of these models.

\subsection{Target Customers}
This depth estimation research is aimed at customers who need precise 3D models for cultural heritage preservation with small-scale level change, and academic study. Museums and cultural heritage organizations can use FocDepthFormer to create highly accurate 3D models of sculptures and artifacts like paintings, enabling detailed digital replicas for exhibitions or digital preservation without physical contact, which is crucial for fragile ancient artifacts at risk of being damaged. Researchers and art historians worldwide can benefit from these digital models for in-depth analysis of these historical artworks, studying intricate details like surface textures or structural patterns that might otherwise be lost to time. Additionally, educational institutions, museums, and virtual exhibition platforms can utilize these 3D reconstructions to provide immersive experiences, allowing global audiences to inspect cultural artifacts in detail. By addressing the challenge of handling arbitrary focal stack sizes, our method offers a flexible solution for these fields, ensuring high-quality depth estimation for diverse cultural business and research needs.

\subsection{Economic Value}
The economic value of FocDepthFormer lies in its ability to produce detailed 3D models through focal stack depth estimation, creating 3D assets in cultural heritage and related industries with just a sweep of images. By licensing this technology to museums and cultural preservation organizations, we hope to find opportunities in the growing market for affordable digital heritage solutions. The technique can expand to research institutes to document artifacts digitally. The creation of digital replicas for exhibitions or online platforms enhances global access to more cultural resources. The method is a cost-effective approach, using pre-training on monocular depth datasets and focal stack datasets to reduce reliance on expensive 3D data or multi-view data, making it an affordable option for smaller institutions, while its ability to preserve the digital twin of degrading and historic sculptures ensures long-term value for society through digitalization. Furthermore, applications in virtual exhibitions and education can generate revenue through subscription-based access to the use of copyright of such digital 3D models, aligning with sustainable preservation practices. By overcoming limitations in traditional depth estimation, this research provides a scalable, high-precision tool that industries can use to advance cultural preservation and reproduction.

\section{Use case of 3D Reconstruction via implicit SDF with Wavelet Feature Prior}

Geometric deep learning techniques can help transform various aspects of the industry, such as robotics \cite{wong2025survey}, self-driving, VR/AR, 3D printing or the gaming industry; here, we will focus on introducing the latter three fields in this part, which are more closely tied to the usage of digital cultural heritage.

\noindent \textbf{VR/AR for Virtual museum.} In this digital era, 3D reconstruction techniques are essential for generating high-quality 3D assets. Cultural heritage artifacts can be digitally preserved eternally through the creation of digital 3D reconstruction, typically reconstructed using non-learning or deep learning models applied to multi-frame point clouds or images. These geometric models serve as the foundation for downstream tasks such as photorealistic rendering and relighting, enabling a more immersive and interactive experience of cultural heritage assets in high-quality virtual environments.

Once the digital twin generation becomes affordable and scalable, cultural heritage artifacts can establish virtual museum databases offering multiple benefits: VR/AR spatial computing creates immersive experiences that surpass traditional 2D displays by enabling visitors to navigate freely through the historical environments; interactive exploration allows users to manipulate virtual artifacts and navigate reconstructed ancient sites without access limits, promoting both educational and entertainment purposes; the mixing of reality with virtual properties offers new analytical perspectives on cultural heritage; global accessibility eliminates expensive travel cost while aiding digital preservation of fragile artifacts for the public to enjoy; educational value is enhanced through powerful interactive tools that place students within historical contexts; and heritage research capabilities are also expanded as VR environments provide platforms for visual analysis and efficient sharing of digital twins, potentially generating more great business insights.

\begin{figure}[!thbp]
\centering
\includegraphics[width=1.0\linewidth]{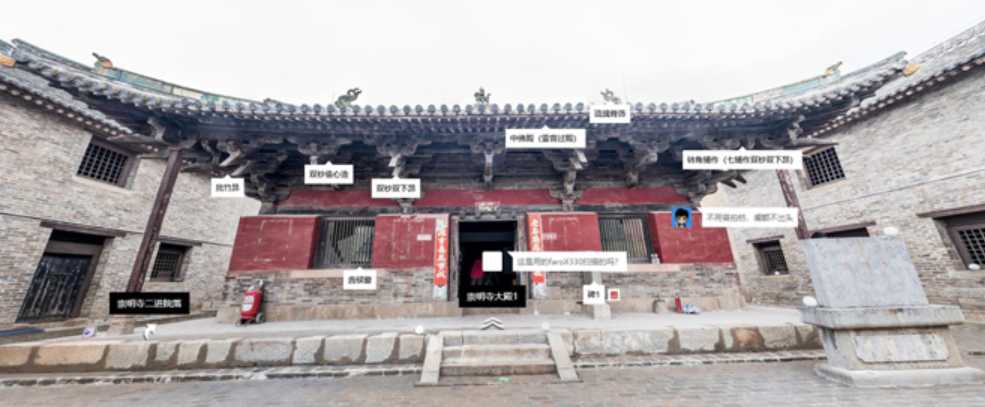}
\caption{3D AR Guides overlaid with the real temple. The image was retrieved from public \href{https://quod.lib.umich.edu/a/ars/13441566.0050.004/--vr-heritage-chinese-cultural-heritage-in-the-digital-age?rgn=main;view=fulltext}{website link}.}
\label{fig:temple-label}  
\end{figure}

By harnessing VR and AR technologies, virtual museums offer visitors immersive experiences that seamlessly mix technology, education, entertainment, and cultural art appreciation in unprecedented ways. Specifically, in these virtual museums, visitors can examine artifacts with close-up checks, view them from any angle, and even manipulate virtual artifacts. The integration of AR technology further enhances the experience by overlaying digital information onto physical spatial spaces or objects in real-time, providing additional guide introduction for tourism, as illustrated in Figure \ref{fig:temple-label}, it shows a traditional Chinese temple complex with architectural labels, illustrating the potential for visualizing cultural heritage sites through 3D digital reconstruction and augmented reality (AR). By overlaying digital information onto the physical structure, AR technology can provide visitors with a more immersive and educational experience, so that it allows for a vivid presentation of the temple components without physically altering the site.

\noindent \textbf{Gaming industry.}
The famous game design company, Ubisoft, has been a long-lasting pioneer in integrating historical 3D reconstructions of buildings and environments into their game products, particularly in the Assassin's Creed series. This approach has evolved to incorporate VR and AR technologies, enhancing the immersive experience for game players and the educational value of their games to learn history and culture.

Ubisoft development teams collaborate closely with historians and archaeologists to create detailed, historically accurate reconstructions of real ancient cities and landmark sites. These range from Ancient Egypt in Assassin's Creed Origins to Renaissance Italy. The company has also experimented with VR adaptations of their reconstructed sites for a better immersive experience. The level of detail and accuracy in these reconstructions has been so impressive that they are now used for educational purposes beyond gaming \cite{bevilacqua20223d}. As such, high-quality digital twins of real cultural heritage sites have great potential for the gaming industry.

\begin{figure}[!thbp]
\centering
\includegraphics[width=1.0\linewidth]{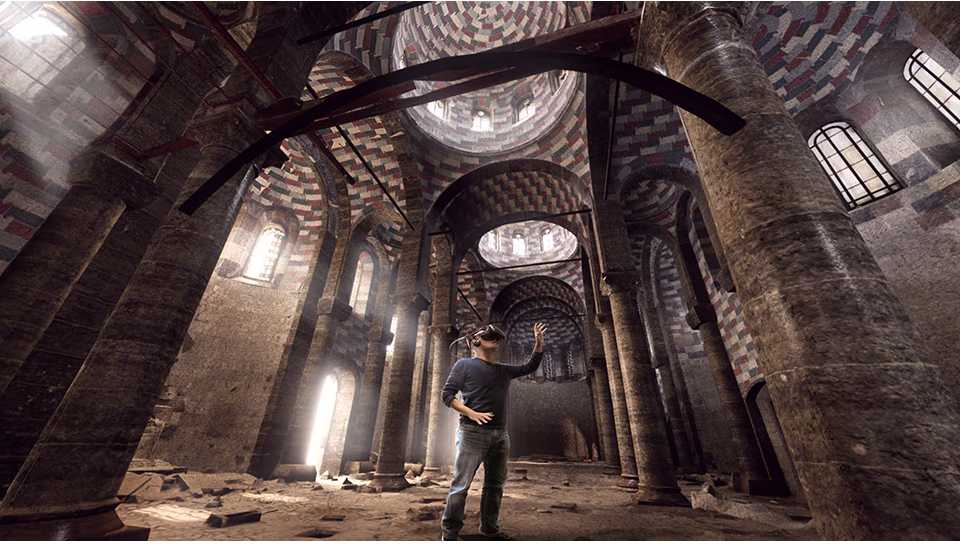}
\caption{The demo image is from public \href{https://news.ubisoft.com/en-us/article/6OUATm5pVU1tO6O72rTAAA/ubisoft-creates-vr-experience-at-smithsonians-ageold-cities-exhibition}{website link}.}
\label{fig:game}  
\end{figure} 

Figure \ref{fig:game} demonstrates the beautiful interior of a historic cathedral with intricate decorations, arched windows, and a huge domed ceiling. The player can be immersed in exploring this digital architectural heritage. Such collaborations between game developers and cultural institutions are opening up new possibilities for education, conservation, and public engagement with historical architecture, making centuries-old buildings accessible to people around the world through the power of technology. Lastly, this collaborative project can be scaled to other cultural relics in global cities.



\begin{figure}[!thbp]
\centering
\includegraphics[width=\linewidth]{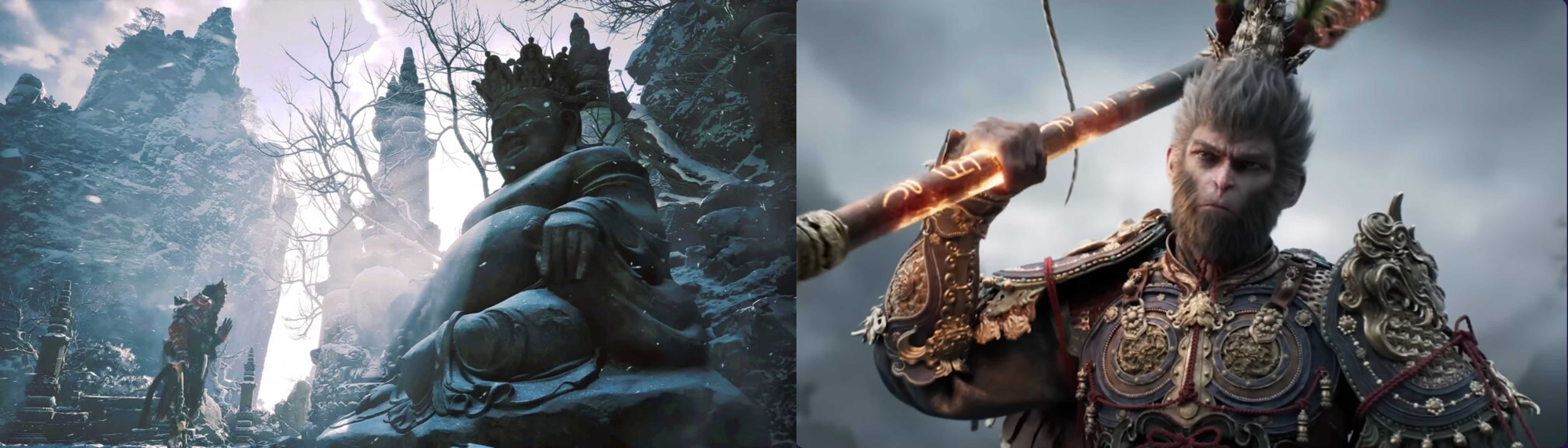}
\caption{Chinese 3A Video Game, Black-myth Wukong, with many cultural heritage environments created from real scans of cultural buildings in \href{https://blog.cdkeys.com/black-myth-wukong/}{Chinese cultural sites}. The images depict the game's scene and character.}
\label{fig:wukong}  
\end{figure}

Another example from the gaming industry is the video game ``Black Myth Wukong'' from China, which has generated considerable attention in the gaming world of 2024. The game story is based on an old Chinese mythology novel published in the Ming dynasty, and the video game has provided quite an immersive experience based on real cultural building scans across China, including traditional Chinese towers, pagodas, tower, and Buddha sculptures, as demonstrated in Figure \ref{fig:wukong}.

These cases demonstrate the strong potential of 3D reconstruction techniques for generating assets in virtual gaming, a rapidly growing consumer market within the broader entertainment industry. 

\noindent \textbf{3D Printing.} 

\begin{figure}[!thbp]
\centering
\includegraphics[width=1.0\linewidth]{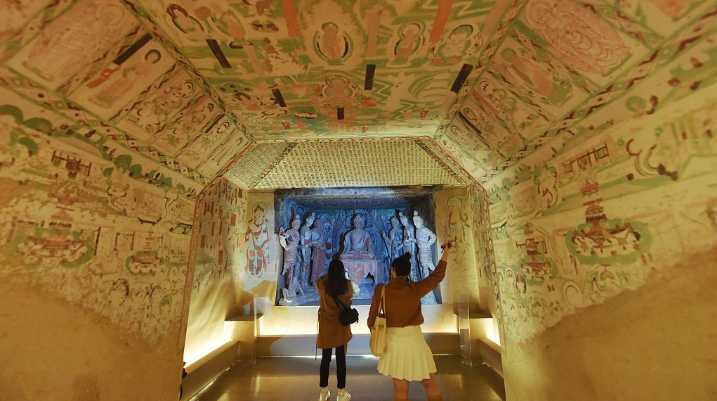}
\caption{3D printed copy of Dunhuang Mogao Grottes inside the mountain caves. The image was retrieved from public \href{https://news.cgtn.com/news/2023-02-14/China-protects-cultural-heritage-with-digitalization--1hpzr0BIBy0/index.html}{website link}.}
\label{fig:dunhuang-grottes}  
\end{figure}

Figure \ref{fig:dunhuang-grottes} exhibits another critical application of how 3D digital reconstruction technology is revolutionizing the reproduction of cultural heritage sites like the Dunhuang Mogao Grottoes. By creating detailed digital models of these ancient cave temples, preservationists can now offer immersive experiences to visitors far from the original location through 3D printing. This technology not only allows tourists to enjoy but also enables the creation of precise 3D-printed replicas for other business purposes. Such reproductions can be displayed in museums worldwide, bringing the intricate art details and historical significance of the Dunhuang Mogao Grottoes to a global audience. This approach balances the need for the preservation of delicate original sites, which often have limited access due to conservation concerns or fragile preservation environments, with the desire to share these cultural treasures through 3D printing. The 3D printing results of such a digital twin of heritage paintings in caves presented here, with its vivid colors and detailed paintings, demonstrate how technology can bridge the gap between remote cultural sites and tourists and art enthusiasts, making these fragile and inaccessible heritage sites available for study and appreciation in new physical ways.

\subsection{Validation}

\begin{figure}[!thbp]
\centering
\includegraphics[width=\linewidth]{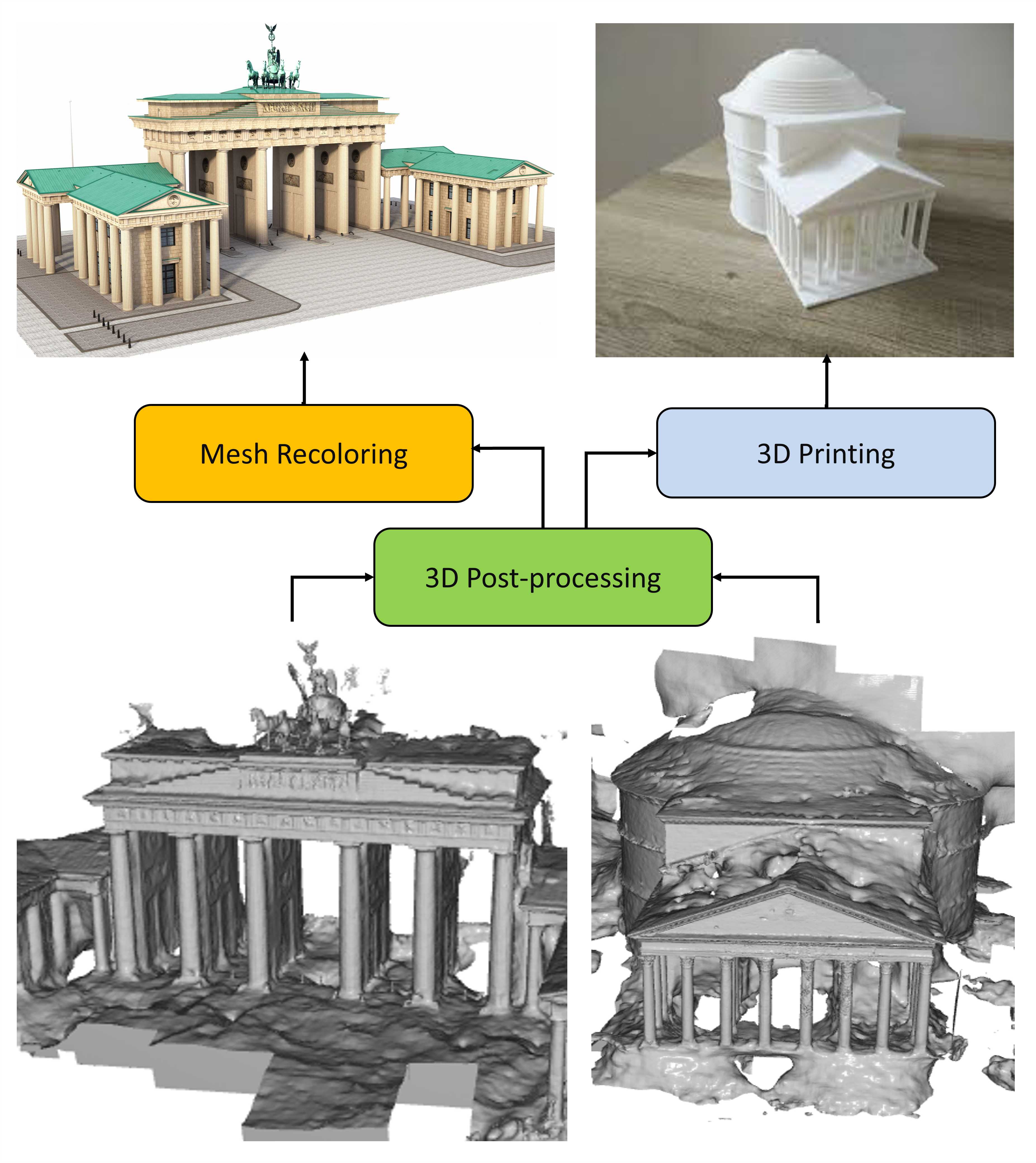}
\caption{Practical application of my 3D reconstruction model: The heritage site mesh is first reconstructed from online images, followed by post-processing and specific downstream tasks. The left part shows recoloring to create a colored mesh, while the right part illustrates preparation for 3D printing.}
\label{fig:valor-recons}  
\end{figure}

Our proposed reconstruction technique in Chapter 6 enables high-fidelity 3D reconstruction with fine-grained geometric details being preserved from multi-view images. Once the reconstruction is converted into a mesh, we can further apply post-processing steps to refine it by completing missing regions and removing noisy artifacts or unwanted background regions, ensuring a clean and accurate mesh representation. This post-processed mesh of cultural heritage sites can then be utilized for various downstream applications, such as high-quality rendering, recoloring (the left branch in Figure \ref{fig:valor-recons}), or 3D printing of digital models (the right branch in Figure \ref{fig:valor-recons}), as illustrated in Figure \ref{fig:valor-recons}. This can prove the great usage potential of our reconstruction method.

\revAdalberto{The digitally reconstructed geometry can serve as a 3D representation primitive, such as a point cloud or mesh surface, for downstream applications like semantic segmentation \cite{abdelreheem2023satr}, colorization \cite{leifman2013pattern}, and texturing \cite{siddiqui2022texturify}. As a fundamental shape representation, it provides shared structural knowledge for various object and target representations, enabling the reuse of geometry for diverse post-processing tasks such as color and texture editing of 3D assets.
}

\subsection{Target Customers}
This 3D reconstruction research targets customers who need high-fidelity 3D models for diverse applications in virtual reality, the gaming industry, and the 3D printing industry. Educational or, museums, and training providers using VR/AR can leverage our wavelet-feature-conditioned implicit SDF model to create detailed, accurate historic architectures or cultural artifacts, enabling students or users to explore reconstructed cultural heritage sites or artifacts in the digital world. Gaming companies, especially those developing historically themed storytelling games, can use our method to generate detailed 3D assets from realistic data like images for digital worlds, improving players' immersive experience with virtual environments in games. Additionally, museums and cultural heritage organizations can benefit from 3D printing the reconstructed digital meshes to produce precise replicas of artifacts or cultural architectures, supporting preservation to make a copy of original heritage properties and public exhibitions to reduce the protection cost and risk for the valuable heritage collections. By addressing the challenge of capturing and learning fine-grained geometric details, our approach offers a comprehensive solution for these three fields, ensuring high-quality reconstructions across various object or scene scales.

\subsection{Economic Value}
The economic value of our 3D reconstruction system lies mainly in its ability to create detailed, accurate 3D models, opening opportunities in high-demand industries like virtual reality, the gaming industry, and 3D printing. By licensing this technology to VR/AR developers, we can penetrate the growing market for immersive training or exhibition programs, reaching a vast market, particularly in Asia, Europe, and the USA, covering a population of over multiple billion people. Such reconstructions of historical sites or cultural artifacts enhance training and educational experiences, which can be offered as a digital subscription service, allowing users to access our model via web or mobile app for a valid period. In the gaming industry, companies can easily integrate our high-fidelity 3D assets into their pipelines, reducing development costs for realistic environments and saving efforts in 3D asset creation, particularly for small or medium game development teams, driven by the growing market, game players, and investments. Furthermore, the ability to 3D print accurate replicas of reconstructed meshes offers economic potential for museums and cultural heritage institutions to broadcast or guide the protection of original, old, and fragile heritage properties, enabling them to sell or exhibit artifact replicas while supporting sustainable preservation practices. By overcoming limitations in capturing fine geometric details, this research provides a cheap, flexible, and high-precision solution that industries can adopt to drive innovation in digital 3D applications and create great value for the mass market in entertainment or cultural research fields.






\bibliographystyle{acm}
\bibliography{allpapers}

\begin{thebibliography}{100}

\bibitem{abdelreheem2023satr}
{\sc Abdelreheem, A., Skorokhodov, I., Ovsjanikov, M., and Wonka, P.}
\newblock Satr: Zero-shot semantic segmentation of 3d shapes.
\newblock In {\em Proceedings of the IEEE/CVF International Conference on Computer Vision\/} (2023), pp.~15166--15179.

\bibitem{slic}
{\sc Achanta, R., Shaji, A., Smith, K., Lucchi, A., Fua, P., and Susstrunk, S.}
\newblock Slic superpixels compared to state-of-the-art superpixel methods.
\newblock {\em IEEE transactions on pattern analysis and machine intelligence 34\/} (05 2012).

\bibitem{agarwal2022depthformer}
{\sc Agarwal, A., and Arora, C.}
\newblock Depthformer: Multiscale vision transformer for monocular depth estimation with global local information fusion.
\newblock In {\em 2022 IEEE International Conference on Image Processing (ICIP)\/} (2022), IEEE, pp.~3873--3877.

\bibitem{alpert1999hybrid}
{\sc Alpert, B.~K.}
\newblock Hybrid gauss-trapezoidal quadrature rules.
\newblock {\em SIAM Journal on Scientific Computing 20}, 5 (1999), 1551--1584.

\bibitem{robot-IMU}
{\sc Alquisiris-Quecha, O., and Martinez-Carranza, J.}
\newblock Video stabilization of the nao robot using imu data.
\newblock In {\em Robot Operating System (ROS)}. Springer, Cham, 2020, pp.~147--162.

\bibitem{amini2022yolopose}
{\sc Amini, A., Selvam~Periyasamy, A., and Behnke, S.}
\newblock Yolopose: Transformer-based multi-object 6d pose estimation using keypoint regression.
\newblock In {\em International Conference on Intelligent Autonomous Systems\/} (2022), Springer, pp.~392--406.

\bibitem{anwar2021deblur}
{\sc Anwar, S., Hayder, Z., and Porikli, F.}
\newblock Deblur and deep depth from single defocus image.
\newblock {\em Machine vision and applications 32}, 1 (2021), 1--13.

\bibitem{ao2021spinnet}
{\sc Ao, S., Hu, Q., Yang, B., Markham, A., and Guo, Y.}
\newblock Spinnet: Learning a general surface descriptor for 3d point cloud registration.
\newblock In {\em Proceedings of the IEEE/CVF conference on computer vision and pattern recognition\/} (2021), pp.~11753--11762.

\bibitem{hybrid-motion}
{\sc Auysakul, J., Xu, H., and Pooneeth, V.}
\newblock A hybrid motion estimation for video stabilization based on an imu sensor.
\newblock {\em Sensors 18}, 8 (2018), 2708.

\bibitem{bae2022multi}
{\sc Bae, G., Budvytis, I., and Cipolla, R.}
\newblock Multi-view depth estimation by fusing single-view depth probability with multi-view geometry.
\newblock In {\em Proceedings of the IEEE/CVF Conference on Computer Vision and Pattern Recognition\/} (2022), pp.~2842--2851.

\bibitem{bai2021pointdsc}
{\sc Bai, X., Luo, Z., Zhou, L., Chen, H., Li, L., Hu, Z., Fu, H., and Tai, C.-L.}
\newblock Pointdsc: Robust point cloud registration using deep spatial consistency.
\newblock In {\em Proceedings of the IEEE/CVF Conference on Computer Vision and Pattern Recognition\/} (2021), pp.~15859--15869.

\bibitem{bai2020d3feat}
{\sc Bai, X., Luo, Z., Zhou, L., Fu, H., Quan, L., and Tai, C.-L.}
\newblock D3feat: Joint learning of dense detection and description of 3d local features.
\newblock In {\em Proceedings of the IEEE/CVF conference on computer vision and pattern recognition\/} (2020), pp.~6359--6367.

\bibitem{shane20allinfocus}
{\sc Barratt, S., and Hannel, B.}
\newblock Extracting the depth and all-in-focus image from a focal stack.
\newblock In {\em Proceedings of the IEEE International Conference on Computer Vision\/} (2015), pp.~3451--3459.

\bibitem{barron2019general}
{\sc Barron, J.~T.}
\newblock A general and adaptive robust loss function.
\newblock In {\em Proceedings of the IEEE/CVF conference on computer vision and pattern recognition\/} (2019), pp.~4331--4339.

\bibitem{barron2021mip}
{\sc Barron, J.~T., Mildenhall, B., Tancik, M., Hedman, P., Martin-Brualla, R., and Srinivasan, P.~P.}
\newblock Mip-nerf: A multiscale representation for anti-aliasing neural radiance fields.
\newblock In {\em Proceedings of the IEEE/CVF international conference on computer vision\/} (2021), pp.~5855--5864.

\bibitem{barron2022mip}
{\sc Barron, J.~T., Mildenhall, B., Verbin, D., Srinivasan, P.~P., and Hedman, P.}
\newblock Mip-nerf 360: Unbounded anti-aliased neural radiance fields.
\newblock In {\em Proceedings of the IEEE/CVF conference on computer vision and pattern recognition\/} (2022), pp.~5470--5479.

\bibitem{sift-video}
{\sc Battiato, S., et~al.}
\newblock Sift features tracking for video stabilization.
\newblock In {\em 14th International Conference on Image Analysis and Processing (ICIAP 2007)\/} (2007), IEEE.

\bibitem{behley2018efficient}
{\sc Behley, J., and Stachniss, C.}
\newblock Efficient surfel-based slam using 3d laser range data in urban environments.
\newblock In {\em Robotics: Science and Systems\/} (2018), vol.~2018, p.~59.

\bibitem{benavides2022phonedepth}
{\sc Benavides, F.~T., Ignatov, A., and Timofte, R.}
\newblock Phonedepth: A dataset for monocular depth estimation on mobile devices.
\newblock In {\em Proceedings of the IEEE/CVF Conference on Computer Vision and Pattern Recognition\/} (2022), pp.~3049--3056.

\bibitem{bevilacqua20223d}
{\sc Bevilacqua, M.~G., Russo, M., Giordano, A., and Spallone, R.}
\newblock 3d reconstruction, digital twinning, and virtual reality: Architectural heritage applications.
\newblock In {\em 2022 IEEE Conference on Virtual Reality and 3D User Interfaces Abstracts and Workshops (VRW)\/} (2022), IEEE, pp.~92--96.

\bibitem{bhatnagar2020loopreg}
{\sc Bhatnagar, B.~L., Sminchisescu, C., Theobalt, C., and Pons-Moll, G.}
\newblock Loopreg: Self-supervised learning of implicit surface correspondences, pose and shape for 3d human mesh registration.
\newblock {\em Advances in Neural Information Processing Systems 33\/} (2020), 12909--12922.

\bibitem{brachmann2018learning}
{\sc Brachmann, E., and Rother, C.}
\newblock Learning less is more-6d camera localization via 3d surface regression.
\newblock In {\em Proceedings of the IEEE conference on computer vision and pattern recognition\/} (2018), pp.~4654--4662.

\bibitem{bronstein2017geometric}
{\sc Bronstein, M.~M., Bruna, J., LeCun, Y., Szlam, A., and Vandergheynst, P.}
\newblock Geometric deep learning: going beyond euclidean data.
\newblock {\em IEEE Signal Processing Magazine 34}, 4 (2017), 18--42.

\bibitem{buchholz2022fourier}
{\sc Buchholz, T.-O., and Jug, F.}
\newblock Fourier image transformer.
\newblock In {\em Proceedings of the IEEE/CVF Conference on Computer Vision and Pattern Recognition\/} (2022), pp.~1846--1854.

\bibitem{ORBSLAM3_TRO}
{\sc Campos, C., Elvira, R., Gomez, J.~J., Montiel, J. M.~M., and Tardos, J.~D.}
\newblock {ORB-SLAM3}: An accurate open-source library for visual, visual-inertial and multi-map {SLAM}.
\newblock {\em IEEE Transactions on Robotics 37}, 6 (2021), 1874--1890.

\bibitem{canny}
{\sc Canny, J., et~al.}
\newblock A computational approach to edge detection.
\newblock {\em IEEE Transactions on Pattern Analysis and Machine Intelligence 6\/} (1986), 679--698.

\bibitem{carrio2018attitude}
{\sc Carrio, A., Bavle, H., and Campoy, P.}
\newblock Attitude estimation using horizon detection in thermal images.
\newblock {\em International Journal of Micro Air Vehicles 10}, 4 (2018), 352--361.

\bibitem{carvalho2018deep}
{\sc Carvalho, M., Le~Saux, B., Trouv{\'e}-Peloux, P., Almansa, A., and Champagnat, F.}
\newblock Deep depth from defocus: how can defocus blur improve 3d estimation using dense neural networks?
\newblock In {\em Proceedings of the European Conference on Computer Vision (ECCV) Workshops\/} (2018), pp.~0--0.

\bibitem{casser2019depth}
{\sc Casser, V., Pirk, S., Mahjourian, R., and Angelova, A.}
\newblock Depth prediction without the sensors: Leveraging structure for unsupervised learning from monocular videos.
\newblock In {\em Proceedings of the AAAI conference on artificial intelligence\/} (2019), vol.~33, pp.~8001--8008.

\bibitem{cesa2022program}
{\sc Cesa, G., Lang, L., and Weiler, M.}
\newblock A program to build e (n)-equivariant steerable cnns.
\newblock In {\em International conference on learning representations\/} (2022).

\bibitem{chabra2020deep}
{\sc Chabra, R., Lenssen, J.~E., Ilg, E., Schmidt, T., Straub, J., Lovegrove, S., and Newcombe, R.}
\newblock Deep local shapes: Learning local sdf priors for detailed 3d reconstruction.
\newblock In {\em Computer Vision--ECCV 2020: 16th European Conference, Glasgow, UK, August 23--28, 2020, Proceedings, Part XXIX 16\/} (2020), Springer, pp.~608--625.

\bibitem{chang2000adaptive}
{\sc Chang, S.~G., Yu, B., and Vetterli, M.}
\newblock Adaptive wavelet thresholding for image denoising and compression.
\newblock {\em IEEE transactions on image processing 9}, 9 (2000), 1532--1546.

\bibitem{chen2024pgsr}
{\sc Chen, D., Li, H., Ye, W., Wang, Y., Xie, W., Zhai, S., Wang, N., Liu, H., Bao, H., and Zhang, G.}
\newblock Pgsr: Planar-based gaussian splatting for efficient and high-fidelity surface reconstruction.
\newblock {\em arXiv preprint arXiv:2406.06521\/} (2024).

\bibitem{chen2021equivariant}
{\sc Chen, H., Liu, S., Chen, W., Li, H., and Hill, R.}
\newblock Equivariant point network for 3d point cloud analysis.
\newblock In {\em Proceedings of the IEEE/CVF conference on computer vision and pattern recognition\/} (2021), pp.~14514--14523.

\bibitem{chen2024map}
{\sc Chen, S., Cavallari, T., Prisacariu, V.~A., and Brachmann, E.}
\newblock Map-relative pose regression for visual re-localization.
\newblock In {\em Proceedings of the IEEE/CVF Conference on Computer Vision and Pattern Recognition\/} (2024), pp.~20665--20674.

\bibitem{chen2018learning}
{\sc Chen, T., Lin, L., Zuo, W., Luo, X., and Zhang, L.}
\newblock Learning a wavelet-like auto-encoder to accelerate deep neural networks.
\newblock In {\em Proceedings of the AAAI Conference on Artificial Intelligence\/} (2018), vol.~32.

\bibitem{cheng2023sdfusion}
{\sc Cheng, Y.-C., Lee, H.-Y., Tulyakov, S., Schwing, A.~G., and Gui, L.-Y.}
\newblock Sdfusion: Multimodal 3d shape completion, reconstruction, and generation.
\newblock In {\em Proceedings of the IEEE/CVF Conference on Computer Vision and Pattern Recognition\/} (2023), pp.~4456--4465.

\bibitem{chibane20ifnet}
{\sc Chibane, J., Alldieck, T., and Pons-Moll, G.}
\newblock Implicit functions in feature space for 3d shape reconstruction and completion.
\newblock In {\em {IEEE} Conference on Computer Vision and Pattern Recognition (CVPR)\/} (jun 2020), {IEEE}.

\bibitem{self-supervised}
{\sc Choi, J., Park, J., S, I., and Kweon}.
\newblock Self-supervised real-time video stabilization, 2021.
\newblock arXiv preprint arXiv:2111.05980.

\bibitem{chou2023diffusion}
{\sc Chou, G., Bahat, Y., and Heide, F.}
\newblock Diffusion-sdf: Conditional generative modeling of signed distance functions.
\newblock In {\em Proceedings of the IEEE/CVF international conference on computer vision\/} (2023), pp.~2262--2272.

\bibitem{choy2020deep}
{\sc Choy, C., Dong, W., and Koltun, V.}
\newblock Deep global registration.
\newblock In {\em CVPR\/} (2020).

\bibitem{choy2019fully}
{\sc Choy, C., Park, J., and Koltun, V.}
\newblock Fully convolutional geometric features.
\newblock In {\em Proceedings of the IEEE/CVF international conference on computer vision\/} (2019), pp.~8958--8966.

\bibitem{choy20163d}
{\sc Choy, C.~B., Xu, D., Gwak, J., Chen, K., and Savarese, S.}
\newblock 3d-r2n2: A unified approach for single and multi-view 3d object reconstruction.
\newblock In {\em Proceedings of the European Conference on Computer Vision ({ECCV})\/} (2016).

\bibitem{citraro2020real}
{\sc Citraro, L., M{\'a}rquez-Neila, P., Savare, S., Jayaram, V., Dubout, C., Renaut, F., Hasfura, A., Ben~Shitrit, H., and Fua, P.}
\newblock Real-time camera pose estimation for sports fields.
\newblock {\em Machine Vision and Applications 31}, 3 (2020), 16.

\bibitem{cohen2018spherical}
{\sc Cohen, T.~S., Geiger, M., K{\"o}hler, J., and Welling, M.}
\newblock Spherical cnns.
\newblock {\em arXiv preprint arXiv:1801.10130\/} (2018).

\bibitem{surfel_geometry}
{\sc Dahl, V., Aanæs, H., and Bærentzen, J.}
\newblock Surfel based geometry reconstruction.
\newblock pp.~39--44.

\bibitem{deng2021vector}
{\sc Deng, C., Litany, O., Duan, Y., Poulenard, A., Tagliasacchi, A., and Guibas, L.~J.}
\newblock Vector neurons: A general framework for so (3)-equivariant networks.
\newblock In {\em Proceedings of the IEEE/CVF International Conference on Computer Vision\/} (2021), pp.~12200--12209.

\bibitem{dosovitskiy2020image}
{\sc Dosovitskiy, A., Beyer, L., Kolesnikov, A., Weissenborn, D., Zhai, X., Unterthiner, T., Dehghani, M., Minderer, M., Heigold, G., Gelly, S., et~al.}
\newblock An image is worth 16x16 words: Transformers for image recognition at scale.
\newblock {\em arXiv preprint arXiv:2010.11929\/} (2020).

\bibitem{du2020dh3d}
{\sc Du, J., Wang, R., and Cremers, D.}
\newblock Dh3d: Deep hierarchical 3d descriptors for robust large-scale 6dof relocalization.
\newblock In {\em Computer Vision--ECCV 2020: 16th European Conference, Glasgow, UK, August 23--28, 2020, Proceedings, Part IV 16\/} (2020), Springer, pp.~744--762.

\bibitem{du2022se}
{\sc Du, W., Zhang, H., Du, Y., Meng, Q., Chen, W., Zheng, N., Shao, B., and Liu, T.-Y.}
\newblock Se(3) equivariant graph neural networks with complete local frames.
\newblock In {\em International Conference on Machine Learning\/} (2022), PMLR, pp.~5583--5608.

\bibitem{fixed-wing}
{\sc Dusha, D., Boles, W., and Walker, R.}
\newblock Fixed-wing attitude estimation using computer vision based horizon detection.
\newblock In {\em Proceedings of AIAC12: 2nd Australasian Unmanned Air Vehicles Conference\/} (2007), Waldron Smith Management.

\bibitem{eigen2014depth}
{\sc Eigen, D., Puhrsch, C., and Fergus, R.}
\newblock Depth map prediction from a single image using a multi-scale deep network.
\newblock {\em Advances in neural information processing systems 27\/} (2014).

\bibitem{fan2017point}
{\sc Fan, H., Su, H., and Guibas, L.~J.}
\newblock A point set generation network for 3d object reconstruction from a single image.
\newblock In {\em Proceedings of the IEEE conference on computer vision and pattern recognition\/} (2017), pp.~605--613.

\bibitem{feng2019unsupervised}
{\sc Feng, Y., Wu, S., K{\"o}p{\"u}kl{\"u}, O., Kang, X., and Tombari, F.}
\newblock Unsupervised monocular depth prediction for indoor continuous video streams.
\newblock {\em arXiv preprint arXiv:1911.08995\/} (2019).

\bibitem{figueiredo2003algorithm}
{\sc Figueiredo, M.~A., and Nowak, R.~D.}
\newblock An em algorithm for wavelet-based image restoration.
\newblock {\em IEEE Transactions on Image Processing 12}, 8 (2003), 906--916.

\bibitem{finzi2020generalizing}
{\sc Finzi, M., Stanton, S., Izmailov, P., and Wilson, A.~G.}
\newblock Generalizing convolutional neural networks for equivariance to lie groups on arbitrary continuous data.
\newblock In {\em International Conference on Machine Learning\/} (2020), PMLR, pp.~3165--3176.

\bibitem{fischler1981random}
{\sc Fischler, M.~A., and Bolles, R.~C.}
\newblock Random sample consensus: a paradigm for model fitting with applications to image analysis and automated cartography.
\newblock {\em Communications of the ACM 24}, 6 (1981), 381--395.

\bibitem{fu2024geowizard}
{\sc Fu, X., Yin, W., Hu, M., Wang, K., Ma, Y., Tan, P., Shen, S., Lin, D., and Long, X.}
\newblock Geowizard: Unleashing the diffusion priors for 3d geometry estimation from a single image.
\newblock In {\em ECCV\/} (2024).

\bibitem{fuchs2020se}
{\sc Fuchs, F., Worrall, D., Fischer, V., and Welling, M.}
\newblock Se (3)-transformers: 3d roto-translation equivariant attention networks.
\newblock {\em Advances in neural information processing systems 33\/} (2020), 1970--1981.

\bibitem{fuchs2021iterative}
{\sc Fuchs, F.~B., Wagstaff, E., Dauparas, J., and Posner, I.}
\newblock Iterative se (3)-transformers.
\newblock In {\em Geometric Science of Information: 5th International Conference, GSI 2021, Paris, France, July 21--23, 2021, Proceedings 5\/} (2021), Springer, pp.~585--595.

\bibitem{fuentes2015visual}
{\sc Fuentes-Pacheco, J., Ruiz-Ascencio, J., and Rend{\'o}n-Mancha, J.~M.}
\newblock Visual simultaneous localization and mapping: a survey.
\newblock {\em Artificial intelligence review 43\/} (2015), 55--81.

\bibitem{fujieda2018wavelet}
{\sc Fujieda, S., Takayama, K., and Hachisuka, T.}
\newblock Wavelet convolutional neural networks.
\newblock {\em arXiv preprint arXiv:1805.08620\/} (2018).

\bibitem{fujimura2023deep}
{\sc Fujimura, Y., Iiyama, M., Funatomi, T., and Mukaigawa, Y.}
\newblock Deep depth from focal stack with defocus model for camera-setting invariance.
\newblock {\em International Journal of Computer Vision\/} (2023), 1--16.

\bibitem{furgale2013unified}
{\sc Furgale, P., Rehder, J., and Siegwart, R.}
\newblock Unified temporal and spatial calibration for multi-sensor systems.
\newblock In {\em 2013 IEEE/RSJ International Conference on Intelligent Robots and Systems\/} (2013), IEEE, pp.~1280--1286.

\bibitem{garg2016unsupervised}
{\sc Garg, R., Bg, V.~K., Carneiro, G., and Reid, I.}
\newblock Unsupervised cnn for single view depth estimation: Geometry to the rescue.
\newblock In {\em Computer Vision--ECCV 2016: 14th European Conference, Amsterdam, The Netherlands, October 11-14, 2016, Proceedings, Part VIII 14\/} (2016), Springer, pp.~740--756.

\bibitem{geiger2012we}
{\sc Geiger, A., Lenz, P., and Urtasun, R.}
\newblock Are we ready for autonomous driving? the kitti vision benchmark suite.
\newblock In {\em 2012 IEEE conference on computer vision and pattern recognition\/} (2012), IEEE, pp.~3354--3361.

\bibitem{glocker2013real}
{\sc Glocker, B., Izadi, S., Shotton, J., and Criminisi, A.}
\newblock Real-time rgb-d camera relocalization.
\newblock In {\em 2013 IEEE International Symposium on Mixed and Augmented Reality (ISMAR)\/} (2013), IEEE, pp.~173--179.

\bibitem{godard2017unsupervised}
{\sc Godard, C., Mac~Aodha, O., and Brostow, G.~J.}
\newblock Unsupervised monocular depth estimation with left-right consistency.
\newblock In {\em Proceedings of the IEEE conference on computer vision and pattern recognition\/} (2017), pp.~270--279.

\bibitem{godard2019digging}
{\sc Godard, C., Mac~Aodha, O., Firman, M., and Brostow, G.~J.}
\newblock Digging into self-supervised monocular depth estimation.
\newblock In {\em Proceedings of the IEEE/CVF International Conference on Computer Vision\/} (2019), pp.~3828--3838.

\bibitem{monodepth2}
{\sc Godard, C., {Mac Aodha}, O., Firman, M., and Brostow, G.~J.}
\newblock Digging into self-supervised monocular depth prediction.

\bibitem{gojcic2020learning}
{\sc Gojcic, Z., Zhou, C., Wegner, J.~D., Guibas, L.~J., and Birdal, T.}
\newblock Learning multiview 3d point cloud registration.
\newblock In {\em Proceedings of the IEEE/CVF conference on computer vision and pattern recognition\/} (2020), pp.~1759--1769.

\bibitem{gordon2019depth}
{\sc Gordon, A., Li, H., Jonschkowski, R., and Angelova, A.}
\newblock Depth from videos in the wild: Unsupervised monocular depth learning from unknown cameras.
\newblock In {\em Proceedings of the IEEE/CVF International Conference on Computer Vision\/} (2019), pp.~8977--8986.

\bibitem{guedon2024sugar}
{\sc Gu{\'e}don, A., and Lepetit, V.}
\newblock Sugar: Surface-aligned gaussian splatting for efficient 3d mesh reconstruction and high-quality mesh rendering.
\newblock In {\em Proceedings of the IEEE/CVF Conference on Computer Vision and Pattern Recognition\/} (2024), pp.~5354--5363.

\bibitem{guo2017deep}
{\sc Guo, T., Seyed~Mousavi, H., Huu~Vu, T., and Monga, V.}
\newblock Deep wavelet prediction for image super-resolution.
\newblock In {\em Proceedings of the IEEE conference on computer vision and pattern recognition workshops\/} (2017), pp.~104--113.

\bibitem{guo2018learning}
{\sc Guo, X., Li, H., Yi, S., Ren, J., and Wang, X.}
\newblock Learning monocular depth by distilling cross-domain stereo networks.
\newblock In {\em Proceedings of the European Conference on Computer Vision (ECCV)\/} (2018), pp.~484--500.

\bibitem{gur2019single}
{\sc Gur, S., and Wolf, L.}
\newblock Single image depth estimation trained via depth from defocus cues.
\newblock In {\em Proceedings of the IEEE/CVF Conference on Computer Vision and Pattern Recognition\/} (2019), pp.~7683--7692.

\bibitem{pf-local}
{\sc Gustafsson, F., et~al.}
\newblock Particle filter theory and practice with positioning applications.
\newblock {\em IEEE Aerospace and Electronic Systems Magazine 25}, 7 (2010), 53--82.

\bibitem{hasson2020leveraging}
{\sc Hasson, Y., Tekin, B., Bogo, F., Laptev, I., Pollefeys, M., and Schmid, C.}
\newblock Leveraging photometric consistency over time for sparsely supervised hand-object reconstruction.
\newblock In {\em Proceedings of the IEEE/CVF conference on computer vision and pattern recognition\/} (2020), pp.~571--580.

\bibitem{hazirbas2018deep}
{\sc Hazirbas, C., Soyer, S.~G., Staab, M.~C., Leal-Taix{\'e}, L., and Cremers, D.}
\newblock Deep depth from focus.
\newblock In {\em Asian conference on computer vision\/} (2018), Springer, pp.~525--541.

\bibitem{he2024lotus}
{\sc He, J., Li, H., Yin, W., Liang, Y., Li, L., Zhou, K., Liu, H., Liu, B., and Chen, Y.-C.}
\newblock Lotus: Diffusion-based visual foundation model for high-quality dense prediction.
\newblock {\em arXiv preprint arXiv:2409.18124\/} (2024).

\bibitem{he2022multi}
{\sc He, R., Hong, H., Fu, B., and Liu, F.}
\newblock Multi-task learning for monocular depth and defocus estimations with real images.
\newblock {\em arXiv preprint arXiv:2208.09848\/} (2022).

\bibitem{he2023fourier}
{\sc He, Z., Yang, M., Feng, M., Yin, J., Wang, X., Leng, J., and Lin, Z.}
\newblock Fourier transformer: Fast long range modeling by removing sequence redundancy with fft operator.
\newblock {\em arXiv preprint arXiv:2305.15099\/} (2023).

\bibitem{hess2016real}
{\sc Hess, W., Kohler, D., Rapp, H., and Andor, D.}
\newblock Real-time loop closure in 2d lidar slam.
\newblock In {\em 2016 IEEE international conference on robotics and automation (ICRA)\/} (2016), IEEE, pp.~1271--1278.

\bibitem{hochreiter1997long}
{\sc Hochreiter, S., and Schmidhuber, J.}
\newblock Long short-term memory.
\newblock {\em Neural computation 9}, 8 (1997), 1735--1780.

\bibitem{honauer2016dataset}
{\sc Honauer, K., Johannsen, O., Kondermann, D., and Goldluecke, B.}
\newblock A dataset and evaluation methodology for depth estimation on 4d light fields.
\newblock In {\em Asian conference on computer vision\/} (2016), Springer, pp.~19--34.

\bibitem{hornauer2022gradient}
{\sc Hornauer, J., and Belagiannis, V.}
\newblock Gradient-based uncertainty for monocular depth estimation.
\newblock In {\em European Conference on Computer Vision\/} (2022), Springer, pp.~613--630.

\bibitem{hu2024neural}
{\sc Hu, J., Hui, K.-H., Liu, Z., Li, R., and Fu, C.-W.}
\newblock Neural wavelet-domain diffusion for 3d shape generation, inversion, and manipulation.
\newblock {\em ACM Transactions on Graphics 43}, 2 (2024), 1--18.

\bibitem{hu2019revisiting}
{\sc Hu, J., Ozay, M., Zhang, Y., and Okatani, T.}
\newblock Revisiting single image depth estimation: Toward higher resolution maps with accurate object boundaries.
\newblock In {\em 2019 IEEE Winter Conference on Applications of Computer Vision (WACV)\/} (2019), IEEE, pp.~1043--1051.

\bibitem{hu2022refractive}
{\sc Hu, Y., Rao, W., Qi, L., Dong, J., Cai, J., and Fan, H.}
\newblock A refractive stereo structured-light 3-d measurement system for immersed object.
\newblock {\em IEEE Transactions on Instrumentation and Measurement 72\/} (2022), 1--13.

\bibitem{huai2015real}
{\sc Huai, J., Zhang, Y., and Yilmaz, A.}
\newblock Real-time large scale 3d reconstruction by fusing kinect and imu data.
\newblock {\em ISPRS Annals of the Photogrammetry, Remote Sensing and Spatial Information Sciences 2\/} (2015), 491--496.

\bibitem{huai2022robocentric}
{\sc Huai, Z., and Huang, G.}
\newblock Robocentric visual-inertial odometry.
\newblock {\em The International Journal of Robotics Research 41}, 7 (2022), 667--689.

\bibitem{huang20242d}
{\sc Huang, B., Yu, Z., Chen, A., Geiger, A., and Gao, S.}
\newblock 2d gaussian splatting for geometrically accurate radiance fields.
\newblock In {\em ACM SIGGRAPH 2024 conference papers\/} (2024), pp.~1--11.

\bibitem{Huang2DGS2024}
{\sc Huang, B., Yu, Z., Chen, A., Geiger, A., and Gao, S.}
\newblock 2d gaussian splatting for geometrically accurate radiance fields.
\newblock In {\em SIGGRAPH 2024 Conference Papers\/} (2024), Association for Computing Machinery.

\bibitem{huang20242d`}
{\sc Huang, B., Yu, Z., Chen, A., Geiger, A., and Gao, S.}
\newblock 2d gaussian splatting for geometrically accurate radiance fields.
\newblock In {\em ACM SIGGRAPH 2024 conference papers\/} (2024), pp.~1--11.

\bibitem{huang2017wavelet}
{\sc Huang, H., He, R., Sun, Z., and Tan, T.}
\newblock Wavelet-srnet: A wavelet-based cnn for multi-scale face super resolution.
\newblock In {\em Proceedings of the IEEE international conference on computer vision\/} (2017), pp.~1689--1697.

\bibitem{huang2024wavedm}
{\sc Huang, Y., Huang, J., Liu, J., Yan, M., Dong, Y., Lyu, J., Chen, C., and Chen, S.}
\newblock Wavedm: Wavelet-based diffusion models for image restoration.
\newblock {\em IEEE Transactions on Multimedia\/} (2024).

\bibitem{huang2020trans}
{\sc Huang, Z., Xu, P., Liang, D., Mishra, A., and Xiang, B.}
\newblock Trans-blstm: Transformer with bidirectional lstm for language understanding.
\newblock {\em arXiv preprint arXiv:2003.07000\/} (2020).

\bibitem{hui2022neural}
{\sc Hui, K.-H., Li, R., Hu, J., and Fu, C.-W.}
\newblock Neural wavelet-domain diffusion for 3d shape generation.
\newblock In {\em SIGGRAPH Asia 2022 Conference Papers\/} (2022), pp.~1--9.

\bibitem{hutchins2022block}
{\sc Hutchins, D., Schlag, I., Wu, Y., Dyer, E., and Neyshabur, B.}
\newblock Block-recurrent transformers.
\newblock {\em arXiv preprint arXiv:2203.07852\/} (2022).

\bibitem{hutchinson2020lietransformer}
{\sc Hutchinson, M., Lan, C.~L., Zaidi, S., Dupont, E., Teh, Y.~W., and Kim, H.}
\newblock Lietransformer: Equivariant self-attention for lie groups, 2020.

\bibitem{izadi2011kinectfusion}
{\sc Izadi, S., Kim, D., Hilliges, O., Molyneaux, D., Newcombe, R., Kohli, P., Shotton, J., Hodges, S., Freeman, D., Davison, A., et~al.}
\newblock Kinectfusion: real-time 3d reconstruction and interaction using a moving depth camera.
\newblock In {\em Proceedings of the 24th annual ACM symposium on User interface software and technology\/} (2011), pp.~559--568.

\bibitem{jenner2022steerable}
{\sc Jenner, E., and Weiler, M.}
\newblock Steerable partial differential operators for equivariant neural networks.
\newblock In {\em International Conference on Learning Representations\/} (2022).

\bibitem{jensen2014large}
{\sc Jensen, R., Dahl, A., Vogiatzis, G., Tola, E., and Aan{\ae}s, H.}
\newblock Large scale multi-view stereopsis evaluation.
\newblock In {\em Proceedings of the IEEE conference on computer vision and pattern recognition\/} (2014), pp.~406--413.

\bibitem{jiang2024gaussianshader}
{\sc Jiang, Y., Tu, J., Liu, Y., Gao, X., Long, X., Wang, W., and Ma, Y.}
\newblock Gaussianshader: 3d gaussian splatting with shading functions for reflective surfaces.
\newblock In {\em Proceedings of the IEEE/CVF Conference on Computer Vision and Pattern Recognition\/} (2024), pp.~5322--5332.

\bibitem{johannsen2017taxonomy}
{\sc Johannsen, O., Honauer, K., Goldluecke, B., Alperovich, A., Battisti, F., Bok, Y., Brizzi, M., Carli, M., Choe, G., Diebold, M., et~al.}
\newblock A taxonomy and evaluation of dense light field depth estimation algorithms.
\newblock In {\em Proceedings of the IEEE Conference on Computer Vision and Pattern Recognition Workshops\/} (2017), pp.~82--99.

\bibitem{kang2024focdepthformer}
{\sc Kang, X., Han, F., Fayjie, A.~R., Vandewalle, P., Khoshelham, K., and Gong, D.}
\newblock Focdepthformer: Transformer with latent lstm for depth estimation from focal stack.
\newblock In {\em Australasian Joint Conference on Artificial Intelligence\/} (2024), Springer, pp.~273--290.

\bibitem{kang2023adaptive}
{\sc Kang, X., Herrera, A., Lema, H., Valencia, E., and Vandewalle, P.}
\newblock Adaptive sampling-based particle filter for visual-inertial gimbal in the wild.
\newblock In {\em 2023 IEEE International Conference on Robotics and Automation (ICRA)\/} (2023), IEEE, pp.~2738--2744.

\bibitem{kang2024equi}
{\sc Kang, X., Luan, Z., Khoshelham, K., and Wang, B.}
\newblock Equi-gspr: Equivariant se (3) graph network model for sparse point cloud registration.
\newblock In {\em European Conference on Computer Vision\/} (2024), Springer, pp.~149--167.

\bibitem{kang2025multi}
{\sc Kang, X., Xiang, Z., Zhang, Z., and Khoshelham, K.}
\newblock Multi-view geometry-aware diffusion transformer for indoor novel view synthesis.
\newblock In {\em ICLR 2025 Workshop on Deep Generative Model in Machine Learning: Theory, Principle and Efficacy\/} (2025).

\bibitem{kang20183d}
{\sc Kang, X., Yin, S., and Fen, Y.}
\newblock 3d reconstruction \& assessment framework based on affordable 2d lidar.
\newblock In {\em 2018 IEEE/ASME International Conference on Advanced Intelligent Mechatronics (AIM)\/} (2018), IEEE, pp.~292--297.

\bibitem{kang2019robust}
{\sc Kang, X., and Yuan, S.}
\newblock Robust data association for object-level semantic slam.
\newblock {\em arXiv preprint arXiv:1909.13493\/} (2019).

\bibitem{kang2023integrated}
{\sc Kang, X., and Yuan, S.}
\newblock Integrated visual-inertial odometry and image stabilization for image processing.
\newblock {\em Google Patents, US Patent App 18}, 035,479 (2023).

\bibitem{Patent1}
{\sc Kang, X., and Yuan, S.}
\newblock Integrated visual-inertial odometry and image stabilization for image processing, December 28 2023.

\bibitem{kang20252d}
{\sc Kang, X., Zhao, H., Khoshelham, K., and Patrick, V.}
\newblock 2d surfel-based 3d point cloud registration with robust equivariant se (3) features.
\newblock In {\em The conference proceedings and published in IEEE Xplore of 2025 IEEE International Geoscience and Remote Sensing Symposium\/} (2025).

\bibitem{ke2024repurposing}
{\sc Ke, B., Obukhov, A., Huang, S., Metzger, N., Daudt, R.~C., and Schindler, K.}
\newblock Repurposing diffusion-based image generators for monocular depth estimation.
\newblock In {\em Proceedings of the IEEE/CVF Conference on Computer Vision and Pattern Recognition\/} (2024), pp.~9492--9502.

\bibitem{kendall2015posenet}
{\sc Kendall, A., Grimes, M., and Cipolla, R.}
\newblock Posenet: A convolutional network for real-time 6-dof camera relocalization.
\newblock In {\em Proceedings of the IEEE international conference on computer vision\/} (2015), pp.~2938--2946.

\bibitem{kerbl3Dgaussians}
{\sc Kerbl, B., Kopanas, G., Leimk{\"u}hler, T., and Drettakis, G.}
\newblock 3d gaussian splatting for real-time radiance field rendering.
\newblock {\em ACM Transactions on Graphics 42}, 4 (July 2023).

\bibitem{kerbl20233d}
{\sc Kerbl, B., Kopanas, G., Leimk{\"u}hler, T., and Drettakis, G.}
\newblock 3d gaussian splatting for real-time radiance field rendering.
\newblock {\em ACM Transactions on Graphics 42}, 4 (2023), 1--14.

\bibitem{khamis2018stereonet}
{\sc Khamis, S., Fanello, S., Rhemann, C., Kowdle, A., Valentin, J., and Izadi, S.}
\newblock Stereonet: Guided hierarchical refinement for real-time edge-aware depth prediction.
\newblock In {\em Proceedings of the European conference on computer vision (ECCV)\/} (2018), pp.~573--590.

\bibitem{khatib2024trinerflet}
{\sc Khatib, R., and Giryes, R.}
\newblock Trinerflet: A wavelet based multiscale triplane nerf representation.
\newblock {\em arXiv preprint arXiv:2401.06191\/} (2024).

\bibitem{kipponen2020defining}
{\sc Kipponen, S.}
\newblock Defining the market potential of industry specific drone software: Case: Metropolia innovation hub of smart mobility.

\bibitem{kirillov2023segany}
{\sc Kirillov, A., Mintun, E., Ravi, N., Mao, H., Rolland, C., Gustafson, L., Xiao, T., Whitehead, S., Berg, A.~C., Lo, W.-Y., Doll{\'a}r, P., and Girshick, R.}
\newblock Segment anything.
\newblock {\em arXiv:2304.02643\/} (2023).

\bibitem{knapitsch2017tanks}
{\sc Knapitsch, A., Park, J., Zhou, Q.-Y., and Koltun, V.}
\newblock Tanks and temples: Benchmarking large-scale scene reconstruction.
\newblock {\em ACM Transactions on Graphics (ToG) 36}, 4 (2017), 1--13.

\bibitem{kong2023vmap}
{\sc Kong, X., Liu, S., Taher, M., and Davison, A.~J.}
\newblock vmap: Vectorised object mapping for neural field slam.
\newblock In {\em Proceedings of the IEEE/CVF Conference on Computer Vision and Pattern Recognition\/} (2023), pp.~952--961.

\bibitem{dataset}
{\sc La~Place, C., Urooj, A., and Borji, A.}
\newblock Segmenting sky pixels in images: Analysis and comparison.
\newblock In {\em 2019 IEEE Winter Conference on Applications of Computer Vision (WACV)\/} (2019), IEEE.

\bibitem{laina2016deeper}
{\sc Laina, I., Rupprecht, C., Belagiannis, V., Tombari, F., and Navab, N.}
\newblock Deeper depth prediction with fully convolutional residual networks.
\newblock In {\em 2016 Fourth international conference on 3D vision (3DV)\/} (2016), IEEE, pp.~239--248.

\bibitem{lavin2016fast}
{\sc Lavin, A., and Gray, S.}
\newblock Fast algorithms for convolutional neural networks.
\newblock In {\em Proceedings of the IEEE conference on computer vision and pattern recognition\/} (2016), pp.~4013--4021.

\bibitem{stabilization-depth}
{\sc Lee, Y.-C., et~al.}
\newblock 3d video stabilization with depth estimation by cnn-based optimization.
\newblock In {\em Proceedings of the IEEE/CVF Conference on Computer Vision and Pattern Recognition\/} (2021).

\bibitem{leifman2013pattern}
{\sc Leifman, G., and Tal, A.}
\newblock Pattern-driven colorization of 3d surfaces.
\newblock In {\em Proceedings of the IEEE Conference on Computer Vision and Pattern Recognition\/} (2013), pp.~241--248.

\bibitem{deep-IMU-stabilization}
{\sc Li, C., et~al.}
\newblock Deep online video stabilization using imu sensors.
\newblock {\em IEEE Transactions on Multimedia\/} (2022).

\bibitem{li2024learning}
{\sc Li, J. C.~L., Liu, C., Huang, B., and Wong, N.}
\newblock Learning spatially collaged fourier bases for implicit neural representation.
\newblock In {\em Proceedings of the AAAI Conference on Artificial Intelligence\/} (2024), vol.~38, pp.~13492--13499.

\bibitem{li2023diffusion}
{\sc Li, M., Duan, Y., Zhou, J., and Lu, J.}
\newblock Diffusion-sdf: Text-to-shape via voxelized diffusion.
\newblock In {\em Proceedings of the IEEE/CVF conference on computer vision and pattern recognition\/} (2023), pp.~12642--12651.

\bibitem{li2025geogaussian}
{\sc Li, Y., Lyu, C., Di, Y., Zhai, G., Lee, G.~H., and Tombari, F.}
\newblock Geogaussian: Geometry-aware gaussian splatting for scene rendering.
\newblock In {\em European Conference on Computer Vision\/} (2025), Springer, pp.~441--457.

\bibitem{li2023neuralangelo}
{\sc Li, Z., M{\"u}ller, T., Evans, A., Taylor, R.~H., Unberath, M., Liu, M.-Y., and Lin, C.-H.}
\newblock Neuralangelo: High-fidelity neural surface reconstruction.
\newblock In {\em Proceedings of the IEEE/CVF Conference on Computer Vision and Pattern Recognition\/} (2023), pp.~8456--8465.

\bibitem{li2020through}
{\sc Li, Z., Yeh, Y.-Y., and Chandraker, M.}
\newblock Through the looking glass: Neural 3d reconstruction of transparent shapes.
\newblock In {\em Proceedings of the IEEE/CVF Conference on Computer Vision and Pattern Recognition\/} (2020), pp.~1262--1271.

\bibitem{lin2024se3et}
{\sc Lin, C.~E., Zhu, M., and Ghaffari, M.}
\newblock Se3et: Se (3)-equivariant transformer for low-overlap point cloud registration.
\newblock {\em IEEE Robotics and Automation Letters\/} (2024).

\bibitem{lin2015depth}
{\sc Lin, H., Chen, C., Kang, S.~B., and Yu, J.}
\newblock Depth recovery from light field using focal stack symmetry.
\newblock In {\em Proceedings of the IEEE International Conference on Computer Vision\/} (2015), pp.~3451--3459.

\bibitem{lindeberg2012scale}
{\sc Lindeberg, T.}
\newblock Scale invariant feature transform.

\bibitem{sift}
{\sc Lindeberg, T., et~al.}
\newblock Scale invariant feature transform.
\newblock In {\em Proceedings of the International Conference on Computer Vision\/} (2012), p.~10491.

\bibitem{lindell2022bacon}
{\sc Lindell, D.~B., Van~Veen, D., Park, J.~J., and Wetzstein, G.}
\newblock Bacon: Band-limited coordinate networks for multiscale scene representation.
\newblock In {\em Proceedings of the IEEE/CVF conference on computer vision and pattern recognition\/} (2022), pp.~16252--16262.

\bibitem{liu2017light}
{\sc Liu, C., Qiu, J., and Jiang, M.}
\newblock Light field reconstruction from focal stack based on landweber iterative scheme.
\newblock In {\em Mathematics in Imaging\/} (2017), Optica Publishing Group, pp.~MM2C--3.

\bibitem{liu2024mirrorgaussian}
{\sc Liu, J., Tang, X., Cheng, F., Yang, R., Li, Z., Liu, J., Huang, Y., Lin, J., Liu, S., Wu, X., et~al.}
\newblock Mirrorgaussian: Reflecting 3d gaussians for reconstructing mirror reflections.
\newblock In {\em European Conference on Computer Vision\/} (2024), Springer, pp.~377--393.

\bibitem{liu2023se}
{\sc Liu, R., Lauze, F., Bekkers, E., Erleben, K., and Darkner, S.}
\newblock Se (3) group convolutional neural networks and a study on group convolutions and equivariance for dwi segmentation.

\bibitem{liu2022neural}
{\sc Liu, Y., Peng, S., Liu, L., Wang, Q., Wang, P., Theobalt, C., Zhou, X., and Wang, W.}
\newblock Neural rays for occlusion-aware image-based rendering.
\newblock In {\em Proceedings of the IEEE/CVF Conference on Computer Vision and Pattern Recognition\/} (2022), pp.~7824--7833.

\bibitem{hybrid-neural-fusion}
{\sc Liu, Y.-L., et~al.}
\newblock Hybrid neural fusion for full-frame video stabilization.
\newblock In {\em Proceedings of the IEEE/CVF International Conference on Computer Vision\/} (2021).

\bibitem{liu2021swin}
{\sc Liu, Z., Lin, Y., Cao, Y., Hu, H., Wei, Y., Zhang, Z., Lin, S., and Guo, B.}
\newblock Swin transformer: Hierarchical vision transformer using shifted windows.
\newblock In {\em Proceedings of the IEEE/CVF International Conference on Computer Vision\/} (2021), pp.~10012--10022.

\bibitem{liu2024finer}
{\sc Liu, Z., Zhu, H., Zhang, Q., Fu, J., Deng, W., Ma, Z., Guo, Y., and Cao, X.}
\newblock Finer: Flexible spectral-bias tuning in implicit neural representation by variable-periodic activation functions.
\newblock In {\em Proceedings of the IEEE/CVF Conference on Computer Vision and Pattern Recognition\/} (2024), pp.~2713--2722.

\bibitem{lorensen1998marching}
{\sc Lorensen, W.~E., and Cline, H.~E.}
\newblock Marching cubes: A high resolution 3d surface construction algorithm.
\newblock In {\em Seminal graphics: pioneering efforts that shaped the field}. 1998, pp.~347--353.

\bibitem{low2004linear}
{\sc Low, K.-L.}
\newblock Linear least-squares optimization for point-to-plane icp surface registration.
\newblock {\em Chapel Hill, University of North Carolina 4}, 10 (2004), 1--3.

\bibitem{lowe2004sift}
{\sc Lowe, G.}
\newblock Sift-the scale invariant feature transform.
\newblock {\em Int. J 2}, 91-110 (2004), 2.

\bibitem{ma2018sparse}
{\sc Ma, F., and Karaman, S.}
\newblock Sparse-to-dense: Depth prediction from sparse depth samples and a single image.
\newblock In {\em 2018 IEEE international conference on robotics and automation (ICRA)\/} (2018), IEEE, pp.~4796--4803.

\bibitem{ma2020end}
{\sc Ma, H., Liu, D., Yan, N., Li, H., and Wu, F.}
\newblock End-to-end optimized versatile image compression with wavelet-like transform.
\newblock {\em IEEE Transactions on Pattern Analysis and Machine Intelligence 44}, 3 (2020), 1247--1263.

\bibitem{madgwick}
{\sc Madgwick, S.}
\newblock An efficient orientation filter for inertial and inertial/magnetic sensor arrays.
\newblock Tech. Rep.~25, x-io and University of Bristol (UK), 2010.

\bibitem{mahendran20173d}
{\sc Mahendran, S., Ali, H., and Vidal, R.}
\newblock 3d pose regression using convolutional neural networks.
\newblock In {\em Proceedings of the IEEE international conference on computer vision workshops\/} (2017), pp.~2174--2182.

\bibitem{mallat1989theory}
{\sc Mallat, S.~G.}
\newblock A theory for multiresolution signal decomposition: the wavelet representation.
\newblock {\em IEEE transactions on pattern analysis and machine intelligence 11}, 7 (1989), 674--693.

\bibitem{mao2021tfpose}
{\sc Mao, W., Ge, Y., Shen, C., Tian, Z., Wang, X., and Wang, Z.}
\newblock Tfpose: Direct human pose estimation with transformers.
\newblock {\em arXiv preprint arXiv:2103.15320\/} (2021).

\bibitem{martinbrualla2020nerfw}
{\sc Martin-Brualla, R., Radwan, N., Sajjadi, M. S.~M., Barron, J.~T., Dosovitskiy, A., and Duckworth, D.}
\newblock {NeRF in the Wild: Neural Radiance Fields for Unconstrained Photo Collections}.
\newblock In {\em CVPR\/} (2021).

\bibitem{matsuki2023gaussian}
{\sc Matsuki, H., Murai, R., Kelly, P.~H., and Davison, A.~J.}
\newblock Gaussian splatting slam.
\newblock {\em arXiv preprint arXiv:2312.06741\/} (2023).

\bibitem{matsushita2005full}
{\sc Matsushita, Y., Ofek, E., Tang, X., and Shum, H.-Y.}
\newblock Full-frame video stabilization.
\newblock In {\em 2005 IEEE computer society conference on computer vision and pattern recognition (CVPR'05)\/} (2005), vol.~1, IEEE, pp.~50--57.

\bibitem{maturana2015voxnet}
{\sc Maturana, D., and Scherer, S.}
\newblock Voxnet: A 3d convolutional neural network for real-time object recognition.
\newblock In {\em 2015 IEEE/RSJ international conference on intelligent robots and systems (IROS)\/} (2015), IEEE, pp.~922--928.

\bibitem{maximov2020focus}
{\sc Maximov, M., Galim, K., and Leal-Taix{\'e}, L.}
\newblock Focus on defocus: bridging the synthetic to real domain gap for depth estimation.
\newblock In {\em Proceedings of the IEEE/CVF Conference on Computer Vision and Pattern Recognition\/} (2020), pp.~1071--1080.

\bibitem{meng2021cornet}
{\sc Meng, X., Fan, C., Ming, Y., and Yu, H.}
\newblock Cornet: Context-based ordinal regression network for monocular depth estimation.
\newblock {\em IEEE Transactions on Circuits and Systems for Video Technology\/} (2021).

\bibitem{meshry2019neural}
{\sc Meshry, M., Goldman, D.~B., Khamis, S., Hoppe, H., Pandey, R., Snavely, N., and Martin-Brualla, R.}
\newblock Neural rerendering in the wild.
\newblock In {\em Proceedings of the IEEE/CVF Conference on Computer Vision and Pattern Recognition\/} (2019), pp.~6878--6887.

\bibitem{miangoleh2021boosting}
{\sc Miangoleh, S. M.~H., Dille, S., Mai, L., Paris, S., and Aksoy, Y.}
\newblock Boosting monocular depth estimation models to high-resolution via content-adaptive multi-resolution merging.
\newblock In {\em Proceedings of the IEEE/CVF Conference on Computer Vision and Pattern Recognition\/} (2021), pp.~9685--9694.

\bibitem{sky}
{\sc Mihail, R. P.~W., Bessinger, S., Jacobs, Z., and Jacobs, N.}
\newblock Sky segmentation in the wild: An empirical study.
\newblock In {\em 2016 IEEE Winter Conference on Applications of Computer Vision (WACV)\/} (March 2016), IEEE, pp.~1--6.

\bibitem{mildenhall2022nerf}
{\sc Mildenhall, B., Hedman, P., Martin-Brualla, R., Srinivasan, P.~P., and Barron, J.~T.}
\newblock Nerf in the dark: High dynamic range view synthesis from noisy raw images.
\newblock In {\em Proceedings of the IEEE/CVF conference on computer vision and pattern recognition\/} (2022), pp.~16190--16199.

\bibitem{mildenhall2020nerf}
{\sc Mildenhall, B., Srinivasan, P.~P., Tancik, M., Barron, J.~T., Ramamoorthi, R., and Ng, R.}
\newblock Nerf: Representing scenes as neural radiance fields for view synthesis.
\newblock In {\em ECCV\/} (2020).

\bibitem{mildenhall2021nerf}
{\sc Mildenhall, B., Srinivasan, P.~P., Tancik, M., Barron, J.~T., Ramamoorthi, R., and Ng, R.}
\newblock Nerf: Representing scenes as neural radiance fields for view synthesis.
\newblock {\em Communications of the ACM 65}, 1 (2021), 99--106.

\bibitem{mishra2020wavelet}
{\sc Mishra, D., Singh, S.~K., and Singh, R.~K.}
\newblock Wavelet-based deep auto encoder-decoder (wdaed)-based image compression.
\newblock {\em IEEE Transactions on Circuits and Systems for Video Technology 31}, 4 (2020), 1452--1462.

\bibitem{mitra2007dynamic}
{\sc Mitra, N.~J., Fl{\"o}ry, S., Ovsjanikov, M., Gelfand, N., Guibas, L.~J., and Pottmann, H.}
\newblock Dynamic geometry registration.
\newblock In {\em Symposium on geometry processing\/} (2007), pp.~173--182.

\bibitem{moeller2015variational}
{\sc Moeller, M., Benning, M., Sch{\"o}nlieb, C., and Cremers, D.}
\newblock Variational depth from focus reconstruction.
\newblock {\em IEEE Transactions on Image Processing 24}, 12 (2015), 5369--5378.

\bibitem{moemen20203}
{\sc Moemen, M.~Y., Elghamrawy, H., Givigi, S.~N., and Noureldin, A.}
\newblock 3-d reconstruction and measurement system based on multimobile robot machine vision.
\newblock {\em IEEE Transactions on Instrumentation and Measurement 70\/} (2020), 1--9.

\bibitem{mohideen2008image}
{\sc Mohideen, S.~K., Perumal, S.~A., and Sathik, M.~M.}
\newblock Image de-noising using discrete wavelet transform.
\newblock {\em International Journal of Computer Science and Network Security 8}, 1 (2008), 213--216.

\bibitem{mur2015orb}
{\sc Mur-Artal, R., Montiel, J. M.~M., and Tardos, J.~D.}
\newblock Orb-slam: a versatile and accurate monocular slam system.
\newblock {\em IEEE transactions on robotics 31}, 5 (2015), 1147--1163.

\bibitem{myronenko2010point}
{\sc Myronenko, A., and Song, X.}
\newblock Point set registration: Coherent point drift.
\newblock {\em IEEE transactions on pattern analysis and machine intelligence 32}, 12 (2010), 2262--2275.

\bibitem{nguyen2022fourierformer}
{\sc Nguyen, T., Pham, M., Nguyen, T., Nguyen, K., Osher, S., and Ho, N.}
\newblock Fourierformer: Transformer meets generalized fourier integral theorem.
\newblock {\em Advances in Neural Information Processing Systems 35\/} (2022), 29319--29335.

\bibitem{five-point}
{\sc Nistér, D.}
\newblock An efficient solution to the five-point relative pose problem.
\newblock {\em IEEE Transactions on Pattern Analysis and Machine Intelligence 26}, 6 (2004), 756--770.

\bibitem{nwoye2019weakly}
{\sc Nwoye, C.~I., Mutter, D., Marescaux, J., and Padoy, N.}
\newblock Weakly supervised convolutional lstm approach for tool tracking in laparoscopic videos.
\newblock {\em International journal of computer assisted radiology and surgery 14}, 6 (2019), 1059--1067.

\bibitem{oechsle2021unisurf}
{\sc Oechsle, M., Peng, S., and Geiger, A.}
\newblock Unisurf: Unifying neural implicit surfaces and radiance fields for multi-view reconstruction.
\newblock In {\em Proceedings of the IEEE/CVF International Conference on Computer Vision\/} (2021), pp.~5589--5599.

\bibitem{ozyecsil2017survey}
{\sc {\"O}zye{\c{s}}il, O., Voroninski, V., Basri, R., and Singer, A.}
\newblock A survey of structure from motion*.
\newblock {\em Acta Numerica 26\/} (2017), 305--364.

\bibitem{pan2024global}
{\sc Pan, L., Bar{\'a}th, D., Pollefeys, M., and Sch{\"o}nberger, J.~L.}
\newblock Global structure-from-motion revisited.
\newblock In {\em European Conference on Computer Vision\/} (2024), Springer, pp.~58--77.

\bibitem{park2017colored}
{\sc Park, J., Zhou, Q.-Y., and Koltun, V.}
\newblock Colored point cloud registration revisited.
\newblock In {\em Proceedings of the IEEE international conference on computer vision\/} (2017), pp.~143--152.

\bibitem{park2019deepsdf}
{\sc Park, J.~J., Florence, P., Straub, J., Newcombe, R., and Lovegrove, S.}
\newblock Deepsdf: Learning continuous signed distance functions for shape representation.
\newblock In {\em Proceedings of the IEEE/CVF conference on computer vision and pattern recognition\/} (2019), pp.~165--174.

\bibitem{park2003accurate}
{\sc Park, S.-Y., and Subbarao, M.}
\newblock An accurate and fast point-to-plane registration technique.
\newblock {\em Pattern Recognition Letters 24}, 16 (2003), 2967--2976.

\bibitem{peng2020convolutional}
{\sc Peng, S., Niemeyer, M., Mescheder, L., Pollefeys, M., and Geiger, A.}
\newblock Convolutional occupancy networks.
\newblock In {\em Computer Vision--ECCV 2020: 16th European Conference, Glasgow, UK, August 23--28, 2020, Proceedings, Part III 16\/} (2020), Springer, pp.~523--540.

\bibitem{pentland1987new}
{\sc Pentland, A.~P.}
\newblock A new sense for depth of field.
\newblock {\em IEEE transactions on pattern analysis and machine intelligence\/} (1987), 523--531.

\bibitem{surfel_primitives}
{\sc Pfister, H., Zwicker, M., Baar, J., and Gross, M.}
\newblock Surfels: Surface elements as rendering primitives.
\newblock {\em Proceedings of the ACM SIGGRAPH Conference on Computer Graphics\/} (05 2000).

\bibitem{surf-tracking}
{\sc Pinto, B., and Anurenjan, P.~R.}
\newblock Video stabilization using speeded up robust features.
\newblock In {\em 2011 International Conference on Communications and Signal Processing\/} (2011), IEEE.

\bibitem{pintore2021slicenet}
{\sc Pintore, G., Agus, M., Almansa, E., Schneider, J., and Gobbetti, E.}
\newblock Slicenet: deep dense depth estimation from a single indoor panorama using a slice-based representation.
\newblock In {\em Proceedings of the IEEE/CVF Conference on Computer Vision and Pattern Recognition\/} (2021), pp.~11536--11545.

\bibitem{pratt2017fcnn}
{\sc Pratt, H., Williams, B., Coenen, F., and Zheng, Y.}
\newblock Fcnn: Fourier convolutional neural networks.
\newblock In {\em Machine Learning and Knowledge Discovery in Databases: European Conference, ECML PKDD 2017, Skopje, Macedonia, September 18--22, 2017, Proceedings, Part I 17\/} (2017), Springer, pp.~786--798.

\bibitem{qi2017pointnet}
{\sc Qi, C.~R., Su, H., Mo, K., and Guibas, L.~J.}
\newblock Pointnet: Deep learning on point sets for 3d classification and segmentation.
\newblock In {\em Proceedings of the IEEE conference on computer vision and pattern recognition\/} (2017), pp.~652--660.

\bibitem{qin2018vins}
{\sc Qin, T., Li, P., and Shen, S.}
\newblock Vins-mono: A robust and versatile monocular visual-inertial state estimator.
\newblock {\em IEEE Transactions on Robotics 34}, 4 (2018), 1004--1020.

\bibitem{qin2023geotransformer}
{\sc Qin, Z., Yu, H., Wang, C., Guo, Y., Peng, Y., Ilic, S., Hu, D., and Xu, K.}
\newblock Geotransformer: Fast and robust point cloud registration with geometric transformer.
\newblock {\em IEEE Transactions on Pattern Analysis and Machine Intelligence 45}, 8 (2023), 9806--9821.

\bibitem{qin2022geometric}
{\sc Qin, Z., Yu, H., Wang, C., Guo, Y., Peng, Y., and Xu, K.}
\newblock Geometric transformer for fast and robust point cloud registration.
\newblock In {\em Proceedings of the IEEE/CVF conference on computer vision and pattern recognition\/} (2022), pp.~11143--11152.

\bibitem{qiu2023looking}
{\sc Qiu, J., Jiang, P.-T., Zhu, Y., Yin, Z.-X., Cheng, M.-M., and Ren, B.}
\newblock Looking through the glass: Neural surface reconstruction against high specular reflections.
\newblock In {\em Proceedings of the IEEE/CVF Conference on Computer Vision and Pattern Recognition\/} (2023), pp.~20823--20833.

\bibitem{ramamonjisoa2021single}
{\sc Ramamonjisoa, M., Firman, M., Watson, J., Lepetit, V., and Turmukhambetov, D.}
\newblock Single image depth prediction with wavelet decomposition.
\newblock In {\em Proceedings of the IEEE/CVF conference on computer vision and pattern recognition\/} (2021), pp.~11089--11098.

\bibitem{ranftl2021vision}
{\sc Ranftl, R., Bochkovskiy, A., and Koltun, V.}
\newblock Vision transformers for dense prediction.
\newblock In {\em Proceedings of the IEEE/CVF international conference on computer vision\/} (2021), pp.~12179--12188.

\bibitem{Ranftl2020}
{\sc Ranftl, R., Lasinger, K., Hafner, D., Schindler, K., and Koltun, V.}
\newblock Towards robust monocular depth estimation: Mixing datasets for zero-shot cross-dataset transfer.
\newblock {\em IEEE Transactions on Pattern Analysis and Machine Intelligence (TPAMI)\/} (2020).

\bibitem{review-stabilization-tech}
{\sc Rawat, P., and Singhai, J.}
\newblock Review of motion estimation and video stabilization techniques for hand held mobile video.
\newblock {\em Signal \& Image Processing: An International Journal (SIPIJ) 2\/} (2011).

\bibitem{rehder2016extending}
{\sc Rehder, J., Nikolic, J., Schneider, T., Hinzmann, T., and Siegwart, R.}
\newblock Extending kalibr: Calibrating the extrinsics of multiple imus and of individual axes.
\newblock In {\em 2016 IEEE International Conference on Robotics and Automation (ICRA)\/} (2016), IEEE, pp.~4304--4311.

\bibitem{ren2024agsmesh}
{\sc Ren, X., Turkulainen, M., Wang, J., Seiskari, O., Melekhov, I., Kannala, J., and Rahtu, E.}
\newblock Ags-mesh: Adaptive gaussian splatting and meshing with geometric priors for indoor room reconstruction using smartphones.
\newblock In {\em International Conference on 3D Vision (3DV)\/} (2025).

\bibitem{rippel2017real}
{\sc Rippel, O., and Bourdev, L.}
\newblock Real-time adaptive image compression.
\newblock In {\em International Conference on Machine Learning\/} (2017), PMLR, pp.~2922--2930.

\bibitem{ruan2021aifnet}
{\sc Ruan, L., Chen, B., Li, J., and Lam, M.-L.}
\newblock Aifnet: All-in-focus image restoration network using a light field-based dataset.
\newblock {\em IEEE Transactions on Computational Imaging 7\/} (2021), 675--688.

\bibitem{rusu2009fast}
{\sc Rusu, R.~B., Blodow, N., and Beetz, M.}
\newblock Fast point feature histograms (fpfh) for 3d registration.
\newblock In {\em 2009 IEEE international conference on robotics and automation\/} (2009), IEEE, pp.~3212--3217.

\bibitem{sadat2024litevae}
{\sc Sadat, S., Buhmann, J., Bradley, D., Hilliges, O., and Weber, R.~M.}
\newblock Litevae: Lightweight and efficient variational autoencoders for latent diffusion models.
\newblock {\em arXiv preprint arXiv:2405.14477\/} (2024).

\bibitem{sarode2019pcrnet}
{\sc Sarode, V., Li, X., Goforth, H., Aoki, Y., Srivatsan, R.~A., Lucey, S., and Choset, H.}
\newblock Pcrnet: Point cloud registration network using pointnet encoding.
\newblock {\em arXiv preprint arXiv:1908.07906\/} (2019).

\bibitem{satorras2021en}
{\sc Satorras, V.~G., Hoogeboom, E., and Welling, M.}
\newblock E(n) equivariant graph neural networks, 2021.

\bibitem{sattler2019understanding}
{\sc Sattler, T., Zhou, Q., Pollefeys, M., and Leal-Taixe, L.}
\newblock Understanding the limitations of cnn-based absolute camera pose regression.
\newblock In {\em Proceedings of the IEEE/CVF conference on computer vision and pattern recognition\/} (2019), pp.~3302--3312.

\bibitem{schelkens2003wavelet}
{\sc Schelkens, P., Munteanu, A., Barbarien, J., Galca, M., Giro-Nieto, X., and Cornelis, J.}
\newblock Wavelet coding of volumetric medical datasets.
\newblock {\em IEEE Transactions on medical Imaging 22}, 3 (2003), 441--458.

\bibitem{schoenberger2016sfm}
{\sc Sch\"{o}nberger, J.~L., and Frahm, J.-M.}
\newblock Structure-from-motion revisited.
\newblock In {\em Conference on Computer Vision and Pattern Recognition (CVPR)\/} (2016).

\bibitem{schonberger2016structure}
{\sc Schonberger, J.~L., and Frahm, J.-M.}
\newblock Structure-from-motion revisited.
\newblock In {\em Proceedings of the IEEE conference on computer vision and pattern recognition\/} (2016), pp.~4104--4113.

\bibitem{schoenberger2016mvs}
{\sc Sch\"{o}nberger, J.~L., Zheng, E., Pollefeys, M., and Frahm, J.-M.}
\newblock Pixelwise view selection for unstructured multi-view stereo.
\newblock In {\em European Conference on Computer Vision (ECCV)\/} (2016).

\bibitem{schops2019surfelmeshing}
{\sc Sch{\"o}ps, T., Sattler, T., and Pollefeys, M.}
\newblock Surfelmeshing: Online surfel-based mesh reconstruction.
\newblock {\em IEEE transactions on pattern analysis and machine intelligence 42}, 10 (2019), 2494--2507.

\bibitem{segal2009generalized}
{\sc Segal, A., Haehnel, D., and Thrun, S.}
\newblock Generalized-icp.
\newblock In {\em Robotics: science and systems\/} (2009), vol.~2, Seattle, WA, p.~435.

\bibitem{serafin2015nicp}
{\sc Serafin, J., and Grisetti, G.}
\newblock Nicp: Dense normal based point cloud registration.
\newblock In {\em 2015 IEEE/RSJ International Conference on Intelligent Robots and Systems (IROS)\/} (2015), IEEE, pp.~742--749.

\bibitem{shabanov2024banf}
{\sc Shabanov, A., Govindarajan, S., Reading, C., Goli, L., Rebain, D., Yi, K.~M., and Tagliasacchi, A.}
\newblock Banf: Band-limited neural fields for levels of detail reconstruction.
\newblock In {\em Proceedings of the IEEE/CVF Conference on Computer Vision and Pattern Recognition\/} (2024), pp.~20571--20580.

\bibitem{boundary}
{\sc Shabayek, A. E.~R., et~al.}
\newblock Vision-based uav attitude estimation: Progress and insights.
\newblock {\em Journal of Intelligent \& Robotic Systems 65}, 1 (2012), 295--308.

\bibitem{shavit2024learning}
{\sc Shavit, Y., Ferens, R., and Keller, Y.}
\newblock Learning single and multi-scene camera pose regression with transformer encoders.
\newblock {\em Computer Vision and Image Understanding 243\/} (2024), 103982.

\bibitem{shavit2022camera}
{\sc Shavit, Y., and Keller, Y.}
\newblock Camera pose auto-encoders for improving pose regression.
\newblock In {\em European Conference on Computer Vision\/} (2022), Springer, pp.~140--157.

\bibitem{shen1999wavelet}
{\sc Shen, K., and Delp, E.~J.}
\newblock Wavelet based rate scalable video compression.
\newblock {\em IEEE transactions on circuits and systems for video technology 9}, 1 (1999), 109--122.

\bibitem{shen2013accurate}
{\sc Shen, S.}
\newblock Accurate multiple view 3d reconstruction using patch-based stereo for large-scale scenes.
\newblock {\em IEEE transactions on image processing 22}, 5 (2013), 1901--1914.

\bibitem{shim2023diffusion}
{\sc Shim, J., Kang, C., and Joo, K.}
\newblock Diffusion-based signed distance fields for 3d shape generation.
\newblock In {\em Proceedings of the IEEE/CVF conference on computer vision and pattern recognition\/} (2023), pp.~20887--20897.

\bibitem{shotton2013scene}
{\sc Shotton, J., Glocker, B., Zach, C., Izadi, S., Criminisi, A., and Fitzgibbon, A.}
\newblock Scene coordinate regression forests for camera relocalization in rgb-d images.
\newblock In {\em Proceedings of the IEEE conference on computer vision and pattern recognition\/} (2013), pp.~2930--2937.

\bibitem{si2023fully}
{\sc Si, H., Zhao, B., Wang, D., Gao, Y., Chen, M., Wang, Z., and Li, X.}
\newblock Fully self-supervised depth estimation from defocus clue.
\newblock In {\em Proceedings of the IEEE/CVF Conference on Computer Vision and Pattern Recognition\/} (2023), pp.~9140--9149.

\bibitem{siddiqui2022texturify}
{\sc Siddiqui, Y., Thies, J., Ma, F., Shan, Q., Nie{\ss}ner, M., and Dai, A.}
\newblock Texturify: Generating textures on 3d shape surfaces.
\newblock In {\em European Conference on Computer Vision\/} (2022), Springer, pp.~72--88.

\bibitem{silberman2012indoor}
{\sc Silberman, N., Hoiem, D., Kohli, P., and Fergus, R.}
\newblock Indoor segmentation and support inference from rgbd images.
\newblock In {\em European conference on computer vision\/} (2012), Springer, pp.~746--760.

\bibitem{sitzmann2020implicit}
{\sc Sitzmann, V., Martel, J., Bergman, A., Lindell, D., and Wetzstein, G.}
\newblock Implicit neural representations with periodic activation functions.
\newblock {\em Advances in neural information processing systems 33\/} (2020), 7462--7473.

\bibitem{smolyanskiy2018importance}
{\sc Smolyanskiy, N., Kamenev, A., and Birchfield, S.}
\newblock On the importance of stereo for accurate depth estimation: An efficient semi-supervised deep neural network approach.
\newblock In {\em Proceedings of the IEEE conference on computer vision and pattern recognition workshops\/} (2018), pp.~1007--1015.

\bibitem{sosnovik2019scale}
{\sc Sosnovik, I., Szmaja, M., and Smeulders, A.}
\newblock Scale-tooltool steerable networks.
\newblock {\em arXiv preprint arXiv:1910.11093\/} (2019).

\bibitem{stuckler2014multi}
{\sc St{\"u}ckler, J., and Behnke, S.}
\newblock Multi-resolution surfel maps for efficient dense 3d modeling and tracking.
\newblock {\em Journal of Visual Communication and Image Representation 25}, 1 (2014), 137--147.

\bibitem{surh2017noise}
{\sc Surh, J., Jeon, H.-G., Park, Y., Im, S., Ha, H., and So~Kweon, I.}
\newblock Noise robust depth from focus using a ring difference filter.
\newblock In {\em Proceedings of the IEEE Conference on Computer Vision and Pattern Recognition\/} (2017), pp.~6328--6337.

\bibitem{suwajanakorn2015depth}
{\sc Suwajanakorn, S., Hernandez, C., and Seitz, S.~M.}
\newblock Depth from focus with your mobile phone.
\newblock In {\em Proceedings of the IEEE Conference on Computer Vision and Pattern Recognition\/} (2015), pp.~3497--3506.

\bibitem{tang2017depth}
{\sc Tang, H., Cohen, S., Price, B., Schiller, S., and Kutulakos, K.~N.}
\newblock Depth from defocus in the wild.
\newblock In {\em Proceedings of the IEEE conference on computer vision and pattern recognition\/} (2017), pp.~2740--2748.

\bibitem{tang2022neural}
{\sc Tang, J., Markhasin, L., Wang, B., Thies, J., and Nie{\ss}ner, M.}
\newblock Neural shape deformation priors.
\newblock {\em Advances in Neural Information Processing Systems 35\/} (2022), 17117--17132.

\bibitem{thomas2018tensor}
{\sc Thomas, N., Smidt, T., Kearnes, S., Yang, L., Li, L., Kohlhoff, K., and Riley, P.}
\newblock Tensor field networks: Rotation-and translation-equivariant neural networks for 3d point clouds.
\newblock {\em arXiv preprint arXiv:1802.08219\/} (2018).

\bibitem{thrun2008simultaneous}
{\sc Thrun, S.}
\newblock Simultaneous localization and mapping.
\newblock In {\em Robotics and cognitive approaches to spatial mapping}. Springer, 2008, pp.~13--41.

\bibitem{tong2023seeing}
{\sc Tong, J., Muthu, S., Maken, F.~A., Nguyen, C., and Li, H.}
\newblock Seeing through the glass: Neural 3d reconstruction of object inside a transparent container.
\newblock In {\em Proceedings of the IEEE/CVF Conference on Computer Vision and Pattern Recognition\/} (2023), pp.~12555--12564.

\bibitem{turkulainen2024dn}
{\sc Turkulainen, M., Ren, X., Melekhov, I., Seiskari, O., Rahtu, E., and Kannala, J.}
\newblock Dn-splatter: Depth and normal priors for gaussian splatting and meshing.
\newblock {\em arXiv preprint arXiv:2403.17822\/} (2024).

\bibitem{vaswani2017attention}
{\sc Vaswani, A., Shazeer, N., Parmar, N., Uszkoreit, J., Jones, L., Gomez, A.~N., Kaiser, {\L}., and Polosukhin, I.}
\newblock Attention is all you need.
\newblock {\em Advances in neural information processing systems 30\/} (2017).

\bibitem{vizzo2023kiss}
{\sc Vizzo, I., Guadagnino, T., Mersch, B., Wiesmann, L., Behley, J., and Stachniss, C.}
\newblock Kiss-icp: In defense of point-to-point icp--simple, accurate, and robust registration if done the right way.
\newblock {\em IEEE Robotics and Automation Letters 8}, 2 (2023), 1029--1036.

\bibitem{video-surveillance}
{\sc Walha, A., Wali, A., and Alimi, A.~M.}
\newblock Video stabilization for aerial video surveillance.
\newblock {\em Aasri Procedia 4\/} (2013), 72--77.

\bibitem{walha2015video}
{\sc Walha, A., Wali, A., and Alimi, A.~M.}
\newblock Video stabilization with moving object detecting and tracking for aerial video surveillance.
\newblock {\em Multimedia Tools and Applications 74\/} (2015), 6745--6767.

\bibitem{wang2023roreg}
{\sc Wang, H., Liu, Y., Hu, Q., Wang, B., Chen, J., Dong, Z., Guo, Y., Wang, W., and Yang, B.}
\newblock Roreg: Pairwise point cloud registration with oriented descriptors and local rotations.
\newblock {\em IEEE Transactions on Pattern Analysis and Machine Intelligence\/} (2023).

\bibitem{wang20223}
{\sc Wang, J., Gong, Z., Tao, B., and Yin, Z.}
\newblock A 3-d reconstruction method for large freeform surfaces based on mobile robotic measurement and global optimization.
\newblock {\em IEEE Transactions on Instrumentation and Measurement 71\/} (2022), 1--9.

\bibitem{wang2019real}
{\sc Wang, K., Gao, F., and Shen, S.}
\newblock Real-time scalable dense surfel mapping.
\newblock In {\em 2019 International conference on robotics and automation (ICRA)\/} (2019), IEEE, pp.~6919--6925.

\bibitem{wang2021bridging}
{\sc Wang, N.-H., Wang, R., Liu, Y.-L., Huang, Y.-H., Chang, Y.-L., Chen, C.-P., and Jou, K.}
\newblock Bridging unsupervised and supervised depth from focus via all-in-focus supervision.
\newblock In {\em Proceedings of the IEEE/CVF International Conference on Computer Vision\/} (2021), pp.~12621--12631.

\bibitem{wang2021neus}
{\sc Wang, P., Liu, L., Liu, Y., Theobalt, C., Komura, T., and Wang, W.}
\newblock Neus: Learning neural implicit surfaces by volume rendering for multi-view reconstruction.
\newblock {\em arXiv preprint arXiv:2106.10689\/} (2021).

\bibitem{wang2019deep}
{\sc Wang, Y., and Solomon, J.~M.}
\newblock Deep closest point: Learning representations for point cloud registration.
\newblock In {\em Proceedings of the IEEE/CVF international conference on computer vision\/} (2019), pp.~3523--3532.

\bibitem{wang2024enhancing}
{\sc Wang, Y., Zhou, P., Geng, G., An, L., and Zhou, M.}
\newblock Enhancing point cloud registration with transformer: cultural heritage protection of the terracotta warriors.
\newblock {\em Heritage Science 12}, 1 (2024), 314.

\bibitem{wang2024puzzlefusion}
{\sc Wang, Z., Chen, J., and Furukawa, Y.}
\newblock Puzzlefusion++: Auto-agglomerative 3d fracture assembly by denoise and verify.
\newblock {\em arXiv preprint arXiv:2406.00259\/} (2024).

\bibitem{e2cnn}
{\sc Weiler, M., and Cesa, G.}
\newblock {General E(2)-Equivariant Steerable CNNs}.
\newblock In {\em Conference on Neural Information Processing Systems (NeurIPS)\/} (2019).

\bibitem{weiler20183d}
{\sc Weiler, M., Geiger, M., Welling, M., Boomsma, W., and Cohen, T.~S.}
\newblock 3d steerable cnns: Learning rotationally equivariant features in volumetric data.
\newblock {\em Advances in Neural Information Processing Systems 31\/} (2018).

\bibitem{weiler2018learning}
{\sc Weiler, M., Hamprecht, F.~A., and Storath, M.}
\newblock Learning steerable filters for rotation equivariant cnns.
\newblock In {\em Proceedings of the IEEE Conference on Computer Vision and Pattern Recognition\/} (2018), pp.~849--858.

\bibitem{whelan2018reconstructing}
{\sc Whelan, T., Goesele, M., Lovegrove, S.~J., Straub, J., Green, S., Szeliski, R., Butterfield, S., Verma, S., Newcombe, R.~A., Goesele, M., et~al.}
\newblock Reconstructing scenes with mirror and glass surfaces.
\newblock {\em ACM Trans. Graph. 37}, 4 (2018), 102.

\bibitem{wong2025survey}
{\sc Wong, L. H.~K., Kang, X., Bai, K., and Zhang, J.}
\newblock A survey of robotic navigation and manipulation with physics simulators in the era of embodied ai.
\newblock {\em arXiv preprint arXiv:2505.01458\/} (2025).

\bibitem{xiang2017posecnn}
{\sc Xiang, Y., Schmidt, T., Narayanan, V., and Fox, D.}
\newblock Posecnn: A convolutional neural network for 6d object pose estimation in cluttered scenes.
\newblock {\em arXiv preprint arXiv:1711.00199\/} (2017).

\bibitem{xiao2021early}
{\sc Xiao, T., Singh, M., Mintun, E., Darrell, T., Doll{\'a}r, P., and Girshick, R.}
\newblock Early convolutions help transformers see better.
\newblock {\em Advances in Neural Information Processing Systems 34\/} (2021), 30392--30400.

\bibitem{xie2016deep3d}
{\sc Xie, J., Girshick, R., and Farhadi, A.}
\newblock Deep3d: Fully automatic 2d-to-3d video conversion with deep convolutional neural networks.
\newblock In {\em European conference on computer vision\/} (2016), Springer, pp.~842--857.

\bibitem{resnet}
{\sc Xie, S., et~al.}
\newblock Aggregated residual transformations for deep neural networks.
\newblock In {\em Proceedings of the IEEE Conference on Computer Vision and Pattern Recognition\/} (2017).

\bibitem{xiong1993depth}
{\sc Xiong, Y., and Shafer, S.~A.}
\newblock Depth from focusing and defocusing.
\newblock In {\em Proceedings of IEEE Conference on Computer Vision and Pattern Recognition\/} (1993), IEEE, pp.~68--73.

\bibitem{xu2018youtube}
{\sc Xu, N., Yang, L., Fan, Y., Yang, J., Yue, D., Liang, Y., Price, B., Cohen, S., and Huang, T.}
\newblock Youtube-vos: Sequence-to-sequence video object segmentation.
\newblock In {\em Proceedings of the European conference on computer vision (ECCV)\/} (2018), pp.~585--601.

\bibitem{xu2019disn}
{\sc Xu, Q., Wang, W., Ceylan, D., Mech, R., and Neumann, U.}
\newblock Disn: Deep implicit surface network for high-quality single-view 3d reconstruction.
\newblock {\em Advances in neural information processing systems 32\/} (2019).

\bibitem{xueyang2023simultaneous}
{\sc Xueyang, K., Xu, L., Zou, Y., Xu, H., and Ma, L.}
\newblock Simultaneous localization and mapping using cameras capturing multiple spectra of light, June~8 2023.
\newblock US Patent App. 18/004,795.

\bibitem{yang2022deep}
{\sc Yang, F., Huang, X., and Zhou, Z.}
\newblock Deep depth from focus with differential focus volume.
\newblock In {\em Proceedings of the IEEE/CVF Conference on Computer Vision and Pattern Recognition\/} (2022), pp.~12642--12651.

\bibitem{pf-tracking}
{\sc Yang, J., Schonfeld, D., and Mohamed, M.}
\newblock Robust video stabilization based on particle filter tracking of projected camera motion.
\newblock {\em IEEE Transactions on Circuits and Systems for Video Technology 19}, 7 (2009), 945--954.

\bibitem{depth_anything_v1}
{\sc Yang, L., Kang, B., Huang, Z., Xu, X., Feng, J., and Zhao, H.}
\newblock Depth anything: Unleashing the power of large-scale unlabeled data.
\newblock In {\em CVPR\/} (2024).

\bibitem{yang2024depth}
{\sc Yang, L., Kang, B., Huang, Z., Xu, X., Feng, J., and Zhao, H.}
\newblock Depth anything: Unleashing the power of large-scale unlabeled data.
\newblock {\em arXiv preprint arXiv:2401.10891\/} (2024).

\bibitem{depth_anything_v2}
{\sc Yang, L., Kang, B., Huang, Z., Zhao, Z., Xu, X., Feng, J., and Zhao, H.}
\newblock Depth anything v2.
\newblock {\em arXiv:2406.09414\/} (2024).

\bibitem{yang2018deep}
{\sc Yang, N., Wang, R., Stuckler, J., and Cremers, D.}
\newblock Deep virtual stereo odometry: Leveraging deep depth prediction for monocular direct sparse odometry.
\newblock In {\em Proceedings of the European conference on computer vision (ECCV)\/} (2018), pp.~817--833.

\bibitem{yariv2021volume}
{\sc Yariv, L., Gu, J., Kasten, Y., and Lipman, Y.}
\newblock Volume rendering of neural implicit surfaces.
\newblock {\em Advances in Neural Information Processing Systems 34\/} (2021), 4805--4815.

\bibitem{yariv2023bakedsdf}
{\sc Yariv, L., Hedman, P., Reiser, C., Verbin, D., Srinivasan, P.~P., Szeliski, R., Barron, J.~T., and Mildenhall, B.}
\newblock Bakedsdf: Meshing neural sdfs for real-time view synthesis.
\newblock In {\em ACM SIGGRAPH 2023 Conference Proceedings\/} (2023), pp.~1--9.

\bibitem{ye20243d}
{\sc Ye, K., Hou, Q., and Zhou, K.}
\newblock 3d gaussian splatting with deferred reflection.
\newblock In {\em ACM SIGGRAPH 2024 Conference Papers\/} (2024), pp.~1--10.

\bibitem{yin2019enforcing}
{\sc Yin, W., Liu, Y., Shen, C., and Yan, Y.}
\newblock Enforcing geometric constraints of virtual normal for depth prediction.
\newblock In {\em Proceedings of the IEEE/CVF International Conference on Computer Vision\/} (2019), pp.~5684--5693.

\bibitem{selfie-tracking}
{\sc Yu, J., et~al.}
\newblock Real-time selfie video stabilization.
\newblock In {\em Proceedings of the IEEE/CVF Conference on Computer Vision and Pattern Recognition\/} (2021).

\bibitem{optical-flow}
{\sc Yu, J., and Ramamoorthi, R.}
\newblock Learning video stabilization using optical flow.
\newblock In {\em Proceedings of the IEEE/CVF Conference on Computer Vision and Pattern Recognition\/} (2020).

\bibitem{yu2021wavefill}
{\sc Yu, Y., Zhan, F., Lu, S., Pan, J., Ma, F., Xie, X., and Miao, C.}
\newblock Wavefill: A wavelet-based generation network for image inpainting.
\newblock In {\em Proceedings of the IEEE/CVF international conference on computer vision\/} (2021), pp.~14114--14123.

\bibitem{Yu2022SDFStudio}
{\sc Yu, Z., Chen, A., Antic, B., Peng, S., Bhattacharyya, A., Niemeyer, M., Tang, S., Sattler, T., and Geiger, A.}
\newblock Sdfstudio: A unified framework for surface reconstruction, 2022.

\bibitem{yuan2022efficient}
{\sc Yuan, C., Xu, W., Liu, X., Hong, X., and Zhang, F.}
\newblock Efficient and probabilistic adaptive voxel mapping for accurate online lidar odometry.
\newblock {\em IEEE Robotics and Automation Letters 7}, 3 (2022), 8518--8525.

\bibitem{zeng20173dmatch}
{\sc Zeng, A., Song, S., Nie{\ss}ner, M., Fisher, M., Xiao, J., and Funkhouser, T.}
\newblock 3dmatch: Learning local geometric descriptors from rgb-d reconstructions.
\newblock In {\em Proceedings of the IEEE conference on computer vision and pattern recognition\/} (2017), pp.~1802--1811.

\bibitem{zeng20163dmatch}
{\sc Zeng, A., Song, S., Nie{\ss}ner, M., Fisher, M., Xiao, J., and Funkhouser, T.}
\newblock 3dmatch: Learning local geometric descriptors from rgb-d reconstructions.
\newblock In {\em CVPR\/} (2017).

\bibitem{zhang2019ga}
{\sc Zhang, F., Prisacariu, V., Yang, R., and Torr, P.~H.}
\newblock Ga-net: Guided aggregation net for end-to-end stereo matching.
\newblock In {\em Proceedings of the IEEE/CVF Conference on Computer Vision and Pattern Recognition\/} (2019), pp.~185--194.

\bibitem{zhang20233d}
{\sc Zhang, X., Yang, J., Zhang, S., and Zhang, Y.}
\newblock 3d registration with maximal cliques.
\newblock In {\em Proceedings of the IEEE/CVF conference on computer vision and pattern recognition\/} (2023), pp.~17745--17754.

\bibitem{zhang2000flexible}
{\sc Zhang, Z.}
\newblock A flexible new technique for camera calibration.
\newblock {\em IEEE Transactions on pattern analysis and machine intelligence 22}, 11 (2000), 1330--1334.

\bibitem{zheng2023locally}
{\sc Zheng, X.-Y., Pan, H., Wang, P.-S., Tong, X., Liu, Y., and Shum, H.-Y.}
\newblock Locally attentional sdf diffusion for controllable 3d shape generation.
\newblock {\em ACM Transactions on Graphics (ToG) 42}, 4 (2023), 1--13.

\bibitem{zhou2024udiff}
{\sc Zhou, J., Zhang, W., Ma, B., Shi, K., Liu, Y.-S., and Han, Z.}
\newblock Udiff: Generating conditional unsigned distance fields with optimal wavelet diffusion.
\newblock In {\em Proceedings of the IEEE/CVF Conference on Computer Vision and Pattern Recognition\/} (2024), pp.~21496--21506.

\bibitem{zhou2016fast}
{\sc Zhou, Q.-Y., Park, J., and Koltun, V.}
\newblock Fast global registration.
\newblock In {\em Computer Vision--ECCV 2016: 14th European Conference, Amsterdam, The Netherlands, October 11-14, 2016, Proceedings, Part II 14\/} (2016), Springer, pp.~766--782.

\bibitem{zhu2023e2pn}
{\sc Zhu, M., Ghaffari, M., Clark, W.~A., and Peng, H.}
\newblock E2pn: Efficient se (3)-equivariant point network.
\newblock In {\em Proceedings of the IEEE/CVF Conference on Computer Vision and Pattern Recognition\/} (2023), pp.~1223--1232.

\bibitem{zhu2022nice}
{\sc Zhu, Z., Peng, S., Larsson, V., Xu, W., Bao, H., Cui, Z., Oswald, M.~R., and Pollefeys, M.}
\newblock Nice-slam: Neural implicit scalable encoding for slam.
\newblock In {\em Proceedings of the IEEE/CVF Conference on Computer Vision and Pattern Recognition\/} (2022), pp.~12786--12796.

\end{thebibliography}

\end{document}